\documentclass[lettersize,journal,twoside]{IEEEtran}
\usepackage{amsmath,amsfonts}
\usepackage{array}
\usepackage[caption=false,font=normalsize,labelfont=sf,textfont=sf]{subfig}
\usepackage{textcomp}
\usepackage{stfloats}
\usepackage{url}
\usepackage{verbatim}
\usepackage{graphicx}
\usepackage{cite}
\usepackage{xfrac}
\usepackage{multirow}
\usepackage[ruled,linesnumbered]{algorithm2e}
\usepackage[section]{placeins}
\graphicspath{{Images/}}
\usepackage[colorlinks=true]{hyperref}
\usepackage[figure, figure*, table, table*]{hypcap}
\usepackage{fancyhdr}
\usepackage{CJK}
\usepackage{indentfirst}
\usepackage{cases}

\begin{document}

\title{Multi-view Point Cloud Registration based on Evolutionary Multitasking with Bi-Channel Knowledge Sharing Mechanism}

\author{Yue~Wu,~\IEEEmembership{Member,~IEEE,} Yibo~Liu, Maoguo~Gong,~\IEEEmembership{Senior Member,~IEEE,} Peiran~Gong, Hao~Li, Zedong~Tang, Qiguang~Miao,~\IEEEmembership{Senior Member,~IEEE,} Wenping~Ma,~\IEEEmembership{Senior Member,~IEEE}
\thanks{\copyright~2022 IEEE.  Personal use of this material is permitted.  Permission from IEEE must be obtained for all other uses, in any current or future media, including reprinting/republishing this material for advertising or promotional purposes, creating new collective works, for resale or redistribution to servers or lists, or reuse of any copyrighted component of this work in other works.}
\thanks{This work was supported by the Key-Area Research and Development Program of Guangdong Province (2020B090921001), National Natural Science Foundation of China (62036006), Natural Science Basic Research Plan in Shaanxi Province of China (2022JM-327), MindSpore, CANN (Compute Architecture for Neural Networks) and Ascend AI Processor. \textit{(Corresponding author: Maoguo~Gong.)}}
\thanks{Yue~Wu, Yibo~Liu, Peiran~Gong and Qiguang~Miao are with the School of Computer Science and Technology, Key Laboratory of Big Data and Intelligent Vision, Xidian University, Xi'an~710071, China (e-mail: \nolinkurl{ywu@xidian.edu.cn}; \nolinkurl{yb_liu@stu.xidian.edu.cn}; \nolinkurl{gpr@stu.xidian.edu.cn}; \nolinkurl{qgmiao@xidian.edu.cn}).}
\thanks{Maoguo~Gong, Hao~Li and Zedong~Tang are with the School of Electronic Engineering, Key Laboratory of Intelligent Perception and Image Understanding of Ministry of Education, Xidian University, Xi'an 710071, China (e-mail: \nolinkurl{gong@ieee.org}; \nolinkurl{omegalihao@gmail.com}; \nolinkurl{omegatangzd@gmail.com}).}
\thanks{Wenping~Ma is with the School of Artificial Intelligence, Key Laboratory of Intelligent Perception and Image Understanding of Ministry of Education, Xidian University, Xi'an 710071, China (e-mail: \nolinkurl{wpma@mail.xidian.edu.cn}).}
}

\fancyhead[L]{\footnotesize THIS PAPER HAS BEEN ACCEPTED AT IEEE TRANSACTIONS ON EMERGING TOPICS IN COMPUTATIONAL INTELLIGENCE}

\maketitle\thispagestyle{fancy}

\begin{abstract}
Multi-view point cloud registration is fundamental in 3D reconstruction. Since there are close connections between point clouds captured from different viewpoints, registration performance can be enhanced if these connections be harnessed properly. Therefore, this paper models the registration problem as multi-task optimization, and proposes a novel bi-channel knowledge sharing mechanism for effective and efficient problem solving. The modeling of multi-view point cloud registration as multi-task optimization are twofold. By simultaneously considering the local accuracy of two point clouds as well as the global consistency posed by all the point clouds involved, a fitness function with an adaptive threshold is derived. Also a framework of the co-evolutionary search process is defined for the concurrent optimization of multiple fitness functions belonging to related tasks. To enhance solution quality and convergence speed, the proposed bi-channel knowledge sharing mechanism plays its role. The intra-task knowledge sharing introduces aiding tasks that are much simpler to solve, and useful information is shared across aiding tasks and the original tasks, accelerating the search process. The inter-task knowledge sharing explores commonalities buried among the original tasks, aiming to prevent tasks from getting stuck to local optima. Comprehensive experiments conducted on model object as well as scene point clouds show the efficacy of the proposed method. 
\end{abstract}

\begin{IEEEkeywords}
Evolutionary multitasking (EMT), knowledge sharing, multi-task optimization (MTO), point cloud registration. 
\end{IEEEkeywords}

\section{Introduction}
\label{secIntroduction}
\begin{figure}
	\centering
	\includegraphics[height=0.25\columnwidth]{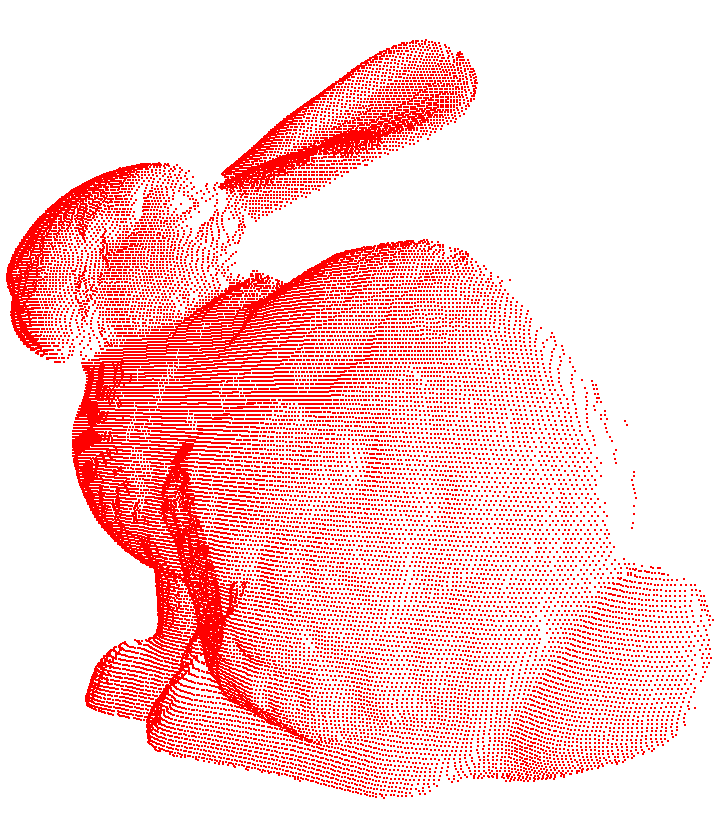}
	\includegraphics[height=0.25\columnwidth]{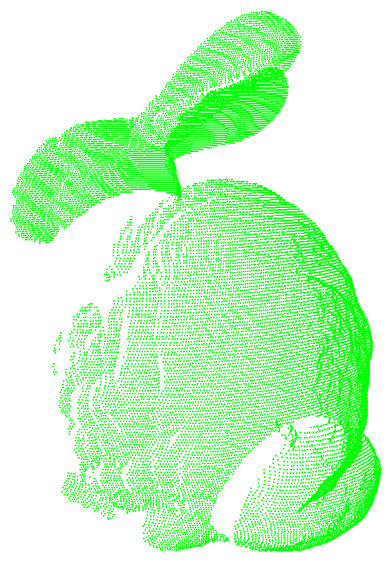}
	\includegraphics[height=0.25\columnwidth]{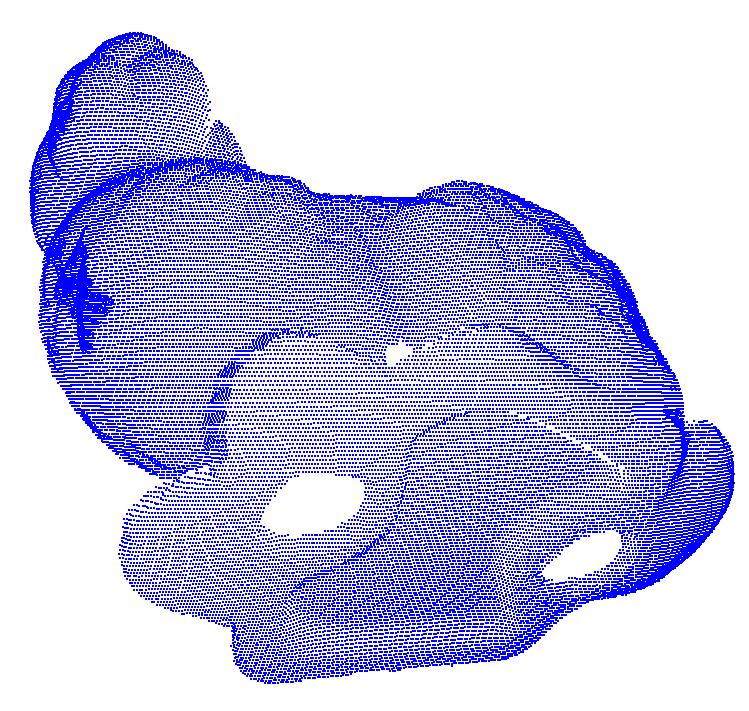}
	\includegraphics[height=0.25\columnwidth]{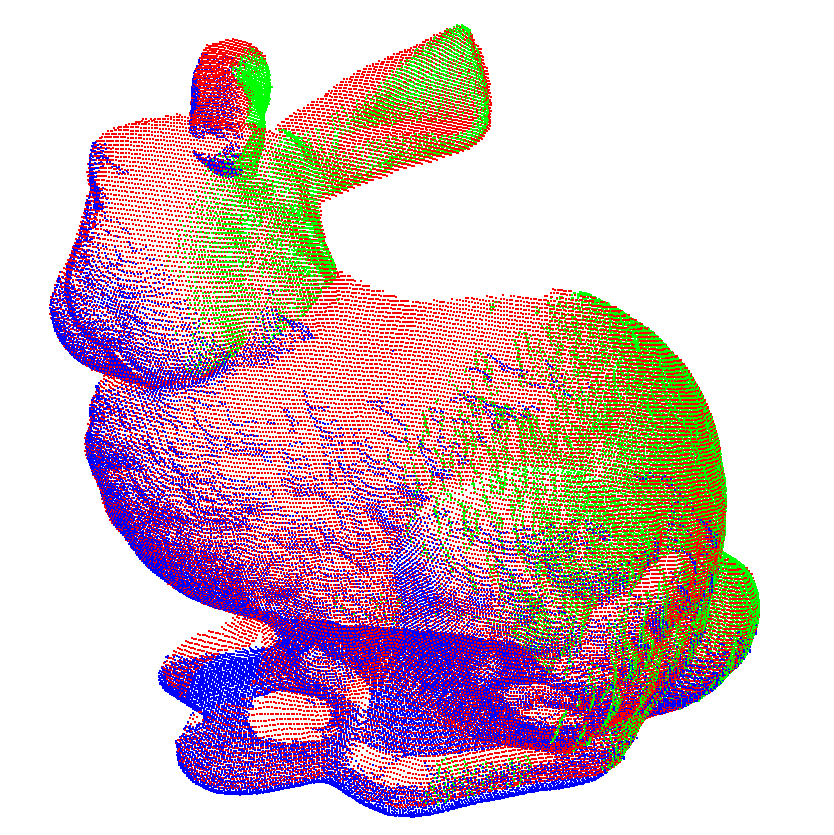}
	\caption{Practical example of multi-view point cloud registration. The first three columns show the original point clouds to be registered, and the last column shows the expected registration result.}
	\label{figRig}
\end{figure}

\IEEEPARstart{T}{he} problem of point cloud registration is fundamental in many domains, such as 3D reconstruction\cite{dai2017scannet,8099512}, computer vision\cite{7368945,7918612}, robotics\cite{7298781,8461063}, etc. Due to occlusion and restricted viewpoint of each scan while obtaining data, a set of point clouds are required to reconstruct the whole object model or scene, and different coordinate frames are attached to each of these point clouds. The objective of pairwise point cloud registration is to find a rigid transformation that brings a source point cloud to the same coordinate frame as the target one. Accordingly, the objective of multi-view point cloud registration is to simultaneously estimate optimal rigid transformations for each point cloud, in order to bring them to the same coordinate frame. Pairwise registration is not trivial because low overlap between two point clouds may be the case, and there is a large search space as well. The problem becomes even more challenging when it comes to multi-view registration, for that global consistency concerning all the point clouds involved is hard to be guaranteed. Fig.~\ref{figRig} shows an example of multi-view point cloud registration with three point clouds involved.

Given two 3D point clouds, the source point cloud $P=\left \{ p_{i}\in \mathbb{R}^{3} \right \} _{i=1}^{N_{p}} $ and the target one $Q=\left \{ q_{i}\in \mathbb{R}^{3} \right \} _{i=1}^{N_{q}} $, the goal of pairwise point cloud registration is to find the optimal rigid transformation $T\in{SE}(3)$\footnote{$SE(3)=\left \{ T=\begin{bmatrix}R & t\\0^T & 1\end{bmatrix} \in \mathbb{R}^{4\times 4}\mid R\in SO(3),t\in \mathbb{R}^3\right \} $}, which is composed of a rotation matrix $R\in{SO}(3)$\footnote{$SO(3)=\left \{ R\in \mathbb{R}^{3\times 3} \mid RR^T=I,det(R)=1 \right \}$} and a translation vector $t\in \mathbb{R}^3$, that brings $P$ to the same coordinate where $Q$ lives. Generally, overlap ratio between two point clouds may vary in a wide range due to captured from different viewpoints, resulting in a varied number of outliers in the registration. Moreover, a large amount of noise may also exist in real scanned data, increasing the difficulty of the problem. Therefore, the registration should be implemented in a robust manner in order to alleviate the negative effect caused by data contamination. The registration between two point clouds can be formulated as
\begin{equation}
	\label{equBasicForm}
	\underset{T \in {SE}(3) }{min} \sum_{i=1}^{N_{p} } \varphi (\varrho_i (T) )
\end{equation}
where $N_p$ is the number of points in $P$, $\varphi(\cdot)$ is a robust loss function, and the error function $\varrho(\cdot)$ between matched points of the two point clouds represents the distance metric used in the registration. In this work, Euclidean distance is considered, i.e.,
\begin{equation}
	\label{equEculideanDistance}
	\varrho _i (T)=\left \| Tp_i - q_j \right \| 
\end{equation}
with $q_{j}$ the closest point in $Q$ of transformed $p_{i}$:
\begin{equation}
	\label{equMatchedPoints}
	q_{j}=\underset{q_{k}\in Q }{argmin} \left \| Tp_i - q_{k} \right \|  
\end{equation}

It has to be mentioned that homogeneous coordinate is used to denote a point. For example, the coordinate of point $p_i$ is actually $[x_i, y_i, z_i, 1]^T$, with the fourth dimension always be $1$. The use of homogeneous coordinate is for convenience of mathematical notation. Function \eqref{equBasicForm} can solve the registration between two point clouds, which considers only local accuracy. In real applications, the problem of multi-view point cloud registration is more common and more challenging, with the objective to simultaneously bring several point clouds acquired from different viewpoints to the same coordinate frame. A simple way to achieve multi-view registration is through concatenating pairwise registration results, which is known to accumulate errors along the process, resulting in a visible gap between the first point cloud and the last one. The error accumulation comes from the violation of global consistency caused by combining registration results with local accuracy maximizing.
Therefore, it is a natural thought to concern both local accuracy and global consistency when solving pairwise registration, in order to strike a balance between local and global conflicts in the problem of multi-view registration. 
Consider the multi-view registration problem with three point clouds involved, and denote them $P^1,P^2,P^3$. Then generally there are three pairwise registration tasks to be solved, which treat $P^1,P^2,P^3$ and $P^2,P^3,P^1$ as source and target, respectively. Accordingly, transformation matrices $T_{12},T_{23},T_{31}$ are solutions to each of the three tasks, with $T_{ij}$ representing the transformation of $P^i$ to the coordinate of $P^j$. Given the settings above, the global consistency between $P^1,P^2,P^3$ should be satisfied to alleviate error accumulation:
\begin{equation}
	\label{equGlobalConsistent}
	T_{31}*T_{23}*T_{12}=I
\end{equation}
where $I$ represents the $4\times4$ identity matrix. Suppose a concrete point in $P^1$, after being transformed by the combination of the transformation matrices $T_{12},T_{23},T_{31}$, should get back to the exact position where it starts. 

\begin{figure}
	\centering
	\includegraphics[width=0.45\columnwidth]{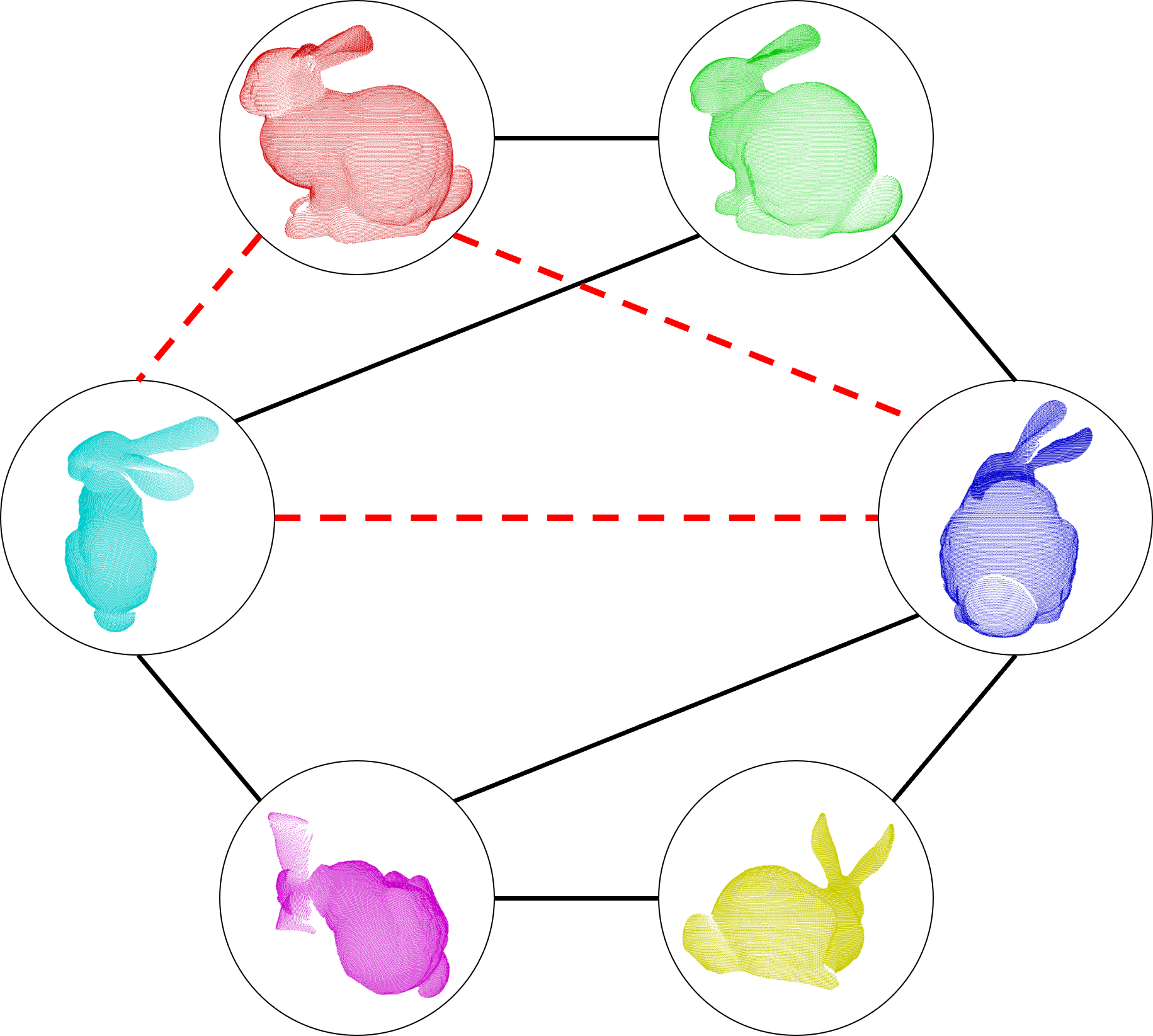}
	\includegraphics[width=0.45\columnwidth]{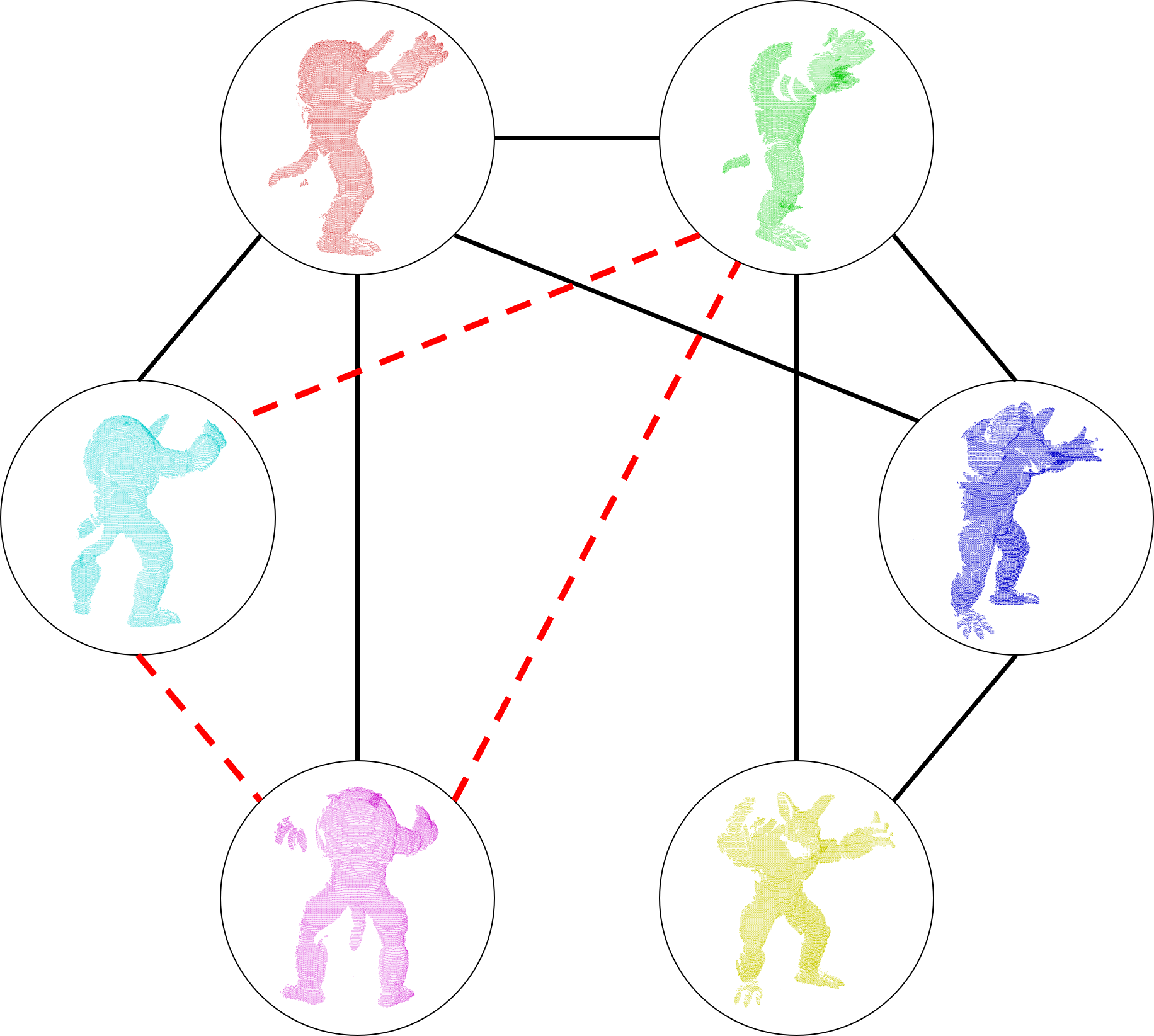}
	\caption{Examples of connections between real scanned point clouds. Overlap areas exist between point clouds connected by lines. Point clouds connected by red dashed lines form a loop.}
	\label{figExamplesofLoop}
\end{figure}

Inspired by the ability of human brain to tackle with multiple related tasks simultaneously, evolutionary multi-task optimization (EMTO) has recently emerged as an increasingly popular research topic in the field of evolutionary computation. Compared with its single task counterparts, i.e., traditional evolutionary algorithms, which solve only one task in a single run, EMTO is able to conduct evolutionary search concurrently on multiple optimization problems possessing varied search spaces and unique function landscapes. The superior performance of EMTO benefits from the knowledge sharing mechanism incorporated, with which useful knowledge is transferred across tasks while the search process online. Compared with traditional EAs, both convergence speed and solution quality of EMTO methods have been shown to be enhanced on continuous optimization as well as discrete optimization problems\cite{7161358,Ong2016EvolutionaryMA,7850039,8401802,7464295}. Despite its satisfying applications in the literature, the potential of EMTO has not been investigated in point cloud registration problems.

Multi-view point cloud registration can be considered as multi-task optimization, because there are close connections between point clouds captured from different viewpoints, and useful knowledge can be shared across tasks if these connections can be properly harnessed. For example, as shown in Fig.~\ref{figExamplesofLoop}, there are two undirected graphs, where each node represents a certain point cloud, and each edge connects two point clouds if there exists overlap area between the two. Loops can be found in the graphs, and red dashed edges in each graph show an example of a loop containing three point clouds. Within each loop, the loop closure constratint, i.e., the global consistent constraint, shoud be satisfied to reach a high quality registration. However, there is no such guarantee in existing EA based methods. Therefore, this work investigates to simultaneously slove the registration of point clouds that form a loop using EMTO techniques, by concurrently optimizing fitness functions which concern both the pairwise registration accuracy and the loop closure constraint.

Existing EMTO methods have developed implicit as well as explicit knowledge sharing mechanisms, for accelerating convergence speed and enhance solution quality. Generally, implicit knowledge sharing based methods employ one single population for solving multiple optimization tasks, and information is shared across tasks through crossover operations between individuals possessing different skill factors. Although there is performance gain, the optimization for solving each task may be overly influenced by genetic materials transferred from other tasks. Due to the randomness of knowledge transfer, benefits of positive knowledge transfer may be canceled out in the presence of negative knowledge transfer, leading to limited enhancement. In contrast to implicit EMTO methods, multiple populations employed in the explicit methods facilitate a better control over the direction and extent of knowledge transfer, and multiple search mechanisms are able to be incorporated to solve different tasks. However, the explicit EMTO methods are designed to be problem dependent, so a direct transfer of existing explicit EMTO methods to point cloud registration is not expected to work well. Therefore, in this paper an explicit EMTO method designed for point cloud registration is proposed.

The proposed method is termed as MTPCR, indicating explicit evolutionary MultiTasking Point Cloud Registration. In MTPCR, each component task has a unique population, and knowledge sharing occurs in two ways. For the purpose of intra-task knowledge sharing, aiding tasks that are much simpler to solve are introduced, and accordingly the tasks being helped (the target tasks) are called original tasks. The intra-task knowledge sharing guarantees that useful information in the form of genetic materials, is transferred between aiding tasks and original tasks, boosting the convergence speed of the original tasks. The inter-task knowledge sharing is proposed to further prevent tasks from getting stuck to local optima, by leveraging the connections posed by loop closure constraint, as shown in (\ref{equGlobalConsistent}). For this purpose, best individuals of the original tasks are selected as the knowledge source, and knowledge being transferred to a certain task is learned by the combination of best individuals from other tasks. Comprehensive experiments are conducted on point clouds from model objects and indoor scenes, validating the effectiveness of the proposed approach.

Main contributions of this paper are summarized as follows. 

\begin{enumerate}
\item This paper solves multi-view point cloud registration in an evolutionary multitasking setting. The modeling of multi-view registration as multi-task optimization enables the concurrence of pairwise accuracy and global consistency, which are both essential in problem solving.

\item An explicit evolutionary multi-task optimization approach with a bi-channel knowledge sharing mechanism is developed. The proposal of intra-task sharing aims to boost convergence speed of each component task, and the idea of inter-task sharing explores commonalities between tasks to avoid local optima.

\item To strike a balance between local accuracy and global consistency, a robust fitness function is derived to evaluate individuals with respect to each task. Local accuracy is guaranteed with an adaptive threshold, and global consistency is considered by the incorporation of loop closure constraint.
\end{enumerate}

The remainder of this paper is organized as follows. Section~\ref{secPreliminary} begins with a literature review on point cloud registration as well as recent advancements in evolutionary multitasking, and then the motivation of this paper is introduced. Section~\ref{secMethodology} presents details of the proposed explicit evolutionary multitasking point cloud registration method. The design of experiments is introduced in Section~\ref{secEcperimentDesign}, then Section~\ref{secResultsandDiscussion} presents and discusses the experimental results. Finally, conclusions are drawn in Section~\ref{secConclusion}.

\section{Preliminary}
\label{secPreliminary}
In this section, literature review of related works for point cloud registration and evolutionary multitasking methods is first presented, and then motivation of this paper is analyzed.

\subsection{Related Work}
\subsubsection{Point Cloud Registration}
A great number of point cloud registration methods have been developed in the literature. Among these methods, the Iterative Closest Point (ICP) algorithm\cite{121791} is the most popular one and inspired a variety of subsequent works\cite{924423,Pomerleau2015ARO}. Popular though ICP is, its assumption, full overlap between two point clouds, can rarely be the case in real scanned data. The trimmed ICP\cite{CHETVERIKOV2005299} was then developed to solve the registration with partial overlap by estimating and updating overlap ratio between two point clouds. However, the ICP family are known to be local convergent, demanding fine initializations to provide satisfactory results. Descriptors based methods are another route to achieve registration, PFH\cite{Rusu2008PersistentPF}, FPFH\cite{5152473} and ISS\cite{5457637} are examples of commonly used local descriptors. After potential matches between points from different point clouds are established using computed descriptors, RANSAC\cite{Fischler1981RandomSC} based approach is usually adopted later to achieve coarse registration. Efficient though these methods are, they are also shown to be sensitive to noise, and performances degrade rapidly with the increase of noise levels. Gaussian Mixture Model (GMM) based methods\cite{agenerativemodel,7954698,9201412} assume that points in different point clouds are drawn from one or several gaussian mixtures, and expectation maximization (EM) algorithm is then used to achieve registration. Although they are more robust than ICP based ones, there is a tendency for them to get stuck to local optima as well, thus initial transformations are also required. With the rapid development of deep learning techniques, some deep learning based registration methods emerged in recent years, though there is a lack of interpretability\cite{9521221}. PointNetLK\cite{aoki2019pointnetlk} is the first attempt in this domain, and some other work follows\cite{9009450,9157132,9577334}. Despite large amounts of labeled data are required during training, these trained models are shown to have preferences over some categories, leading to problems in terms of generalization. Moreover, these models are trained mainly on point clouds with full or relatively high overlap, and the ground truth rotations as well as translations are small, leading to inferior performance compared with traditional methods when it comes to real scanned data.

There are also some prior works solving registration problems using evolutionary algorithms (EAs)\cite{LOMONOSOV20061201,SILVA2007114,Zhu2014RobustRO,s17091979}. Due to the advantage brought by population based search, initial transformations are not required in EA based methods. Furthermore, there is no restriction that the fitness function be convex, facilitating accurate evaluation specialized for point cloud registration which may be somewhat complex. For example, \cite{SILVA2007114} proposed a surface interpenetration measure (SIM) as the optimization objective between two point clouds, which is accurate but not computationally efficient. Existing EA based methods were designed to solve pairwise registration problems where only local accuracy need to be considered, while multi-view point cloud registration which also requires the incorporation of global consistency is more general in real applications. Therefore, this paper solves the problem of multi-view registration by striking a balance between local accuracy and global consistency.

\subsubsection{Evolutionary multitasking}
Evolutionary multi-task optimization (EMTO) has become an increasingly attracting research topic in the field of evolutionary optimization and has been applied in many scientific and pritical engineering problems\cite{8114198,8616832,8405560,8666053}. The proposal of EMTO aims to simtaneously solve several optimization tasks, which have been considered distinct ones traditionally. Without loss of generality, assume $k$ optimization tasks to be solved, the goal of EMTO is to search optimal solutions for each of the $k$ tasks, i.e., $\{x_1^*, x_2^*, \dots, x_k^*\} = argmin\{f_1(x_1), f_2(x_2), \dots, f_k(x_k)\}$, where $x_j^* \in X_j$ for any $j=1,2,\dots,k$, and $X_j$ is a $d_j$-dimensional search space for the $j^{th}$ task. In the literature, multifactorial evolutionary algorithm (MFEA) was proposed by Gupta \textit{et al.}\cite{7161358} as the early EMTO paradigm. MFEA uses a unified search space for all the tasks involved, with the dimension of the unified search space the same as the largest one among the $k$ tasks. A homogenous population is shared by all the $k$ tasks, and individuals belonging to each task are distinguished by their skill factors. Knowledge transfer happens implicitly through crossover operations between individuals possessing different skill factors as the whole population evolves. 

Inspired by the idea of MFEA, many subsequent EMTO algorithms have been developed to handle complex optization tasks. On the basis of MFEA, Gupta \textit{et al.}\cite{7464295} further extended it to the multiobjective optimization domain. Bali \textit{et al.}\cite{7969454} proposed a linearized domain adapation MFEA which transforms the search space of a simple task to the search space of its constitutive complex task to alleviate negative knowledge transfer. Wen and Ting\cite{7969596} proposed to detect parting ways where knowledge sharing become less effective and extended MFEA to many task optimization. Zhou \textit{et al.}\cite{astudyofsim} presented to measure the similarity between tasks from different perspectives in MFEA. 
Ding \textit{et al.}\cite{8231172} proposed a generalized MFEA which is comprised of a dicision variable translation strategy as well as a decision variable shuffling strategy to facilitate knowledge sharing between tasks whose optimums lie in different locations in the search space. 
More recently, Chen \textit{et al.}\cite{8727933} proposed a many-task evolutionary framework, which is able to solve dynamic many-task optimization where tasks arrive at different time. Zhou \textit{et al.}\cite{9027113} incorporated different crossover operators in the evolutionary process, and proposed a MFEA varient with an adaptive knowledge transfer strategy. 
In the above works, a homogenous population is shared by all the tasks involved, and knowledge sharing is achieved through crossover operations between individuals belonging to different tasks in an implicit manner.

In addition to the above implicit EMTO methods, explicit EMTO is a newly emerging paradigm proposed by Feng \textit{et al.}\cite{8401802}. Compared to implicit algorithms where knowledge transfer happens through crossovers between individuals belonging to different tasks, knowledge transfer happens in a more controlled manner in explicit methods. In \cite{8401802}, best solutions found by each task can be transferred explicitly across tasks by leveraging a denoising autoencoder. After this work, Li \textit{et al.}\cite{Li2020MultifactorialOV} proposed an explicit multipopulation evolutionary framework to exploit positive knowledge transfer as well as to prevent negative knowledge transfer. 
Wu \textit{et al.}\cite{8967000} proposed multitasking genetic algorithm which relys on estimating and using the bias among tasks and used the method to fuzzy system optimization. More recently, Tang \textit{et al.}\cite{9195010} introduced an explicit knowledge transfer strategy implemented in low-dimension space via a learnable alignment matrix. Wang \textit{et al.}\cite{9385398} proposed a multi-task evolutionary algorithm based on anomaly dection, where anomaly detection is a model designed to learn relationship between individuals of different tasks. Feng \textit{et al.}\cite{9023952} further investigated the explicit implementation of EMTO for combinatorial optimization problems, and especially capaciated vehicle routing problem is studied. Compared with implicit EMTO methods, explicit ones have the following advantages. First, due to the independent population of each task, task-specific solution encodings can be adopted in each task, which is essential to the process of problem solving\cite{7947122,8338097}. Next, the simplicity of incorporating different evolutionary operators in each task facilitates evolutionary search, as different evolutionary solvers have varied biases\cite{Herrera2003ATF,9027113}. Further, knowledge transfer can happen in a controlled manner, both solution selection schemes such as elite selection and the direction of knowledge transfer are more flexible.

\subsection{Motivation}
\label{subsecMotivation}

\subsubsection{Motivation of multitasking in multi-view point cloud registration}
Existing EA based point cloud registration algorithms were designed for the registration between two point clouds, i.e., pairwise registration. Although the whole object model or scene can be reconstructed by combining pairwise registration results, errors will accumulate along the process, leading to a visible gap between the first and the last scan. The main reason of error accumulation is the violation of global consistency among all the point clouds involved. Therefore, in this paper, a fitness function concerning both local accuracy and global consistency is derived, with the aim of striking a balance between local and global conflicts. However, the optimization of such a fitness function requires the concurrence of solutions from related pairwise registration tasks, which cannot be implemented in the traditional single-task manner. Naturally, in this paper, multi-view point cloud registration is studied in a multitasking setting.

\subsubsection{Motivation of the proposed explicit evolutionary multitasking point cloud registration}
Existing loss functions for pairwise registration are known to be hard to optimize due to the appearance of local optima, and optimization becomes more challenging when it comes to the above mentioned loss function which also concerns relationship between solutions of related tasks. Implicit EMTO methods maintain a single population for solving all the tasks involved, where knowledge sharing happens randomly through crossovers between individuals possessing different skill factors, thus the relation between tasks cannot be fully utilized. Explicit EMTO methods have shown their potential in multi-task optimization, so the proposed method is implemented explicitly. To be specific, aiding tasks that are much simpler to solve are constructed as the knowledge source of intra-task knowledge sharing, aiming to boost the convergence of the original tasks to be solved. Furthermore, the design of inter-task knowledge sharing makes full use of relationship between tasks posed by global consistent constraint, with the aim of reducing probalility of getting stuck to local optima. The two knowledge sharing mechanisms combined together can enhance performance in terms of both convergence speed and solution quality, as shown in Section~\ref{secResultsandDiscussion}.  

\section{Methodology}
\label{secMethodology}
In this section, the proposed explict evolutionary multitasking point cloud registration is presented in detail. To better understand the whole structure, basic concepts and the framework of the proposed method are first introduced. Then the bi-channel knowledge sharing mechanism is presented. Fitness function concerning both local accuracy and global consistency is finally derived.

\subsection{Framework}
As discribed earlier, it is difficult to optimize the function which takes into account both local accuracy and global consistency at the same time. The proposed method aims to optimize such a function effectively and efficiently by introducing a bi-channel knowledge sharing mechanism in multitasking setting. To better understand the co-evolutionary process of the proposed method, basic notations and the framework for multi-view point cloud registration are presented.

\begin{enumerate}
\item 
\textsl{Original task:} 
the optimization of aforementioned function which concerns local accuracy and global consistency at the same time is called the original task. The three original tasks to be solved simultaneously are denoted $J_1$-$o,J_2$-$o,J_3$-$o$. 
Details of the function used in original tasks are derived in Section~\ref{subsecFunction}. 

\item 
\textsl{Aiding task:} 
each aiding task solves registration between the same two point clouds as its corresponding original task, but with different fitness function which is much simpler to optimize. The three aiding tasks are denoted $J_1$-$a,J_2$-$a,J_3$-$a$. The function used in aiding tasks is set according to \cite{SILVA2007114}.

\item
\textsl{Intra-task sharing:}
each aiding task and its corresponding original task solve registration between the same two point clouds with different fitness functions, i.e., they cooperate to solve the same pairwise registration. Therefore, knowledge sharing happens within each task pair is termed intra-task knowledge sharing.

\item
\textsl{Inter-task sharing:}
$J_1,J_2,J_3$ are the three registration tasks which solve the registration between $P^1$ and $P^2$, $P^2$ and $P^3$, $P^3$ and $P^1$, respectively. Accordingly, knowledge sharing happens across these tasks is termed inter-task knowledge sharing.
\end{enumerate}

Keep the above concepts in mind, the framework of the proposed method for point cloud registration is shown in Algorithm~\ref{algStructure}. Each task is assigned with an independent population, and the evolutionary process starts from generating $n$ individuals for each population belonging to an original task or an aiding task, according to the range of values representing the search space. After fitness evalution, $topR*n$ top individuals are recorded for knowledge learning and sharing. Both intra-task and inter-task knowledge sharing happens in a probabilistic manner along each generation. Populations are updated with an elist selection strategy at the end of each generation. Since differential evolution and simulated binary crossover (SBX) are incorporated in the framework, readers can refer to \cite{4632146} and \cite{Deb1995SimulatedBC} for more details.

With the introduction of aiding tasks, there is actually an increase of fitness evaluations in each generation. However, the additional fitness evaluations are worthwhile. As can be seen in Section~\ref{secResultsandDiscussion}, the performances on the original tasks are very poor if intra-task knowledge sharing is not used. In other words, the original tasks cannot be solved without intra-task knowledge sharing. By contrast, with the help from aiding tasks, both solution quality and convergence speed of the original tasks are enhanced by a large margin. Therefore, the introduction of aiding tasks is essential for problem solving.

\begin{algorithm}
	\caption{Framework of MTPCR}
	\label{algStructure}
	\KwIn{\\
		\hspace*{1.5ex}$J_1$-$o,J_2$-$o,J_3$-$o$: original tasks; \\
		\hspace*{1.5ex}$J_1$-$a,J_2$-$a,J_3$-$a$: aiding tasks; \\
		\hspace*{1.5ex}$n$: population size for each task; \\
		\hspace*{1.5ex}$p$-$intra$: intra-task knowledge sharing probability; \\
		\hspace*{1.5ex}$p$-$inter$: inter-task knowledge sharing probability; \\
		\hspace*{1.5ex}$topR$: ratio of top indididuals for knowledge sharing.
	}
	\KwOut{\\
		\hspace*{1.5ex}$T_{12}, T_{23}, T_{31}$: transformations for $J_1$-$o,J_2$-$o,J_3$-$o$.
	}
	Initialize each population $pop_j$-$a$ and $pop_j$-$o$ with $n$ individuals, $j=1,2,3$; \\
	Evaluate each individual in $pop_j$-$a$ and $pop_j$-$o$ on task $J_j$-$a$ and task $J_j$-$o$, respectively;\\
	Record $topR*n$ top individuals for task $T_j$-$a$ and task $T_j$-$o$, respectively; \\
	\While{stopping creterion is not satisfied}{
		\For{$j=1$ to $3$}{
			Generate $n$ mutant vectors w.r.t. $T_j$-$a$ and $T_j$-$o$ by applying mutant operation;\\
			\If{$rand < p$-$intra$}{
				Intra-task knowledge sharing according to Algorithm~\ref{algIntra};
			}
			\If{$rand < p$-$inter$}{
				Inter-task knowledge sharing according to Algorithm~\ref{algInter};
			}
			Apply crossover operation (SBX) between the updated set of $n$ mutant vectors and the current population to generate $n$ trial vectors;\\
			Evaluate the $n$ generated trial vectors on task $T_j$-$a$ and task $T_j$-$o$, respectively;\\
			Update $pop_j$-$a$ and $pop_j$-$o$ via elitist selection between the current population and the newly generated set of trial vectors;\\
			Update the best individual found so far for task $T_j$-$a$ and task $T_j$-$o$; \\
			Update $topR*n$ top individuals maintained by task $T_j$-$a$ and task $T_j$-$o$.
		}
	}
\end{algorithm}

\subsection{Knowledge Sharing}
The process of knowledge sharing is to capture useful traits embeded in each component task, which can be transferred among tasks to enhance performance. As aforementioned, the final objective is to optimize $J_1$-$o,J_2$-$o,J_3$-$o$ simultaneously along the evolutionary search. For this purpose, a bi-channel knowledge sharing mechanism is developed, which is comprised of intra-task and inter-task knowledge sharing. 

\subsubsection{Solution selection} \label{subsubsecSolutionSelection}
Knowledge is represented in the form of solutions, i.e., individuals, so the choice of solutions used for knowledge learning and sharing has an essential impact on the improvement of the performance of the proposed bi-channel knowledge sharing mechanism. Negative knowledge sharing is considered to deteriorate performance, thus should be avoided. In the multitasking point cloud registration setting, high quality solutions carry more useful information than inferior ones. 
Therefore, a proportion of top solutions in terms of objective values are maintained for knowledge learning and sharing. As shown in Algorithm~\ref{algStructure}, the input $topR$ specifies the ratio of top individuals to be maintained in each population.

\subsubsection{Intra-task knowledge sharing}\label{subsubsecIntra} 
The idea of intra-task knowledge sharing is inspired from the evolutionary multasking community that genetic materials from easy tasks may aid the search process of hard ones through knowledge learning and transfer. So here the concept of aiding task comes into being, whose purpose is to help the optimization of the original task. The solution of the newly constructed aiding task should be near to the original task in the search space to facilitate positive knowledge sharing, and in this way useful knowledge, i.e., genetic materials, can be transferred from aiding task to the original task in a straightforward manner. For the registration of two point clouds, the aiding task that considers only local accuracy needs not to provide a precise registration result, but is expected to produce a coarse solution near the optimal transformation. So in this paper the loss function used in \cite{SILVA2007114} for coarse registration is chosen as the function to be optimized in the aiding task. 
As both the aiding task and the original task solve the registration between the same two point clouds, in other words, they cooperate to solve the same pairwise registration, knowledge sharing occurs between aiding task and original task is consequently termed as intra-task knowledge sharing. As shown in line~7 of Algorithm~\ref{algStructure}, intra-task knowledge sharing happens in a probabilistic manner at each generation, controlled by parameter $p$-$intra$. Algorithm~\ref{algIntra} demonstrates the process of intra-task knowledge sharing. Generally genetic materials from aiding task can help speed up the convergence of original task, but in turn the aiding task may also be aided by original task. Therefore, intra-task knowledge sharing between aiding task and original task proceeds in both directions.

\begin{algorithm}
	\caption{Intra-task knowledge sharing}
	\label{algIntra}
	Randomly select $topR*n$ mutant vectors from $J_j$-$o$; \\
	Replace selected mutant vectors from $J_j$-$o$ with $topR*n$ top individuals maintained by $J_j$-$a$; \\
	Randomly select $topR*n$ mutant vectors from $J_j$-$a$; \\
	Replace selected mutant vectors from $J_j$-$a$ with $topR*n$ top individuals maintained by $J_j$-$o$.
\end{algorithm}

\begin{algorithm}
	\caption{Inter-task knowledge sharing}
	\label{algInter}
	Randomly select $topR*n$ mutant vectors from $J_j$-$a$ and $J_j$-$o$, respectively; \\
	Decode best individuals from $J_{(j+1)\%3}$-$o$ and $J_{(j+2)\%3}$-$o$ into $T_{(j+1)\%3,(j+2)\%3}$ and $T_{(j+2)\%3,(j+3)\%3}$, according to (\ref{equR}) and (\ref{equT}); \\
	Calculate $T_{j,(j+1)\%3}^{'}$ according to \eqref{equThreeTrans}; \\
	Encode calculated $T_{j,(j+1)\%3}^{'}$ back into the form of individual according to (\ref{equV}); \\
	Replace selected $topR*n$ mutant vectors from $J_j$-$a$ and $J_j$-$o$ with encoded six dimensional vector.
\end{algorithm}

\subsubsection{Inter-task knowledge sharing}\label{subsubsecInter}
The proposal of intra-task knowledge sharing aims to speed up the convergence of the evolutionary process, but the aiding task itself may also get stuck to local optima sometimes, resulting in less helpful or even negative knowledge sharing. Therefore, the proposal of inter-task knowledge sharing is to reduce the probability of local optima, by leveraging useful traits buried in each task. Recall the global consistent constraint in \eqref{equGlobalConsistent}, which actually connects the three registration tasks closely. If the connections posed by the constraint can be harnessed properly, knowledge can be learned and shared across tasks to further enhance performance. With this idea the inter-task knowledge sharing mechanism is proposed. Using basic operations of matrix multiplication, the relationships between transformation matrices of the three registration tasks can be derived:
\begin{equation}
	\label{equThreeTrans}
	\left\{
	\begin{matrix}T_{12}^{'}=(T_{31}*T_{23})^{-1}
		\\T_{23}^{'}=(T_{12}*T_{31})^{-1}
		\\T_{31}^{'}=(T_{23}*T_{12})^{-1}
	\end{matrix}\right.
\end{equation}
where $T_{i,(i+1)\%3}^{'}$ is the supposed transformation matrix of task $J_i$-$o$ combined by transformation matrices from the other two tasks. $T_{i,(i+1)\%3}$ is the transformation matrix of task $J_i$-$o$, which brings point cloud $P^i$ to the same coordinate of $P^{(i+1)\%3}$. \eqref{equThreeTrans} reflects the relationship between transformation matrices of the three registration tasks. By combining best solutions from the other two tasks, useful knowledge can be learned and transferred. Accordingly, knowledge sharing occurs in this way is termed as inter-task knowledge sharing. As shown in line~10 of Algorithm~\ref{algStructure}, inter-task knowledge sharing also happens in a probabilistic manner at each generation, controlled by parameter $p$-$inter$. Algorithm~\ref{algInter} exhibits the process of inter-task knowledge sharing. 

\subsubsection{Comparison between intra-task and inter-task knowledge sharing}
There are two main differences between intra-task and inter-task knowledge sharing. On the one hand, intra-task knowledge sharing uses a proportion of top individuals as knowledge source, while inter-task knowledge sharing chooses the best individual of each original task for knowledge learning. The different selection mechanism comes from that inter-task knowledge sharing plays its role mainly in the later stage of the evolutionary search. The combination of best individuals from the other two original tasks suffers less uncertainty than the combination of two individuals randomly selected from top individuals maintained by the other two original tasks. On the other hand, knowledge in the form of individuals can be directly transferred between aiding task and the original task in intra-task knowledge sharing, because they share the same individual representation. By contrast, inter-task knowledge sharing needs an additional decoding and encoding process, for that the learning of knowledge is through transformation matrix multiplication as shown in (\ref{equThreeTrans}). 

Individuals in each population are in the form of six dimensional vectors, $[r_x,r_y,r_z,t_x,t_y,t_z]$. Here $r_x$, $r_y$ and $r_z$ denote the rotation angles along $x,y,z$ axis, and $t_x$, $t_y$, $t_z$ denote the translations along each axis. The rotation matrix $R$ can be constructed via the multiplication of rotation matrices along each axis:
\begin{multline}
	\label{equR}
	R = 
	\begin{bmatrix}
		cos(r_z) & -sin(r_z) & 0\\
		sin(r_z) & cos(r_z) & 0\\
		0 & 0 & 1
	\end{bmatrix} * 
	\begin{bmatrix}
		cos(r_y) & 0 & sin(r_y)\\
		0 & 1 & 0\\
		-sin(r_y) & 0 & cos(r_y)
	\end{bmatrix} \\
	*
	\begin{bmatrix}
		1 \!& 0 & 0\\
		0 \!& cos(r_x) & -sin(r_x)\\
		0 \!& sin(r_x) & cos(r_x)
	\end{bmatrix} 
\end{multline}
where the rotation is performed along the order of $x,y,z$ axis with angles $r_x$, $r_y$ and $r_z$, respectively. After obtaining the $3\times3$ rotation matrix, the $4\times4$ transformation matrix can also be obtained by the combination of rotation matrix $R$ and translation vector $[t_x,t_y,t_z]^T$:
\begin{equation}
	\label{equT}
	T = \begin{bmatrix}R & \begin{matrix}t_x\\t_y\\t_z\end{matrix}\\0^T & 1\end{bmatrix}
\end{equation}

After the learning of knowledge in the form of matrix multiplication, inter-task knowledge sharing requires the learned knowledge to be encoded back to the form of six dimensional vector to facilitate knowledge transfer according to Algorithm~\ref{algInter}. Given the learned transformation matrix $T^{'}$, the translation vector $[t_x^{'}, t_y^{'}, t_z^{'}]^T$ can be directly accessed owing to the relationship between $T$ and $t_x,t_y,t_z$ as shown in (\ref{equT}). The three rotation angles can be calculated according to (\ref{equV}), where $T^{'}[i,j]$ is the element located at $i$th row and $j$th column of transformation matrix $T^{'}$. Note that the row and column indices $i$ and $j$ start from 1.
\begin{align}
	\label{equV}
	\begin{cases}
		r_x^{'}=arctan(T^{'}[3,2]/T^{'}[3,3])\\
		r_y^{'}=arctan(-T^{'}[3,1]/\sqrt{(T^{'}[3,2])^2 + (T^{'}[3,3])^2})\\
		r_z^{'}=arctan(T^{'}[2,1]/T^{'}[1,1])
	\end{cases}
\end{align}

\subsection{Fitness function}\label{subsecFunction}
\begin{figure}[ht]
	\centering
	\subfloat[]{
		\label{figBest}
		\includegraphics[height=0.48\columnwidth]{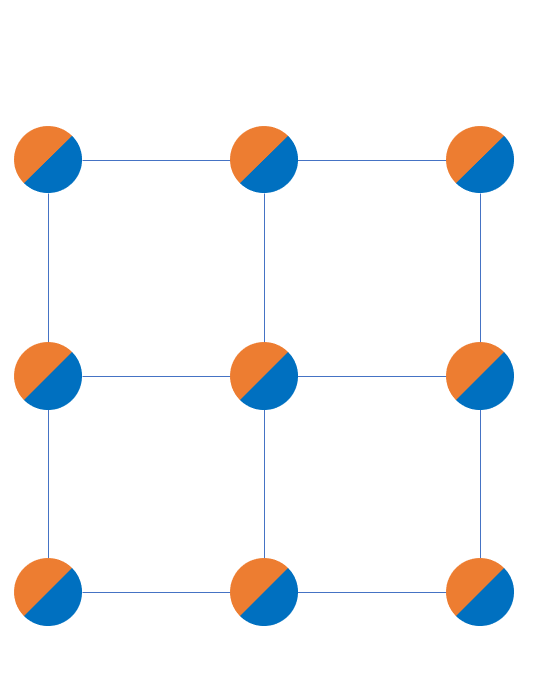}
	}
	\subfloat[]{
		\label{figWorst}
		\includegraphics[height=0.48\columnwidth]{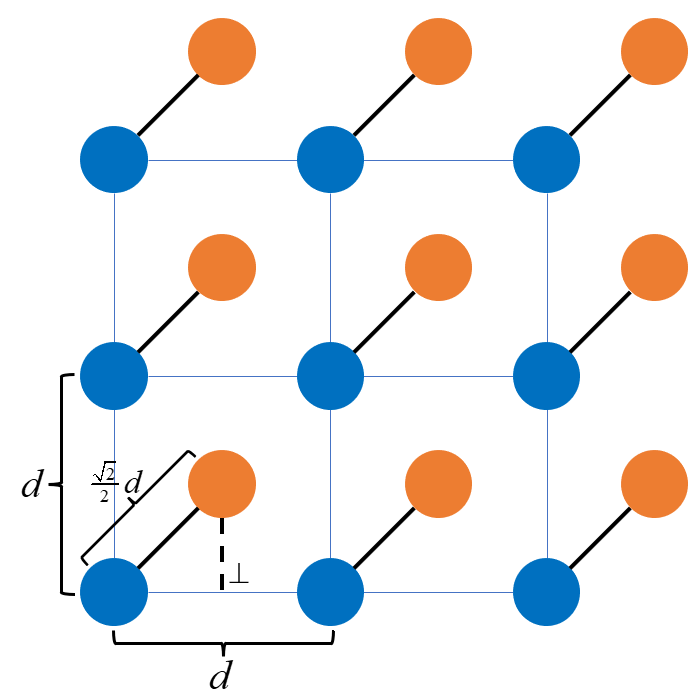}
	}
	\caption{Demo registration of uniformly distributed point clouds in 2D case, where distance between each nearest neighbor pair is denoted $d$. (a)~Perfect registration. (b)~The worst registration without changing correspondence relationships (points connected by black lines) between paired points.}
	\label{figDemoRegistration}
\end{figure}
In this section, a fitness function considers only local accuracy with an adaptive threshold first kicks in as the basic part, and it is further combined with global consistent constraint to obtain the final form used in original tasks. 
To alleviate negative impact caused by noise and outliers, point cloud registration is treated as the problem of maximizing the consensus size in this paper, i.e., the minimization of maximum consensus loss is considered, as it is supposed to diminish the influence of outliers and tolerate noise in a given level \cite{9528069}. With the Euclidean distance metric in the form of \eqref{equEculideanDistance}, $\varphi(\cdot)$ in \eqref{equBasicForm} can be defined as:
\begin{equation}
	\varphi(x)=1-\mathbb{I}(x<\varepsilon )
\end{equation}
where $\mathbb{I}(\cdot)$ is the indicator function that returns 1 if its input is true and 0 otherwise. The parameter $\varepsilon$ is a threshold defining the maximum distance allowed for paired points from different point clouds to be considered inliers. Note that the relationship between paired points is not bi-directional, for example, if $q_j$ in $Q$ is the nearest point of $p_i$ in $P$, $p_i$ is not necessarily in turn the nearest point of $q_j$. Take into account mappings from both directions, as well as the number of points in both point clouds, the optimization objective can be formulated as:
\begin{equation}
	\label{equFirstPartLocalLoss}
	\underset{T\in {SE}(3)}{min}\frac{1}{N_p+N_q}(\sum_{i=1}^{N_p}\varphi (\varrho_i(T)) + \sum_{j=1}^{N_q}\varphi (\varrho_j(T^{-1}) ) )
\end{equation}
where $N_p$ and $N_q$ denote the number of points of point cloud $P$ and $Q$, respectively. $T^{-1}$ is the reverse of transformation matrix $T$, representing the transformation that brings $Q$ to the same coordinate of $P$. The parameter $\varepsilon$ needs tuning for different data to obtain satisfying registration results, due to varied scale of different point clouds. This paper proposes to calculate the value of the threshold with regard to the data used for registration, and the calculation can be carried out by analyzing demo registration results shown in Fig.~\ref{figDemoRegistration}. There are two uniformly distributed point clouds in 2D case to be registered, colored by orange and blue. Fig.~\ref{figBest} and Fig.~\ref{figWorst} show the best and worst registration results, respectively. For the best registration result, distances between paired points are all equal to 0, meaning that the two point clouds are perfectly registered. However, such perfect matching is not likely to happen when register partially overlapping point clouds contaminated by noise and outliers. Therefore, it is reasonable to investigate the worst registration result where the registration result can still be considered as a successful try, as shown in Fig.~\ref{figWorst}. Without changing correct correspondence relationships between paired points, black lines in Fig.~\ref{figWorst} show the maximum shift allowed between matched point pairs, and the parameter $\varepsilon$ is determined using the maximum shift. Mathematically, given the distance $d$ between nearest neighbors in the original point clouds, the maximum allowed shift can be calculated by means of the Pythagorean theorem:
\begin{equation}
	\label{equThreshold}
	\varepsilon = \frac{1}{2} \sqrt{d^2 + d^2} = \frac{\sqrt{2}}{2} d.
\end{equation}

Simple though it is, the way of calculating threshold $\varepsilon$ is shown to work well in 3D, where $d$ is averaged over all the distances of each point to its nearest neighbor in the same point cloud. 

By far, \eqref{equFirstPartLocalLoss} with the proposed adaptive threshold can be well used in pairwise registration problems, but not enough to handle accumulated errors when it comes to multi-view registration. As discussed in Section~\ref{secIntroduction}, the main reason of error accumulation is caused by the violiation of global consistency. Therefore, it is a natural idea to take into account both local accuracy (\ref{equFirstPartLocalLoss}) and global consistency (\ref{equGlobalConsistent}) to further improve the fitness function. To achieve a better balance between local and global conflicts, controlling parameters are added in the combination. Then the final fitness function is reformulated in the form:
\begin{multline}
	\label{equFinalForm}
	\underset{T\in {SE}(3)}{min}\frac{1}{N_p+N_q}(\sum_{i=1}^{N_p}\varphi (\varrho_i(T)) + \sum_{j=1}^{N_q}\varphi (\varrho_j(T^{-1}) ) ) \\
	+ \alpha *\gamma (\left \| T_{31}*T_{23}*T_{12}-I \right \|_{F} )
\end{multline}
where $\left \| \cdot  \right \| _F$ is the Frobenius norm. $\gamma(\cdot)$ is a function that maps its input value to the range $[0, 1]$, which ensures the same scale of local loss and the Frobenius norm of global constraint. The parameter $\alpha$ is used to tune the proportion of the two parts.

\section{Experiment Design}\label{secEcperimentDesign}
This section presents the design of experiments, including comparison methods, registration problems for validation and parameter settings.

\subsection{Comparison Methods}
\label{subsecComparisonMethods}
To illustrate the accuracy and robustness of the proposed fitness function, four registration methods are compared in terms of registration accuracy. To illustrate the efficacy of the proposed bi-channel knowledge sharing mechanism, two multi-task optimization methods as well as the proposed method in different parameter settings are compared in terms of solution quality and convergence trends.

The four registration methods compared in terms of registration accuracy are: the trimmed iterative closest point algorithm\cite{CHETVERIKOV2005299},  the K-means clustering based approach\cite{ZHU2019205}, the expectation-maximization perspective for multi-view registration\cite{9201412}, and the coherent point drift with local surface geometry
for point cloud registration\cite{lsgcpd}, which are abbreviated TrICP, KMEANS, EMPMR, and LSG-CPD, respectively. It has to be mentioned that the four approaches are all locally convergent, thus initial transformations near the ground truth are provided. By contrast, no initial transformations are provided in MTPCR. 
For accurate comparison, the rotation errors and translation errors compared with the ground truth are reported, which are defined as $Err\_R=\frac{1}{N} {\textstyle \sum_{i=1}^{N}}\left \| R_{est,i}-R_{gt,i} \right \|_F$ , $Err\_T=\frac{1}{N} {\textstyle \sum_{i=1}^{N}}\left \| t_{est,i}-t_{gt,i} \right \|$ , respectively. Here $\left \{ R_{gt,i},t_{gt,i} \right \} $ denotes the ground truth of the $i$th transformation, $\left \{ R_{est,i},t_{est,i} \right \} $ indicates the one estimated by the above approaches. $N$ equals three since the registration of three point clouds are investigated in the experiments.

The two multi-task optimization methods compared in terms of solution quality and convergence trends are the multifactorial evolutionary algorithm\cite{7161358} and the generalized multifactorial evolutionary algorithm\cite{8231172}, which are abbreviated MFEA and GMFEA, respectively. The proposed MTPCR utilizes a bi-channel knowledge sharing mechanism, to illustrate the effectiveness of intra-task and inter-task knowledge sharing, counterparts of MTPCR with different settings are also compared. MTPCR-Intra denotes MTPCR with only intra-task knowledge sharing, MTPCR-Inter denotes MTPCR with only inter-task knowledge sharing, and MTPCR-NKS denote the version with no knowledge sharing.

\subsection{Validation Problems}
Point clouds used in the experiments are from Stanford 3D Scanning Repository\footnote{\url{http://graphics.stanford.edu/data/3Dscanrep/}} , Aim@Shape project\footnote{\url{http://visionair.ge.imati.cnr.it/}} and Augmented ICL-NUIM Dataset\footnote{\url{http://redwood-data.org/indoor/dataset.html}}. Table~\ref{tabDataProperty} tabulates the properties of point clouds used in experiments. Each model dataset contains point clouds acquired from different viewpoints of one object model, and each scene dataset contains partial views of indoor scenes. In addition to different types of model or scene, the scale also differs greatly. The `Overlap Ratio' column provides overlap ratios between the three point clouds used in each problem, with the lowest ratio among the three in bold font. For simplicity of notation, each problem is assigned a tag, with a, b, c or d added to the data name, and the lowest overlap ratio goes down from a to d. Ground truth transformations to recover the whole object model or scene are available, providing convenience for fair comparison. 

\begin{table}\caption{Property Summary of Data Utilized in Experiments}\label{tabDataProperty}
	\centering
	\renewcommand\arraystretch{1.2}
	\begin{tabular}{l|l|l|l}
		\hline
		Type & Data & Overlap Ratio & Tag \\
		\hline
		\multirow{15}{*}{Model}
		& \multirow{3}{*}{Bunny} 
		& 85, \textbf{48}, 70 & Bunny-a \\
		& & \textbf{33}, 51, 49 & Bunny-b \\
		& & \textbf{26}, 72, 41 & Bunny-c \\
		\cline{2-4}
		& \multirow{3}{*}{Dragon} 
		& 84, 71, \textbf{56} & Dragon-a \\
		& & 67, 64, \textbf{32} & Dragon-b \\
		& & 66, 61, \textbf{29} & Dragon-c \\
		\cline{2-4}
		& \multirow{3}{*}{Armadillo}
		& 75, 75, \textbf{48} & Armadillo-a \\
		& & 72, 55, \textbf{30} & Armadillo-b \\
		& & 72, 41, \textbf{22} & Armadillo-c \\
		\cline{2-4}
		& \multirow{3}{*}{Buddha} 
		& 81, 70, \textbf{53} & Buddha-a \\
		& & \textbf{33}, 62, 59 & Buddha-b \\
		& &53, \textbf{19}, 44 & Buddha-c \\
		\cline{2-4}
		& \multirow{3}{*}{Buste}
		& 75, \textbf{45}, 53 & Buste-a \\
		& & \textbf{35}, 72, 77 & Buste-b \\
		& & 68, \textbf{29}, 71 & Buste-c \\
		\hline
		\multirow{8}{*}{Scene}
		& \multirow{4}{*}{Room}
		& 71, \textbf{58}, 74 & Room-a \\
		& & \textbf{46}, 60, 49 & Room-b \\
		& & 60, \textbf{42}, 42 & Room-c \\
		& & 75, 42, \textbf{34} & Room-d \\
		\cline{2-4}
		& \multirow{4}{*}{Office}
		& 71, 83, \textbf{68} & Office-a \\
		& & 62, 70, \textbf{55} & Office-b \\
		& & 71, 77, \textbf{54} & Office-c \\
		& & 66, 49, \textbf{29} & Office-d \\
		\hline
	\end{tabular}
\end{table}

\subsection{Parameter Setting}

Parameters for the proposed MTPCR: the population size $n$ is set 100 for each of the six populations attached to optimize each task. The maximal number of generation is 60. The probability of intra-task and inter-task knowledge sharing $p$-$intra$ and $p$-$inter$ are both set to be 0.5. The ratio $topR$ of top individuals maintained for knowledge learning is 0.1. The scale factor used in differential evolution is 0.5. The parameter in simulated binary corssover goes from 2 to 10 linearly with the evolving process. Since there is no prior information about the point clouds to be registered, the range of angles for searching is set to $[-\pi, \pi]$ along each axis, and the range of shifts is $[-shift, shift]$ with the point clouds decentralized first, where $shift$ is the maximum length of bounding box of the point clouds involved. The search space can cover all the possibility of solutions to be found.

Parameters for counterparts of MTPCR: The only difference between MTPCR and its different versions lie in the parameter $p$-$intra$ and $p$-$inter$. They are set 0.5 and 0 in MTPCR-Intra, 0 and 0.5 in MTPCR-Inter, both 0 in MTPCR-NKS.

Parameters for MFEA and GMFEA: there is only one population for optimizing all the tasks, so the population size of the two methods are both 600, ensuring the same number of new individuals evaluated in each generation. The maximal number of generation is 60. The probability allowing individuals possessing different skill factors to go through crossover is 0.5. The parameter in simulated binary crossover also goes from 2 to 10 linearly. There are two more parameters in GMFEA controlling the start generation to calculate directions and the interval to update directions, respectively. They are set 10 and 5.

Parameters for fitness function: the form of fitness function used in original tasks is given in \eqref{equFinalForm}. Threshold $\varepsilon$ is set automatically according to point clouds in each registration problem using $\varepsilon=\frac{\sqrt{2}}{2}d$, where $d$ is averaged over the three mean distances of each point cloud, with each mean distance averaged over the distances of each point to its nearest neighbor. The $tanh$ function is used as $\gamma(\cdot)$, mapping the value of its input to the range $[-1,~1]$. Due to the non-negativity of Frobenius norm, the value is actually mapped to the range $[0,~1]$. $\alpha$ is set 0.05 to keep a balance between local accuracy and global consistency. Fitness function used in aiding tasks is set according to \cite{SILVA2007114}. Experiments are conducted with the help of MindSpore Lite tool\footnote{MindSpore. \url{https://www.mindspore.cn/}}.

\section{Results and Discussion}
\label{secResultsandDiscussion}
This section presents, analyzes and discusses the performance of the proposed method by comparing it with registration methods as well as multi-task optimization methods over the criteria of registration accuracy and robustness, solution quality and search efficiency, respectively.

\begin{figure}[!htbp]
	\centering
	\subfloat{
		\centering
		\includegraphics[width=0.25\columnwidth]{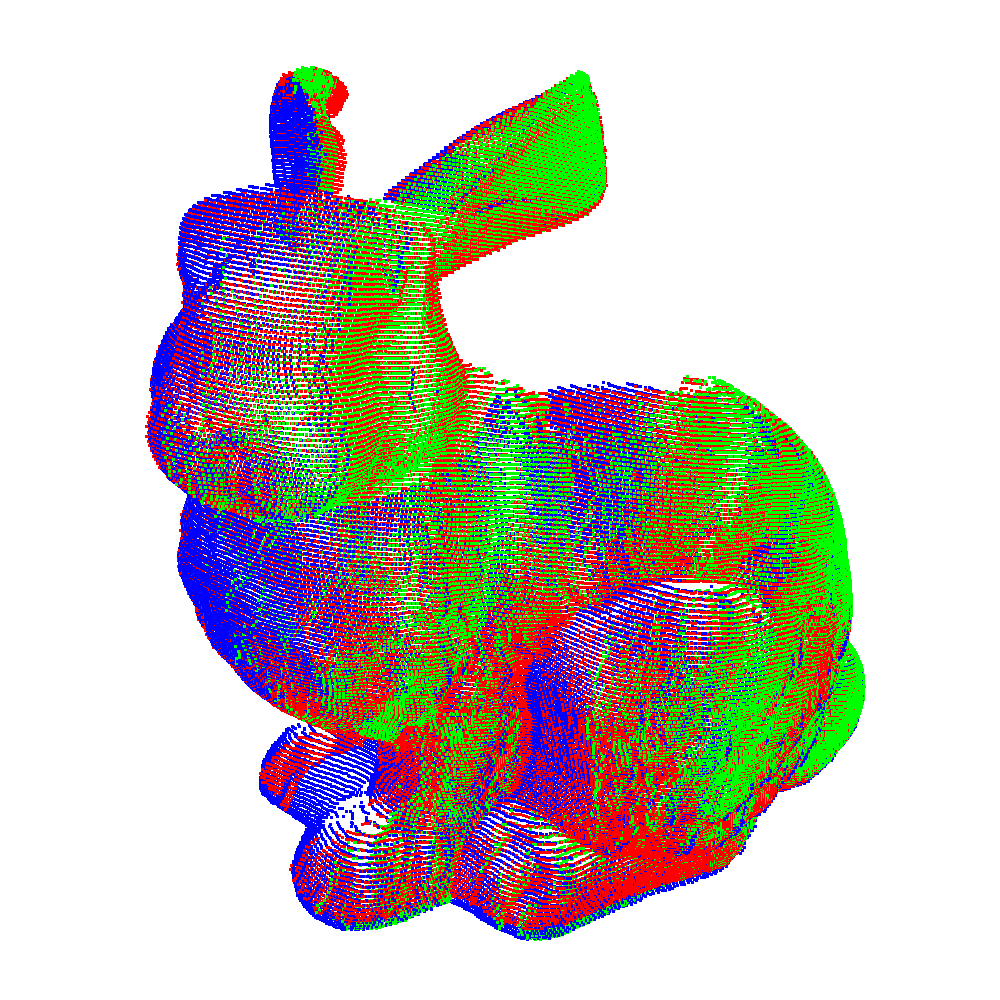}
		\includegraphics[width=0.25\columnwidth]{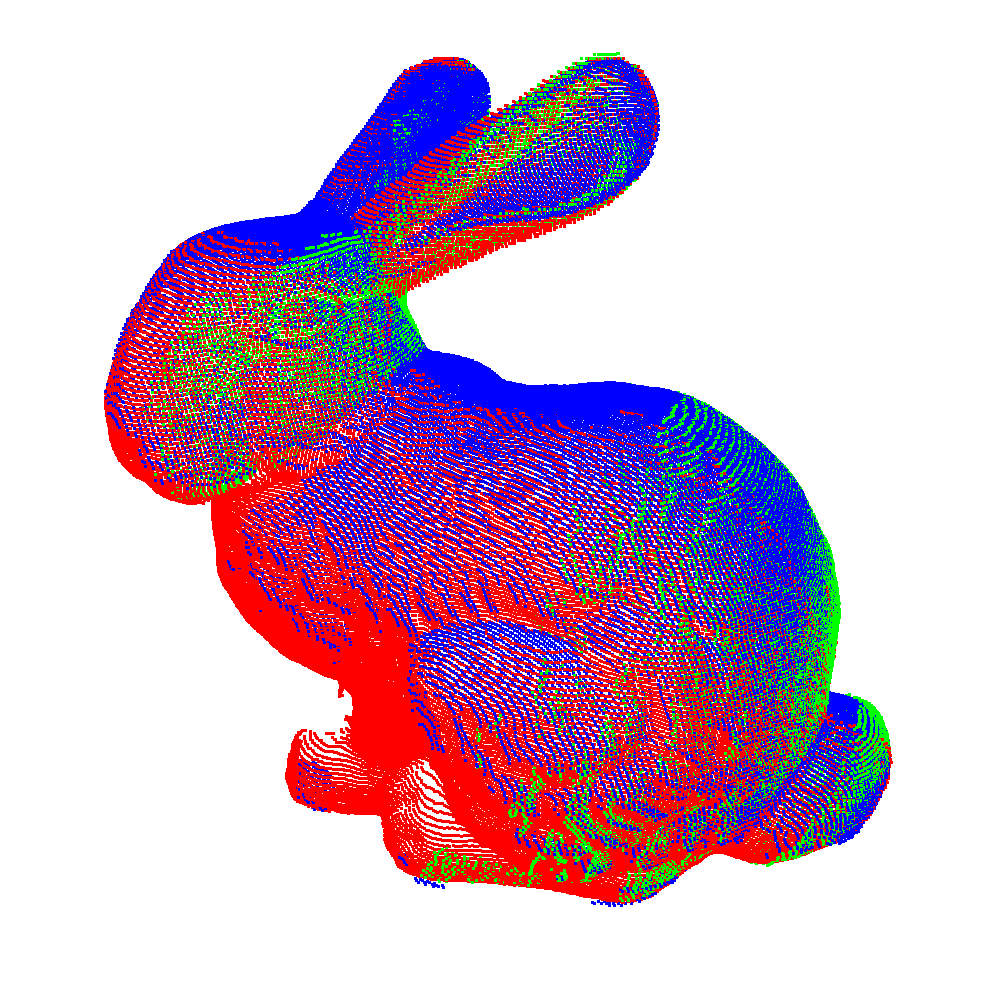}
		\includegraphics[width=0.25\columnwidth]{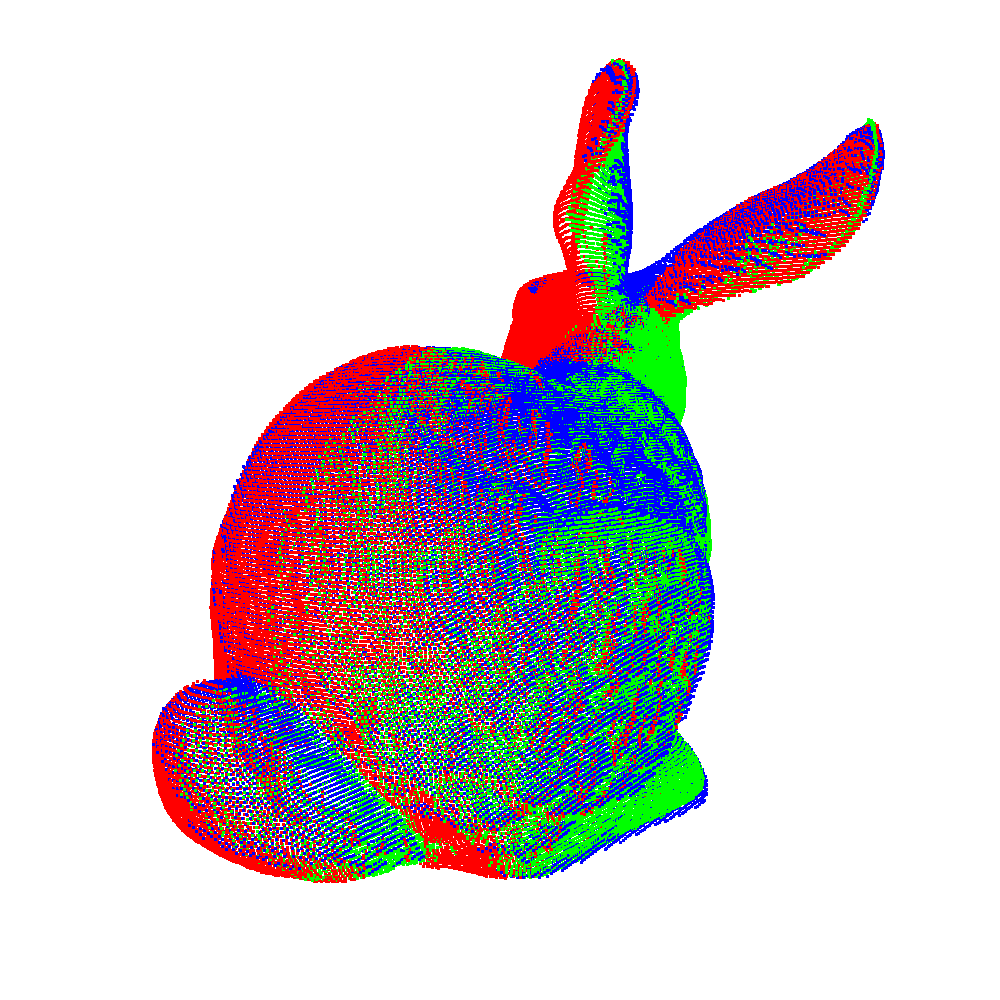}
	} \\
	\vspace{-3ex}
	\subfloat{
		\centering
		\includegraphics[width=0.25\columnwidth]{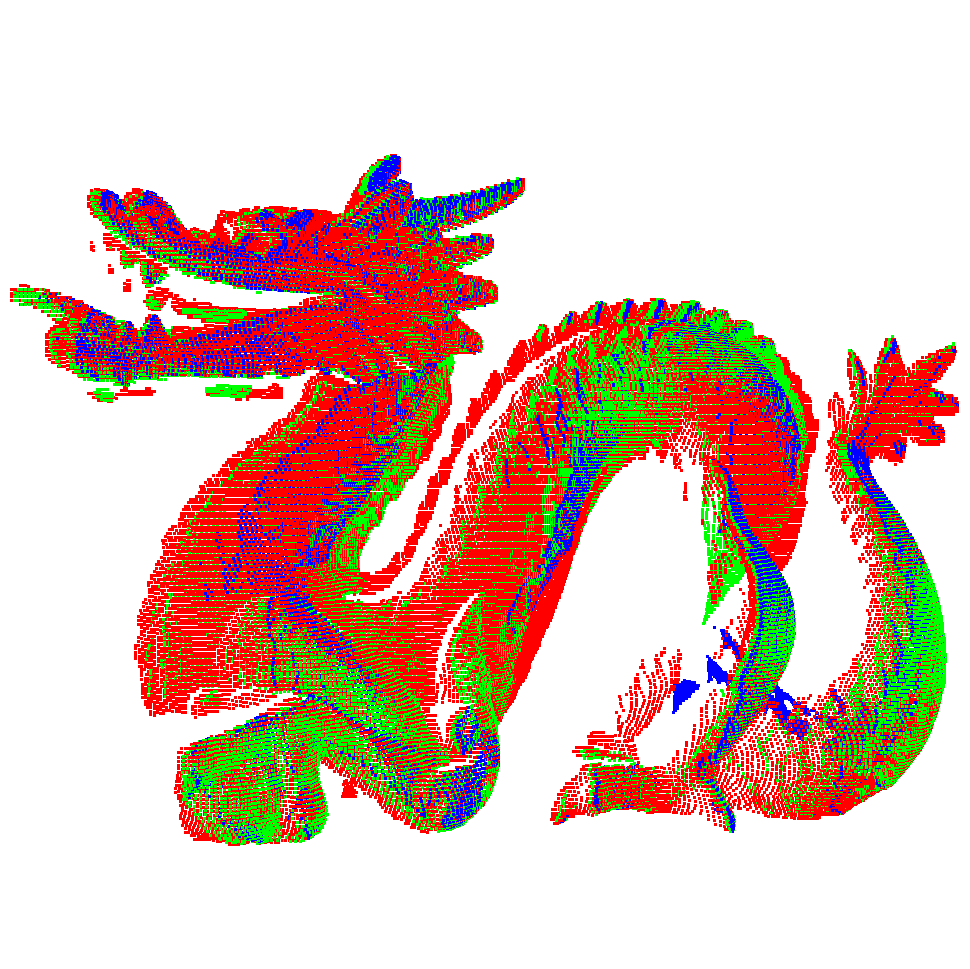}
		\includegraphics[width=0.25\columnwidth]{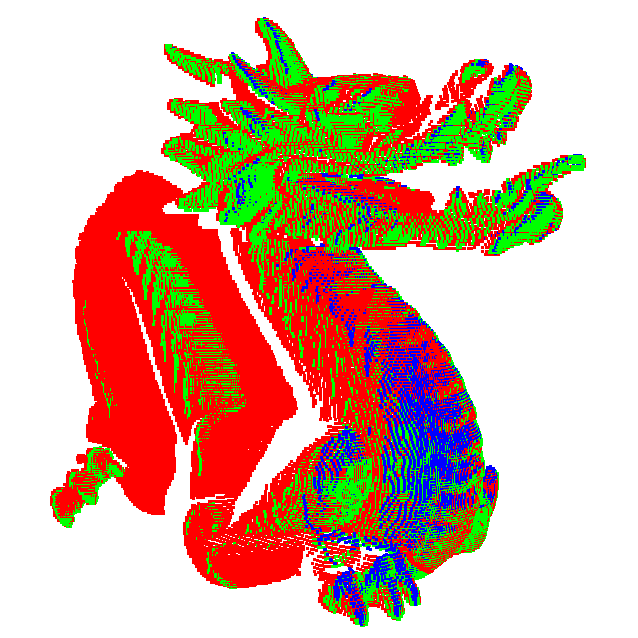}
		\includegraphics[width=0.25\columnwidth]{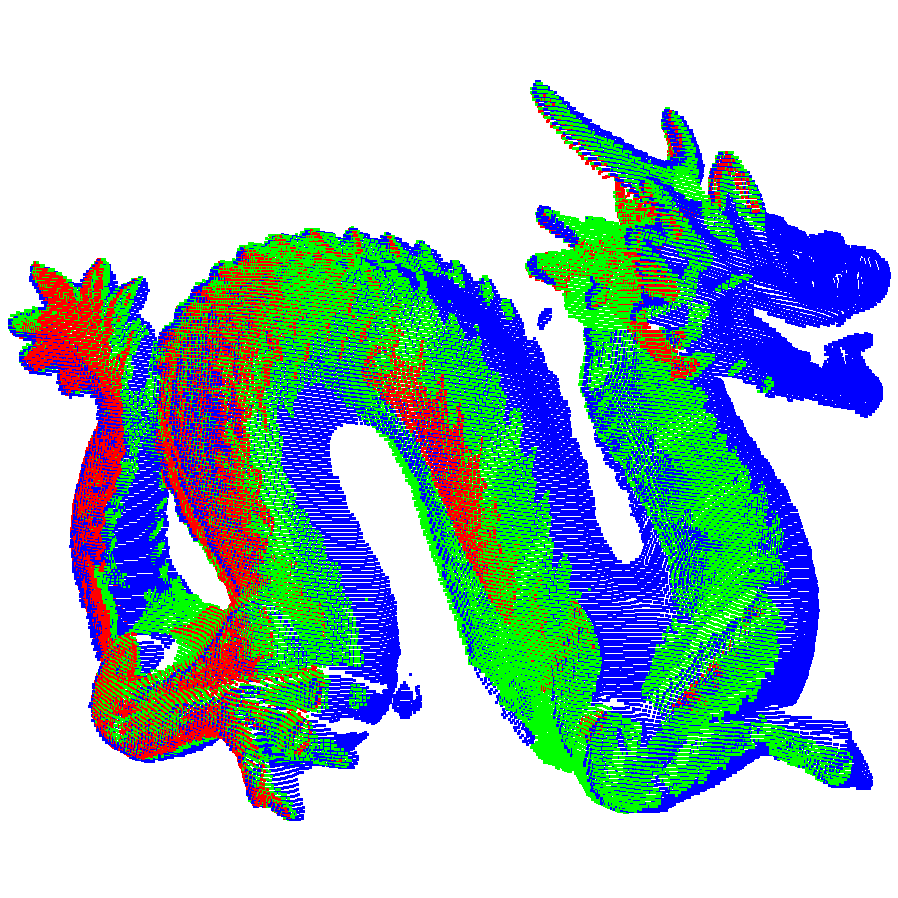}
	} \\
	\vspace{-2ex}
	\subfloat{
		\centering
		\includegraphics[width=0.25\columnwidth]{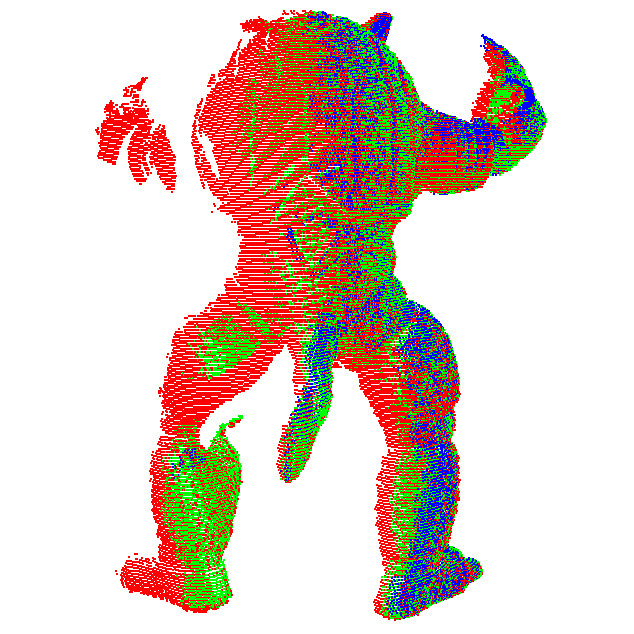}
		\includegraphics[width=0.25\columnwidth]{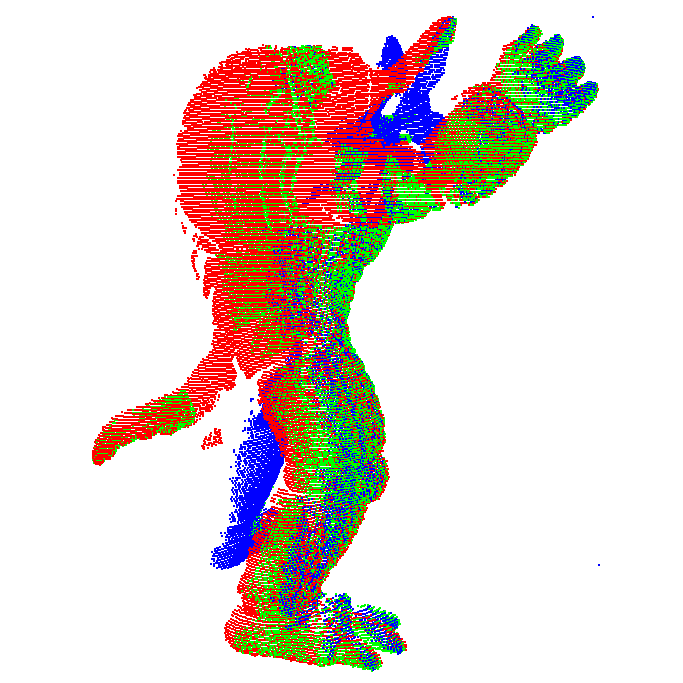}
		\includegraphics[width=0.25\columnwidth]{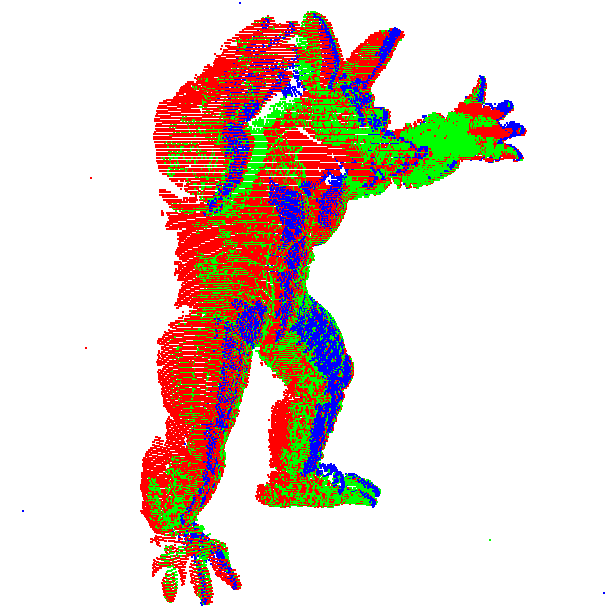}
	} \\
	\vspace{-1ex}
	\subfloat{
		\centering
		\includegraphics[width=0.25\columnwidth]{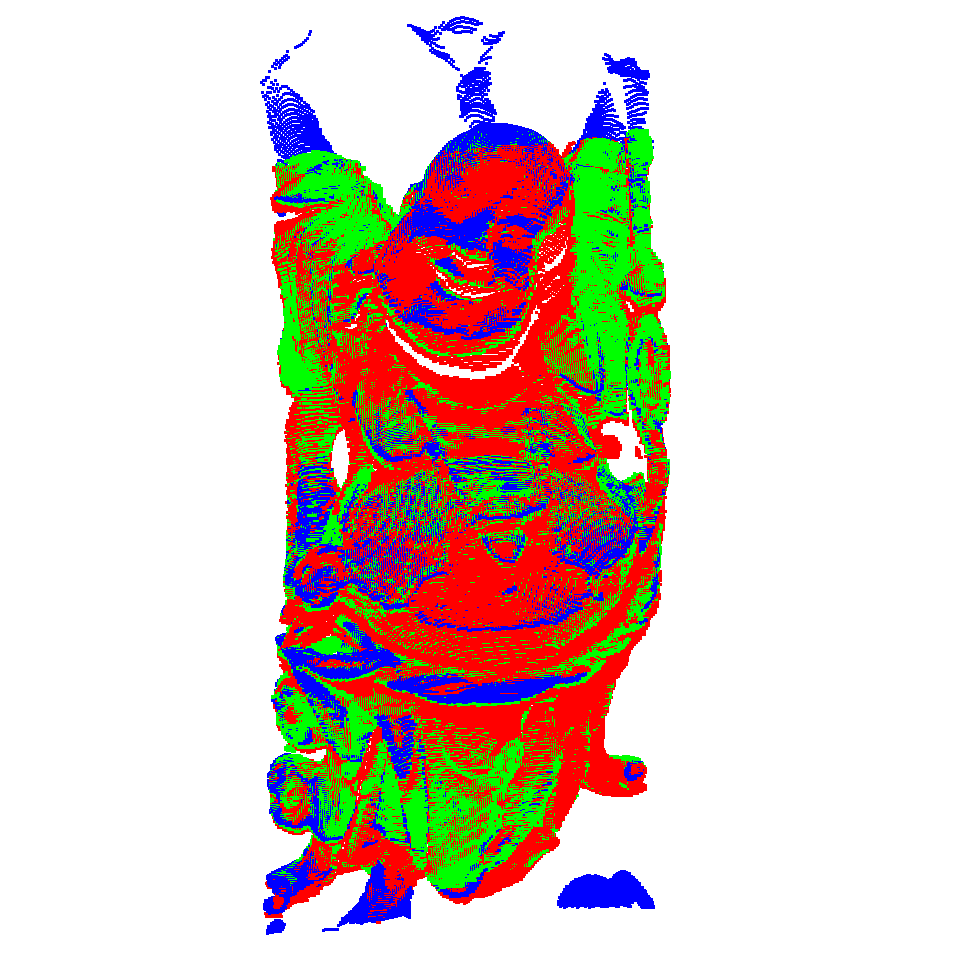}
		\includegraphics[width=0.25\columnwidth]{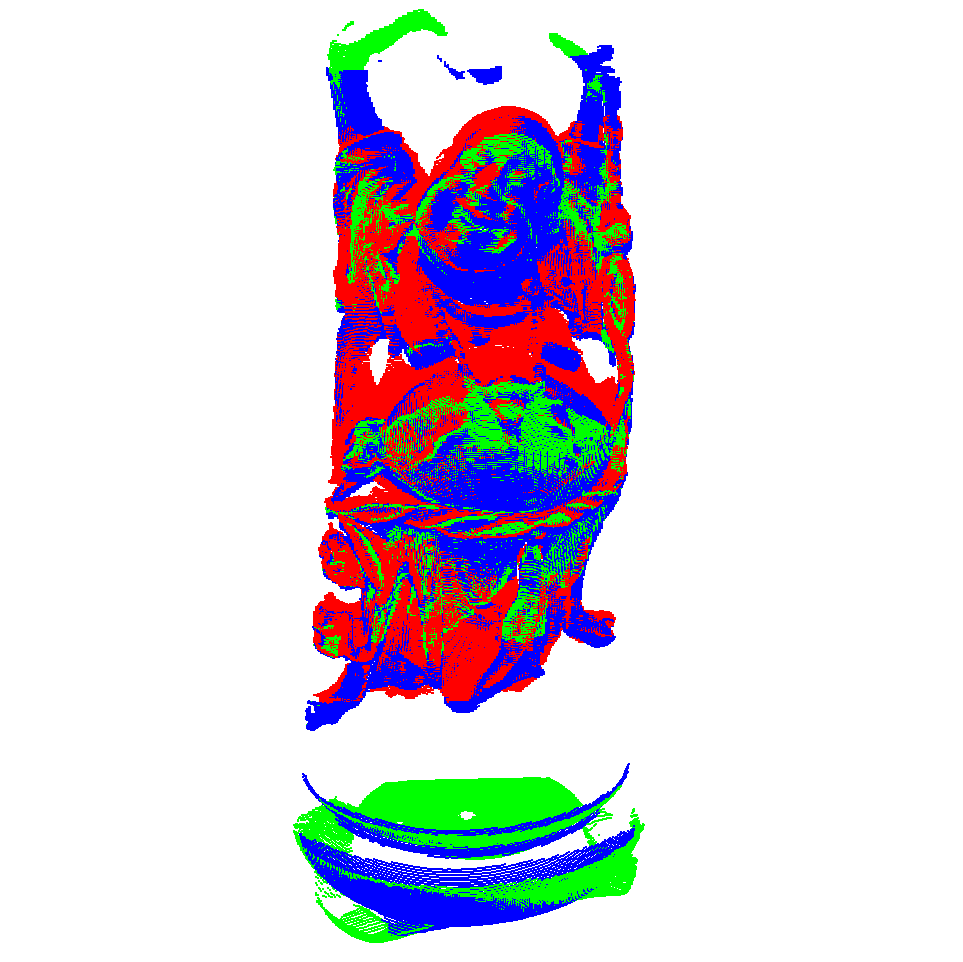}
		\includegraphics[width=0.25\columnwidth]{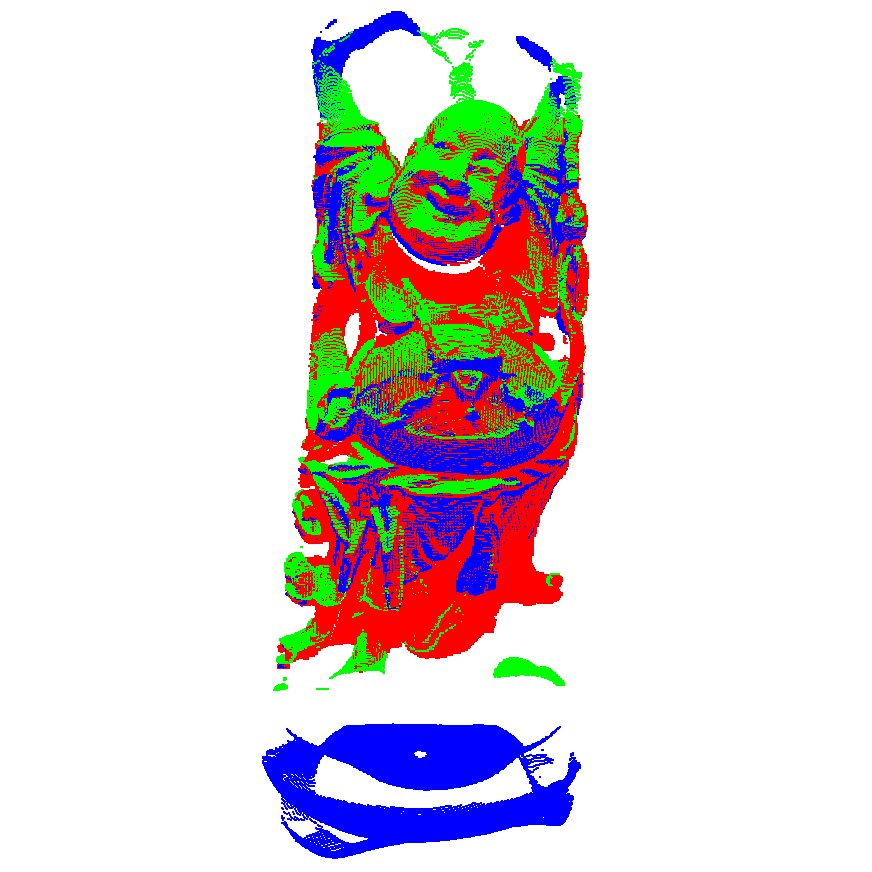}
	} \\
	\vspace{-1ex}
	\subfloat{
		\centering
		\includegraphics[width=0.25\columnwidth]{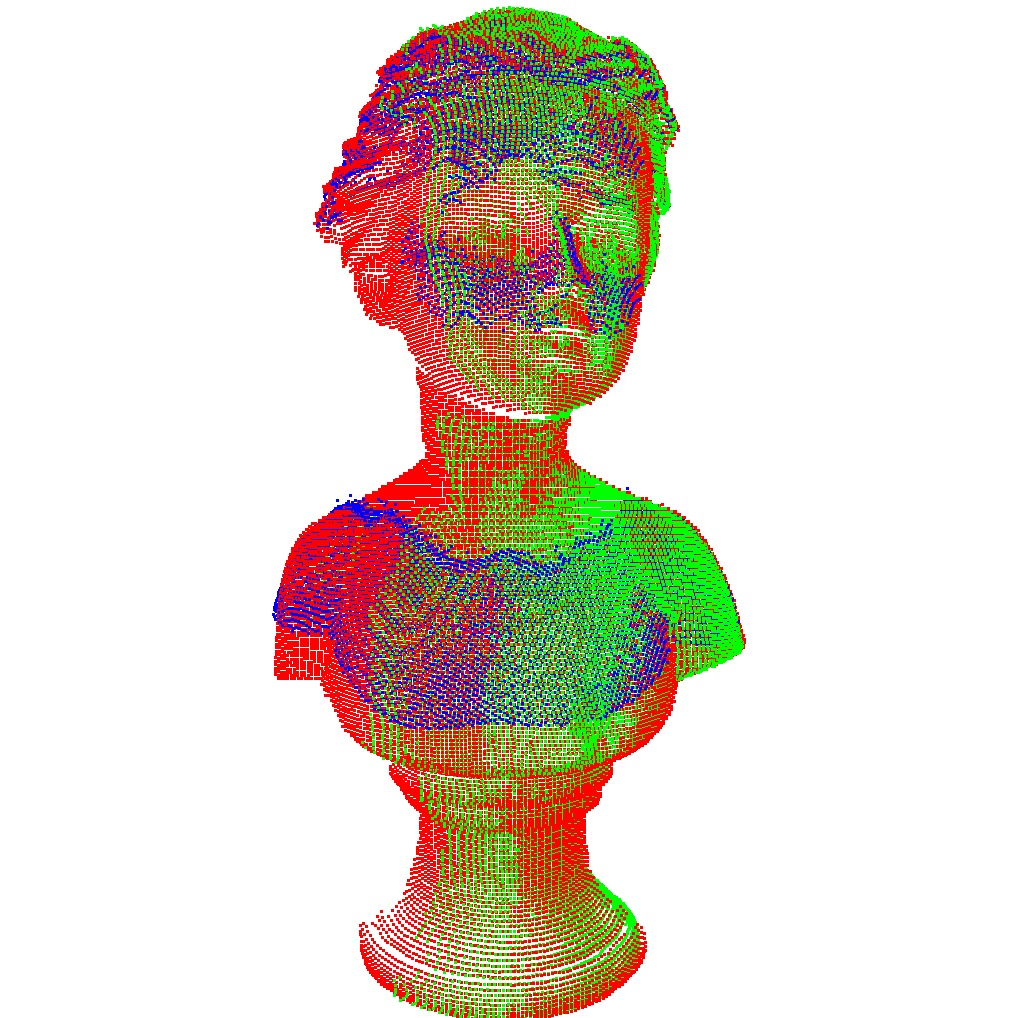}
		\includegraphics[width=0.25\columnwidth]{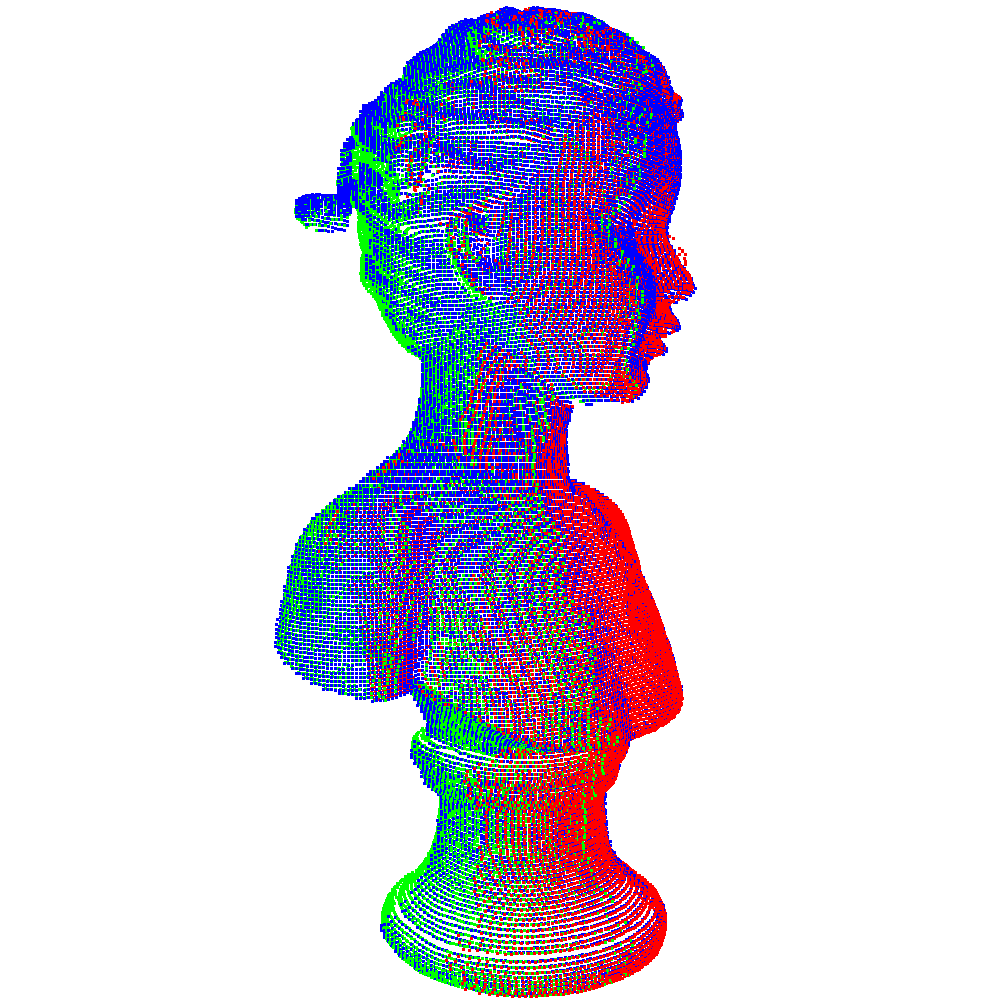}
		\includegraphics[width=0.25\columnwidth]{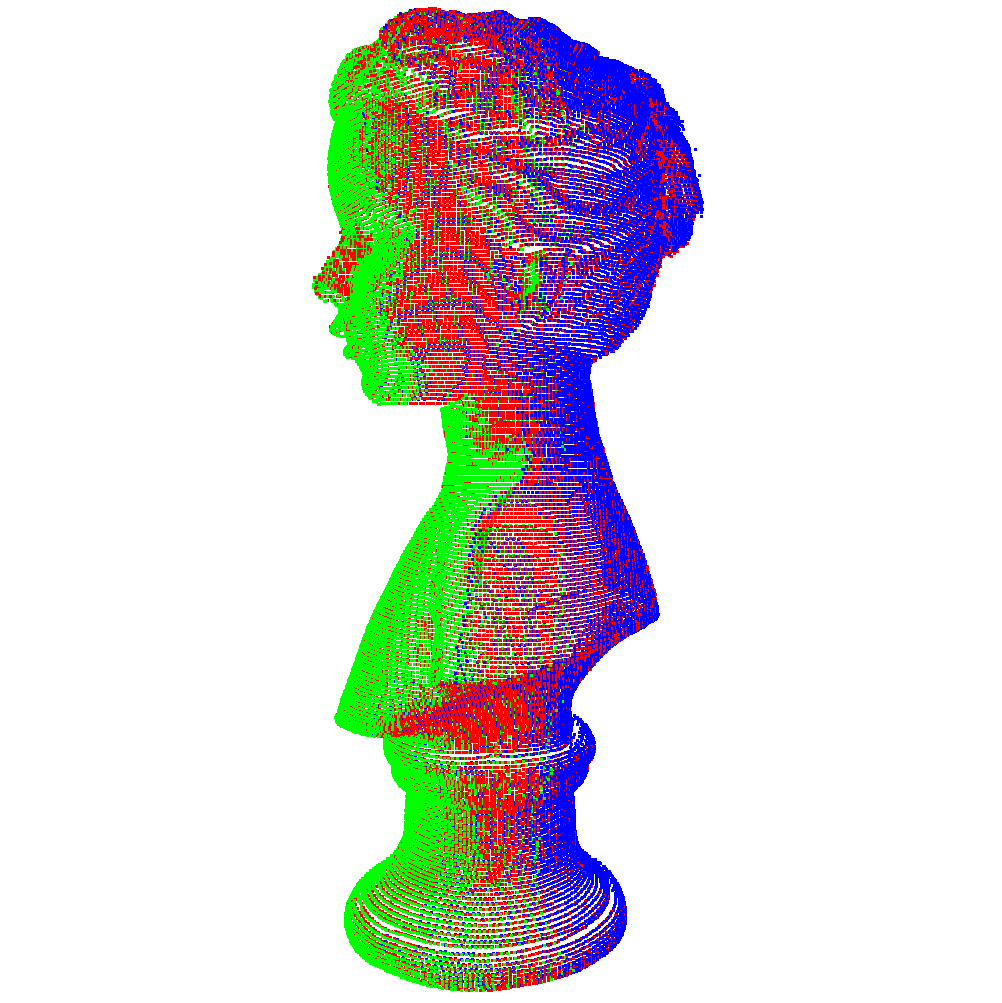}
	}
	\caption{Registration results of MTPCR on model objects. From top to bottom: Bunny, Dragon, Armadillo, Buddha, Buste. From left to right: registration problems with tag a, b and c.}
	\label{figModelObjects}
\end{figure}

\subsection{Registration Accuracy}
\begin{figure*}[htbp]
	\centering
	\subfloat{
		\centering
		\includegraphics[width=0.42\columnwidth]{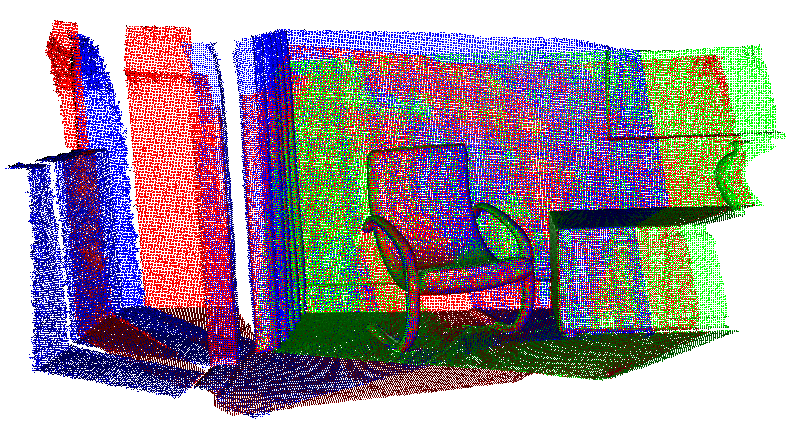}
		\includegraphics[width=0.42\columnwidth]{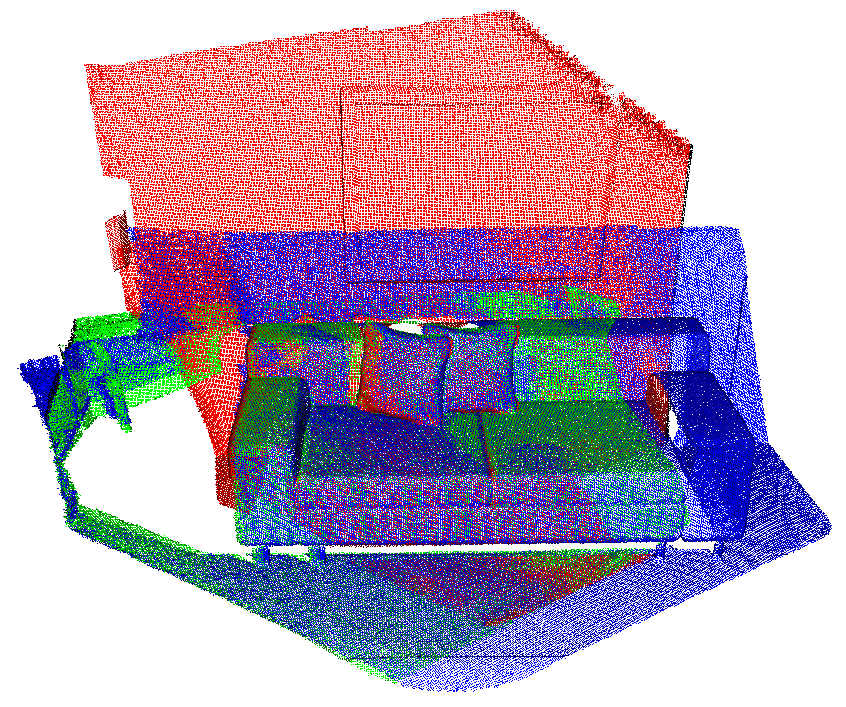}
		\includegraphics[width=0.42\columnwidth]{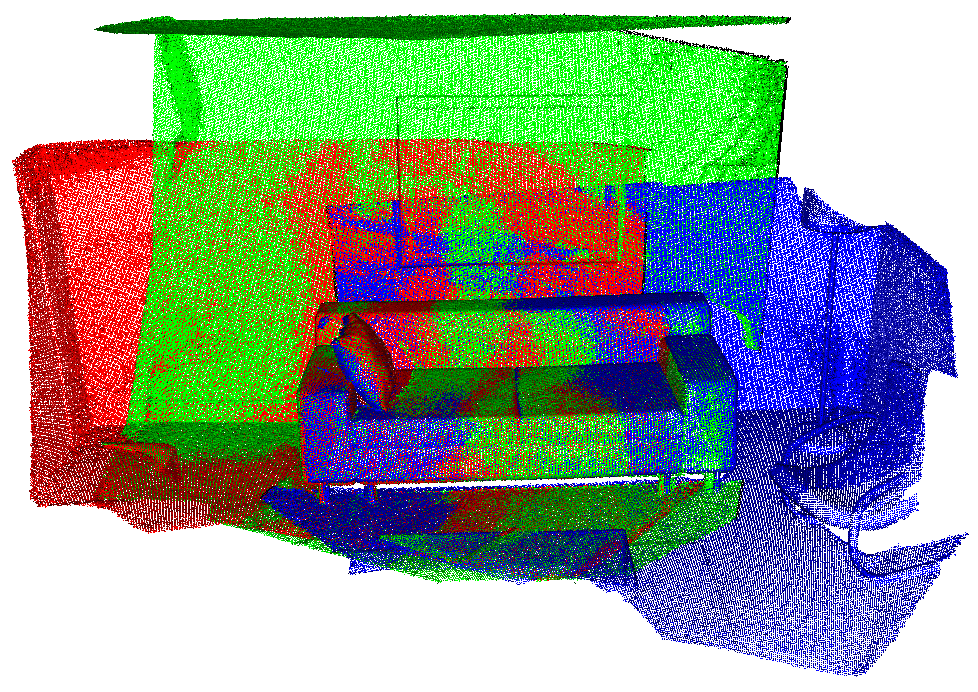}
		\includegraphics[width=0.42\columnwidth]{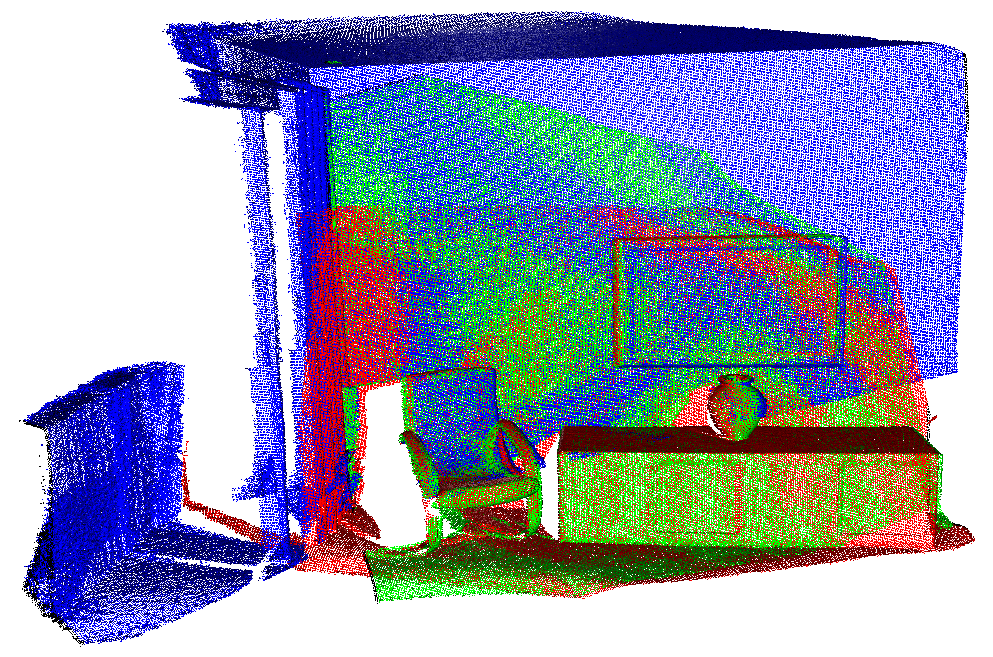}
	} \\
	\vspace{-3ex}
	\subfloat{
		\centering
		\includegraphics[width=0.42\columnwidth]{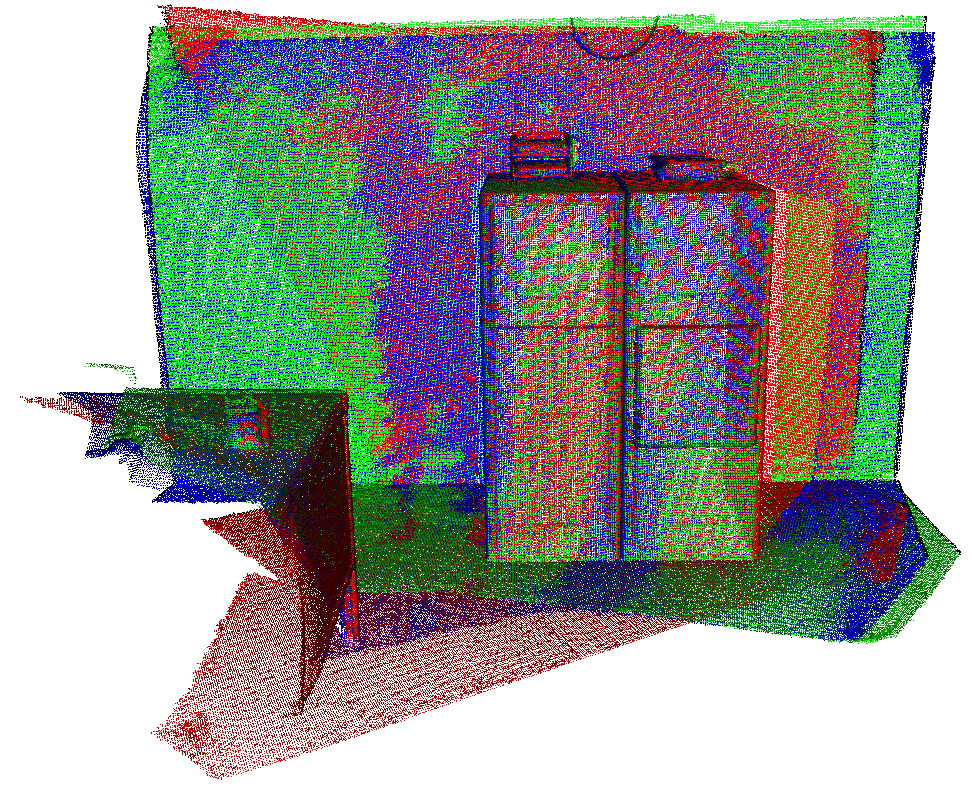}
		\includegraphics[width=0.42\columnwidth]{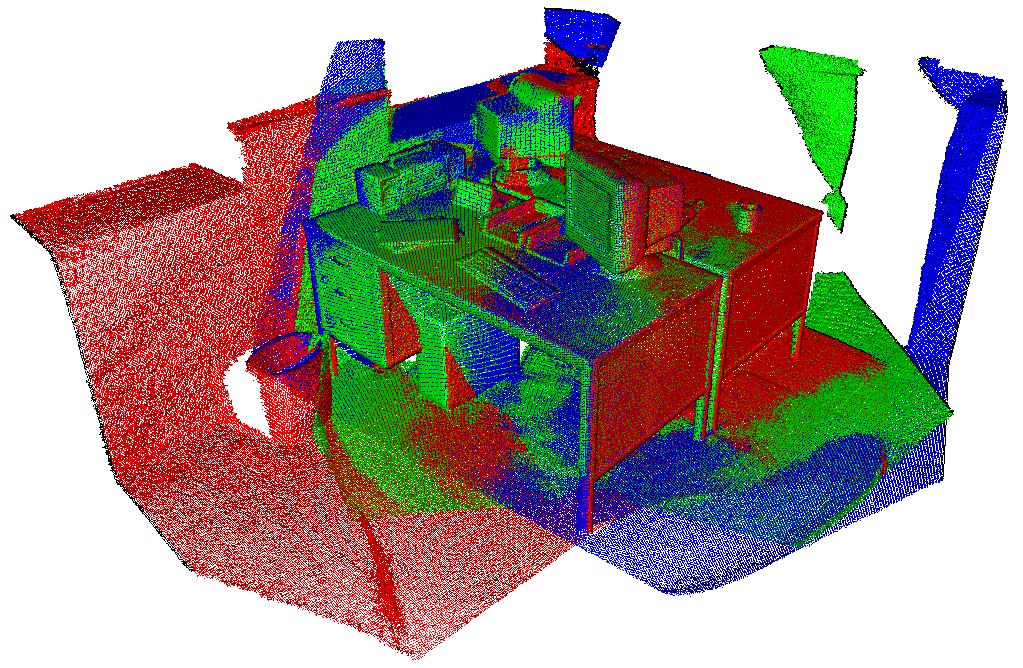}
		\includegraphics[width=0.42\columnwidth]{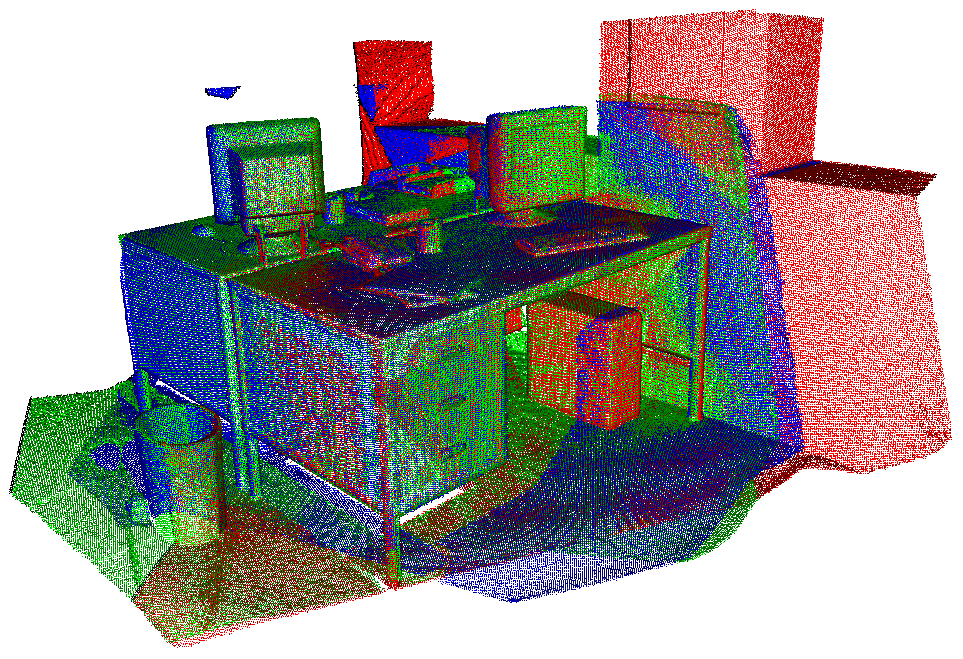}
		\includegraphics[width=0.42\columnwidth]{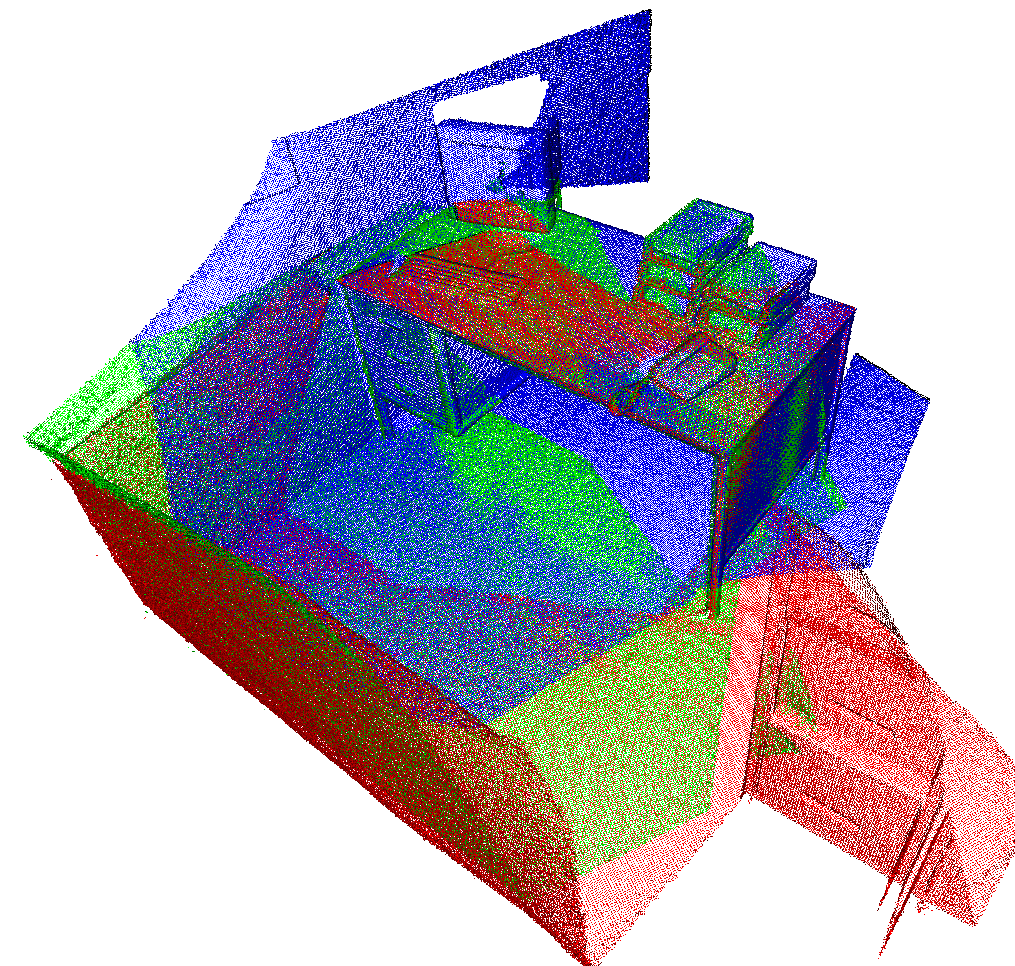}
	}
	\caption{Registration results of MTPCR on indoor scenes. First row: room. Second row: office. From left to right: registration problems with tag a, b, c and d.}
	\label{figIndoorScene}
\end{figure*}

Registration accuracy of the proposed MTPCR is compared in terms of rotation errors and translation errors with TrICP, KMEANS, EMPMR, and LSG-CPD. Table~\ref{tabErrorwithoutNoise} presents the results obtained by registration methods averaged over 10 independent runs. In the table, lowest overlap ratios among the three of each problem goes down with the symbols ``a'', ``b'', ``c'' and ``d'' in the ``Problem'' column. The column ``Err\_R'' and ``Err\_T'' give the rotation error and translation error compared with the ground truth values. For convenient comparison, all the errors as well as their standard deviations are magnified 1000 times. Smaller errors demonstrate better registration accuracy. Since rotation error is the main concern in registration problems, superior rotation results of each problem are highlighted in bold font. Fig.~\ref{figModelObjects} and Fig.~\ref{figIndoorScene} shows the visualization of the registration results obtained by the proposed MTPCR on model objects and indoor scenes, with different colors denote points from different point clouds.

\begin{table*}[t]
	\caption{Registration Errors of TRICP, KMEANS, EMPMR, LSG-CPD and MTPCR on Corresponding Problems. All Values Are Magnified One Thousand Times. Lower Vaules Denote Better Performance. Best Results of Each Problem Is Highlighted in Bold Font.}
	\label{tabErrorwithoutNoise}
	\centering
	\renewcommand\arraystretch{1.2}
	\resizebox{\textwidth}{!}{
		\begin{tabular}{|l|l|l|l|l|l|l|l|l|l|l|}
			\hline
			\multirow{2}{*}{Problem} &
			\multicolumn{2}{c|}{TrICP} &
			\multicolumn{2}{c|}{KMEANS} &
			\multicolumn{2}{c|}{EMPMR} &
			\multicolumn{2}{c|}{LSG-CPD} &
			\multicolumn{2}{c|}{MTPCR} \\ 
			\cline{2-11}
			& Err\_R & Err\_T & Err\_R & Err\_T & Err\_R & Err\_T & Err\_R & Err\_T & Err\_R & Err\_T \\
			\hline
			Bunny-a & 203.24 	$\pm$	65.94 	&	5.57 	$\pm$	3.54 	&	119.93 	$\pm$	107.36 	&	2.96 	$\pm$	2.46 	&	6.23 	$\pm$	0.10 	&	0.25 	$\pm$	0.01 	&	\textbf{1.20 	$\pm$	0.01} 	&	0.07 	$\pm$	0.00 	&	3.73 	$\pm$	0.64 	&	0.19 	$\pm$	0.05 \\
			Bunny-b & 300.08 	$\pm$	160.11 	&	12.17 	$\pm$	8.66 	&	180.74 	$\pm$	191.58 	&	7.04 	$\pm$	9.59 	&	98.89 	$\pm$	235.09 	&	3.59 	$\pm$	8.89 	&	147.15 	$\pm$	140.51 	&	7.36 	$\pm$	7.12 	&	\textbf{4.76 	$\pm$	1.38} 	&	0.24 	$\pm$	0.06 \\
			Bunny-c & 251.12 	$\pm$	119.21 	&	9.59 	$\pm$	5.64 	&	160.13 	$\pm$	150.13 	&	4.74 	$\pm$	4.95 	&	62.38 	$\pm$	0.17 	&	1.89 	$\pm$	0.01 	&	105.59 	$\pm$	91.85 	&	4.83 	$\pm$	4.37 	&	\textbf{6.63 	$\pm$	1.65} 	&	0.28 	$\pm$	0.07 \\
			\hline
			Dragon-a & 132.06 	$\pm$	39.91 	&	3.45 	$\pm$	1.82 	&	112.78 	$\pm$	144.46 	&	3.54 	$\pm$	4.61 	&	\textbf{2.73 	$\pm$	0.05} 	&	0.24 	$\pm$	0.00 	&	24.46 	$\pm$	48.74 	&	1.22 	$\pm$	2.29 	&	4.36 	$\pm$	1.00 	&	0.48 	$\pm$	0.10 \\
			Dragon-b & 196.13 	$\pm$	67.20 	&	7.32 	$\pm$	4.33 	&	219.10 	$\pm$	143.51 	&	8.92 	$\pm$	5.69 	&	11.05 	$\pm$	0.04 	&	1.37 	$\pm$	0.01 	&	33.05 	$\pm$	58.54 	&	2.15 	$\pm$	2.72 	&	\textbf{8.47 	$\pm$	0.24} 	&	0.89 	$\pm$	0.01 \\
			Dragon-c & 114.90 	$\pm$	52.30 	&	4.89 	$\pm$	1.92 	&	137.96 	$\pm$	123.04 	&	4.35 	$\pm$	3.48 	&	11.44 	$\pm$	0.11 	&	1.16 	$\pm$	0.02 	&	19.74 	$\pm$	26.25 	&	1.31 	$\pm$	0.88 	&	\textbf{5.78 	$\pm$	1.21} 	&	0.68 	$\pm$	0.12 \\
			\hline
			Armadillo-a & 166.92 	$\pm$	77.26 	&	4.74 	$\pm$	2.74 	&	162.72 	$\pm$	140.55 	&	3.79 	$\pm$	3.66 	&	9.35 	$\pm$	0.13 	&	0.35 	$\pm$	0.01 	&	63.24 	$\pm$	118.26 	&	3.18 	$\pm$	5.71 	&	\textbf{2.64 	$\pm$	0.34} 	&	0.19 	$\pm$	0.06 \\
			Armadillo-b & 146.18 	$\pm$	92.73 	&	5.25 	$\pm$	3.18 	&	165.02 	$\pm$	137.31 	&	4.45 	$\pm$	3.24 	&	66.52 	$\pm$	64.16 	&	1.34 	$\pm$	1.61 	&	82.90 	$\pm$	117.50 	&	2.41 	$\pm$	2.90 	&	\textbf{3.00 	$\pm$	0.69} 	&	0.33 	$\pm$	0.08 \\
			Armadillo-c & 162.75 	$\pm$	76.46 	&	5.77 	$\pm$	3.54 	&	220.59 	$\pm$	141.23 	&	6.73 	$\pm$	4.16 	&	28.73 	$\pm$	0.19 	&	1.04 	$\pm$	0.00 	&	29.06 	$\pm$	60.02 	&	2.94 	$\pm$	4.23 	&	\textbf{2.36 	$\pm$	0.41} 	&	0.24 	$\pm$	0.06 \\
			\hline
			Buddha-a & 196.94 	$\pm$	75.50 	&	7.71 	$\pm$	6.64 	&	158.55 	$\pm$	118.75 	&	6.43 	$\pm$	6.12 	&	\textbf{1.00 	$\pm$	0.01} 	&	0.16 	$\pm$	0.01 	&	75.72 	$\pm$	120.79 	&	5.53 	$\pm$	8.69 	&	7.76 	$\pm$	0.58 	&	17.23 	$\pm$	2.62 \\
			Buddha-b & 140.37 	$\pm$	90.44 	&	6.59 	$\pm$	4.17 	&	198.25 	$\pm$	131.50 	&	6.74 	$\pm$	6.07 	&	\textbf{7.42 	$\pm$	0.02} 	&	0.16 	$\pm$	0.01 	&	25.44 	$\pm$	26.44 	&	2.63 	$\pm$	2.96 	&	9.62 	$\pm$	0.79 	&	38.07 	$\pm$	3.89 \\
			Buddha-c & 182.69 	$\pm$	86.50 	&	8.76 	$\pm$	6.63 	&	147.72 	$\pm$	133.58 	&	8.19 	$\pm$	6.85 	&	42.92 	$\pm$	30.58 	&	9.90 	$\pm$	7.97 	&	66.65 	$\pm$	62.95 	&	5.21 	$\pm$	5.08 	&	\textbf{12.37 	$\pm$	1.06} 	&	15.28 	$\pm$	3.97 \\
			\hline
			Buste-a & 105.42 	$\pm$	61.80 	&	2972.65 	$\pm$	1738.03 	&	166.88 	$\pm$	138.12 	&	2996.79 	$\pm$	1860.02 	&	137.52 	$\pm$	129.46 	&	2015.51 	$\pm$	1537.41 	&	\textbf{21.07 	$\pm$	38.16} 	&	871.08 	$\pm$	1594.33 	&	46.00 	$\pm$	0.84 	&	88.32 	$\pm$	2.51 \\
			Buste-b & 201.43 	$\pm$	208.50 	&	2479.02 	$\pm$	2559.52 	&	236.67 	$\pm$	246.80 	&	1900.73 	$\pm$	1730.13 	&	213.98 	$\pm$	212.48 	&	1363.61 	$\pm$	1219.36 	&	179.27 	$\pm$	208.01 	&	1519.04 	$\pm$	2021.03 	&	\textbf{3.24 	$\pm$	1.15} 	&	0.27 	$\pm$	0.04 \\
			Buste-c & 181.41 	$\pm$	122.58 	&	1411.48 	$\pm$	748.45 	&	219.22 	$\pm$	67.10 	&	747.79 	$\pm$	226.50 	&	121.72 	$\pm$	162.36 	&	452.98 	$\pm$	506.38 	&	\textbf{0.78 	$\pm$	0.22} 	&	55.35 	$\pm$	1.69 	&	7.14 	$\pm$	1.07 	&	0.29 	$\pm$	0.04 \\
			\hline
			Room-a & 215.77 	$\pm$	94.84 	&	316.39 	$\pm$	154.60 	&	148.82 	$\pm$	131.56 	&	244.41 	$\pm$	193.17 	&	16.72 	$\pm$	21.78 	&	87.87 	$\pm$	209.68 	&	92.11 	$\pm$	112.13 	&	192.11 	$\pm$	263.88 	&	\textbf{5.77 	$\pm$	0.49} 	&	0.27 	$\pm$	0.04 \\
			Room-b & 197.51 	$\pm$	180.28 	&	363.37 	$\pm$	309.66 	&	170.54 	$\pm$	127.11 	&	257.85 	$\pm$	152.94 	&	159.41 	$\pm$	236.71 	&	291.79 	$\pm$	405.72 	&	260.18 	$\pm$	301.72 	&	427.36 	$\pm$	422.79 	&	\textbf{23.45 	$\pm$	6.58} 	&	502.80 	$\pm$	68.89 \\
			Room-c & 142.80 	$\pm$	124.39 	&	370.72 	$\pm$	264.74 	&	173.63 	$\pm$	114.61 	&	205.80 	$\pm$	101.74 	&	96.06 	$\pm$	122.83 	&	384.29 	$\pm$	460.37 	&	107.61 	$\pm$	111.35 	&	477.28 	$\pm$	278.49 	&	\textbf{14.23 	$\pm$	5.85} 	&	489.81 	$\pm$	222.52 \\
			Room-d & 204.22 	$\pm$	120.85 	&	329.93 	$\pm$	157.65 	&	182.30 	$\pm$	141.90 	&	184.24 	$\pm$	128.46 	&	\textbf{11.16 	$\pm$	6.75} 	&	72.41 	$\pm$	65.17 	&	48.59 	$\pm$	72.76 	&	141.18 	$\pm$	84.80 	&	17.35 	$\pm$	6.04 	&	359.17 	$\pm$	81.78 \\
			\hline
			Office-a & 163.01 	$\pm$	98.22 	&	306.19 	$\pm$	133.18 	&	129.94 	$\pm$	125.31 	&	223.49 	$\pm$	168.58 	&	13.25 	$\pm$	5.08 	&	52.67 	$\pm$	79.85 	&	24.40 	$\pm$	29.10 	&	83.96 	$\pm$	74.04 	&	\textbf{8.52 	$\pm$	1.17} 	&	13.56 	$\pm$	1.03 \\
			Office-b & 144.50 	$\pm$	103.06 	&	285.25 	$\pm$	214.12 	&	148.21 	$\pm$	134.04 	&	98.16 	$\pm$	74.55 	&	8.11 	$\pm$	0.04 	&	17.08 	$\pm$	0.19 	&	30.02 	$\pm$	21.45 	&	130.26 	$\pm$	133.42 	&	\textbf{7.07 	$\pm$	1.21} 	&	21.09 	$\pm$	2.59 \\
			Office-c & 130.42 	$\pm$	55.23 	&	204.05 	$\pm$	77.23 	&	43.52 	$\pm$	28.47 	&	74.82 	$\pm$	57.08 	&	\textbf{12.77 	$\pm$	0.23} 	&	6.23 	$\pm$	0.13 	&	43.75 	$\pm$	47.53 	&	64.61 	$\pm$	53.90 	&	13.47 	$\pm$	1.24 	&	9.68 	$\pm$	1.98 \\
			Office-d & 189.44 	$\pm$	344.33 	&	324.44 	$\pm$	496.90 	&	119.35 	$\pm$	181.21 	&	198.53 	$\pm$	201.89 	&	42.23 	$\pm$	91.45 	&	103.50 	$\pm$	206.18 	&	81.23 	$\pm$	159.40 	&	243.76 	$\pm$	195.92 	&	\textbf{21.20 	$\pm$	1.31} 	&	100.08 	$\pm$	12.50 \\
			\hline
		\end{tabular}
	}
\end{table*}

It can be observed from Table~\ref{tabErrorwithoutNoise} that the proposed MTPCR provides superior or competitive performance on almost all the problems possessing varied overlap ratios. For the problem Bunny-a, LSG-CPD enjoys the smallest rotation and translation error, with MTPCR follows very closely. But LSG-CPD does not seem to be stable, for the rotation errors increase rapidly when it comes to Bunny-b and Bunny-c. In contrast, although not the best on Bunny-a, MTPCR continues to perform fairly good on Bunny-b and Bunny-c, and exceeds other methods by a large margin. Situations are very similar on the three problems of Dragon, where EMPMR performs best on Dragon-a with MTPCR follows very closely, and the performance of MTPCR exceeds all others on Dragon-b and Dragon-c. It has to be mentioned that the lowest overlap ratios of Bunny-a and Dragon-a are about 50\%, and decrease to below 30\% on Bunny-c and Dragon-c, demonstrating the proposed fitness function are less sensitive to outliers compared with other methods. Analyzing overlap ratios of the problems where MTPCR does not perform best, it is easy to find that the lowest overlap ratios are relatively high, meaning that less outliers exist in the problem. Buste-c is an exception, on which LSG-CPD obtains abnormally low rotation error but high translation error. Though not perform best on several of the problems, the errors of MTPCR follows closely with the best results. By contrast, on the majority of problems where MTPCR provides best rotation error, performance of other methods vary greatly. Through results presented in Table~\ref{tabErrorwithoutNoise}, the effectiveness of the proposed fitness function that considers both local accuracy and global consistency can be verified.

\begin{table*}[ht]
	\caption{Registration Errors of TRICP, KMEANS, EMPMR, LSG-CPD and MTPCR on Representative Problems with Different Gaussian Noise Levels. All Values Are Magnified One Thousand Times. Lower Vaules Denote Better Performance. Best Results of Each Problem Is Highlighted in Bold Font.}
	\label{tabErrorwithNoise}
	\centering
	\renewcommand\arraystretch{1.2}
	\resizebox{\textwidth}{!}{
		\begin{tabular}{|l|l|l|l|l|l|l|l|l|l|l|l|}
			\hline
			\multirow{2}{*}{Problem} &
			\multirow{2}{*}{Noise} & 
			\multicolumn{2}{c|}{TrICP} &
			\multicolumn{2}{c|}{KMEANS} &
			\multicolumn{2}{c|}{EMPMR} &
			\multicolumn{2}{c|}{LSG-CPD} &
			\multicolumn{2}{c|}{MTPCR} \\
			\cline{3-12}
			& & Err\_R & Err\_T & Err\_R & Err\_T & Err\_R & Err\_T & Err\_R & Err\_T & Err\_R & Err\_T \\
			\hline
			\multirow{3}{*}{Bunny-a} 
			& 0.01 & 205.04 	$\pm$	64.79 	&	4.92 	$\pm$	3.41 	&	133.41 	$\pm$	125.13 	&	3.02 	$\pm$	2.81 	&	23.87 	$\pm$	0.11 	&	0.56 	$\pm$	0.00 	&	\textbf{4.10 	$\pm$	0.72} 	&	0.22 	$\pm$	0.03 	&	9.36 	$\pm$	2.82 	&	0.47 	$\pm$	0.25 \\
			& 0.02 & 208.76 	$\pm$	63.25 	&	3.87 	$\pm$	3.24 	&	98.68 	$\pm$	112.96 	&	2.01 	$\pm$	2.16 	&	47.00 	$\pm$	0.74 	&	1.24 	$\pm$	0.06 	&	33.66 	$\pm$	15.54 	&	1.04 	$\pm$	0.46 	&	\textbf{26.08 	$\pm$	1.97} 	&	1.45 	$\pm$	0.13 \\
			& 0.03 & 213.52 	$\pm$	64.66 	&	2.74 	$\pm$	2.75 	&	65.49 	$\pm$	35.97 	&	1.59 	$\pm$	0.70 	&	67.91 	$\pm$	0.57 	&	1.83 	$\pm$	0.03 	&	66.61 	$\pm$	18.97 	&	2.06 	$\pm$	0.90 	&	\textbf{27.24 	$\pm$	4.79} 	&	0.97 	$\pm$	0.46 \\
			\hline
			\multirow{3}{*}{Bunny-b}
			& 0.01 & 296.39 	$\pm$	158.40 	&	11.39 	$\pm$	8.91 	&	161.18 	$\pm$	152.36 	&	5.56 	$\pm$	6.05 	&	157.60 	$\pm$	217.80 	&	5.60 	$\pm$	8.35 	&	237.43 	$\pm$	191.97 	&	11.51 	$\pm$	9.56 	&	\textbf{17.03 	$\pm$	18.99} 	&	0.68 	$\pm$	0.74 \\
			& 0.02 & 286.65 	$\pm$	149.11 	&	9.87 	$\pm$	8.63 	&	145.44 	$\pm$	177.06 	&	6.51 	$\pm$	8.54 	&	243.29 	$\pm$	186.33 	&	8.67 	$\pm$	6.61 	&	301.22 	$\pm$	208.87 	&	13.31 	$\pm$	10.39 	&	\textbf{19.43 	$\pm$	7.54} 	&	1.08 	$\pm$	0.41 \\
			& 0.03 & 276.06 	$\pm$	130.03 	&	7.82 	$\pm$	7.56 	&	180.17 	$\pm$	163.88 	&	7.25 	$\pm$	7.48 	&	306.85 	$\pm$	164.48 	&	10.50 	$\pm$	5.53 	&	340.10 	$\pm$	213.08 	&	14.42 	$\pm$	10.58 	&	\textbf{34.08 	$\pm$	8.37} 	&	1.90 	$\pm$	0.32 \\
			\hline
			\multirow{3}{*}{Bunny-c}
			& 0.01 & 252.96 	$\pm$	113.86 	&	8.70 	$\pm$	5.60 	&	143.77 	$\pm$	139.33 	&	3.85 	$\pm$	4.31 	&	106.16 	$\pm$	3.09 	&	2.41 	$\pm$	0.11 	&	137.14 	$\pm$	100.56 	&	5.90 	$\pm$	5.03 	&	\textbf{27.66 	$\pm$	15.65} 	&	1.56 	$\pm$	0.72 \\
			& 0.02 & 255.36 	$\pm$	114.11 	&	7.80 	$\pm$	5.67 	&	117.41 	$\pm$	130.67 	&	3.76 	$\pm$	4.19 	&	122.18 	$\pm$	2.50 	&	2.55 	$\pm$	0.02 	&	209.69 	$\pm$	113.08 	&	7.08 	$\pm$	5.02 	&	\textbf{47.77 	$\pm$	11.91} 	&	2.46 	$\pm$	0.88 \\
			& 0.03 & 250.03 	$\pm$	102.35 	&	5.92 	$\pm$	4.76 	&	142.09 	$\pm$	140.88 	&	4.70 	$\pm$	5.15 	&	185.69 	$\pm$	2.48 	&	4.30 	$\pm$	0.22 	&	231.30 	$\pm$	120.23 	&	7.38 	$\pm$	4.60 	&	\textbf{74.98 	$\pm$	8.33} 	&	3.74 	$\pm$	0.53 \\
			\hline
			\multirow{3}{*}{Buddha-a}
			& 0.01 & 200.37 	$\pm$	73.38 	&	7.31 	$\pm$	6.48 	&	158.48 	$\pm$	141.64 	&	7.03 	$\pm$	6.24 	&	\textbf{3.13 	$\pm$	0.07} 	&	0.34 	$\pm$	0.01 	&	79.88 	$\pm$	114.08 	&	6.07 	$\pm$	8.57 	&	7.60 	$\pm$	1.73 	&	0.64 	$\pm$	0.23 \\
			& 0.02 & 204.01 	$\pm$	67.45 	&	6.21 	$\pm$	5.57 	&	140.08 	$\pm$	140.46 	&	7.72 	$\pm$	8.18 	&	\textbf{15.34 	$\pm$	0.96} 	&	1.05 	$\pm$	0.08 	&	61.29 	$\pm$	85.01 	&	4.55 	$\pm$	5.88 	&	17.80 	$\pm$	2.07 	&	1.61 	$\pm$	0.32 \\
			& 0.03 & 202.45 	$\pm$	66.77 	&	5.13 	$\pm$	4.93 	&	115.31 	$\pm$	108.42 	&	7.13 	$\pm$	5.74 	&	34.88 	$\pm$	0.79 	&	3.10 	$\pm$	0.07 	&	84.62 	$\pm$	84.85 	&	5.91 	$\pm$	5.94 	&	\textbf{29.94 	$\pm$	4.23} 	&	2.56 	$\pm$	0.71 \\
			\hline
			\multirow{3}{*}{Buddha-b}
			& 0.01 & 146.71 	$\pm$	80.09 	&	6.88 	$\pm$	3.70 	&	186.50 	$\pm$	112.71 	&	5.95 	$\pm$	6.61 	&	33.94 	$\pm$	0.80 	&	1.45 	$\pm$	0.07 	&	18.25 	$\pm$	11.21 	&	2.46 	$\pm$	1.39 	&	\textbf{10.22 	$\pm$	1.09} 	&	0.99 	$\pm$	0.12 \\
			& 0.02 & 153.54 	$\pm$	82.37 	&	6.14 	$\pm$	3.71 	&	167.98 	$\pm$	111.61 	&	6.46 	$\pm$	5.78 	&	53.65 	$\pm$	3.01 	&	1.71 	$\pm$	0.52 	&	28.37 	$\pm$	8.88 	&	3.83 	$\pm$	1.37 	&	\textbf{15.69 	$\pm$	3.22} 	&	1.44 	$\pm$	0.42 \\
			& 0.03 & 162.68 	$\pm$	77.95 	&	5.37 	$\pm$	3.44 	&	161.12 	$\pm$	126.88 	&	7.20 	$\pm$	5.59 	&	57.92 	$\pm$	2.54 	&	2.31 	$\pm$	0.30 	&	39.09 	$\pm$	8.60 	&	4.88 	$\pm$	1.61 	&	\textbf{27.99 	$\pm$	3.57} 	&	2.74 	$\pm$	1.18 \\
			\hline
			\multirow{3}{*}{Buddha-c}
			& 0.01 & 183.12 	$\pm$	82.74 	&	8.05 	$\pm$	6.82 	&	128.08 	$\pm$	127.80 	&	6.44 	$\pm$	5.88 	&	57.66 	$\pm$	32.66 	&	10.69 	$\pm$	7.96 	&	56.78 	$\pm$	72.09 	&	4.25 	$\pm$	5.25 	&	\textbf{10.86 	$\pm$	3.44} 	&	1.11 	$\pm$	0.34 \\
			& 0.02 & 187.43 	$\pm$	76.92 	&	7.13 	$\pm$	6.33 	&	111.83 	$\pm$	106.05 	&	9.99 	$\pm$	9.69 	&	81.80 	$\pm$	27.78 	&	14.67 	$\pm$	6.77 	&	53.29 	$\pm$	68.01 	&	4.84 	$\pm$	5.41 	&	\textbf{23.10 	$\pm$	3.90} 	&	2.28 	$\pm$	0.60 \\
			& 0.03 & 183.88 	$\pm$	71.02 	&	6.04 	$\pm$	5.37 	&	169.80 	$\pm$	150.03 	&	12.43 	$\pm$	10.33 	&	85.78 	$\pm$	13.76 	&	15.69 	$\pm$	5.54 	&	71.30 	$\pm$	55.39 	&	6.18 	$\pm$	4.58 	&	\textbf{42.72 	$\pm$	6.26} 	&	3.16 	$\pm$	0.95 \\
			\hline
		\end{tabular}
	}
\end{table*}

\begin{figure*}[!ht]
	\centering
	\subfloat{
		\centering
		\includegraphics[width=0.3\columnwidth]{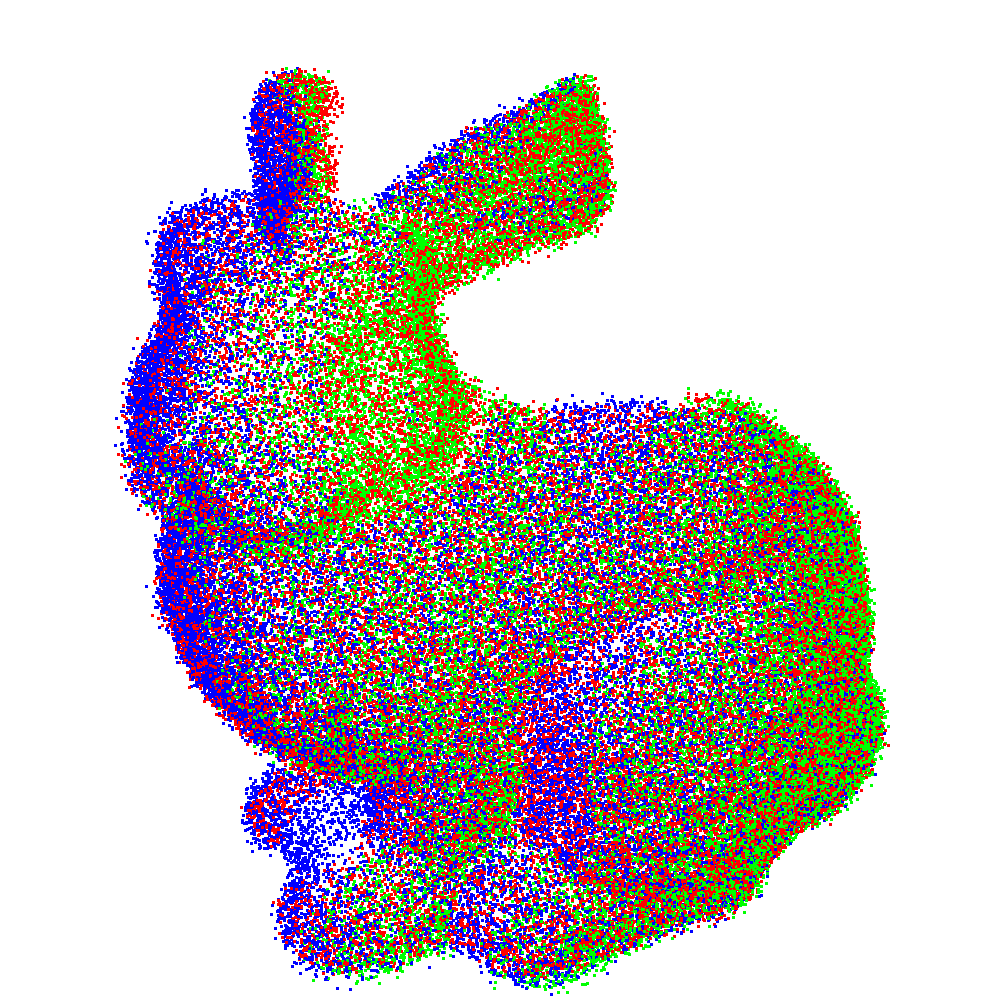}
		\includegraphics[width=0.3\columnwidth]{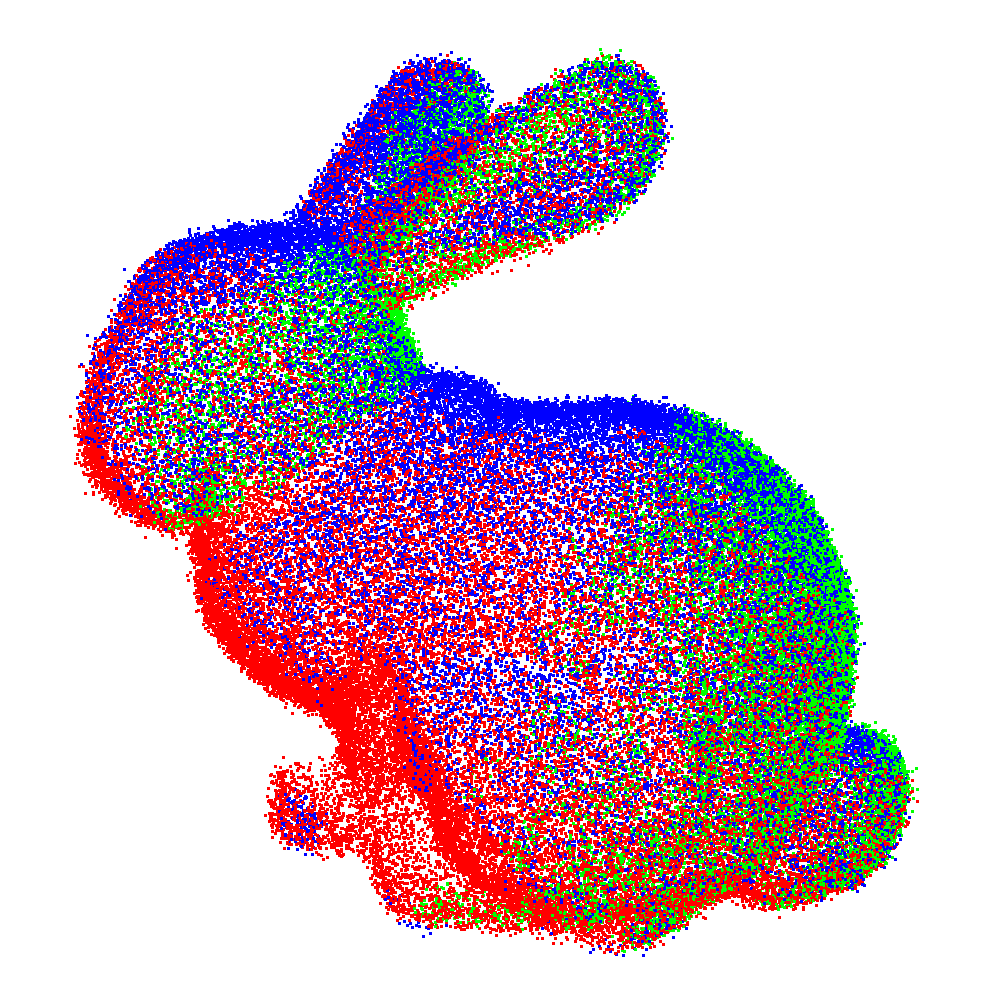}
		\includegraphics[width=0.3\columnwidth]{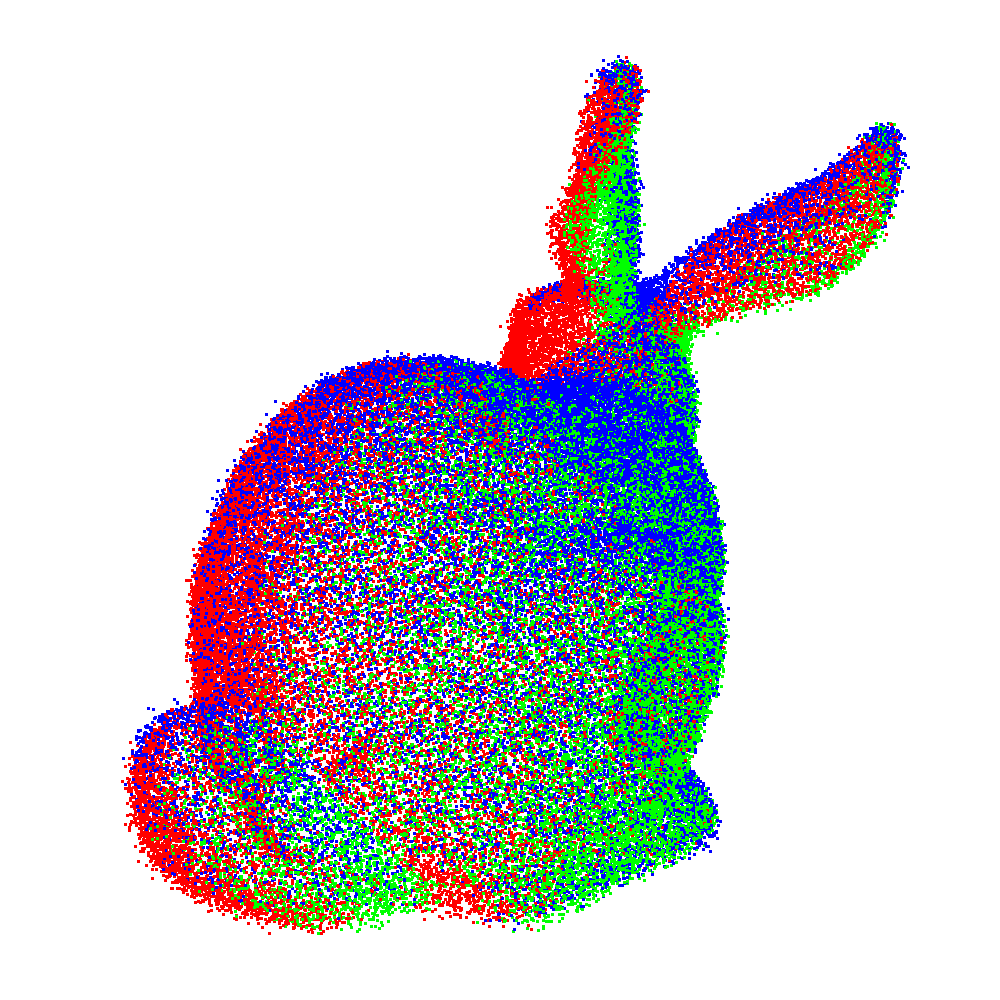}
		\includegraphics[width=0.3\columnwidth]{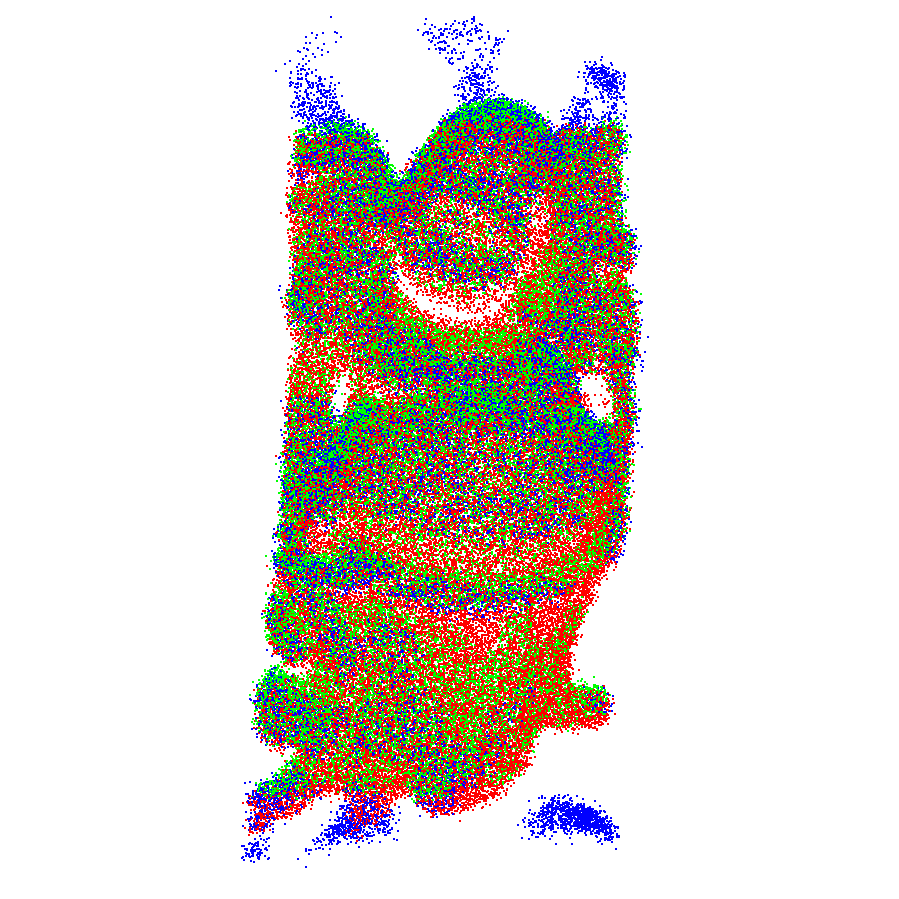}
		\includegraphics[width=0.3\columnwidth]{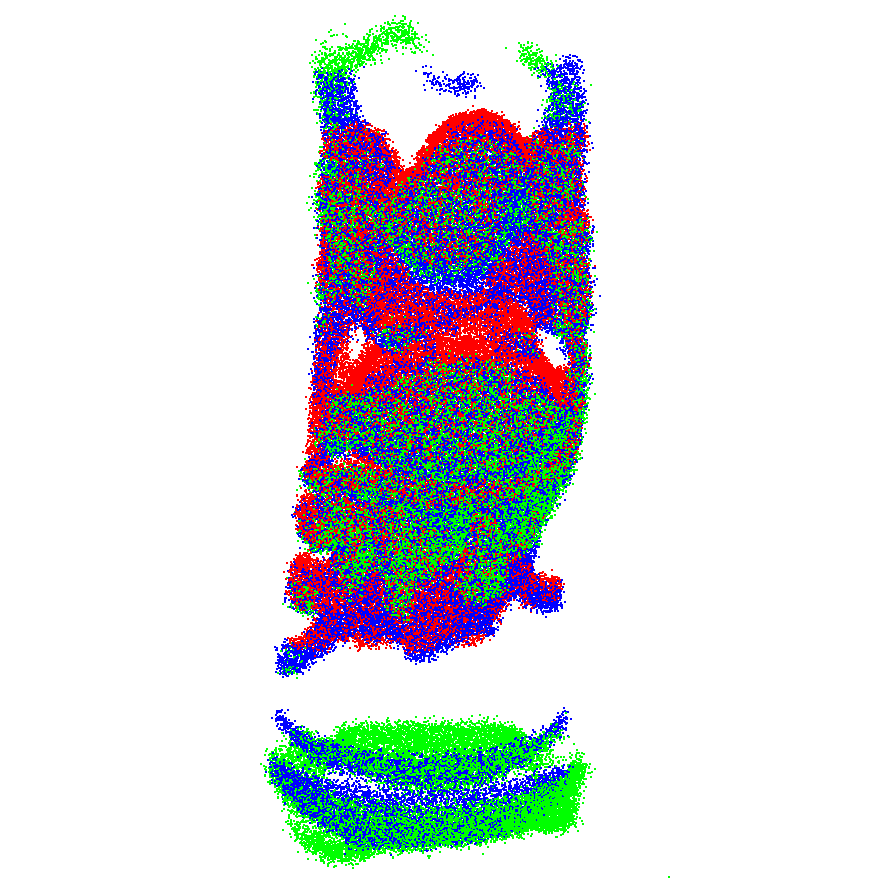}
		\includegraphics[width=0.3\columnwidth]{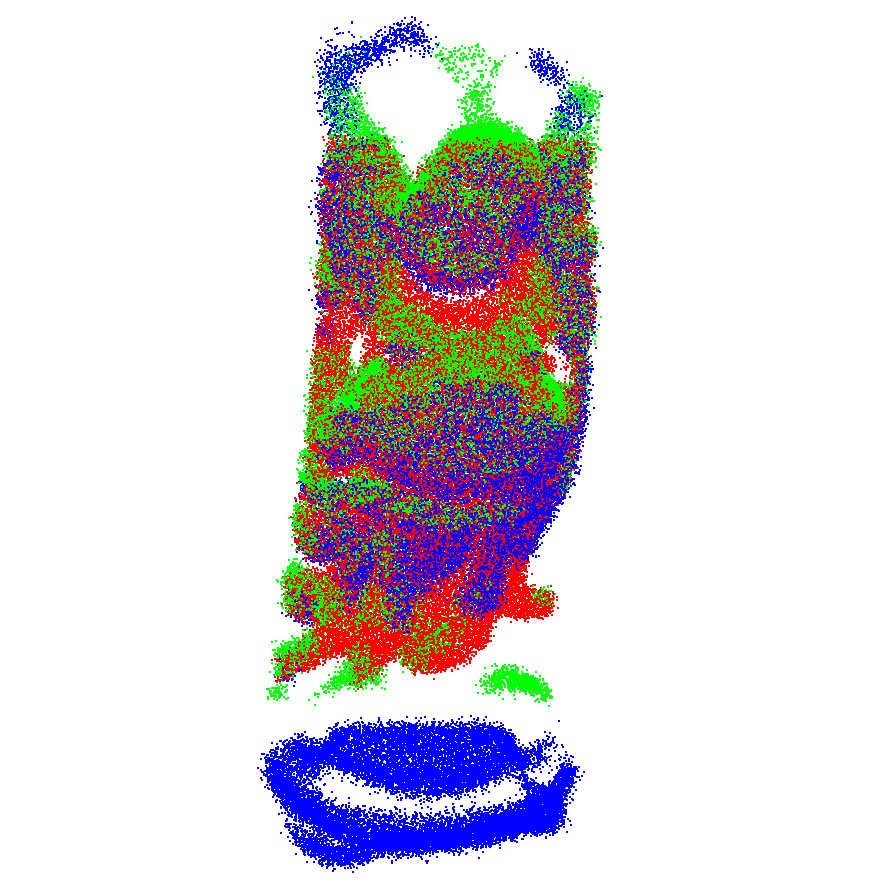}
	}\\
	\vspace{-1ex}
	\subfloat{
		\centering
		\includegraphics[width=0.3\columnwidth]{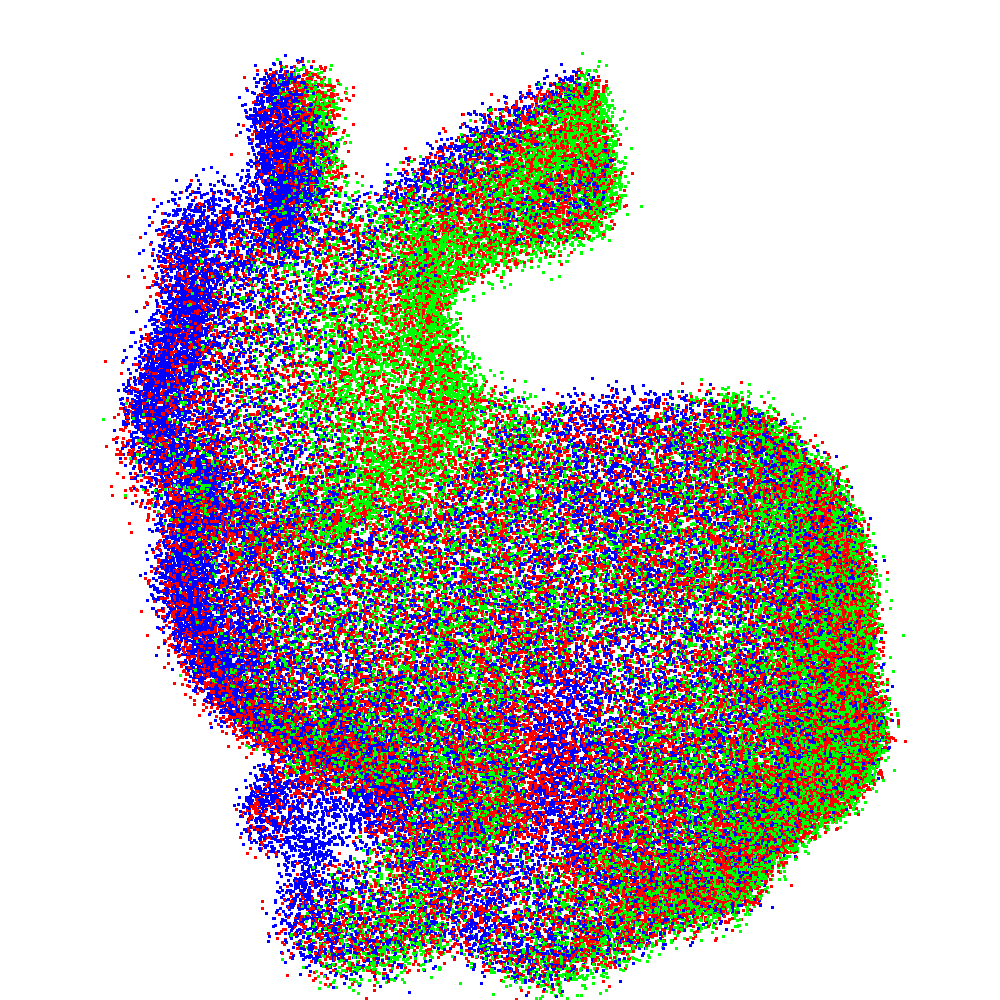}
		\includegraphics[width=0.3\columnwidth]{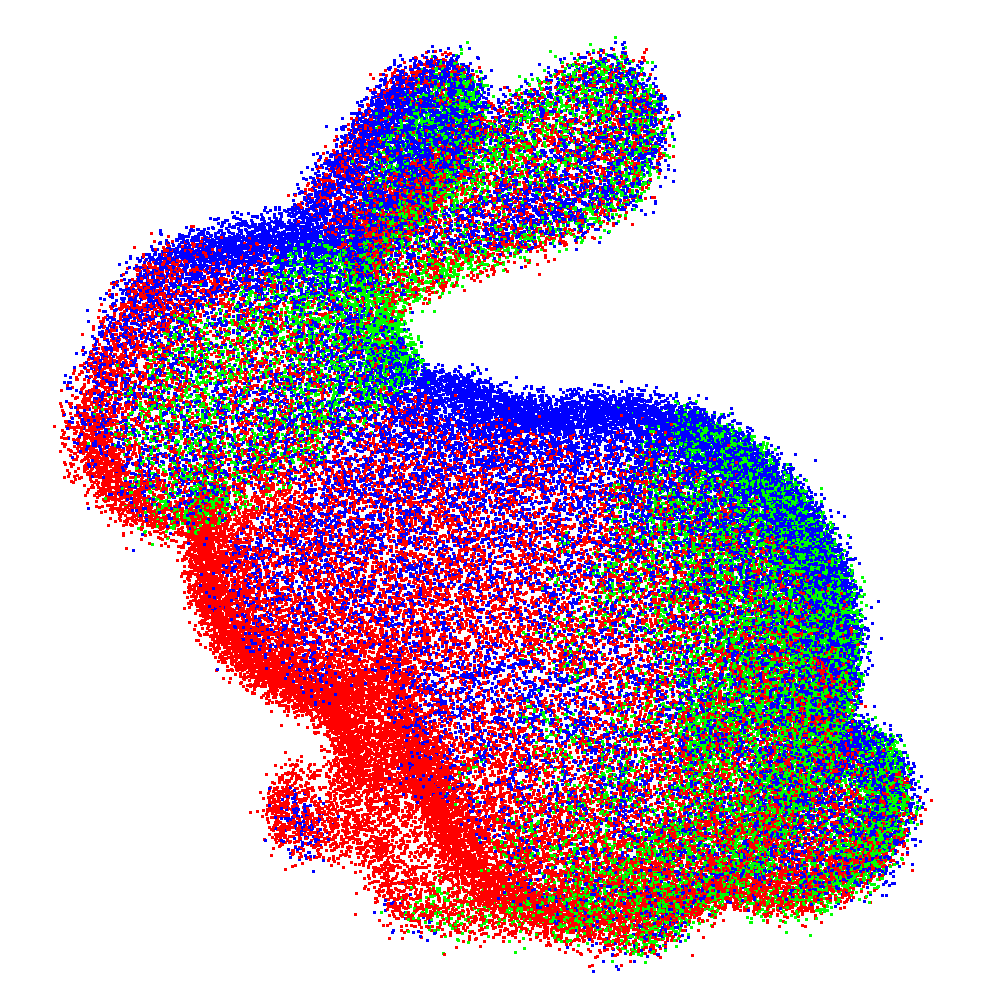}
		\includegraphics[width=0.3\columnwidth]{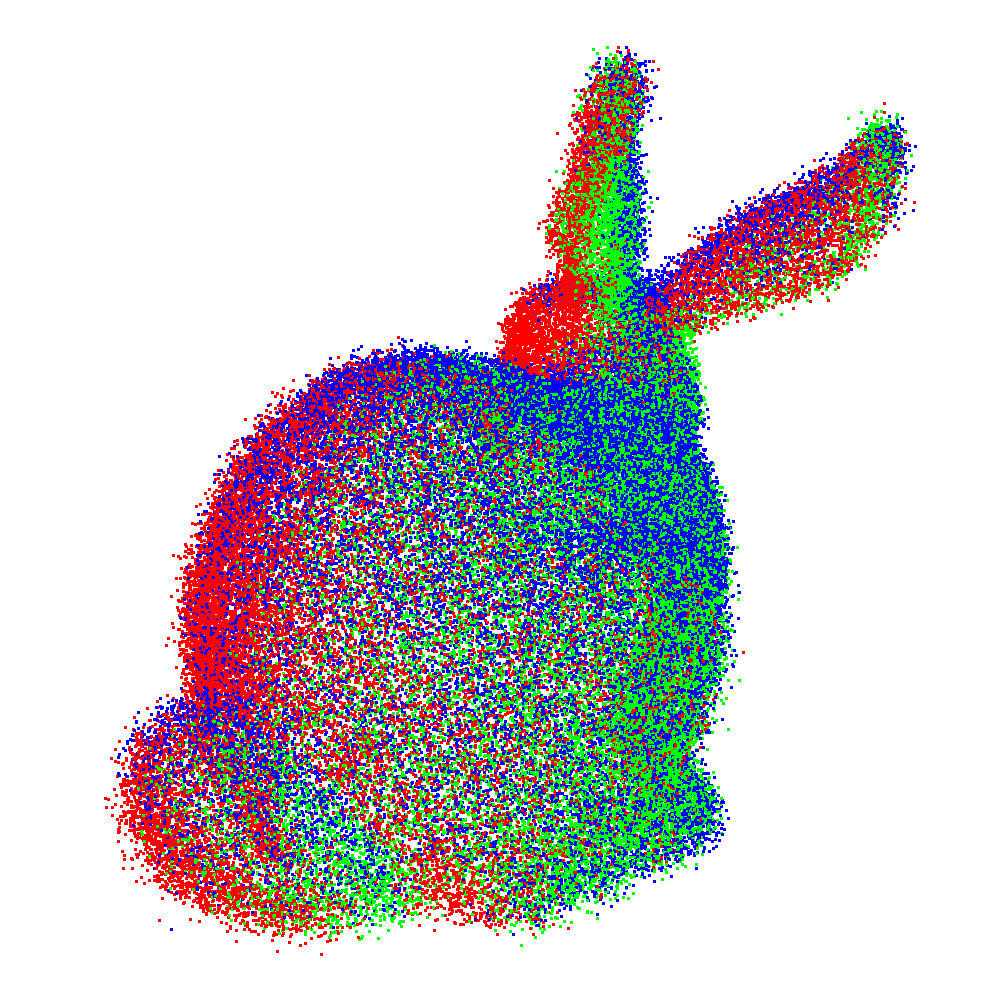}
		\includegraphics[width=0.3\columnwidth]{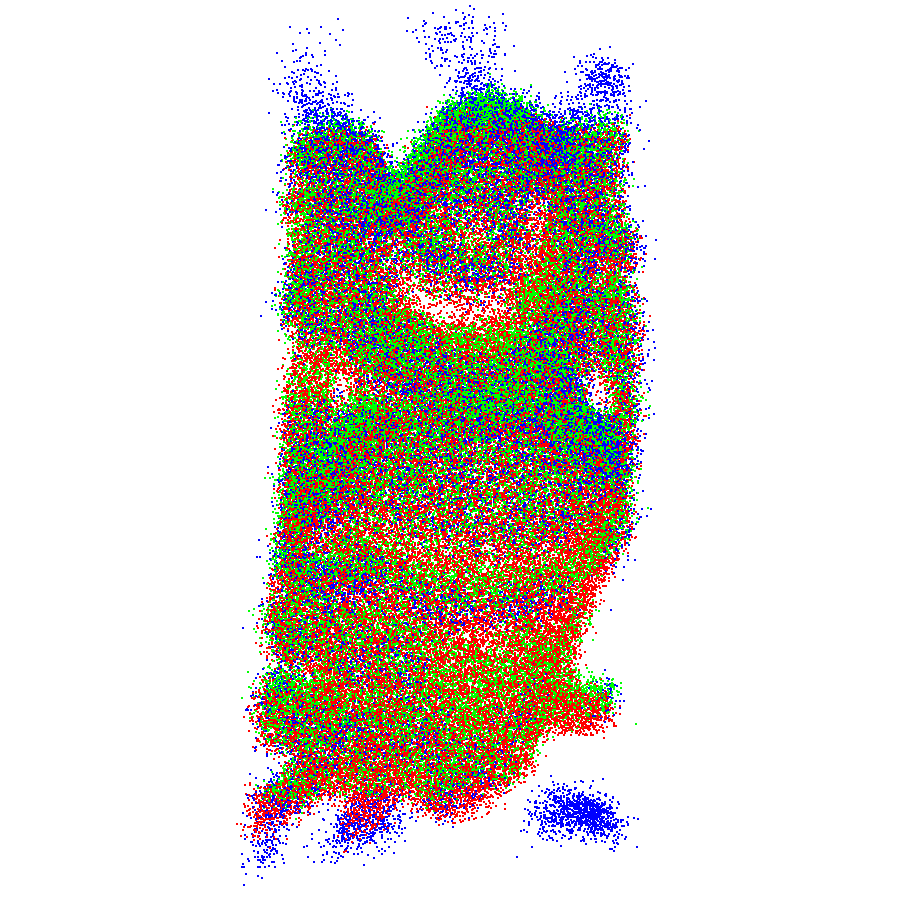}
		\includegraphics[width=0.3\columnwidth]{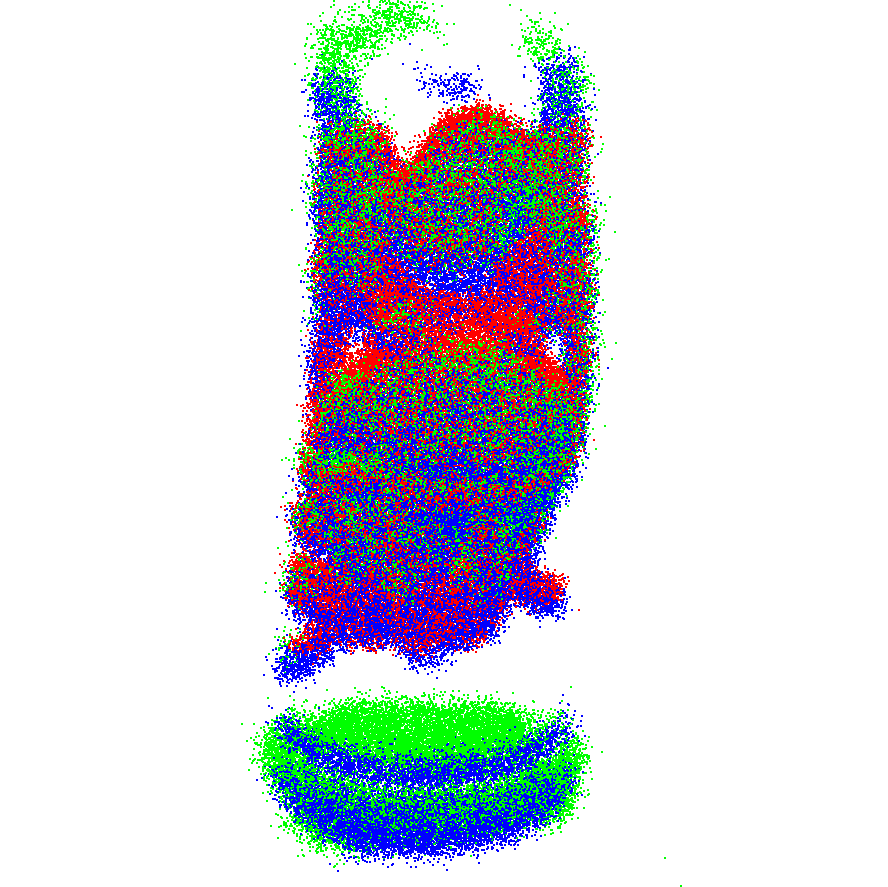}
		\includegraphics[width=0.3\columnwidth]{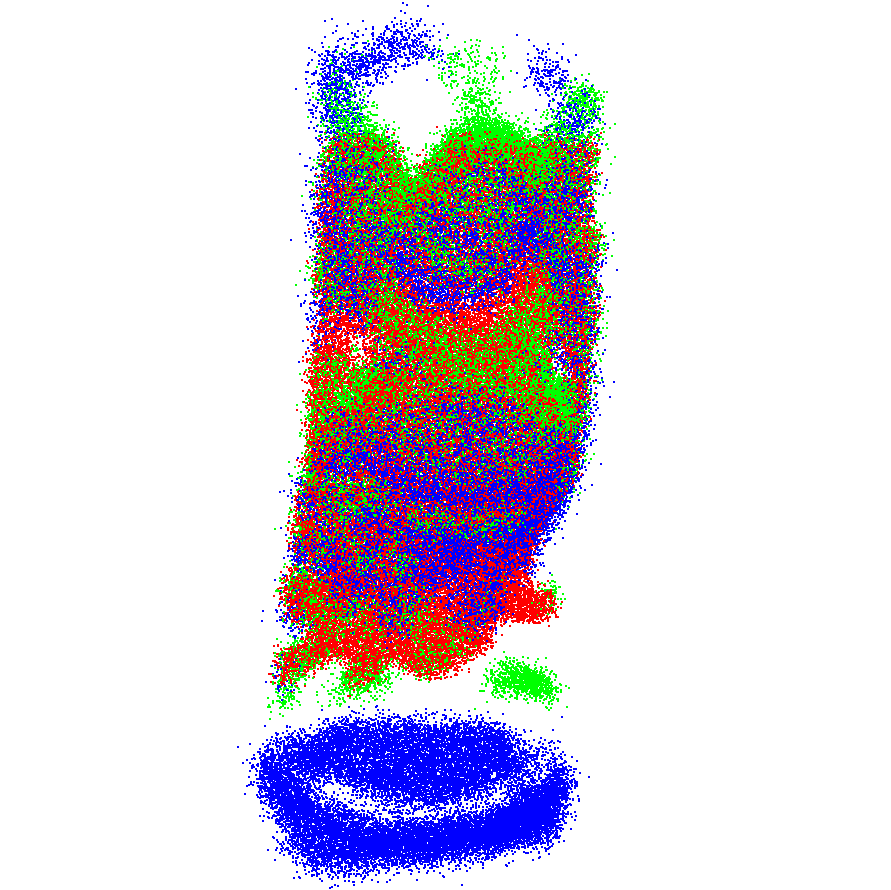}
	}\\
	\vspace{-1ex}
	\subfloat{
		\centering
		\includegraphics[width=0.3\columnwidth]{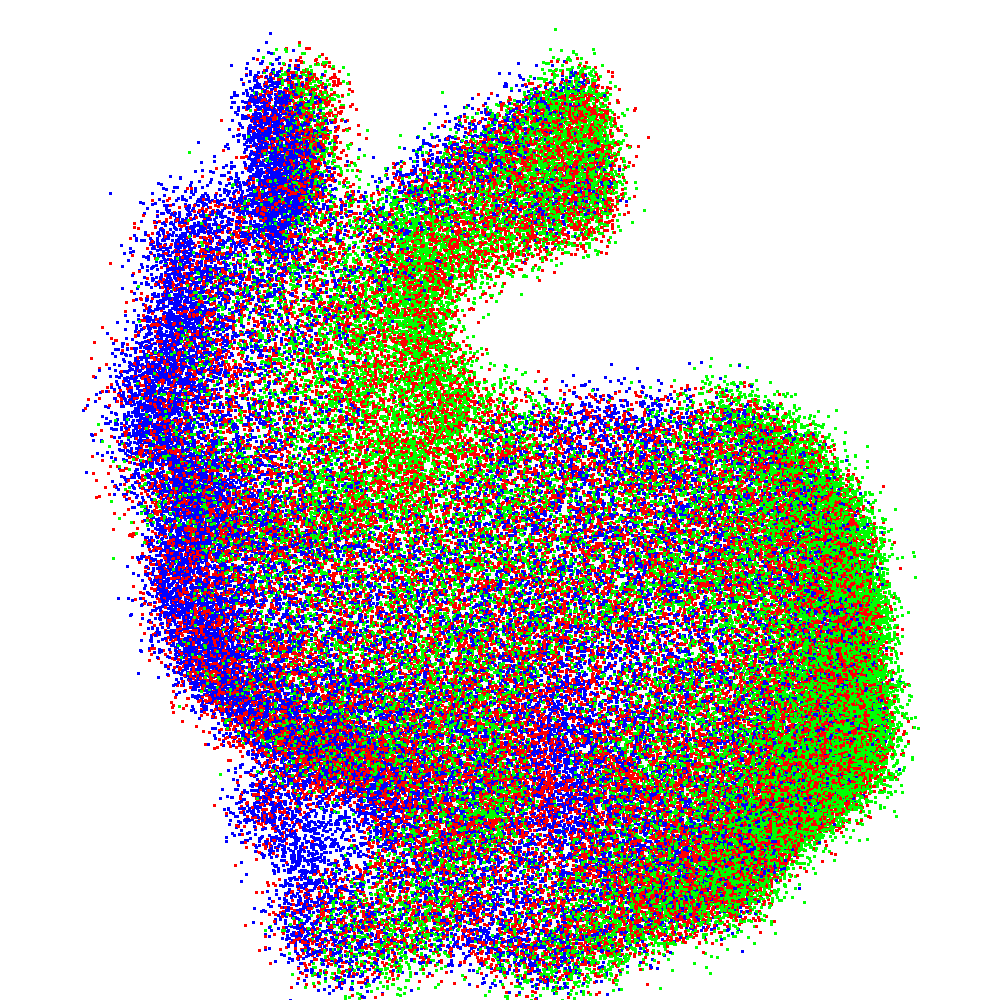}
		\includegraphics[width=0.3\columnwidth]{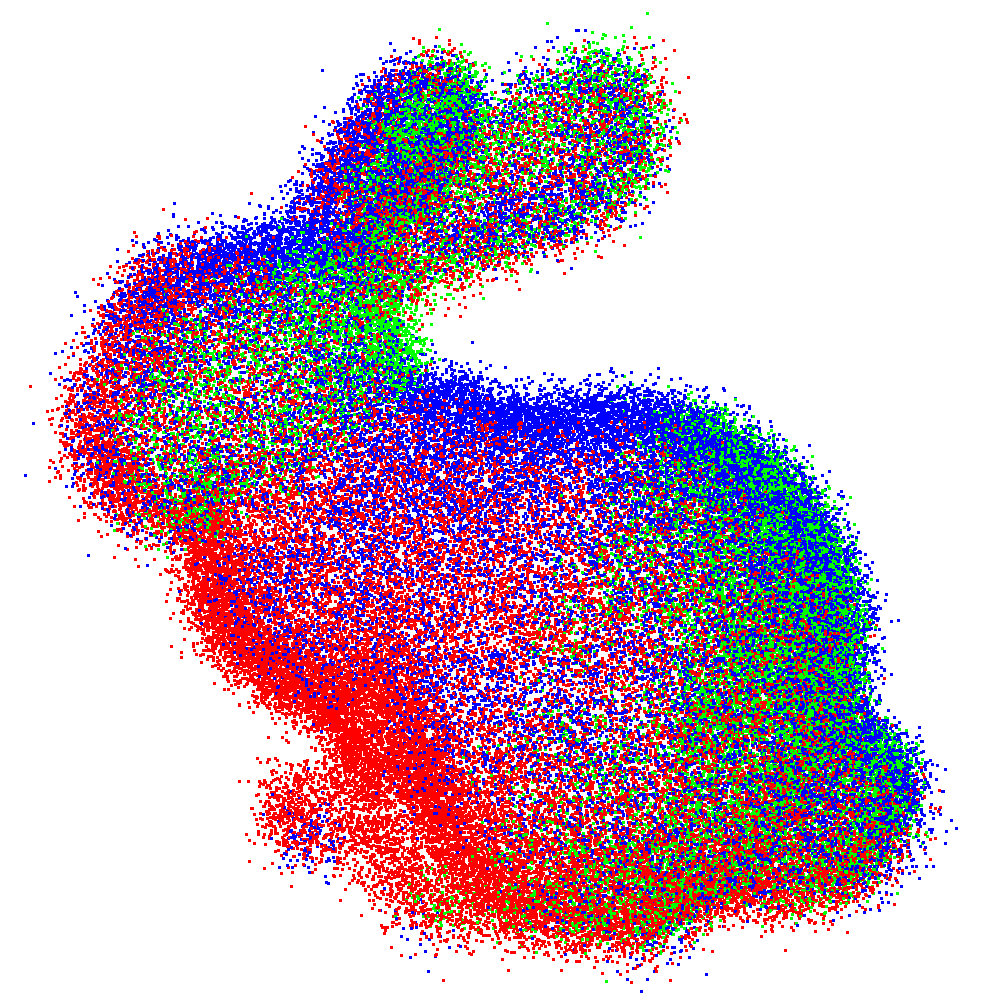}
		\includegraphics[width=0.3\columnwidth]{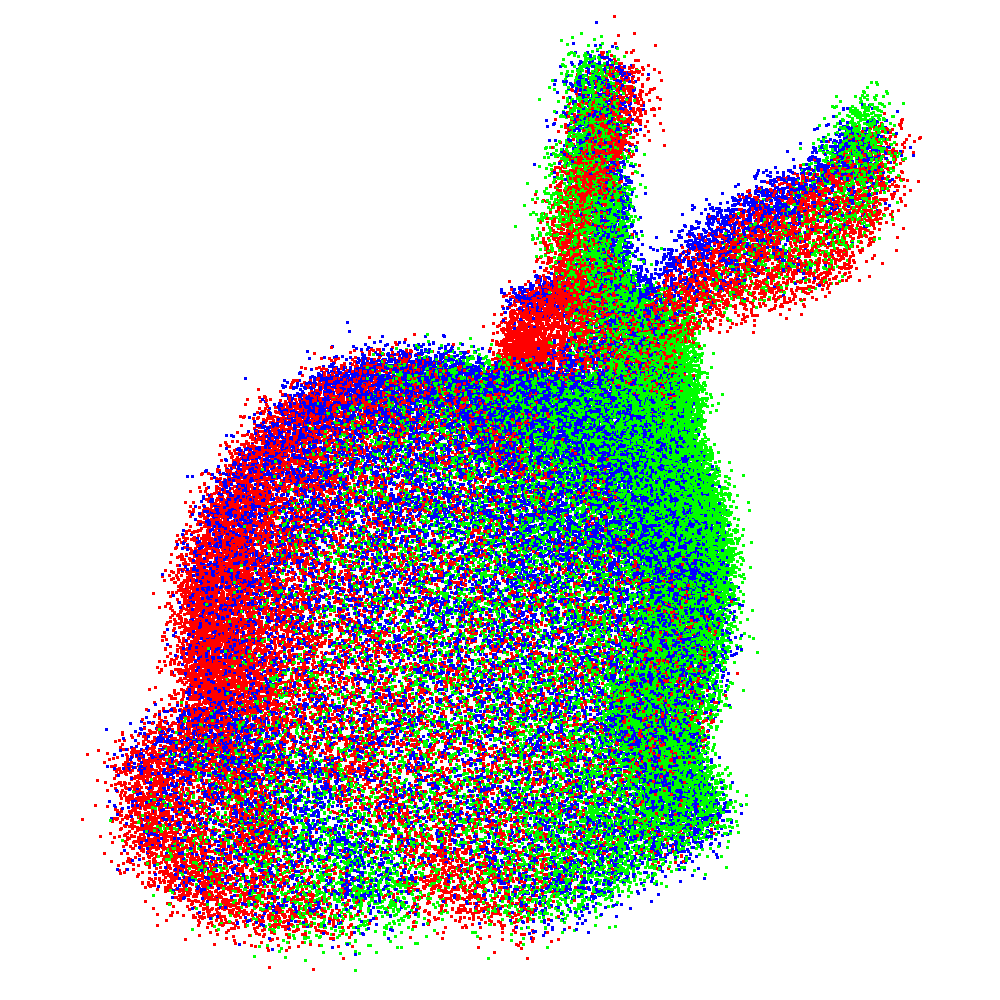}
		\includegraphics[width=0.3\columnwidth]{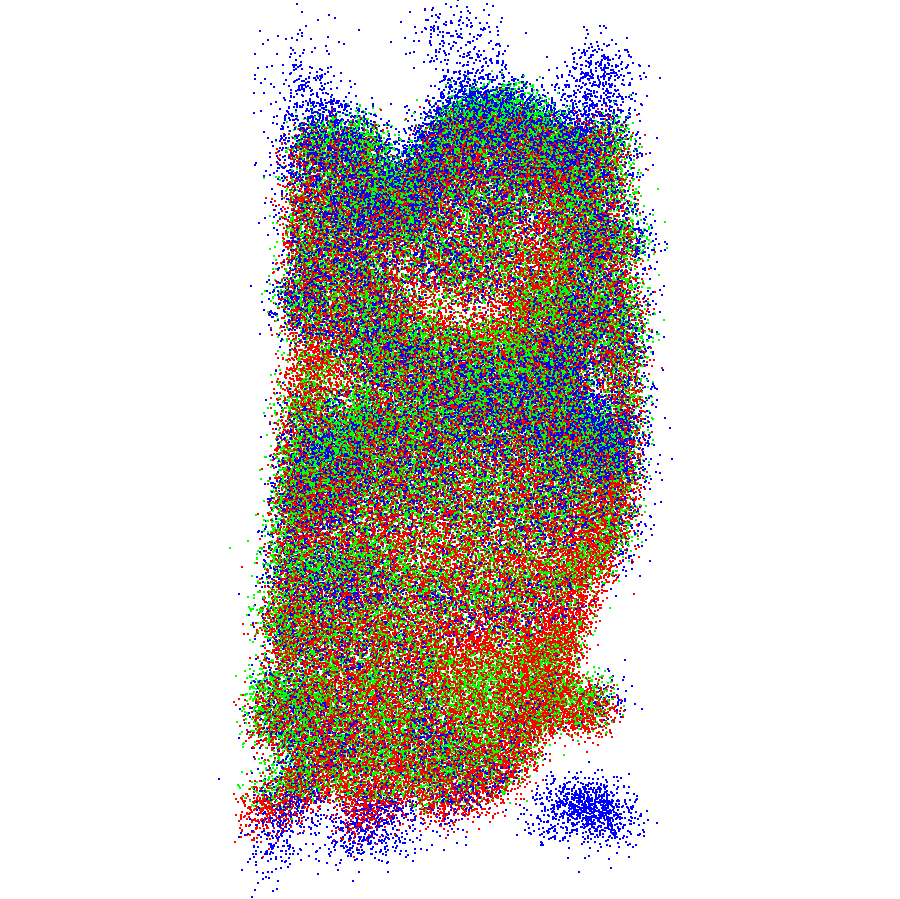}
		\includegraphics[width=0.3\columnwidth]{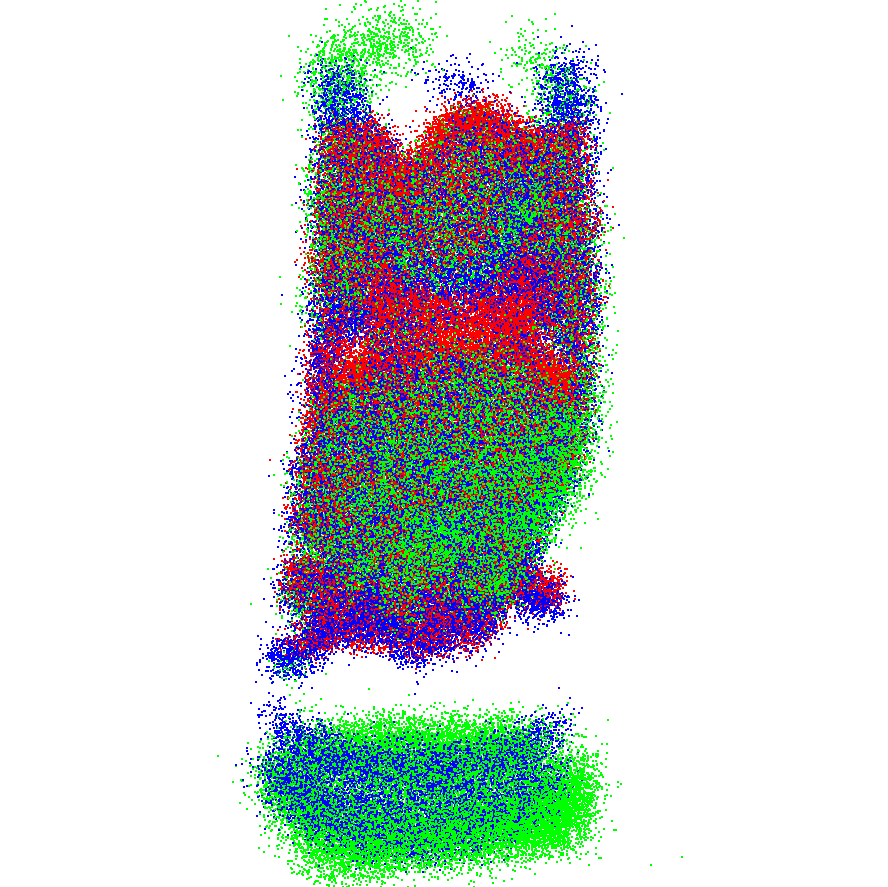}
		\includegraphics[width=0.3\columnwidth]{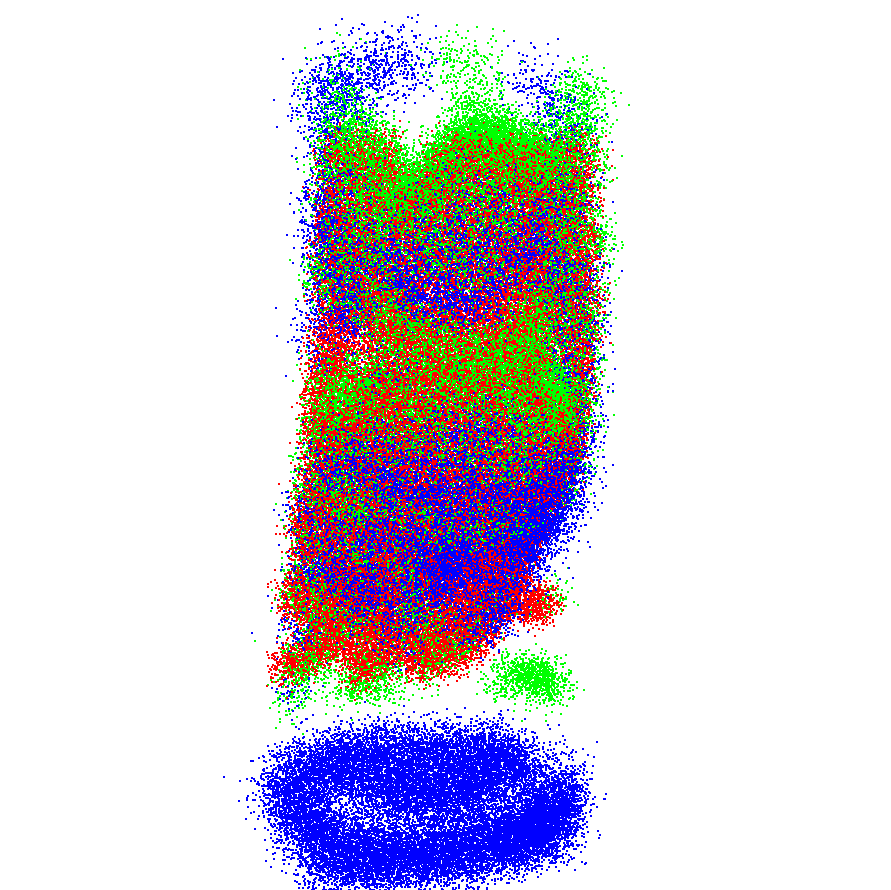}
	}
	\caption{Registration results of MTPCR on Bunny and Buddha with different gaussian noise levels. From top to bottom: gaussian noise levels with standard deviation 0.01, 0.02 and 0.03. From left to right: Bunny-a, Bunny-b, Bunny-c, Buddha-a, Buddha-b, and Buddha-c. }
	\label{figModelObjectsNoisy}
\end{figure*}

\subsection{Robustness}
In addition to accuracy comparison on the registration problems without noise, experitments are also conducted on representative problems with different gaussian noise levels to demonstrate the robust of the proposed fitness function. In order to obtain noisy data, the original point clouds are first normalized to the range  $[-1,~1]$, and gaussian noise with zero mean and standard deviation $\sigma$ is added to the normalized point clouds before recovering to their original scale. In the experiments, the values 0.01, 0.02 and 0.03 are used as deviations to generate noisy point clouds. Table~\ref{tabErrorwithNoise} tabulates the registration errors on Bunny and Buddha with random gaussian noise. Fig.~\ref{figModelObjectsNoisy} shows the visualization of registration results obtained by the proposed MTPCR on noisy point clouds, with different colors denote different point clouds.

Compared to their noise free counterparts as shown in Table~\ref{tabErrorwithoutNoise}, the accuracy of all the methods tend to decrease with the increase of noise levels, except for the KMEANS method. This may be ascribed to the clustering process used in this approach, where a fixed number of points near each cluster center are selected to estimate a new center, thus it is less sensitive to noise. The performance of TrICP seems similiar whether there is noise or not, meaning that it is also less sensitive to noise. However, none of the best results are obtained by TrICP or KMEANS. On the problem Bunny-a with 0.01 gaussian noise, LSG-CPD achieves best rotation error, as in the noise free case. But with the increase of noise levels, the performance of MTPCR exceeds LSG-CPD on Bunny-a. On Buddha-a with 0.01 and 0.02 gaussian noise, EMPMR achieves best rotation error as in the noise free case, while it has also been exceeded when noise arises to 0.03. For all the problems listed in Table~\ref{tabErrorwithNoise} with varied overlap ratios and different gaussian noise levels, MTPCR outperforms other methods on 15 out of 18, demonstrating the superiority in registering noisy point clouds.

\begin{table*}
	\centering
	\caption{Solution Quality of MTPCR, MFEA, GMFEA, MTPCR-Intra, MTPCR-Inter, and MTPCR-NKS on Corresponding Problems. Superior Quality Solutions (Best Solution and Solutions Within One Percent Away From the Best) of Each Respective Problem Instance Are Highlighted in Bold Font.}
	\label{tabSolutionQuality}
	\renewcommand\arraystretch{1.1}
	\resizebox{\textwidth}{!}{
		\begin{tabular}{l|l|ll|ll|ll|ll|ll|ll}
			\hline
			\multirow{2}{*}{Problem} &
			\multirow{2}{*}{Task No.} &
			\multicolumn{2}{c|}{MTPCR} &
			\multicolumn{2}{c|}{MFEA} &
			\multicolumn{2}{c|}{GMFEA} &
			\multicolumn{2}{c|}{MTPCR-Intra} &
			\multicolumn{2}{c|}{MTPCR-Inter} &
			\multicolumn{2}{c}{MTPCR-NKS} \\
			& & $B.Cost$ & $Avg.Cost$ &  $B.Cost$ & $Avg.Cost$ &  $B.Cost$ & $Avg.Cost$ &  $B.Cost$ & $Avg.Cost$ &  $B.Cost$ & $Avg.Cost$ &  $B.Cost$ & $Avg.Cost$  \\
			\hline
			\hline
			Bunny-a & $J_1$-$o$ & \textbf{0.2004} 	&	\textbf{0.2016 	$\pm$	0.0006} 	&	0.2939 	&	0.4273 	$\pm$	0.0834 	&	0.2616 	&	0.4561 	$\pm$	0.1309 	&	\textbf{0.2005} 	&	0.3072 	$\pm$	0.2390 	&	\textbf{0.2011} 	&	0.7205 	$\pm$	0.2984 	&	0.2150 	&	0.8091 	$\pm$	0.1999 \\
			Bunny-a & $J_2$-$o$ & \textbf{0.4614} 	&	\textbf{0.4621 	$\pm$	0.0008} 	&	0.4915 	&	0.6759 	$\pm$	0.0990 	&	0.4842 	&	0.6642 	$\pm$	0.1069 	&	\textbf{0.4613} 	&	0.5107 	$\pm$	0.1238 	&	\textbf{0.4614} 	&	0.7980 	$\pm$	0.1912 	&	0.5136 	&	0.9051 	$\pm$	0.1027 \\
			Bunny-a & $J_3$-$o$ & \textbf{0.3198} 	&	\textbf{0.3235 	$\pm$	0.0116} 	&	0.3885 	&	0.5304 	$\pm$	0.0888 	&	0.4047 	&	0.5610 	$\pm$	0.1044 	&	\textbf{0.3199} 	&	0.3369 	$\pm$	0.0298 	&	\textbf{0.3204} 	&	0.7868 	$\pm$	0.2225 	&	0.3674 	&	0.8047 	$\pm$	0.2240 \\
			Bunny-b & $J_1$-$o$ & \textbf{0.6158} 	&	\textbf{0.6812 	$\pm$	0.1015} 	&	0.7634 	&	0.8566 	$\pm$	0.0465 	&	0.7034 	&	0.8274 	$\pm$	0.0648 	&	\textbf{0.6168} 	&	0.8509 	$\pm$	0.0725 	&	0.8384 	&	0.9119 	$\pm$	0.0418 	&	0.8502 	&	0.9437 	$\pm$	0.0436 \\
			Bunny-b & $J_2$-$o$ & \textbf{0.4694} 	&	\textbf{0.5019 	$\pm$	0.0898} 	&	0.5672 	&	0.7447 	$\pm$	0.0708 	&	0.5657 	&	0.7559 	$\pm$	0.0798 	&	\textbf{0.4694} 	&	0.5694 	$\pm$	0.0890 	&	0.7722 	&	0.8896 	$\pm$	0.0580 	&	0.5449 	&	0.8811 	$\pm$	0.1073 \\
			Bunny-b & $J_3$-$o$ & \textbf{0.4196} 	&	\textbf{0.5153 	$\pm$	0.1711} 	&	0.6666 	&	0.8078 	$\pm$	0.0696 	&	0.6084 	&	0.7708 	$\pm$	0.0827 	&	\textbf{0.4203} 	&	0.5662 	$\pm$	0.1598 	&	0.8269 	&	0.8898 	$\pm$	0.0362 	&	0.8270 	&	0.9151 	$\pm$	0.0469 \\
			Bunny-c & $J_1$-$o$ & \textbf{0.6553} 	&	\textbf{0.7073 	$\pm$	0.0820} 	&	0.6901 	&	0.8150 	$\pm$	0.0684 	&	0.7051 	&	0.8183 	$\pm$	0.0705 	&	\textbf{0.6554} 	&	0.7188 	$\pm$	0.0892 	&	0.8348 	&	0.8798 	$\pm$	0.0329 	&	0.6646 	&	0.8779 	$\pm$	0.0853 \\
			Bunny-c & $J_2$-$o$ & \textbf{0.2973} 	&	\textbf{0.3777 	$\pm$	0.1761} 	&	0.4955 	&	0.6934 	$\pm$	0.1361 	&	0.3378 	&	0.6003 	$\pm$	0.1805 	&	\textbf{0.2976} 	&	0.4125 	$\pm$	0.2044 	&	\textbf{0.2982} 	&	0.8138 	$\pm$	0.1647 	&	0.3304 	&	0.8240 	$\pm$	0.1771 \\
			Bunny-c & $J_3$-$o$ & \textbf{0.5519} 	&	\textbf{0.6306 	$\pm$	0.1304} 	&	0.6449 	&	0.7952 	$\pm$	0.0691 	&	0.6378 	&	0.7922 	$\pm$	0.0768 	&	\textbf{0.5519} 	&	0.6384 	$\pm$	0.1279 	&	0.8108 	&	0.8787 	$\pm$	0.0451 	&	0.5716 	&	0.8791 	$\pm$	0.0956 \\
			Dragon-a & $J_1$-$o$ & \textbf{0.1871} 	&	\textbf{0.2214 	$\pm$	0.1472} 	&	0.3530 	&	0.5754 	$\pm$	0.1189 	&	0.3507 	&	0.5819 	$\pm$	0.1625 	&	\textbf{0.1871} 	&	0.3105 	$\pm$	0.2517 	&	0.9135 	&	0.9557 	$\pm$	0.0236 	&	0.8477 	&	0.9598 	$\pm$	0.0364 \\
			Dragon-a & $J_2$-$o$ & \textbf{0.3176} 	&	\textbf{0.3185 	$\pm$	0.0008} 	&	0.4396 	&	0.6295 	$\pm$	0.1097 	&	0.5079 	&	0.6864 	$\pm$	0.1227 	&	\textbf{0.3173} 	&	0.4555 	$\pm$	0.2446 	&	0.9446 	&	0.9693 	$\pm$	0.0130 	&	0.9751 	&	0.9862 	$\pm$	0.0062 \\
			Dragon-a & $J_3$-$o$ & \textbf{0.4239} 	&	\textbf{0.4483 	$\pm$	0.1036} 	&	0.4973 	&	0.7502 	$\pm$	0.1400 	&	0.4402 	&	0.6982 	$\pm$	0.1343 	&	\textbf{0.4237} 	&	0.5392 	$\pm$	0.2058 	&	0.9212 	&	0.9586 	$\pm$	0.0192 	&	0.9631 	&	0.9839 	$\pm$	0.0064 \\
			Dragon-b & $J_1$-$o$ & \textbf{0.3260} 	&	\textbf{0.3271 	$\pm$	0.0007} 	&	0.3579 	&	0.5493 	$\pm$	0.1159 	&	0.3999 	&	0.5955 	$\pm$	0.1089 	&	\textbf{0.3262} 	&	0.4112 	$\pm$	0.1921 	&	0.8831 	&	0.9235 	$\pm$	0.0238 	&	0.8988 	&	0.9634 	$\pm$	0.0273 \\
			Dragon-b & $J_2$-$o$ & \textbf{0.3774} 	&	\textbf{0.3784 	$\pm$	0.0006} 	&	0.4218 	&	0.5735 	$\pm$	0.0872 	&	0.4758 	&	0.6066 	$\pm$	0.0915 	&	\textbf{0.3774}	    &	0.3868 	$\pm$	0.0176 	&	\textbf{0.3781 }	&	0.8836 	$\pm$	0.1274 	&	0.7357 	&	0.9435 	$\pm$	0.0624 \\
			Dragon-b & $J_3$-$o$ & \textbf{0.6206 }	&	\textbf{0.6212 	$\pm$	0.0003} 	&	0.6666 	&	0.7891 	$\pm$	0.0839 	&	0.6641 	&	0.8334 	$\pm$	0.0937 	&	\textbf{0.6205}	   &	0.6690 	$\pm$	0.1078 	&	0.7984 	&	0.9140 	$\pm$	0.0430 	&	0.8829 	&	0.9615 	$\pm$	0.0283 \\
			Dragon-c & $J_1$-$o$ & \textbf{0.3539} 	&	\textbf{0.4368 	$\pm$	0.1932} 	&	0.3962 	&	0.6505 	$\pm$	0.1419 	&	0.4411 	&	0.6646 	$\pm$	0.1318 	&	\textbf{0.3545} 	&	0.4999 	$\pm$	0.2201 	&	\textbf{0.3547} 	&	0.9170 	$\pm$	0.1327 	&	0.9261 	&	0.9772 	$\pm$	0.0155 \\
			Dragon-c & $J_2$-$o$ & \textbf{0.3785} 	&	\textbf{0.5005 	$\pm$	0.2089} 	&	0.5419 	&	0.7144 	$\pm$	0.1028 	&	0.5055 	&	0.7233 	$\pm$	0.1016 	&	\textbf{0.3795} 	&	0.6250 	$\pm$	0.2427 	&	\textbf{0.3799} 	&	0.9072 	$\pm$	0.1268 	&	0.8368 	&	0.9654 	$\pm$	0.0441 \\
			Dragon-c & $J_3$-$o$ & \textbf{0.6332} 	&	\textbf{0.7424 	$\pm$	0.1327} 	&	0.6578 	&	0.9002 	$\pm$	0.0667 	&	0.7288 	&	0.8681 	$\pm$	0.0708 	&	\textbf{0.6335} 	&	0.8273 	$\pm$	0.1379 	&	0.8825 	&	0.9523 	$\pm$	0.0264 	&	0.9437 	&	0.9829 	$\pm$	0.0107 \\
			Armadillo-a & $J_1$-$o$ & \textbf{0.2546} 	&	\textbf{0.2553 	$\pm$	0.0004} 	&	0.3308 	&	0.5234 	$\pm$	0.1257 	&	0.2920 	&	0.4423 	$\pm$	0.1283 	&	\textbf{0.2546} 	&	0.3188 	$\pm$	0.1825 	&	\textbf{0.2559} 	&	0.8572 	$\pm$	0.1505 	&	0.8445 	&	0.9474 	$\pm$	0.0480 \\
			Armadillo-a & $J_2$-$o$ & \textbf{0.2601} 	&	\textbf{0.2928 	$\pm$	0.1384} 	&	0.2719 	&	0.4897 	$\pm$	0.1339 	&	0.2881 	&	0.4509 	$\pm$	0.1064 	&	\textbf{0.2603} 	&	0.2979 	$\pm$	0.1379 	&	\textbf{0.2604} 	&	0.8353 	$\pm$	0.1870 	&	0.8025 	&	0.9399 	$\pm$	0.0533 \\
			Armadillo-a & $J_3$-$o$ & \textbf{0.4623} 	&	\textbf{0.4851 	$\pm$	0.0972} 	&	0.5839 	&	0.7062 	$\pm$	0.0823 	&	0.5039 	&	0.7041 	$\pm$	0.1082 	&	\textbf{0.4624} 	&	\textbf{0.4839 	$\pm$	0.0600} 	&	\textbf{0.4639} 	&	0.8568 	$\pm$	0.1368 	&	0.5137 	&	0.9090 	$\pm$	0.1070 \\
			Armadillo-b & $J_1$-$o$ & \textbf{0.2702} 	&	\textbf{0.2939 	$\pm$	0.1001} 	&	0.4552 	&	0.6046 	$\pm$	0.0840 	&	0.3616 	&	0.6421 	$\pm$	0.1209 	&	0.3083 	&	0.4082 	$\pm$	0.2135 	&	0.8726 	&	0.9225 	$\pm$	0.0263 	&	0.9119 	&	0.9818 	$\pm$	0.0228 \\
			Armadillo-b & $J_2$-$o$ & \textbf{0.4152} 	&	0.6024 	$\pm$	0.2292 	&	0.4900 	&	0.6990 	$\pm$	0.1112 	&	0.5013 	&	0.7761 	$\pm$	0.0992 	&	0.4200 	&	\textbf{0.5090 	$\pm$	0.1425} 	&	0.8387 	&	0.9307 	$\pm$	0.0289 	&	0.9601 	&	0.9867 	$\pm$	0.0094 \\
			Armadillo-b & $J_3$-$o$ & \textbf{0.6213} 	&	\textbf{0.7250 	$\pm$	0.1276} 	&	0.7948 	&	0.8952 	$\pm$	0.0333 	&	0.7731 	&	0.8818 	$\pm$	0.0374 	&	0.6738 	&	0.9006 	$\pm$	0.0571 	&	\textbf{0.6222} 	&	0.9077 	$\pm$	0.0724 	&	0.8957 	&	0.9657 	$\pm$	0.0305 \\
			Armadillo-c & $J_1$-$o$ & \textbf{0.2343} 	&	\textbf{0.2679 	$\pm$	0.1427} 	&	0.5117 	&	0.6277 	$\pm$	0.1034 	&	0.3276 	&	0.5215 	$\pm$	0.1050 	&	\textbf{0.2346} 	&	0.3026 	$\pm$	0.2023 	&	0.8689 	&	0.9121 	$\pm$	0.0229 	&	0.9342 	&	0.9828 	$\pm$	0.0164 \\
			Armadillo-c & $J_2$-$o$ & \textbf{0.4906} 	&	\textbf{0.5083 	$\pm$	0.0732} 	&	0.6187 	&	0.7118 	$\pm$	0.0672 	&	0.6228 	&	0.6967 	$\pm$	0.0441 	&	\textbf{0.4906} 	&	0.5162 	$\pm$	0.0842 	&	0.8712 	&	0.9270 	$\pm$	0.0259 	&	0.9092 	&	0.9765 	$\pm$	0.0233 \\
			Armadillo-c & $J_3$-$o$ & \textbf{0.6405} 	&	\textbf{0.6518 	$\pm$	0.0443} 	&	0.6674 	&	0.7452 	$\pm$	0.0505 	&	0.6813 	&	0.7426 	$\pm$	0.0386 	&	\textbf{0.6405} 	&	\textbf{0.6564 	$\pm$	0.0439} 	&	0.6532 	&	0.8846 	$\pm$	0.0966 	&	0.6836 	&	0.9593 	$\pm$	0.0703 \\
			Buddha-a & $J_1$-$o$ & \textbf{0.1837} 	&	\textbf{0.1846 	$\pm$	0.0005} 	&	0.2102 	&	0.4527 	$\pm$	0.1316 	&	0.2692 	&	0.4173 	$\pm$	0.0944 	&	\textbf{0.1835} 	&	0.3134 	$\pm$	0.2575 	&	0.1856 	&	0.7977 	$\pm$	0.2618 	&	0.1868 	&	0.6681 	$\pm$	0.3165 \\
			Buddha-a & $J_2$-$o$ & \textbf{0.2883} 	&	\textbf{0.2892 	$\pm$	0.0007} 	&	0.3317 	&	0.5048 	$\pm$	0.1375 	&	0.3037 	&	0.4886 	$\pm$	0.1021 	&	\textbf{0.2885} 	&	0.2970 	$\pm$	0.0173 	&	0.8708 	&	0.9248 	$\pm$	0.0302 	&	0.4203 	&	0.9028 	$\pm$	0.1586 \\
			Buddha-a & $J_3$-$o$ & \textbf{0.4229} 	&	\textbf{0.4239 	$\pm$	0.0020} 	&	0.4532 	&	0.6445 	$\pm$	0.1181 	&	0.4565 	&	0.6778 	$\pm$	0.1163 	&	\textbf{0.4230} 	&	0.4394 	$\pm$	0.0357 	&	\textbf{0.4253} 	&	0.9024 	$\pm$	0.1131 	&	0.4738 	&	0.9089 	$\pm$	0.1090 \\
			Buddha-b & $J_1$-$o$ & \textbf{0.6113} 	&	\textbf{0.6120 	$\pm$	0.0005} 	&	0.6989 	&	0.8321 	$\pm$	0.0685 	&	0.6780 	&	0.8188 	$\pm$	0.0740 	&	\textbf{0.6117} 	&	0.6967 	$\pm$	0.1146 	&	0.8585 	&	0.9320 	$\pm$	0.0284 	&	0.9411 	&	0.9827 	$\pm$	0.0129 \\
			Buddha-b & $J_2$-$o$ & \textbf{0.3909} 	&	\textbf{0.4165 	$\pm$	0.1086} 	&	0.6052 	&	0.7606 	$\pm$	0.0863 	&	0.4710 	&	0.7565 	$\pm$	0.1207 	&	\textbf{0.3910} 	&	0.6226 	$\pm$	0.2640 	&	\textbf{0.3917} 	&	0.9208 	$\pm$	0.1234 	&	0.9484 	&	0.9808 	$\pm$	0.0126 \\
			Buddha-b & $J_3$-$o$ & \textbf{0.3652} 	&	\textbf{0.3894 	$\pm$	0.1022} 	&	0.4910 	&	0.6844 	$\pm$	0.1093 	&	0.4230 	&	0.6909 	$\pm$	0.1239 	&	\textbf{0.3653} 	&	0.4565 	$\pm$	0.1717 	&	\textbf{0.3668} 	&	0.9008 	$\pm$	0.1253 	&	0.4152 	&	0.9159 	$\pm$	0.1683 \\
			Buddha-c & $J_1$-$o$ & \textbf{0.4231} 	&	\textbf{0.4456 	$\pm$	0.0925} 	&	0.4791 	&	0.7084 	$\pm$	0.0940 	&	0.5561 	&	0.6808 	$\pm$	0.0891 	&	\textbf{0.4234} 	&	0.4611 	$\pm$	0.0997 	&	0.8763 	&	0.9355 	$\pm$	0.0259 	&	0.9024 	&	0.9728 	$\pm$	0.0225 \\
			Buddha-c & $J_2$-$o$ & \textbf{0.7192} 	&	\textbf{0.7290 	$\pm$	0.0365} 	&	0.8166 	&	0.8808 	$\pm$	0.0343 	&	0.7776 	&	0.8582 	$\pm$	0.0402 	&	\textbf{0.7191} 	&	0.7472 	$\pm$	0.0563 	&	0.9055 	&	0.9506 	$\pm$	0.0184 	&	0.9593 	&	0.9856 	$\pm$	0.0094 \\
			Buddha-c & $J_3$-$o$ & \textbf{0.5189} 	&	\textbf{0.5553 	$\pm$	0.1075} 	&	0.5729 	&	0.8413 	$\pm$	0.0889 	&	0.6917 	&	0.8517 	$\pm$	0.0745 	&	\textbf{0.5187} 	&	0.7321 	$\pm$	0.1904 	&	0.9131 	&	0.9505 	$\pm$	0.0149 	&	0.9162 	&	0.9787 	$\pm$	0.0178 \\
			Buste-a & $J_1$-$o$ & \textbf{0.3063} 	&	\textbf{0.3801 	$\pm$	0.1651} 	&	0.4454 	&	0.6474 	$\pm$	0.0901 	&	0.3859 	&	0.5981 	$\pm$	0.1431 	&	\textbf{0.3056} 	&	0.3996 	$\pm$	0.1199 	&	\textbf{0.3066} 	&	0.7227 	$\pm$	0.2701 	&	0.3559 	&	0.8635 	$\pm$	0.1460 \\
			Buste-a & $J_2$-$o$ & \textbf{0.5416} 	&	\textbf{0.6239 	$\pm$	0.1326} 	&	0.6511 	&	0.8543 	$\pm$	0.0865 	&	0.6162 	&	0.8580 	$\pm$	0.0985 	&	\textbf{0.5438} 	&	0.7507 	$\pm$	0.1609 	&	0.7933 	&	0.8950 	$\pm$	0.0543 	&	0.8659 	&	0.9308 	$\pm$	0.0326 \\
			Buste-a & $J_3$-$o$ & \textbf{0.4959} 	&	\textbf{0.5381 	$\pm$	0.0948} 	&	0.5723 	&	0.7961 	$\pm$	0.0915 	&	0.6253 	&	0.7687 	$\pm$	0.0840 	&	\textbf{0.4975} 	&	0.6985 	$\pm$	0.1595 	&	0.5452 	&	0.8652 	$\pm$	0.0908 	&	0.8109 	&	0.9091 	$\pm$	0.0551 \\
			Buste-b & $J_1$-$o$ & \textbf{0.6053} 	&	\textbf{0.6421 	$\pm$	0.0679} 	&	0.7548 	&	0.7955 	$\pm$	0.0295 	&	0.7103 	&	0.7831 	$\pm$	0.0299 	&	0.7240 	&	0.7476 	$\pm$	0.0350 	&	0.6605 	&	0.8276 	$\pm$	0.0848 	&	0.6648 	&	0.8865 	$\pm$	0.0767 \\
			Buste-b & $J_2$-$o$ & \textbf{0.3377} 	&	\textbf{0.3696 	$\pm$	0.1042} 	&	0.4136 	&	0.5947 	$\pm$	0.1053 	&	0.3964 	&	0.5635 	$\pm$	0.1036 	&	0.3858 	&	0.4470 	$\pm$	0.1418 	&	0.3445 	&	0.7602 	$\pm$	0.2100 	&	0.8020 	&	0.9219 	$\pm$	0.0563 \\
			Buste-b & $J_3$-$o$ & \textbf{0.2972} 	&	\textbf{0.3530 	$\pm$	0.1411} 	&	0.3585 	&	0.5460 	$\pm$	0.1079 	&	0.3792 	&	0.5140 	$\pm$	0.1006 	&	0.3455 	&	0.4042 	$\pm$	0.1406 	&	0.3073 	&	0.8125 	$\pm$	0.1507 	&	0.3481 	&	0.7318 	$\pm$	0.2296 \\
			Buste-c & $J_1$-$o$ & \textbf{0.3715} 	&	\textbf{0.4254 	$\pm$	0.1045} 	&	0.4379 	&	0.5938 	$\pm$	0.0968 	&	0.4478 	&	0.6420 	$\pm$	0.1057 	&	0.4206 	&	0.5071 	$\pm$	0.1328 	&	0.4120 	&	0.7735 	$\pm$	0.1716 	&	0.4218 	&	0.7709 	$\pm$	0.1849 \\
			Buste-c & $J_2$-$o$ & \textbf{0.6437} 	&	\textbf{0.6823 	$\pm$	0.0477} 	&	0.7203 	&	0.7852 	$\pm$	0.0207 	&	0.7608 	&	0.7986 	$\pm$	0.0247 	&	0.7449 	&	0.7497 	$\pm$	0.0089 	&	0.6712 	&	0.8172 	$\pm$	0.0793 	&	0.7455 	&	0.8726 	$\pm$	0.0593 \\
			Buste-c & $J_3$-$o$ & \textbf{0.3397} 	&	\textbf{0.4044 	$\pm$	0.1245} 	&	0.4581 	&	0.5925 	$\pm$	0.0786 	&	0.5280 	&	0.6696 	$\pm$	0.0740 	&	0.3872 	&	0.4643 	$\pm$	0.1540 	&	0.3510 	&	0.7490 	$\pm$	0.1998 	&	0.3876 	&	0.8260 	$\pm$	0.1856 \\
			Room-a & $J_1$-$o$ & \textbf{0.2827} 	&	0.3030 	$\pm$	0.0844 	&	0.2952 	&	0.3804 	$\pm$	0.0928 	&	0.3279 	&	0.4050 	$\pm$	0.0670 	&	\textbf{0.2831} 	&	\textbf{0.2868 	$\pm$	0.0079} 	&	\textbf{0.2832} 	&	0.4237 	$\pm$	0.1919 	&	\textbf{0.2829} 	&	0.4228 	$\pm$	0.1896 \\
			Room-a & $J_2$-$o$ & \textbf{0.3701} 	&	\textbf{0.3840 	$\pm$	0.0559} 	&	0.3878 	&	0.4987 	$\pm$	0.0767 	&	0.3864 	&	0.4673 	$\pm$	0.0631 	&	\textbf{0.3697} 	&	\textbf{0.3857 	$\pm$	0.0643} 	&	\textbf{0.3701} 	&	0.4621 	$\pm$	0.1392 	&	\textbf{0.3699} 	&	0.5250 	$\pm$	0.1655 \\
			Room-a & $J_3$-$o$ & \textbf{0.2668} 	&	\textbf{0.2681 	$\pm$	0.0023} 	&	0.2855 	&	0.3773 	$\pm$	0.0704 	&	0.2856 	&	0.3532 	$\pm$	0.0463 	&	\textbf{0.2669} 	&	\textbf{0.2700 	$\pm$	0.0084} 	&	\textbf{0.2667} 	&	0.3971 	$\pm$	0.1999 	&	\textbf{0.2672} 	&	0.3543 	$\pm$	0.1739 \\
			Room-b & $J_1$-$o$ & \textbf{0.5068} 	&	\textbf{0.6893 	$\pm$	0.1251} 	&	0.6212 	&	0.7557 	$\pm$	0.0668 	&	0.6972 	&	0.7822 	$\pm$	0.0391 	&	\textbf{0.5059} 	&	0.7684 	$\pm$	0.0890 	&	\textbf{0.5071} 	&	0.7963 	$\pm$	0.1115 	&	\textbf{0.5106} 	&	0.7632 	$\pm$	0.1348 \\
			Room-b & $J_2$-$o$ & \textbf{0.3643} 	&	0.5068 	$\pm$	0.1901 	&	0.4011 	&	0.5551 	$\pm$	0.1044 	&	0.4091 	&	0.5280 	$\pm$	0.0951 	&	\textbf{0.3648} 	&	\textbf{0.4179 	$\pm$	0.0914} 	&	\textbf{0.3647} 	&	0.7019 	$\pm$	0.2113 	&	\textbf{0.3642} 	&	0.6898 	$\pm$	0.2247 \\
			Room-b & $J_3$-$o$ & \textbf{0.5054} 	&	\textbf{0.6403 	$\pm$	0.1258} 	&	0.5817 	&	0.7098 	$\pm$	0.0718 	&	0.5270 	&	0.7237 	$\pm$	0.0873 	&	\textbf{0.5053} 	&	0.7020 	$\pm$	0.1327 	&	\textbf{0.5059} 	&	0.8201 	$\pm$	0.0966 	&	\textbf{0.5068} 	&	0.7722 	$\pm$	0.1451 \\
			Room-c & $J_1$-$o$ & \textbf{0.3985} 	&	\textbf{0.4311 	$\pm$	0.0814} 	&	0.4258 	&	0.5095 	$\pm$	0.0659 	&	0.4438 	&	0.4872 	$\pm$	0.0360 	&	\textbf{0.4013} 	&	0.4738 	$\pm$	0.1253 	&	\textbf{0.3988} 	&	0.6885 	$\pm$	0.1523 	&	\textbf{0.3995} 	&	0.6677 	$\pm$	0.1941 \\
			Room-c & $J_2$-$o$ & \textbf{0.5944} 	&	\textbf{0.6386 	$\pm$	0.0303} 	&	0.6660 	&	0.7223 	$\pm$	0.0437 	&	0.6446 	&	0.7210 	$\pm$	0.0437 	&	0.6221 	&	0.7065 	$\pm$	0.0799 	&	\textbf{0.5926} 	&	0.7400 	$\pm$	0.0878 	&	0.6158 	&	0.7397 	$\pm$	0.0830 \\
			Room-c & $J_3$-$o$ & \textbf{0.5466} 	&	\textbf{0.5835 	$\pm$	0.0426} 	&	0.5821 	&	0.6729 	$\pm$	0.0406 	&	0.5820 	&	0.6508 	$\pm$	0.0516 	&	0.5592 	&	0.6308 	$\pm$	0.0673 	&	\textbf{0.5458} 	&	0.6620 	$\pm$	0.1054 	&	0.5806 	&	0.6948 	$\pm$	0.0874 \\
			Room-d & $J_1$-$o$ & \textbf{0.2599} 	&	\textbf{0.2632 	$\pm$	0.0041} 	&	0.3262 	&	0.4463 	$\pm$	0.0649 	&	0.2988 	&	0.4311 	$\pm$	0.1055 	&	\textbf{0.2608} 	&	0.2908 	$\pm$	0.0226 	&	\textbf{0.2610} 	&	0.6654 	$\pm$	0.2096 	&	0.2974 	&	0.6079 	$\pm$	0.2428 \\
			Room-d & $J_2$-$o$ & \textbf{0.5784} 	&	\textbf{0.6517 	$\pm$	0.0634} 	&	0.6833 	&	0.7421 	$\pm$	0.0220 	&	0.6176 	&	0.7289 	$\pm$	0.0432 	&	\textbf{0.5789} 	&	0.7064 	$\pm$	0.0671 	&	\textbf{0.5790} 	&	0.7660 	$\pm$	0.0869 	&	0.6284 	&	0.7845 	$\pm$	0.0776 \\
			Room-d & $J_3$-$o$ & \textbf{0.6090} 	&	\textbf{0.6941 	$\pm$	0.0769} 	&	0.7038 	&	0.7853 	$\pm$	0.0412 	&	0.6800 	&	0.7694 	$\pm$	0.0437 	&	\textbf{0.6113} 	&	0.7254 	$\pm$	0.0698 	&	0.6153 	&	0.8021 	$\pm$	0.0888 	&	0.6655 	&	0.8197 	$\pm$	0.0965 \\
			Office-a & $J_1$-$o$ & \textbf{0.3565} 	&	\textbf{0.3585 	$\pm$	0.0028} 	&	0.3729 	&	0.4279 	$\pm$	0.0531 	&	0.3722 	&	0.4290 	$\pm$	0.0544 	&	\textbf{0.3563} 	&	\textbf{0.3588 	$\pm$	0.0044} 	&	\textbf{0.3569} 	&	0.5373 	$\pm$	0.1917 	&	\textbf{0.3571} 	&	0.5306 	$\pm$	0.1808 \\
			Office-a & $J_2$-$o$ & \textbf{0.1428} 	&	\textbf{0.1454 	$\pm$	0.0021} 	&	0.1682 	&	0.2452 	$\pm$	0.0707 	&	0.1601 	&	0.2463 	$\pm$	0.0661 	&	\textbf{0.1428} 	&	0.1506 	$\pm$	0.0096 	&	\textbf{0.1430} 	&	0.3210 	$\pm$	0.2707 	&	\textbf{0.1436} 	&	0.2315 	$\pm$	0.1739 \\
			Office-a & $J_3$-$o$ & \textbf{0.3637} 	&	\textbf{0.3648 	$\pm$	0.0010} 	&	0.3738 	&	0.4290 	$\pm$	0.0442 	&	0.3741 	&	0.4423 	$\pm$	0.0490 	&	\textbf{0.3639} 	&	0.3732 	$\pm$	0.0367 	&	\textbf{0.3638} 	&	0.5258 	$\pm$	0.1896 	&	\textbf{0.3635} 	&	0.5328 	$\pm$	0.1459 \\
			Office-b & $J_1$-$o$ & \textbf{0.3323} 	&	\textbf{0.3337 	$\pm$	0.0017} 	&	0.3389 	&	0.4451 	$\pm$	0.0795 	&	0.3427 	&	0.4400 	$\pm$	0.0771 	&	\textbf{0.3326} 	&	0.3383 	$\pm$	0.0151 	&	\textbf{0.3325} 	&	0.5184 	$\pm$	0.2094 	&	\textbf{0.3325} 	&	0.5043 	$\pm$	0.2019 \\
			Office-b & $J_2$-$o$ & \textbf{0.2953} 	&	0.3163 	$\pm$	0.0864 	&	0.3080 	&	0.3742 	$\pm$	0.0591 	&	0.3072 	&	0.3683 	$\pm$	0.0463 	&	\textbf{0.2955} 	&	\textbf{0.3024 	$\pm$	0.0149} 	&	\textbf{0.2953} 	&	0.5132 	$\pm$	0.2203 	&	\textbf{0.2957} 	&	0.4238 	$\pm$	0.1921 \\
			Office-b & $J_3$-$o$ & \textbf{0.4177} 	&	\textbf{0.4332 	$\pm$	0.0618} 	&	0.4508 	&	0.5343 	$\pm$	0.0653 	&	0.4280 	&	0.5295 	$\pm$	0.0663 	&	\textbf{0.4180} 	&	0.4494 	$\pm$	0.0908 	&	\textbf{0.4178} 	&	0.6004 	$\pm$	0.1491 	&	\textbf{0.4183} 	&	0.5695 	$\pm$	0.1532 \\
			Office-c & $J_1$-$o$ & \textbf{0.2711} 	&	\textbf{0.2731 	$\pm$	0.0028} 	&	0.2908 	&	0.3587 	$\pm$	0.0564 	&	0.2838 	&	0.3235 	$\pm$	0.0339 	&	\textbf{0.2715} 	&	\textbf{0.2734 	$\pm$	0.0023} 	&	\textbf{0.2713} 	&	0.4399 	$\pm$	0.2045 	&	\textbf{0.2717} 	&	0.4291 	$\pm$	0.2068 \\
			Office-c & $J_2$-$o$ & \textbf{0.2358} 	&	\textbf{0.2364 	$\pm$	0.0005} 	&	0.2459 	&	0.3086 	$\pm$	0.0472 	&	0.2528 	&	0.3182 	$\pm$	0.0570 	&	\textbf{0.2357} 	&	\textbf{0.2369 	$\pm$	0.0011} 	&	\textbf{0.2357} 	&	0.3269 	$\pm$	0.1802 	&	\textbf{0.2362} 	&	0.3913 	$\pm$	0.2032 \\
			Office-c & $J_3$-$o$ & \textbf{0.3644} 	&	\textbf{0.3652 	$\pm$	0.0005} 	&	0.3813 	&	0.4507 	$\pm$	0.0449 	&	0.3785 	&	0.4506 	$\pm$	0.0514 	&	\textbf{0.3643} 	&	\textbf{0.3662 	$\pm$	0.0030} 	&	\textbf{0.3647} 	&	0.5263 	$\pm$	0.1792 	&	\textbf{0.3650} 	&	0.5355 	$\pm$	0.1827 \\
			Office-d & $J_1$-$o$ & \textbf{0.3177} 	&	\textbf{0.3397 	$\pm$	0.0684} 	&	0.3360 	&	0.4263 	$\pm$	0.0874 	&	0.3570 	&	0.4444 	$\pm$	0.0765 	&	\textbf{0.3206} 	&	0.3582 	$\pm$	0.0163 	&	\textbf{0.3190} 	&	0.5743 	$\pm$	0.1763 	&	0.3659 	&	0.5356 	$\pm$	0.1455 \\
			Office-d & $J_2$-$o$ & \textbf{0.3871} 	&	\textbf{0.4152 	$\pm$	0.0303} 	&	0.4393 	&	0.5169 	$\pm$	0.0558 	&	0.4168 	&	0.5299 	$\pm$	0.0559 	&	0.3942 	&	0.4517 	$\pm$	0.0571 	&	\textbf{0.3903} 	&	0.5488 	$\pm$	0.1393 	&	\textbf{0.3908} 	&	0.5569 	$\pm$	0.1263 \\
			Office-d & $J_3$-$o$ & \textbf{0.6053} 	&	\textbf{0.6196 	$\pm$	0.0232} 	&	0.6486 	&	0.7252 	$\pm$	0.0314 	&	0.6209 	&	0.7158 	$\pm$	0.0372 	&	\textbf{0.6061} 	&	0.6768 	$\pm$	0.0363 	&	\textbf{0.6074} 	&	0.6906 	$\pm$	0.0652 	&	0.6463 	&	0.7418 	$\pm$	0.0388 \\
			\hline
			No. of Best & - & 69 & 65 & 0 & 0 & 0 & 0 & 57 & 12 & 38 & 0 & 17 & 0 \\
			\hline
		\end{tabular}
	}
\end{table*}

\subsection{Solution Quality}
To evaluate the solution quality of the proposed approach, Table~\ref{tabSolutionQuality} tabulates all the results obtained by MTPCR, MFEA, GMFEA, MTPCR-Intra, MTPCR-Inter, and MTPCR-NKS over 20 independent runs. In the table, solutions of the original tasks are presented, and the column ``Task No.'' represents each of the three original tasks for each problem. The column ``B.Cost'' and ``Avg.Cost'' show the best fitness values found in 20 independent runs and fitness values averaged over 20 independent runs, respectively. Superior solutions of each task are highlighted in bold font, and the last row `No. of Best' counts the number of superior solutions produced by each of these methods.

It can be observed from Table~\ref{tabSolutionQuality} that in terms of \textit{B.Cost}, MTPCR can always obtain solutions with the best quality on all the 69 tasks. MTPCR-Intra provides comparative best solutions on 57 tasks out of the 69 tasks, while MTPCR-Inter finds best solutions on 38 tasks, which is a little bit more than a half of the tasks. With no knowledge sharing incorporated, MTPCR-NKS can only finds best solutions on 17 of the tasks. By contrast, MFEA and GMFEA achieves best quality solutions on none of the tasks, although their solutions are near to the best on some tasks. For example, on problem ``Office-a'', ``Office-b'' and ``Office-c'', best solutions of MFEA and GMFEA are close to high quality solutions highlighted in bold font. The inability of MFEA and GMFEA for finding high quality solutions mainly come from their inability of local search, in that they can get close to high quality solutions but never reach the exact position where high quality solutions are. Due to the randomness of crossovers between individuals in the homogenous population of MFEA and GMFEA, the optimization process of one task may be overly influenced by genetic materials from other tasks, as registration between different point clouds has unique landscapes. Since independent populations are maintained in MTPCR-NKS, population attached to each component task guarantees the search process towards the direction of decreased function values. Although it suffers the risk of getting stuck to local optima, it has the chance of finding global optimum, and this confirms the effectiveness of conducting unique population for each task. Qualities of solutions found by MTPCR-Inter are better than that of MTPCR-NKS, and the improvement benefits from inter-task knowledge sharing. However, without the help of intra-task knowledge sharing, the potential of inter-task sharing can not be fully explored, resulting in a limited improvement. As can be seen in Table~\ref{tabSolutionQuality}, MTPCR-Intra is able to achieve high quality solutions on most of the tasks, validating the boon of constructing aiding tasks as knowledge source. With the combination of intra-task learning and inter-task learning, performance of MTPCR is further enhanced.

On the other hand, in terms of \textit{Avg.Cost} which reflects the stableness of the optimization methods over 20 independent runs, it can be seen from Table~\ref{tabSolutionQuality} that MTPCR takes the lead without doubt. The only four tasks that MTPCR-Intra exceeds MTPCR are ``Armadillo-b J2-o'', ``Room-a J1-o'', ``Room-b J2-o'' and ``Office-b J2-0'', with MTPCR follows closely on two out of the four. Solution quality of MFEA, GMFEA, MTPCR-Inter and MTPCR-NKS are far from satisfactory. The main reason that MFEA and GMFEA perform poorly is similar as before, in that knowledge sharing happens through crossovers between individuals possessing different skill factors introduced much randomness in evolutionary search, and negative knowledge transfer will cancel out the benefits brought by positive transfer. This again confirms the boon of conducting explicit multitasking methods. MTPCR-Intra achieves the second best performance, while MTPCR-Inter performs poorly, this once more confirms the usefulness of knowledge transferred from aiding tasks. However, only intra-task knowledge sharing cannot avoid local optima, thus the average performance of MTPCR-Intra is not good enough. Although there is no significant improvement of MTPCR-Inter compared with MTPCR-NKS, inter-task knowledge sharing is very helpful to avoid local optima when combined with intra-task knowledge sharing. Therefore, the performance of MTPCR with bi-channel knowledge sharing benefits from both intra-task sharing and inter-task sharing greatly. In summary, neither intra-task knowledge sharing nor inter-task knowledge sharing is sufficient for the solving of point cloud registration, solution qualities in Table~\ref{tabSolutionQuality} verify the effectiveness of proposed bi-channel knowledge sharing implemented in an explicit manner.

\subsection{Search Efficiency}
\begin{figure*}[th]
	\centering
	\subfloat[]{\label{figConvergenceA1}
		\begin{minipage}{0.46\columnwidth}
			\includegraphics[width=\columnwidth]{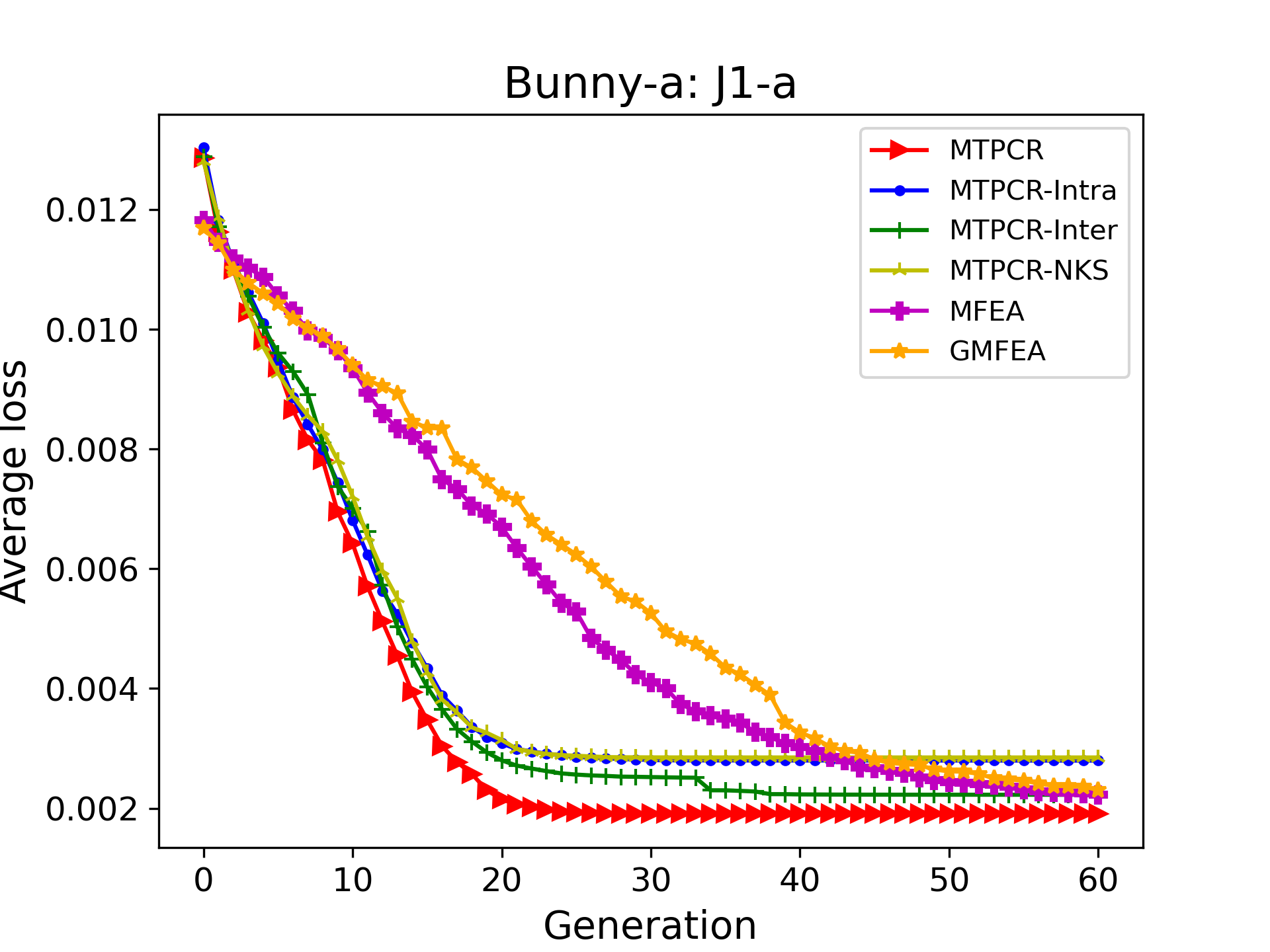}\vspace{1ex}
			\includegraphics[width=\columnwidth]{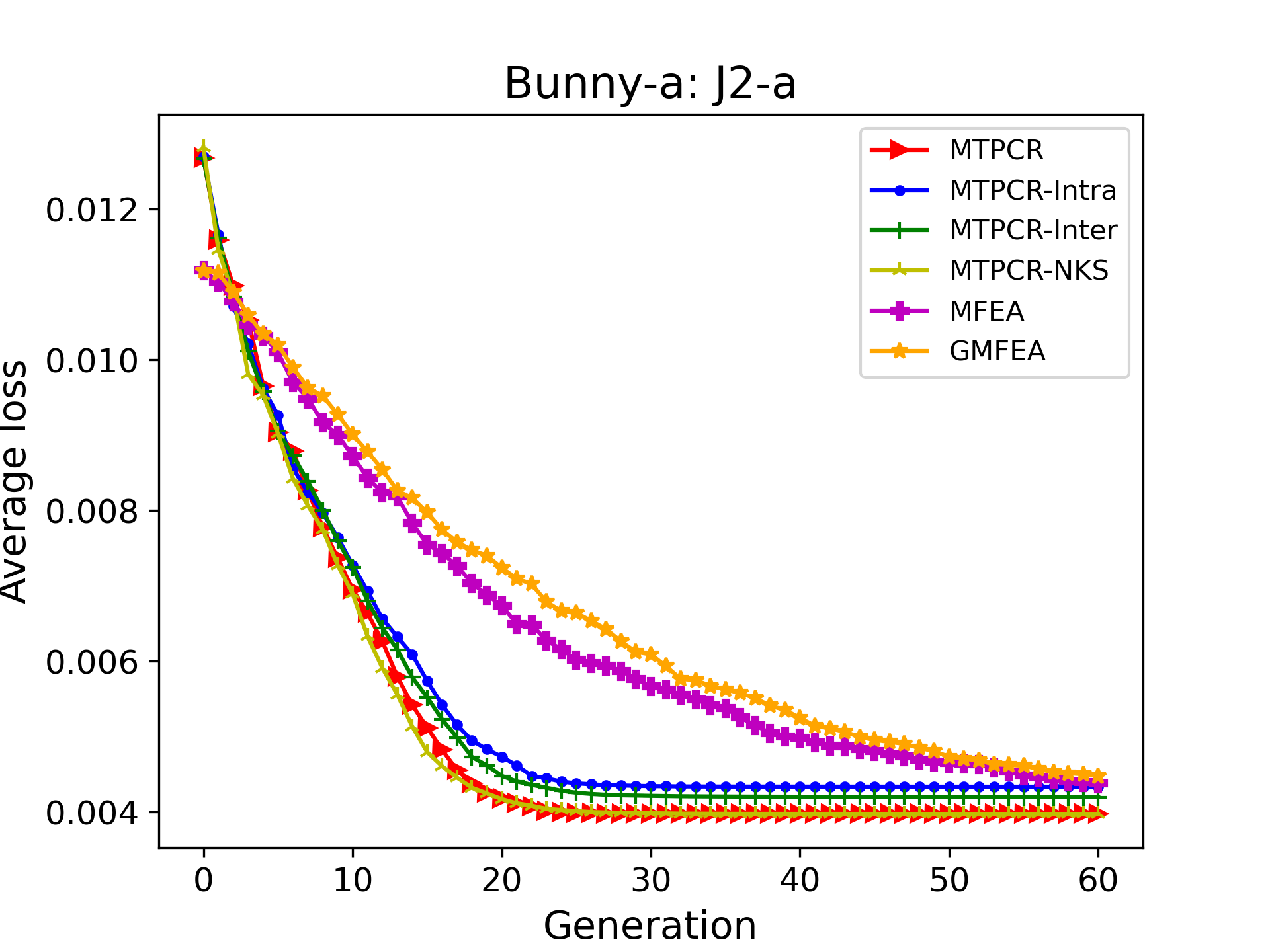}\vspace{1ex}
			\includegraphics[width=\columnwidth]{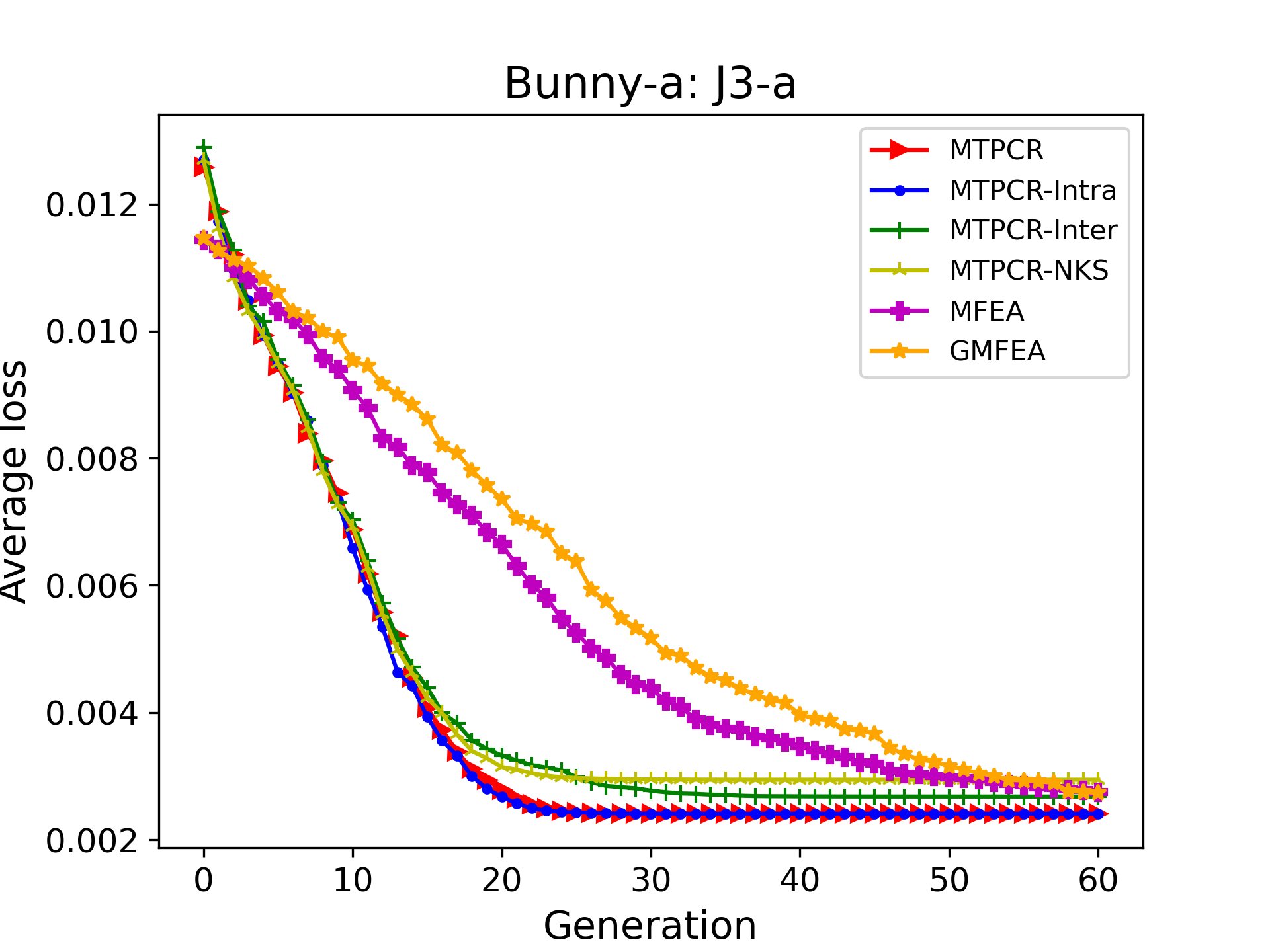}
		\end{minipage}
		\begin{minipage}{0.46\columnwidth}
			\includegraphics[width=\columnwidth]{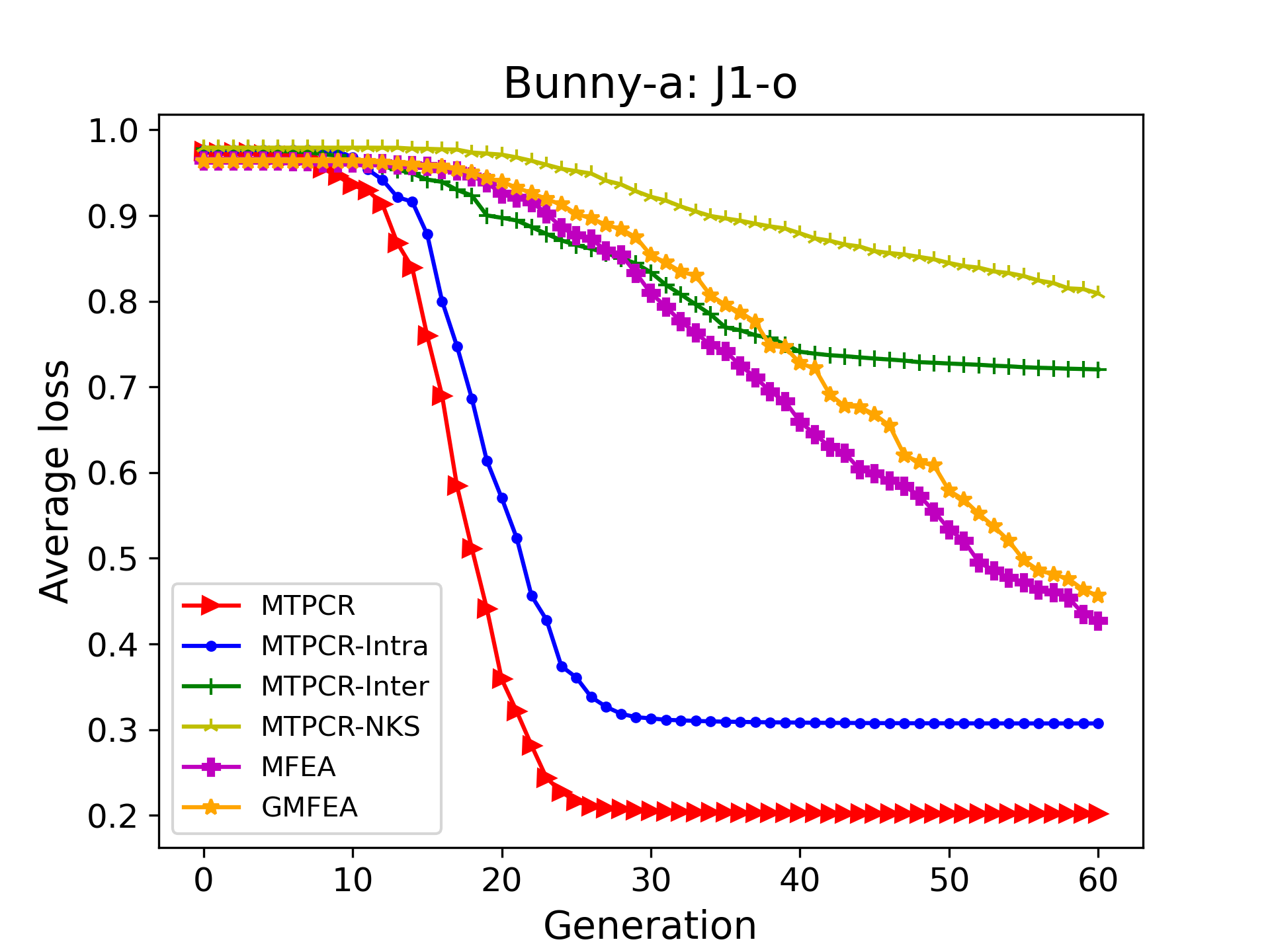}\vspace{1ex}
			\includegraphics[width=\columnwidth]{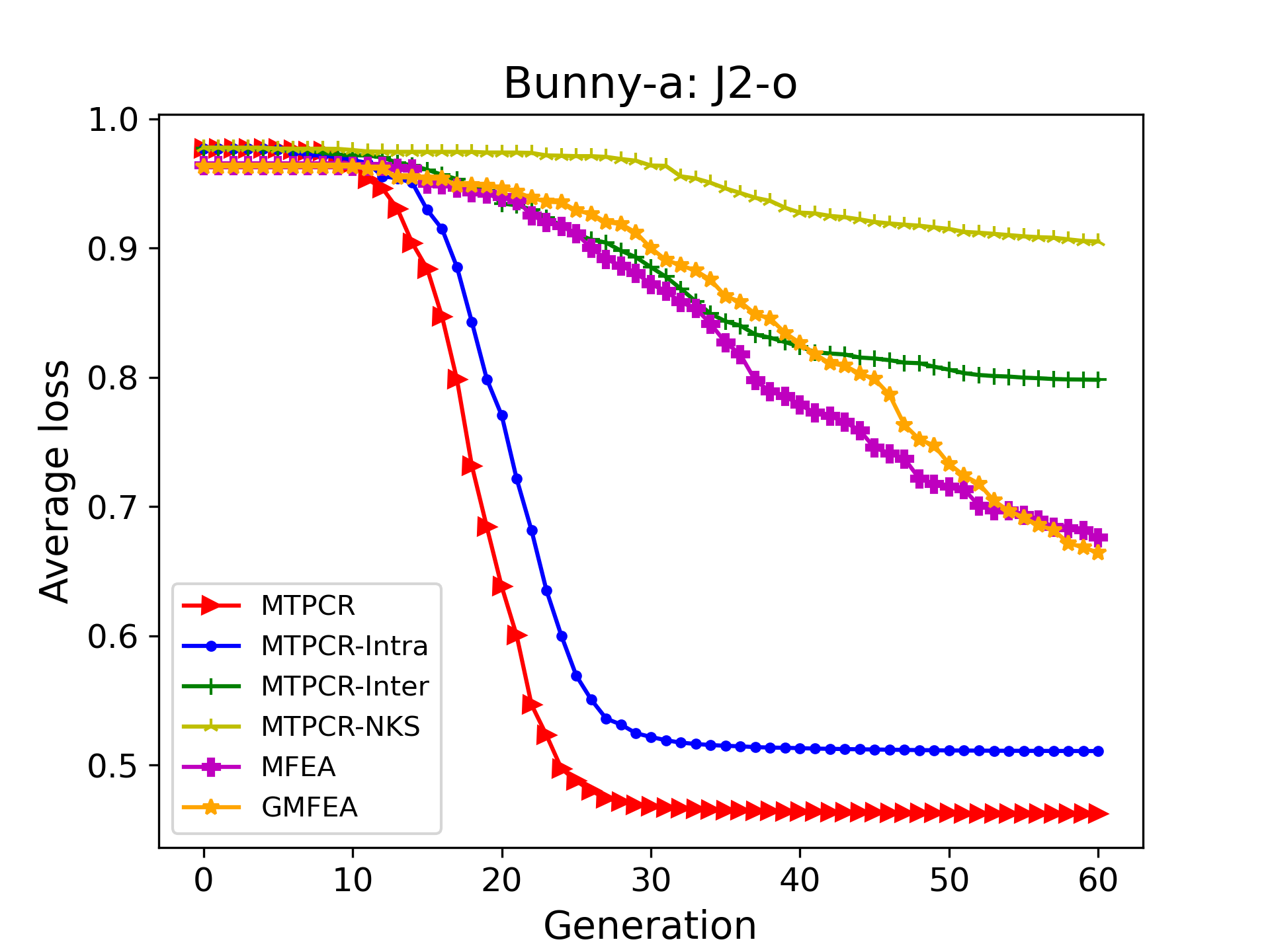}\vspace{1ex}
			\includegraphics[width=\columnwidth]{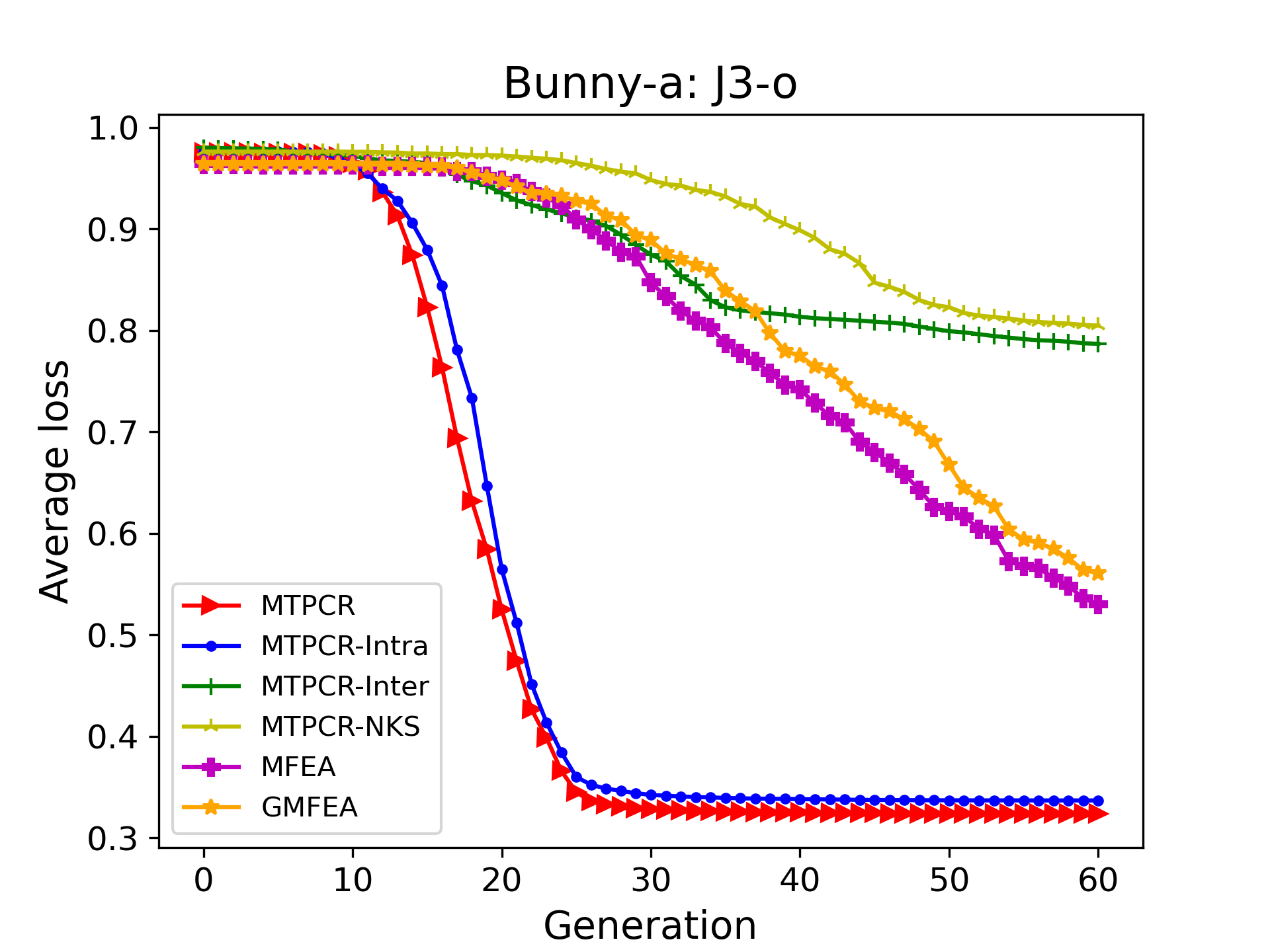}
		\end{minipage}
	}
	\subfloat[]{\label{figConvergenceA2}
		\begin{minipage}{0.46\columnwidth}
			\includegraphics[width=\columnwidth]{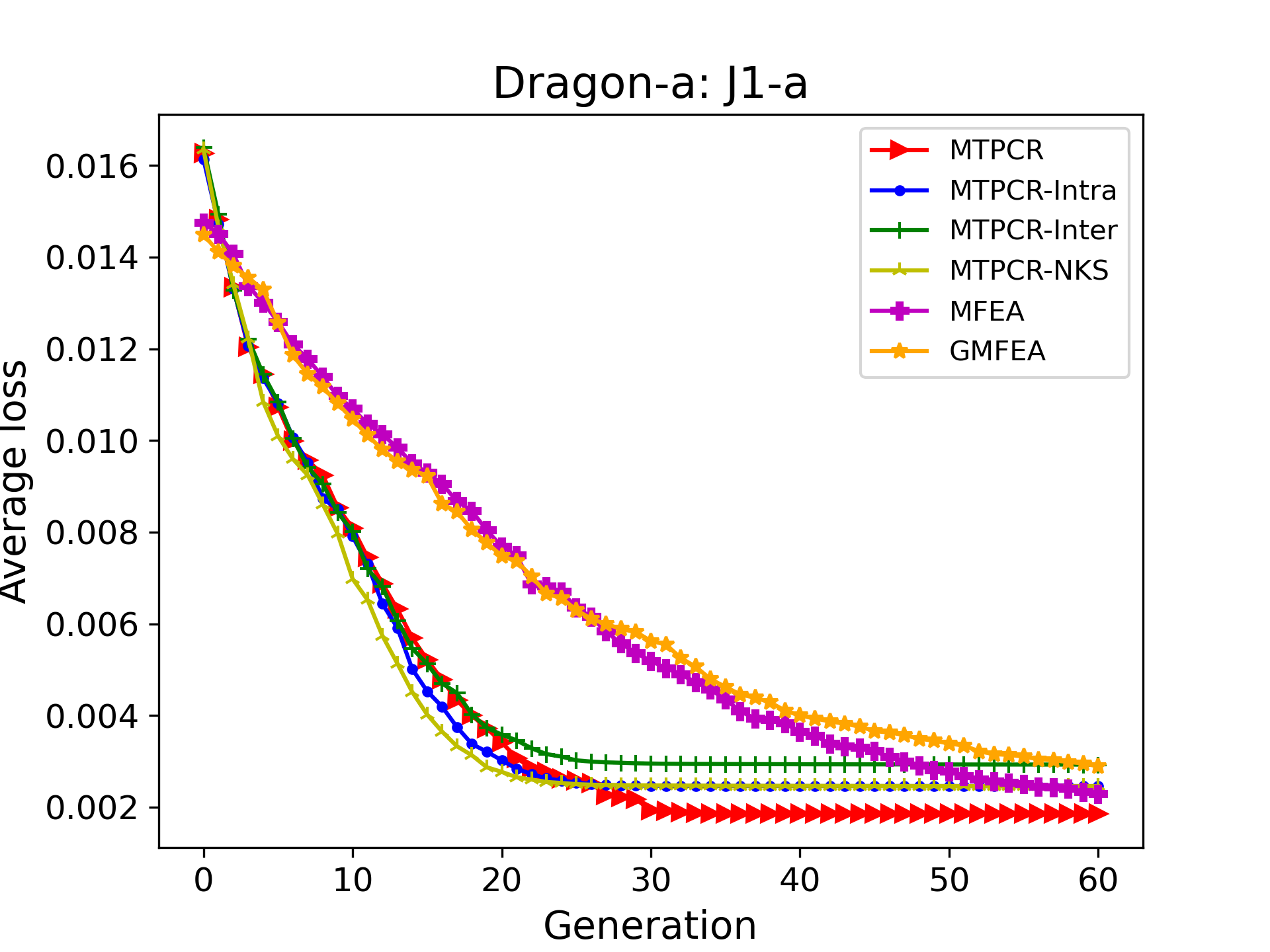}\vspace{1ex}
			\includegraphics[width=\columnwidth]{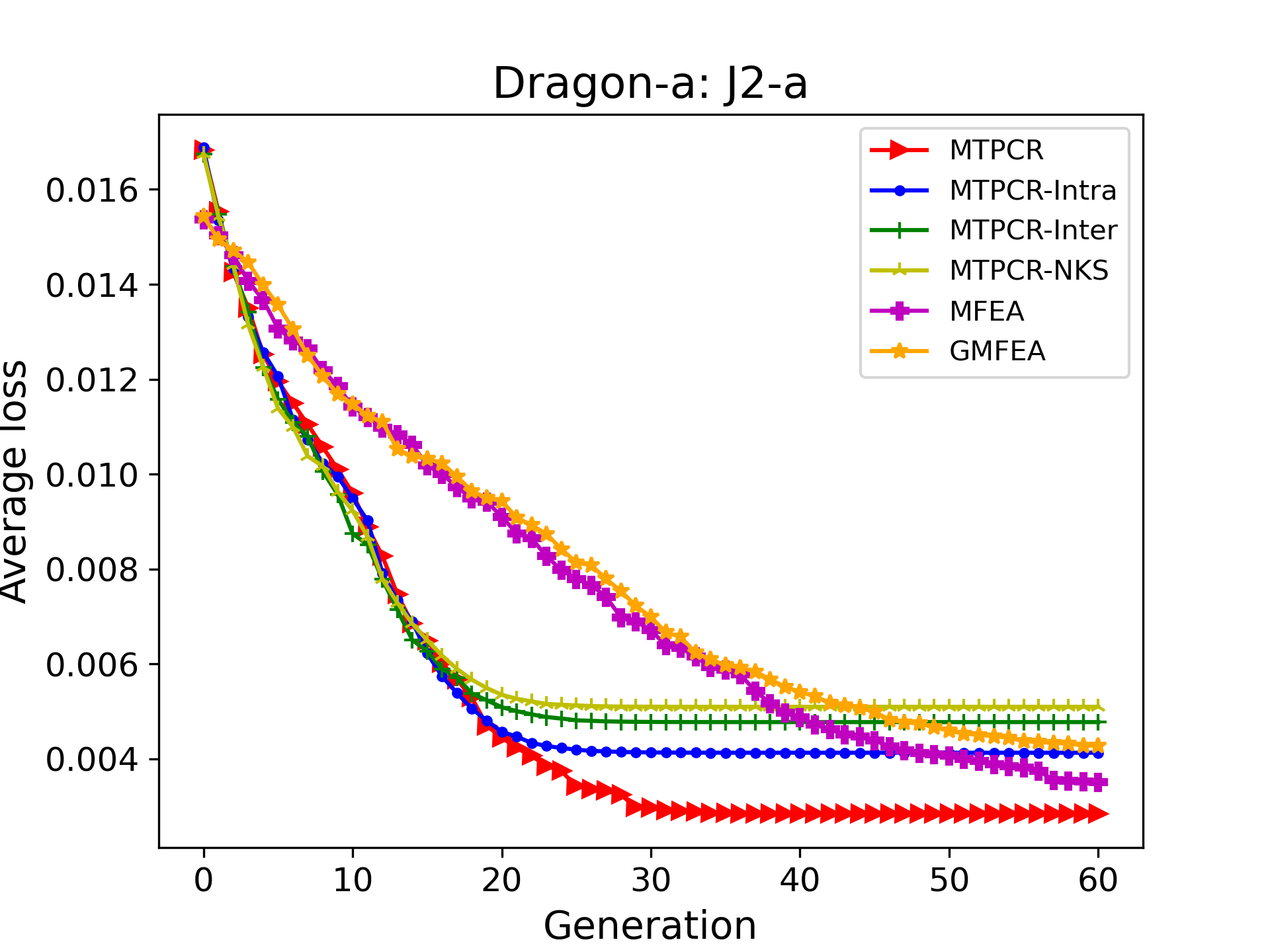}\vspace{1ex}
			\includegraphics[width=\columnwidth]{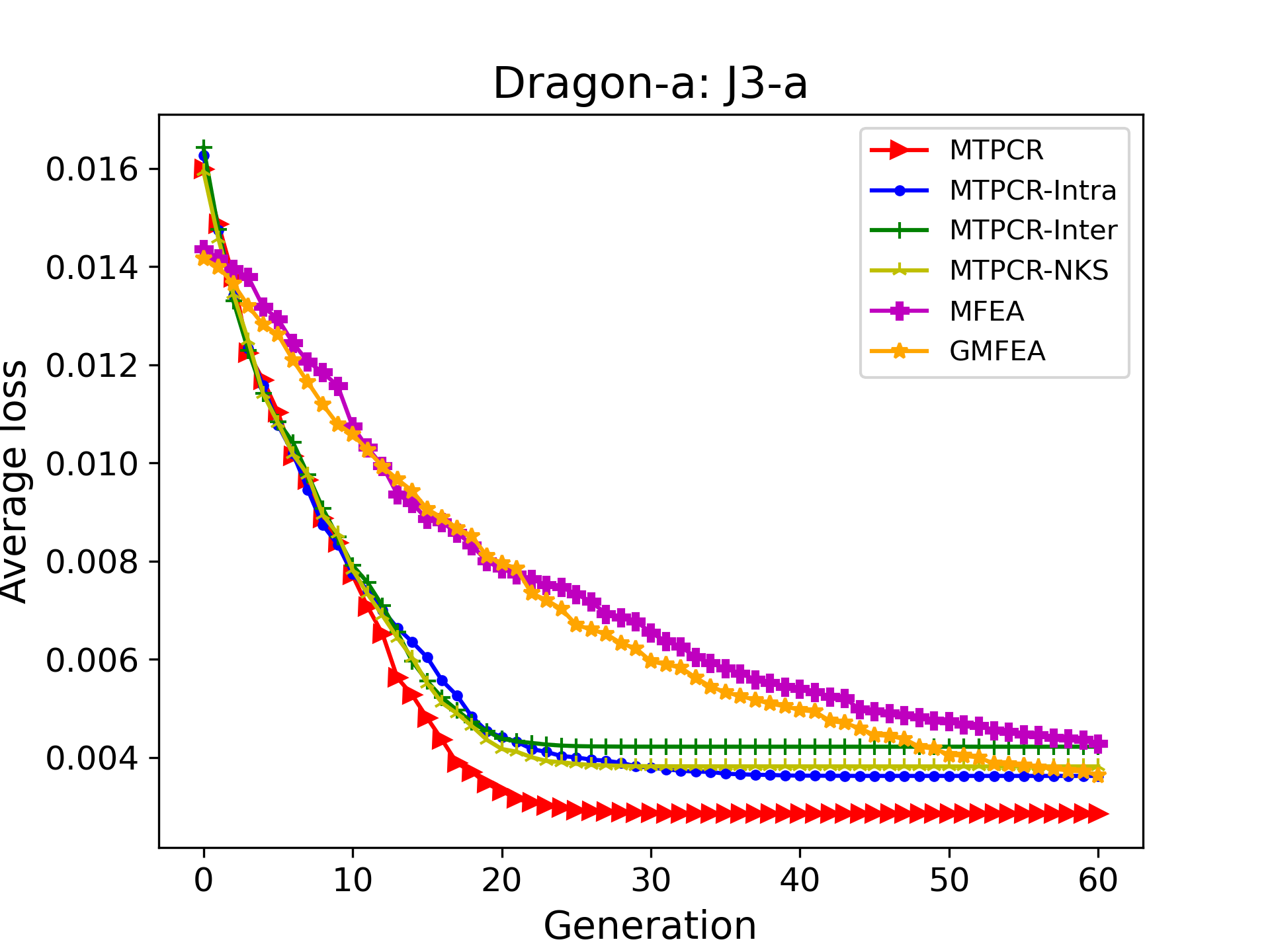}
		\end{minipage}
		\begin{minipage}{0.46\columnwidth}
			\includegraphics[width=\columnwidth]{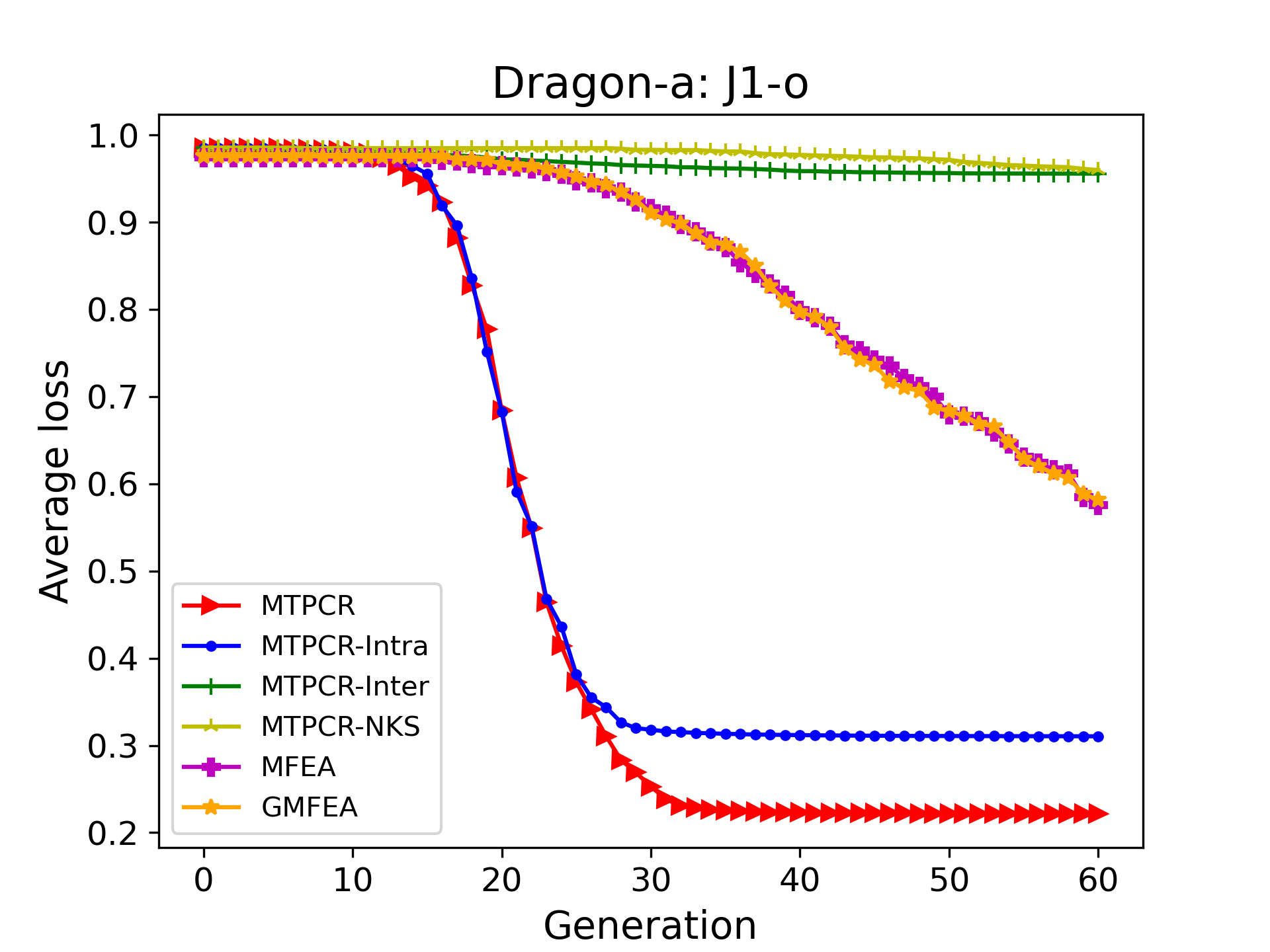}\vspace{1ex}
			\includegraphics[width=\columnwidth]{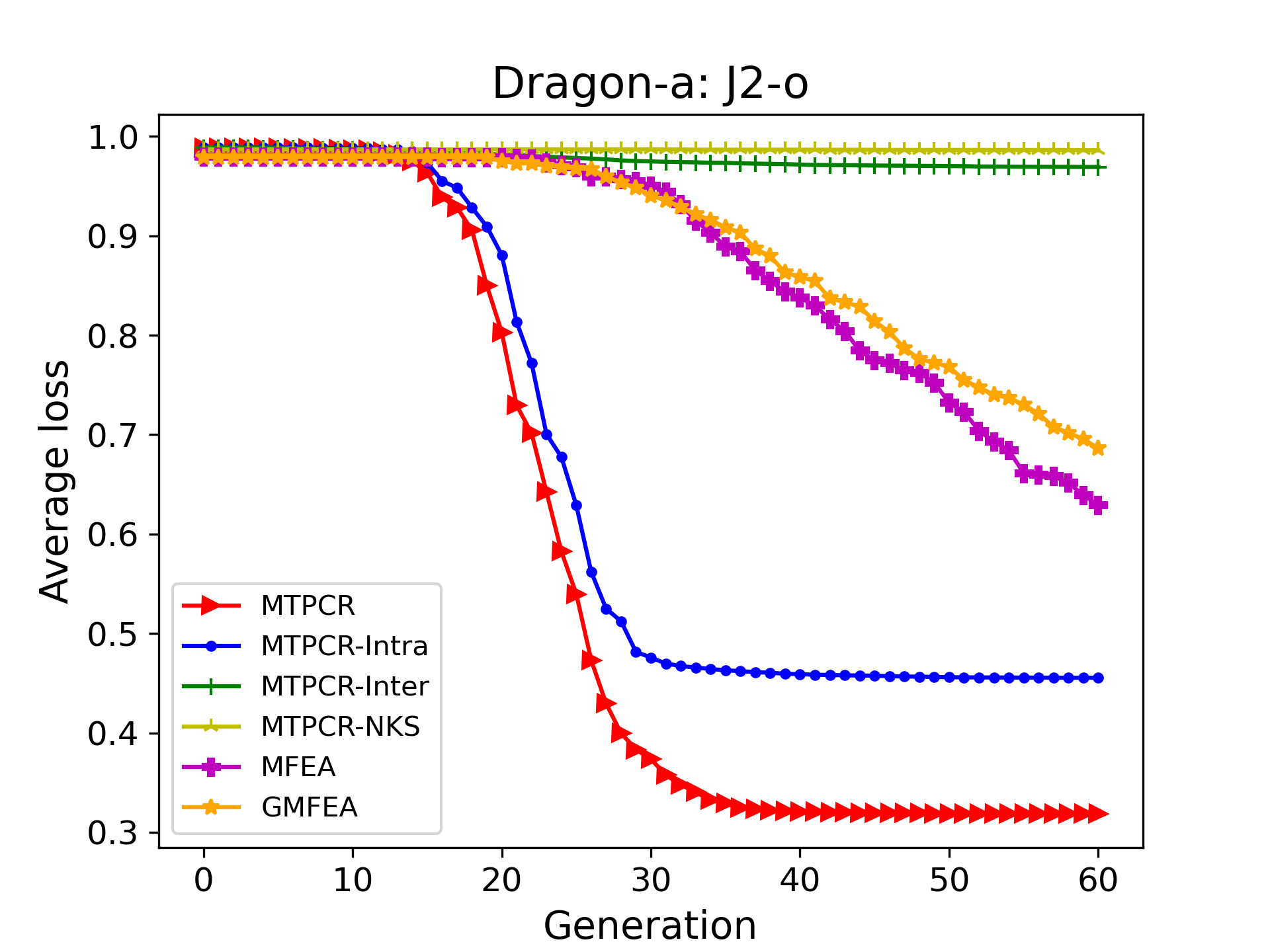}\vspace{1ex}
			\includegraphics[width=\columnwidth]{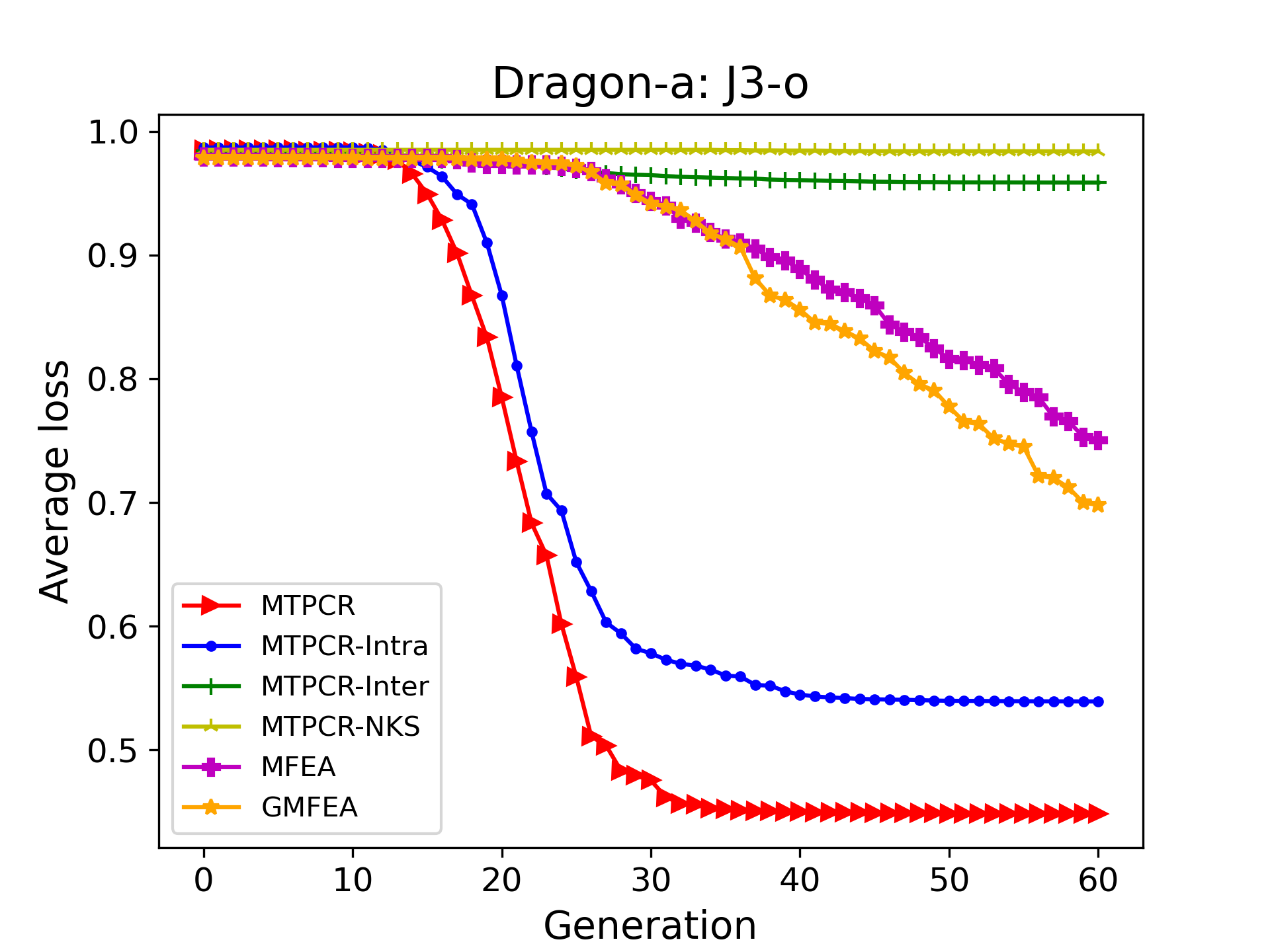}
		\end{minipage}
	}
	\caption{Convergence curves of MTPCR versus MFEA, GMFEA and MTPCR-Intra, MTPCR-Inter, MTPCR-NKS on representative registration tasks. y-axis: loss averaged over 20 independent runs; x-axis: Generation. The left column of each subfigure represents aiding tasks, and the right column of each subfigure represents original tasks. (a) Convergence curves of Bunny-a. (b) Convergence curves of Dragon-a.}
	\label{figConvergenceA}
\end{figure*}

\begin{figure*}[!htbp]
	\centering
	\subfloat[]{\label{figConvergenceB1}
		\begin{minipage}{0.46\columnwidth}
			\includegraphics[width=\columnwidth]{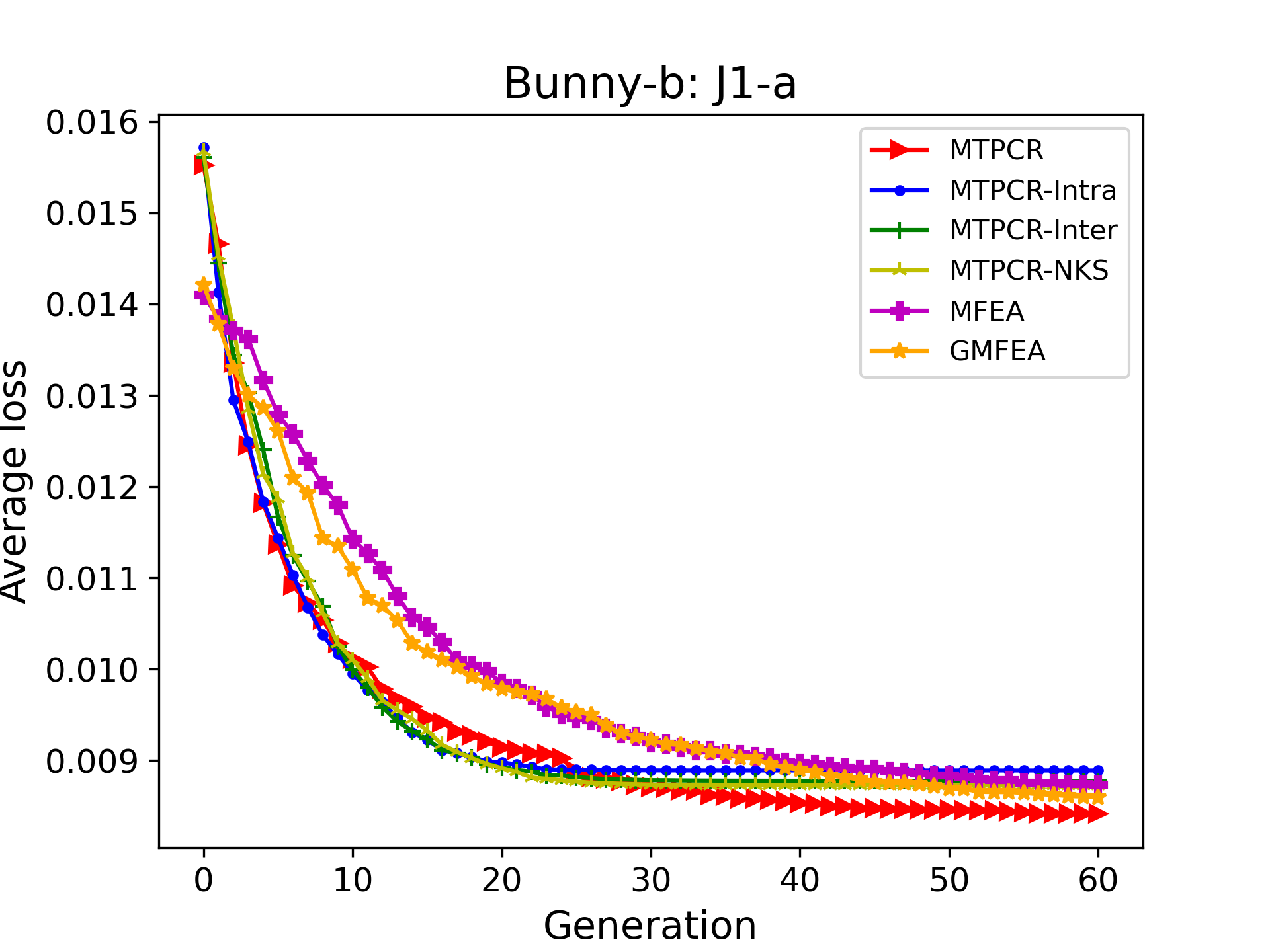}\vspace{1ex}
			\includegraphics[width=\columnwidth]{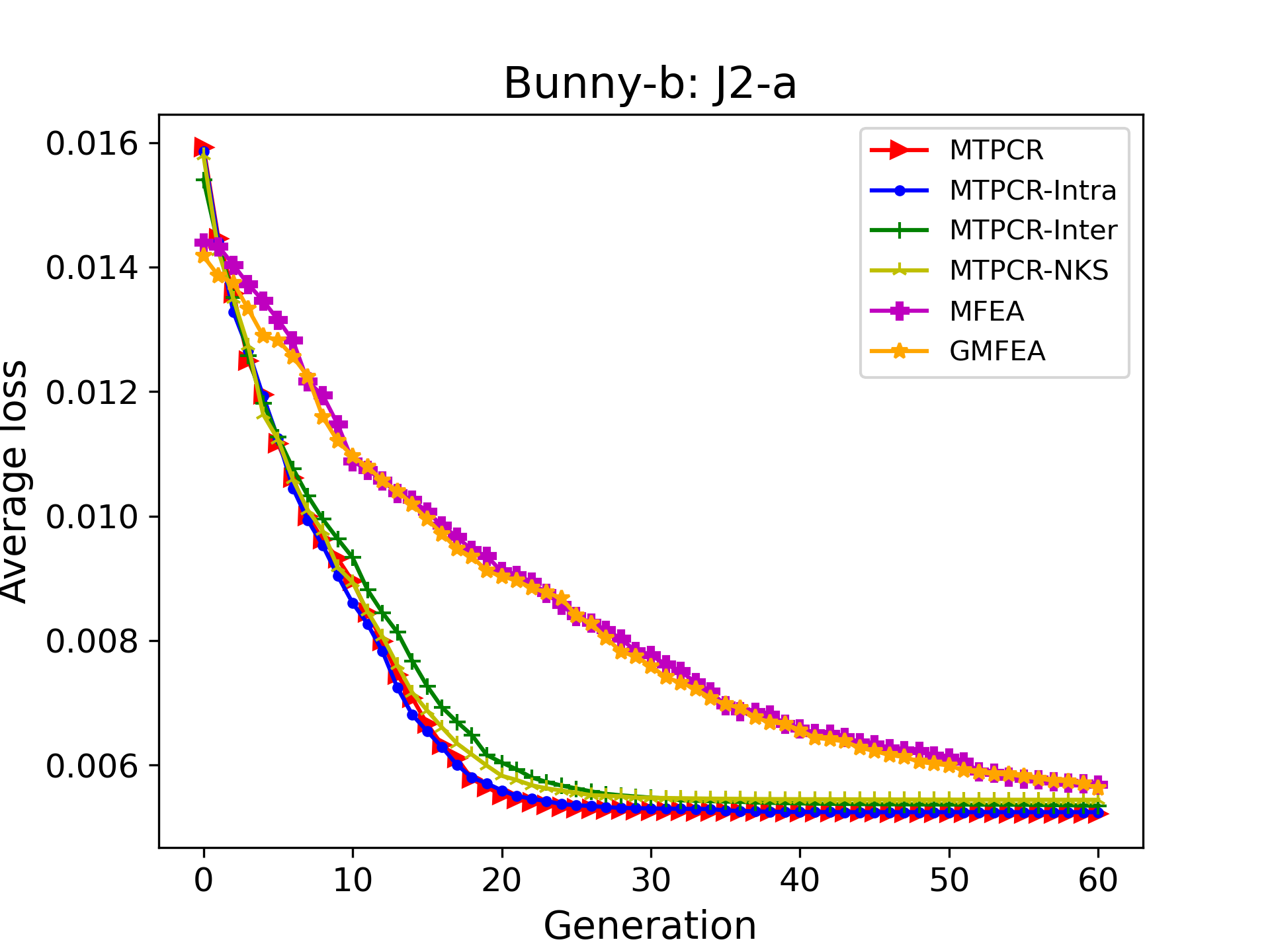}\vspace{1ex}
			\includegraphics[width=\columnwidth]{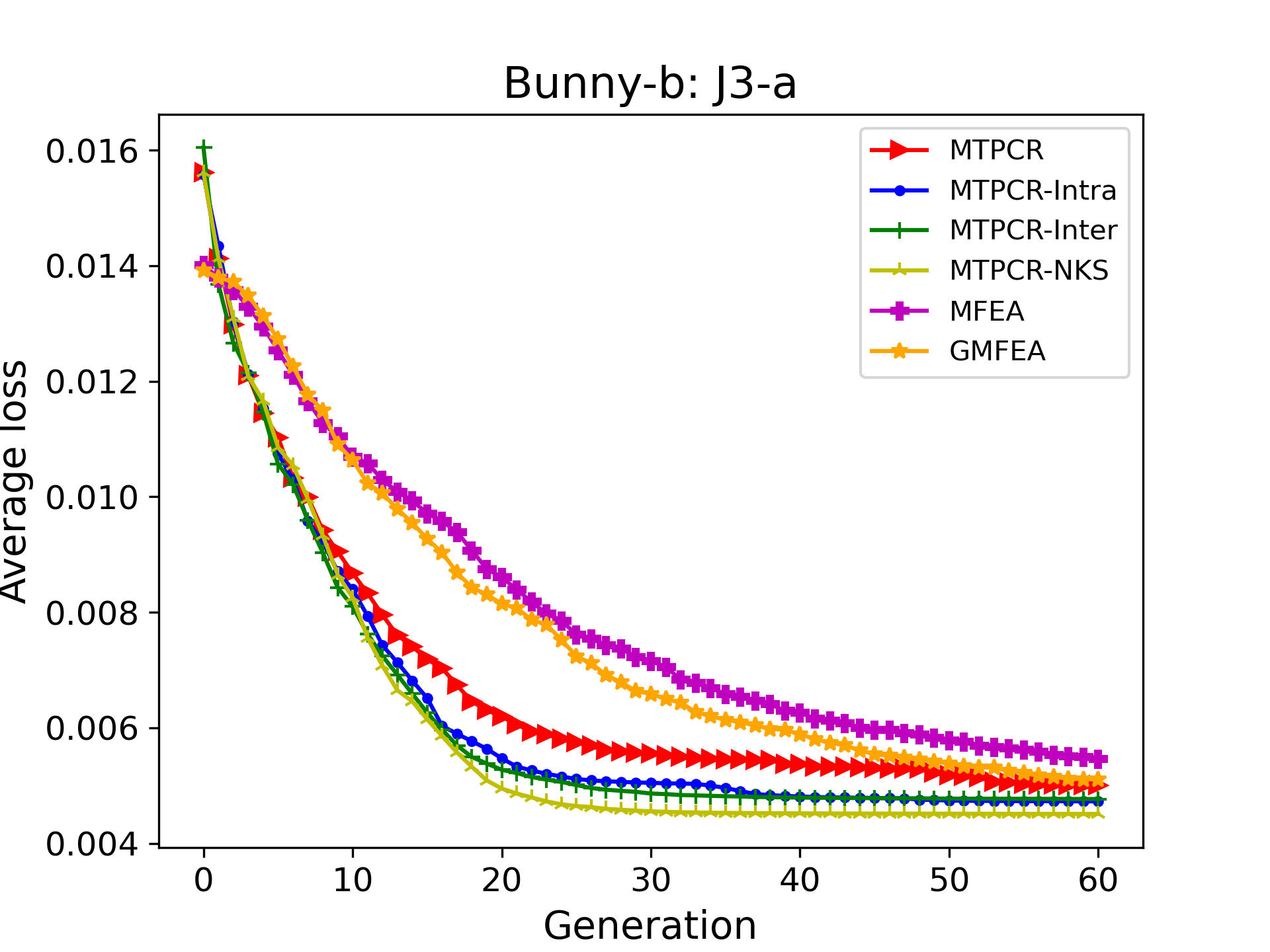}
		\end{minipage}
		\begin{minipage}{0.46\columnwidth}
			\includegraphics[width=\columnwidth]{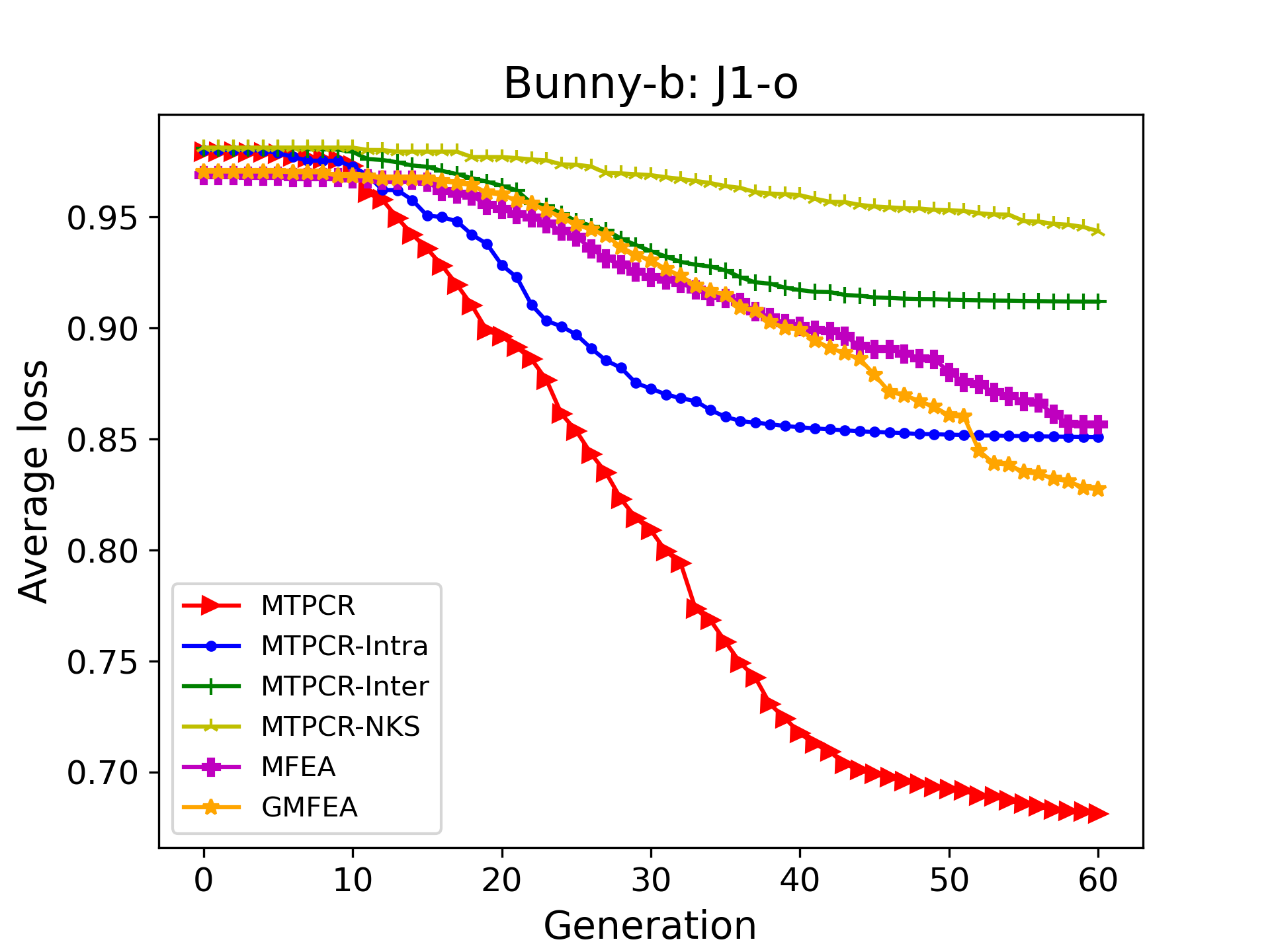}\vspace{1ex}
			\includegraphics[width=\columnwidth]{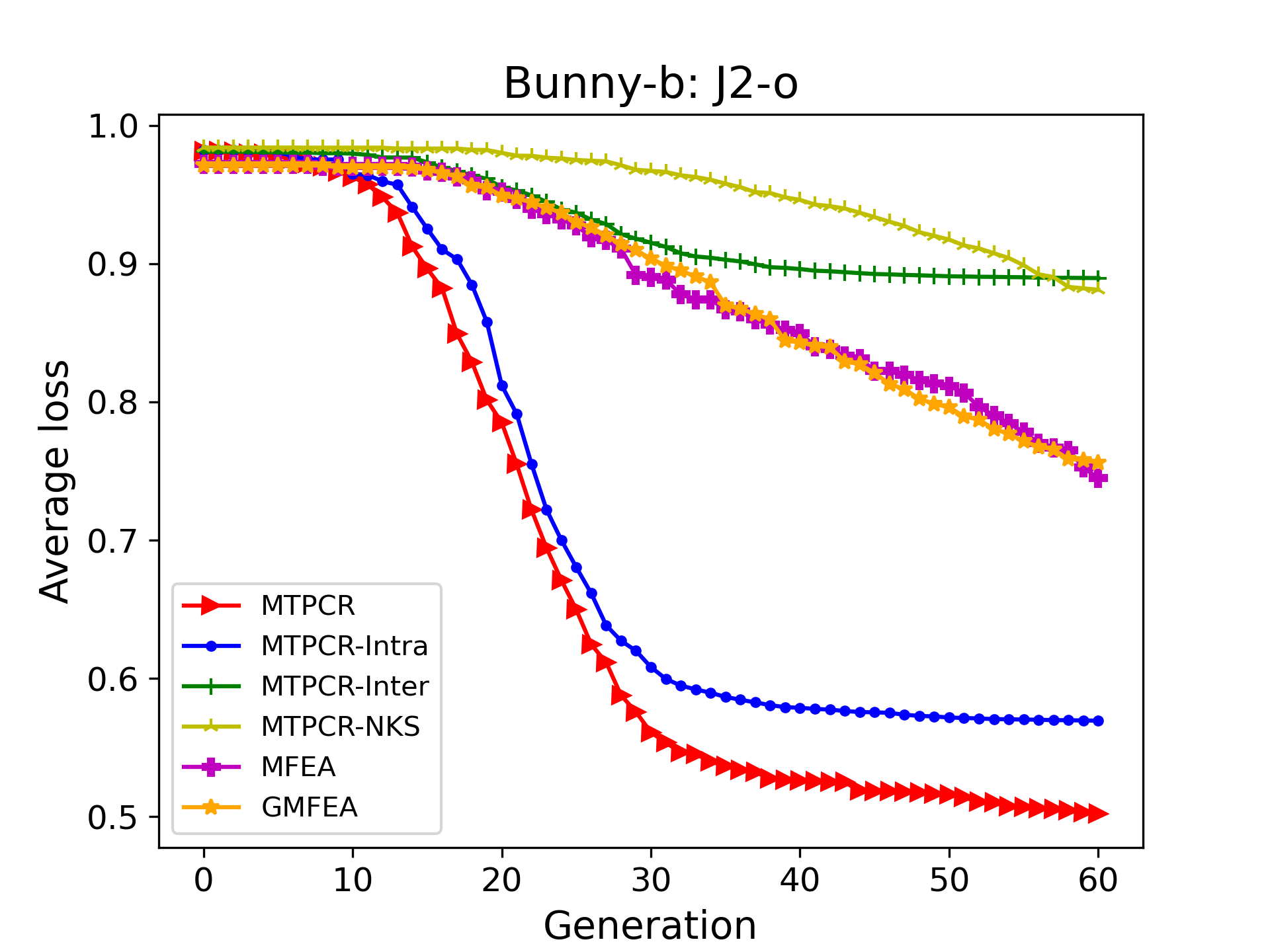}\vspace{1ex}
			\includegraphics[width=\columnwidth]{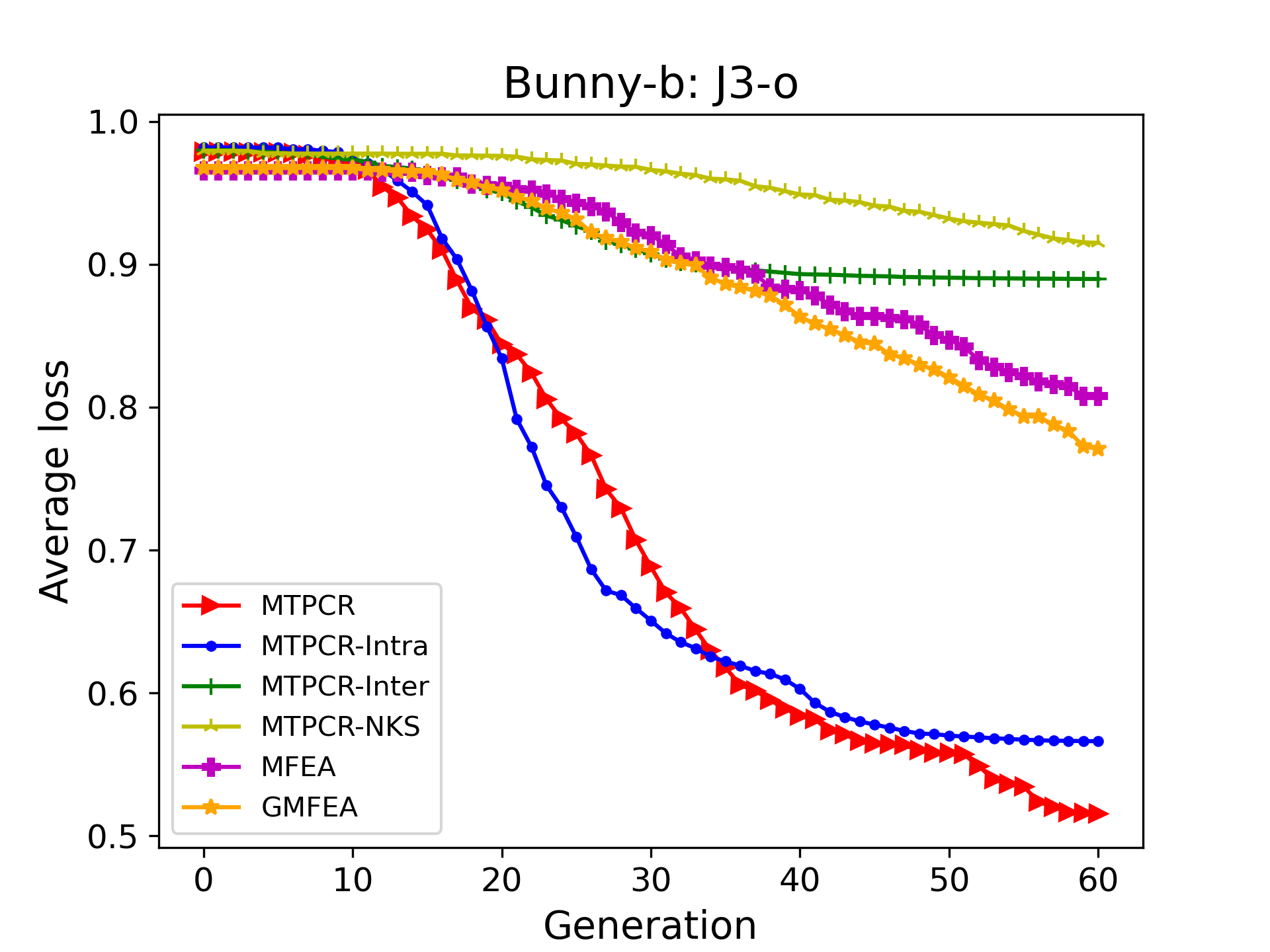}
		\end{minipage}
	}
	\subfloat[]{\label{figConvergenceB2}
		\begin{minipage}{0.46\columnwidth}
			\includegraphics[width=\columnwidth]{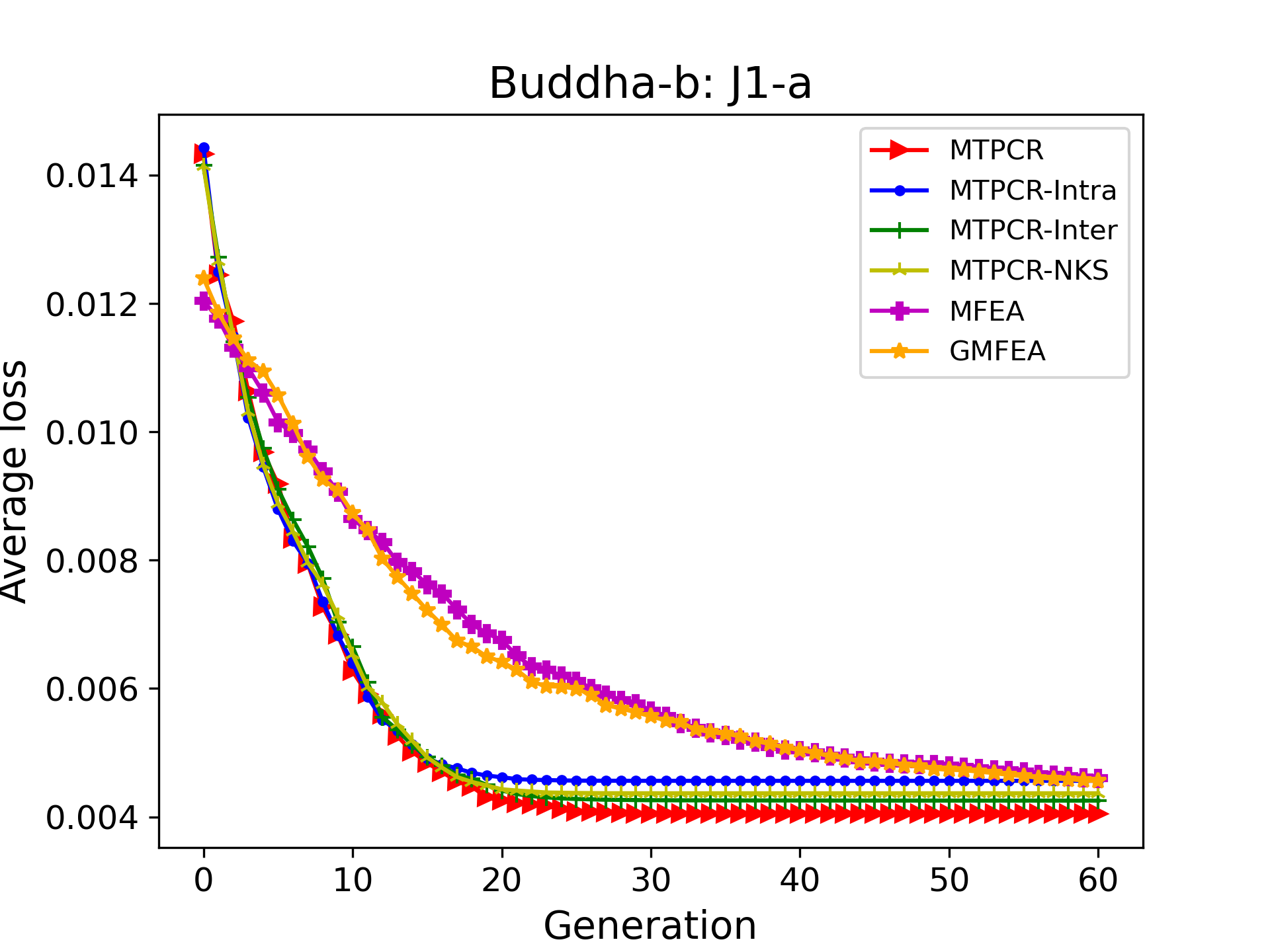}\vspace{1ex}
			\includegraphics[width=\columnwidth]{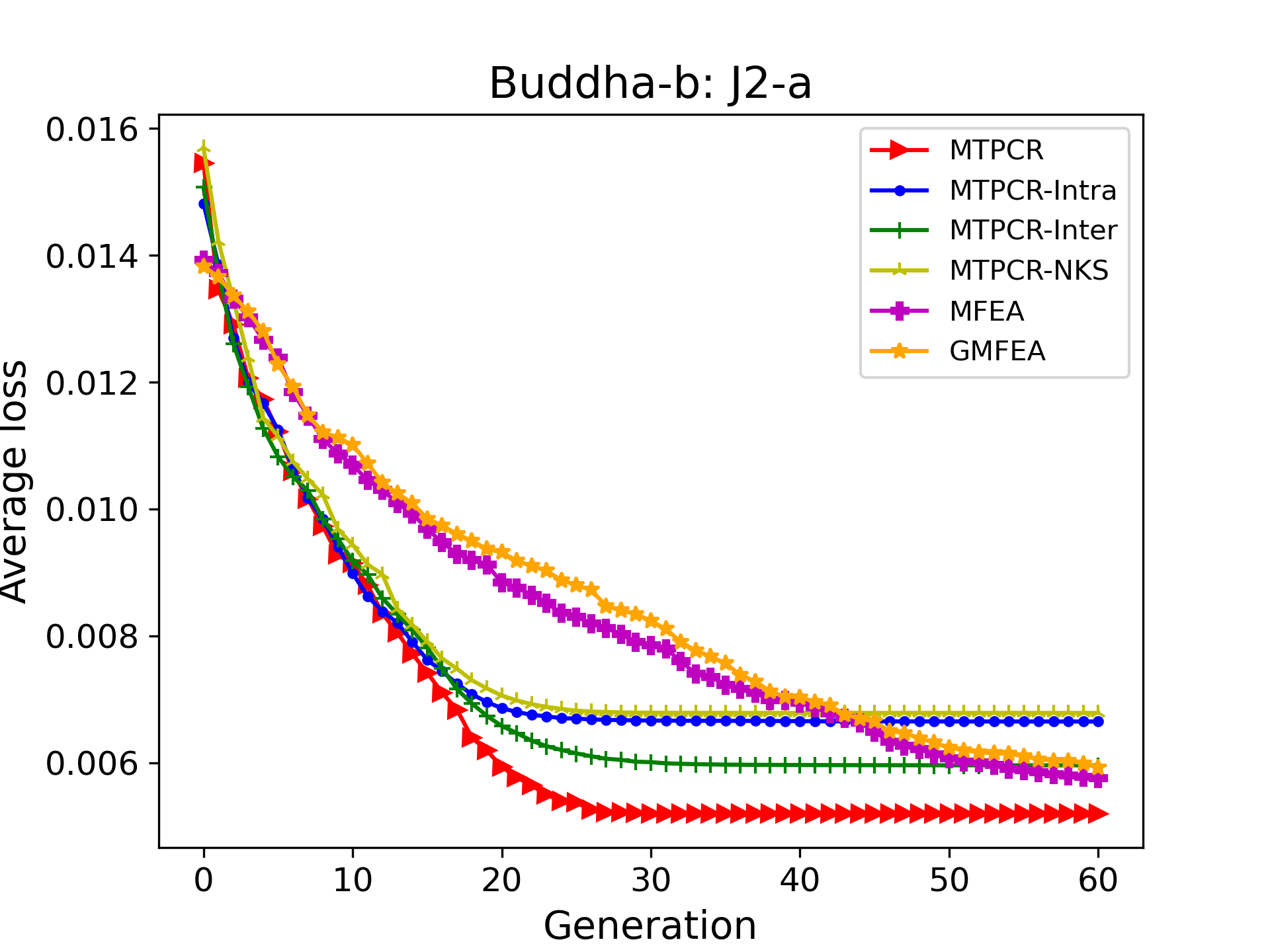}\vspace{1ex}
			\includegraphics[width=\columnwidth]{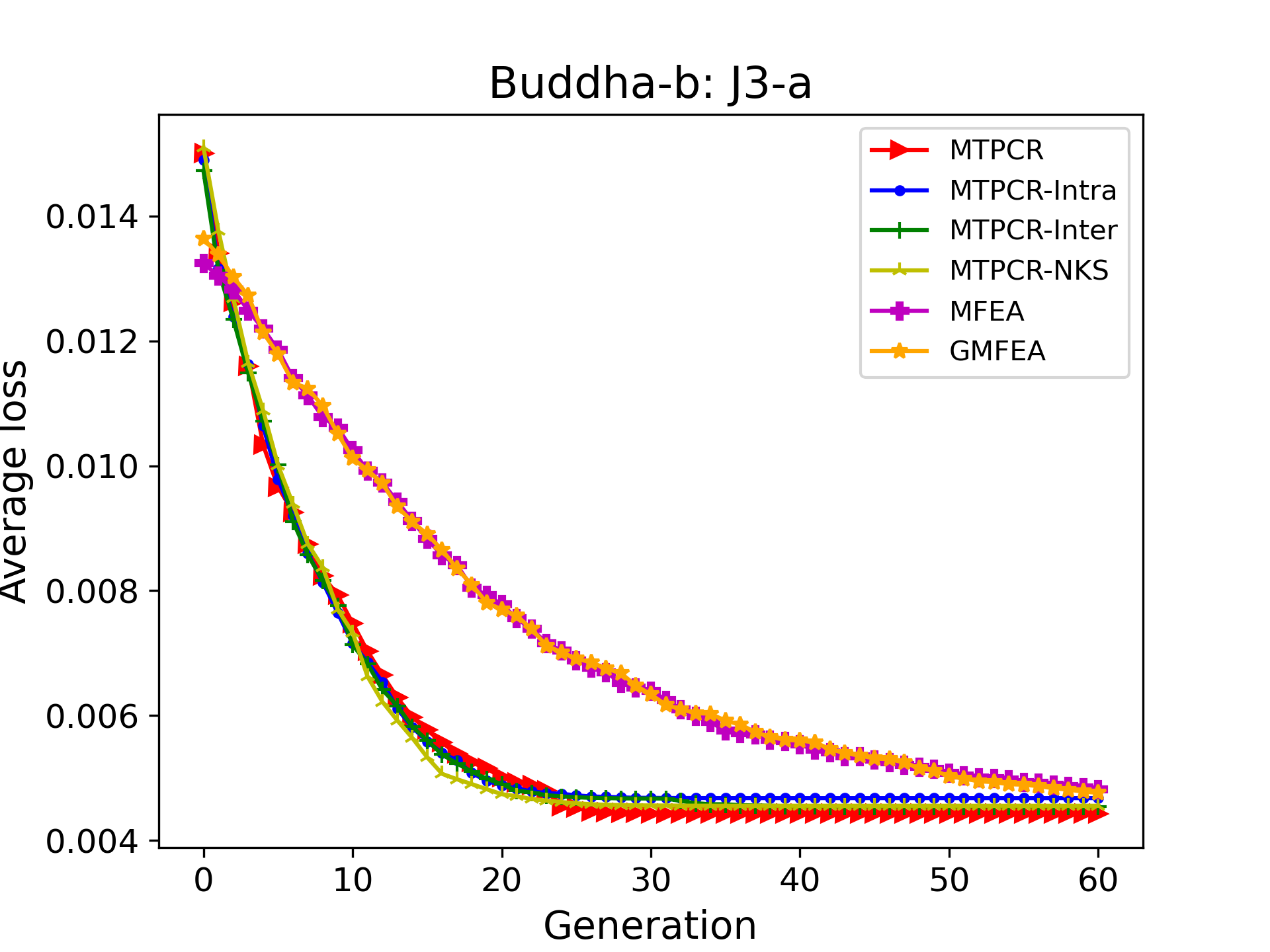}
		\end{minipage}
		\begin{minipage}{0.46\columnwidth}
			\includegraphics[width=\columnwidth]{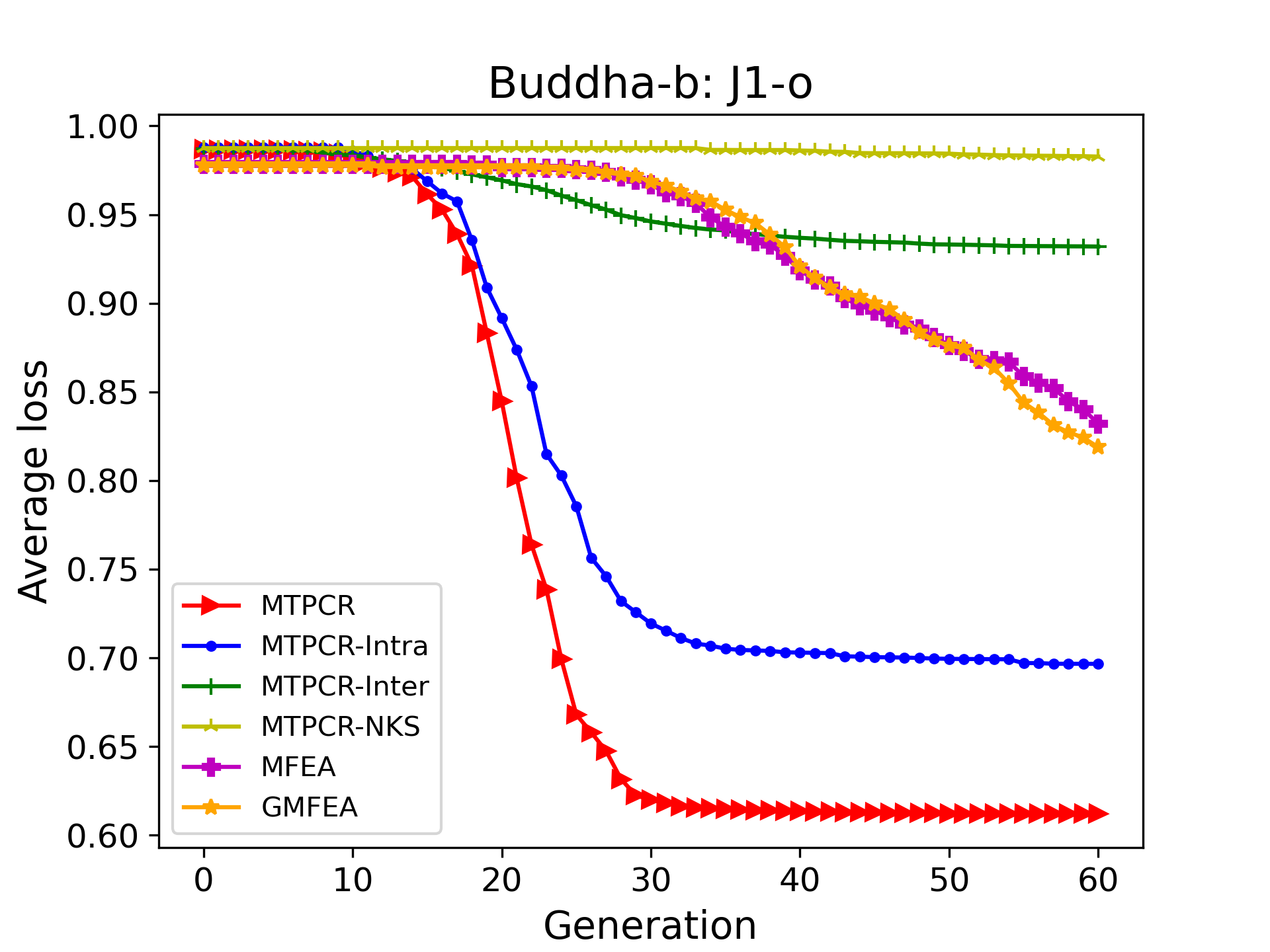}\vspace{1ex}
			\includegraphics[width=\columnwidth]{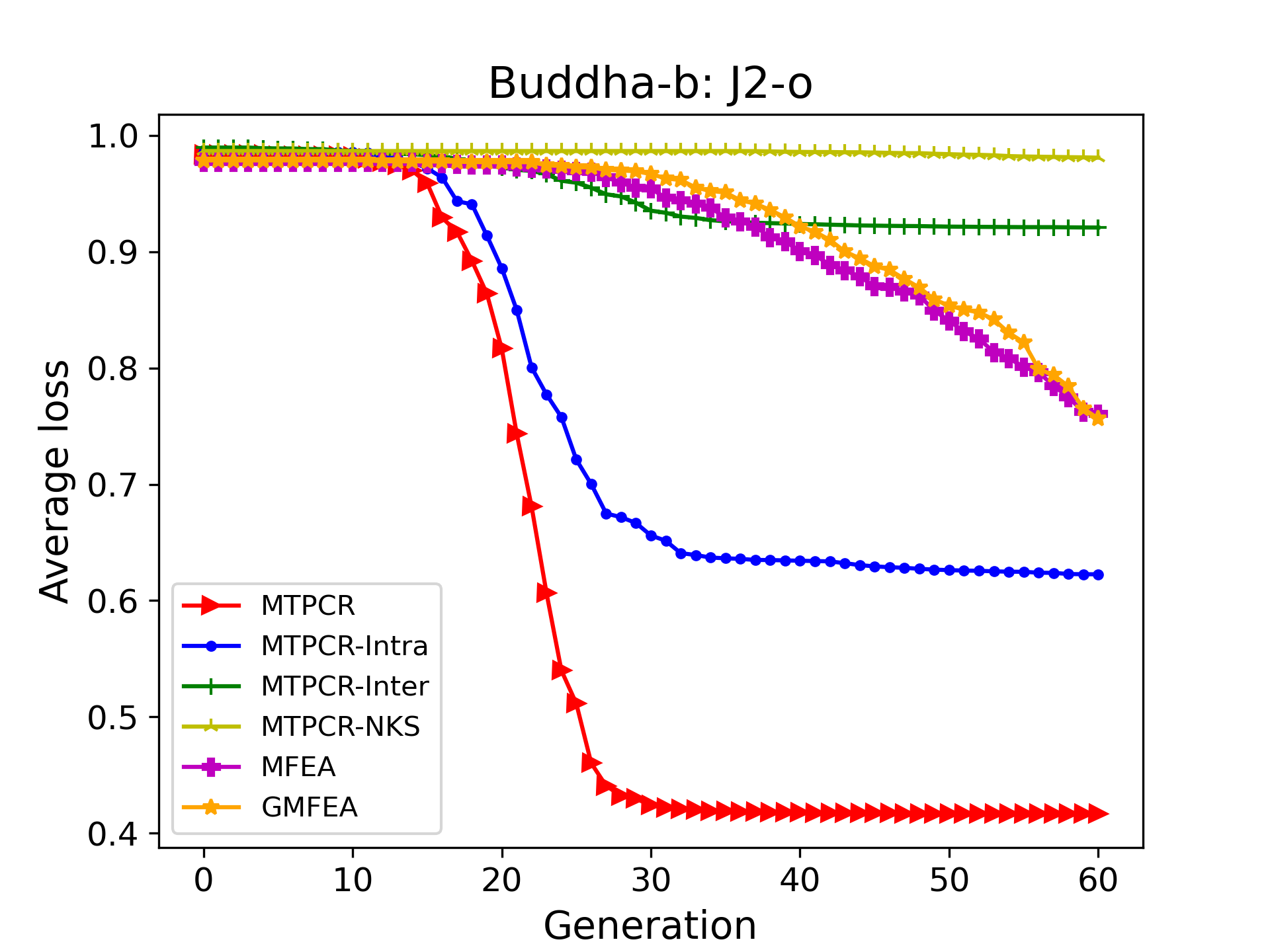}\vspace{1ex}
			\includegraphics[width=\columnwidth]{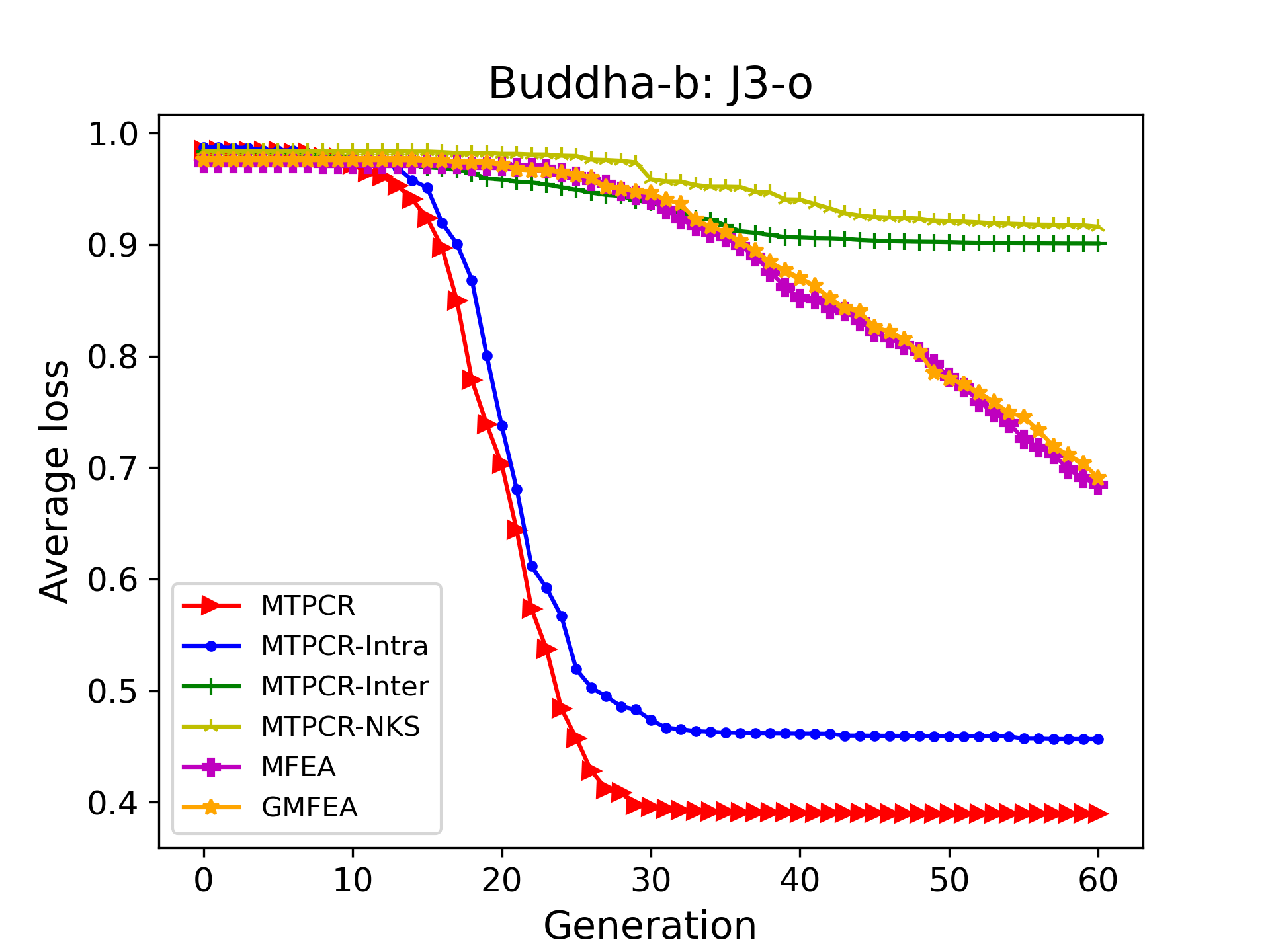}
		\end{minipage}
	}
	\caption{Convergence curves of MTPCR versus MFEA, GMFEA and MTPCR-Intra, MTPCR-Inter, MTPCR-NKS on representative registration tasks. y-axis: loss averaged over 20 independent runs; x-axis: Generation. The left column of each subfigure represents aiding tasks, and the right column of each subfigure represents original tasks. (a) Convergence curves of Bunny-b. (b) Convergence curves of Buddha-b.}
	\label{figConvergenceB}
\end{figure*}

\begin{figure*}[!htbp]
	\centering
	\subfloat[]{\label{figConvergenceC1}
		\begin{minipage}{0.46\columnwidth}
			\includegraphics[width=\columnwidth]{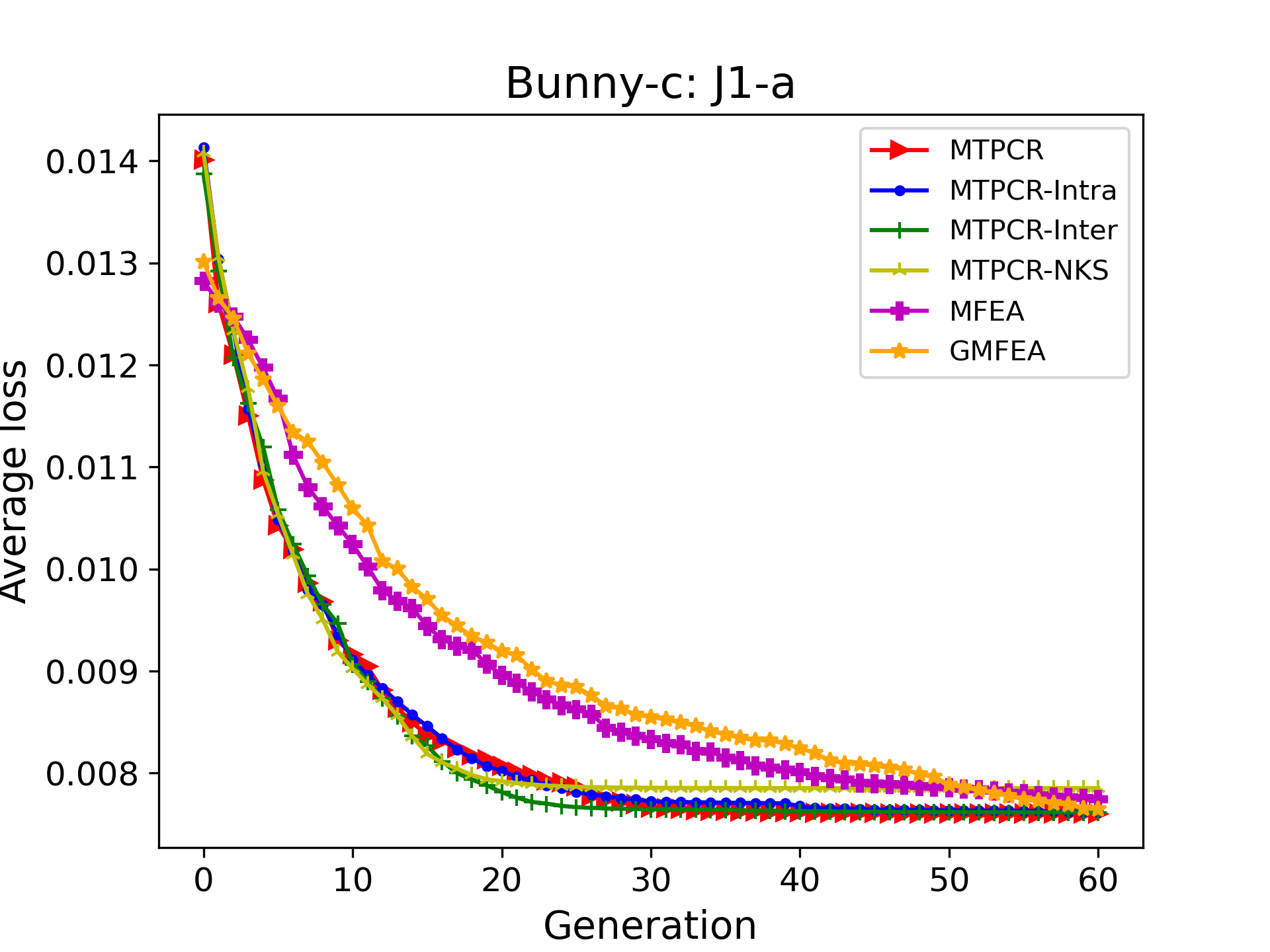}\vspace{1ex}
			\includegraphics[width=\columnwidth]{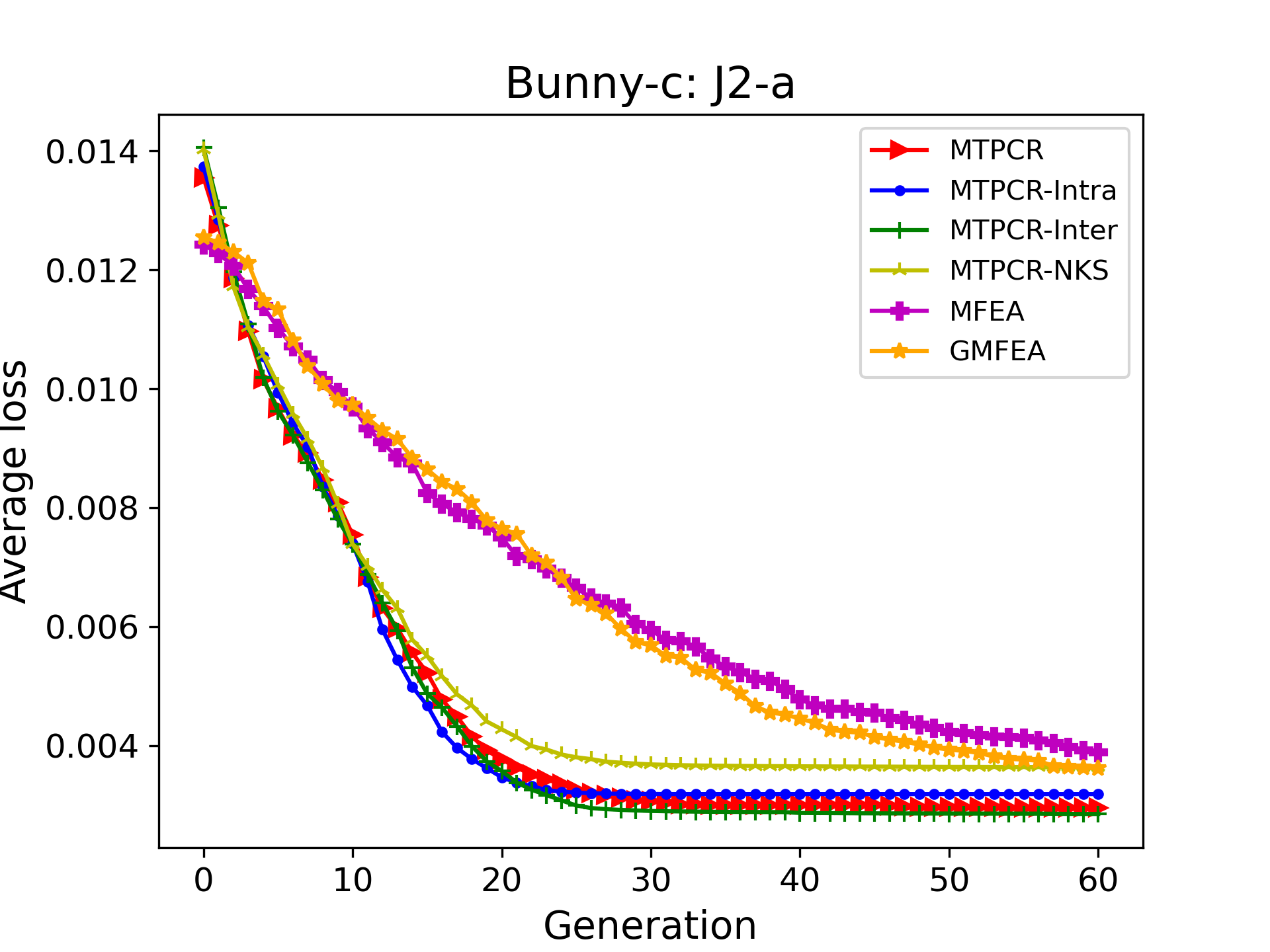}\vspace{1ex}
			\includegraphics[width=\columnwidth]{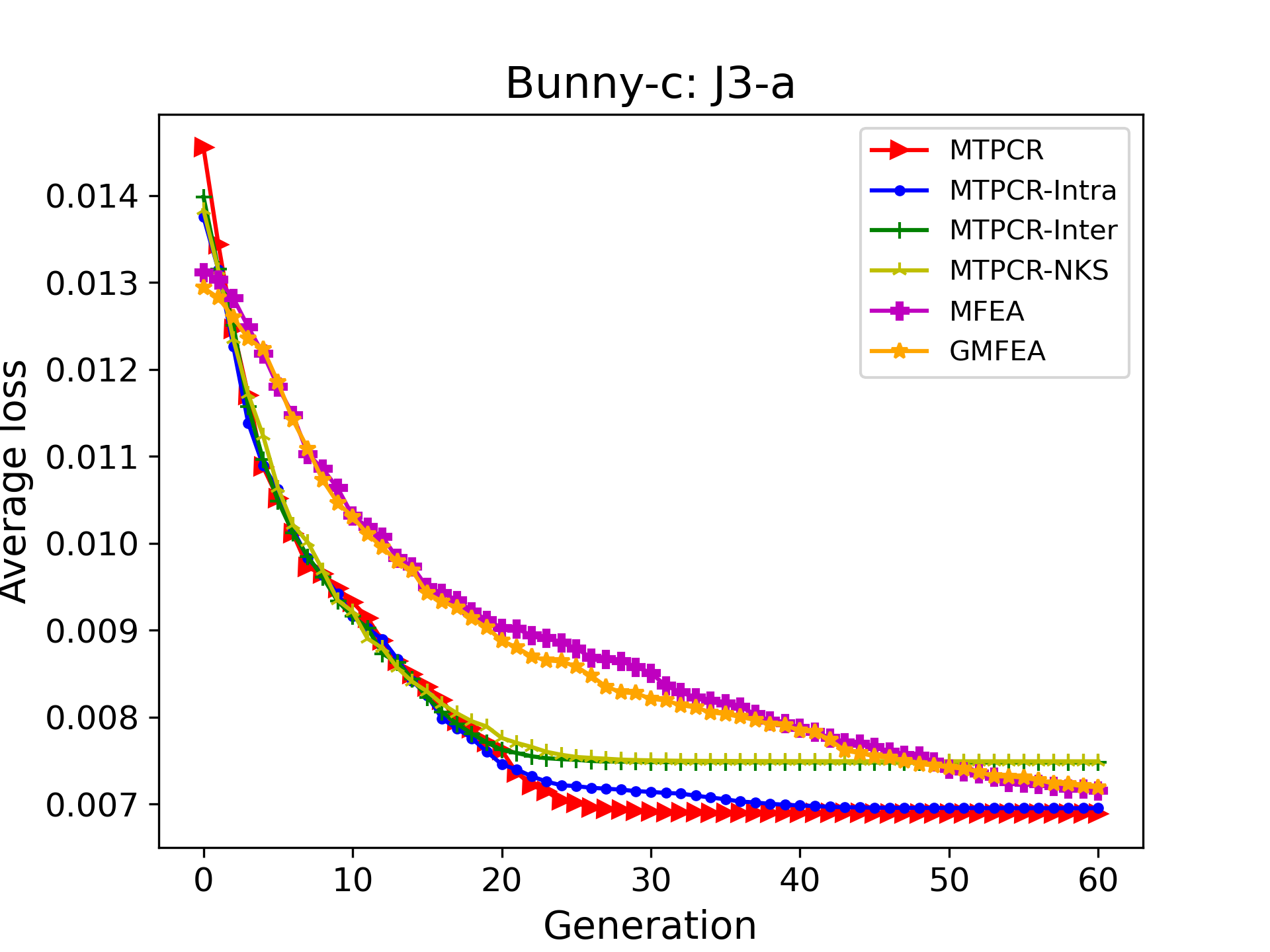}
		\end{minipage}
		\begin{minipage}{0.46\columnwidth}
			\includegraphics[width=\columnwidth]{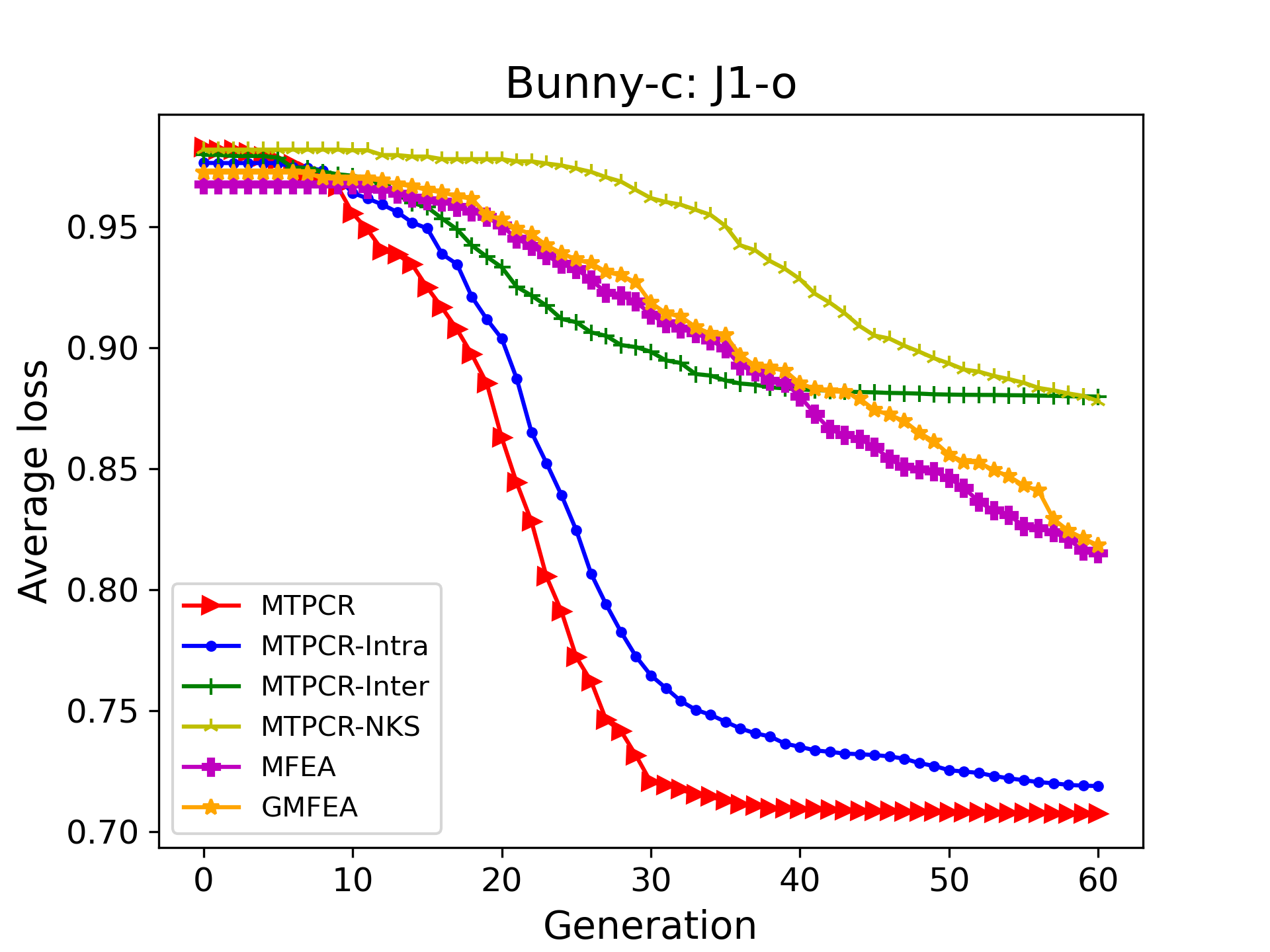}\vspace{1ex}
			\includegraphics[width=\columnwidth]{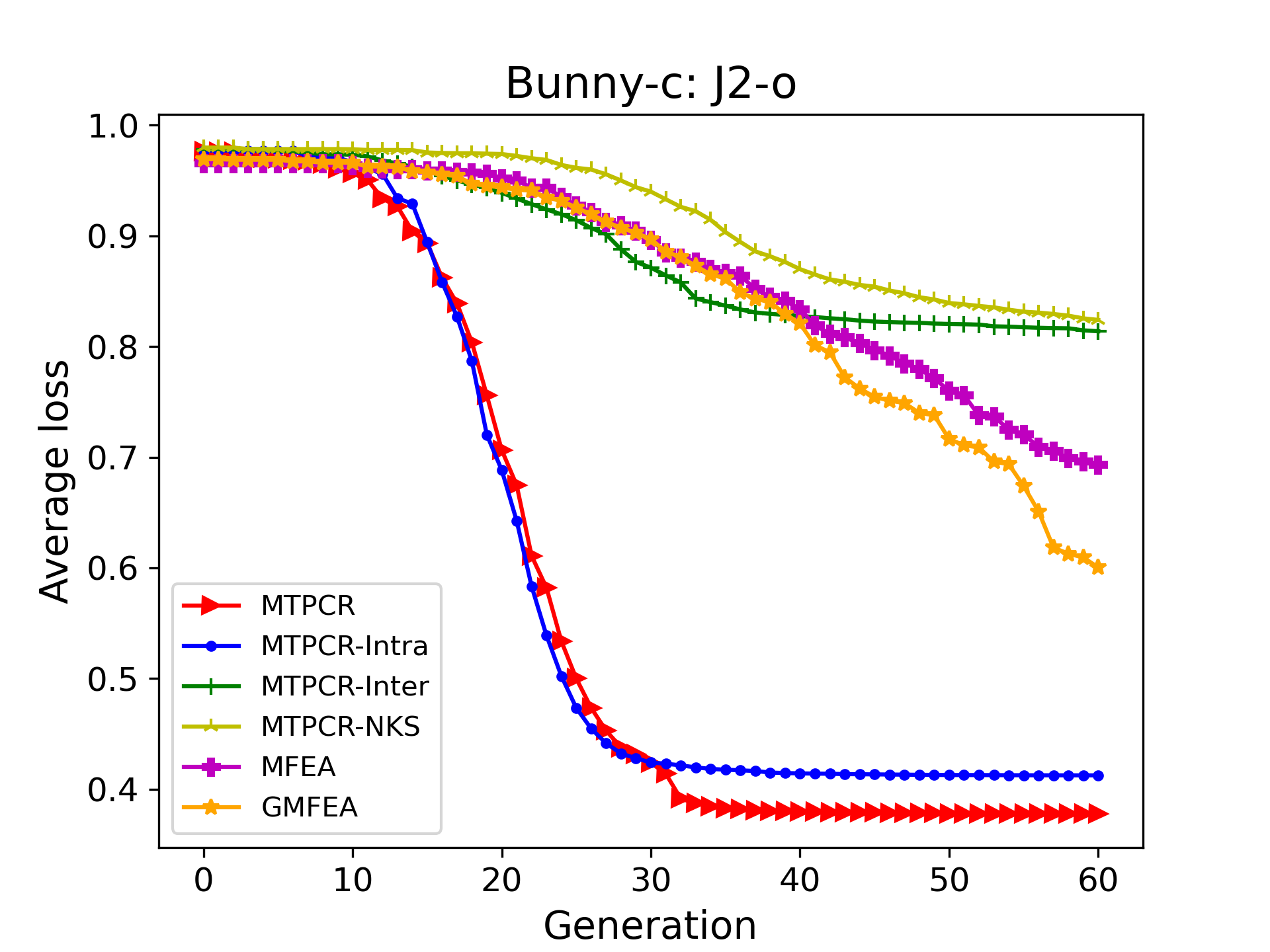}\vspace{1ex}
			\includegraphics[width=\columnwidth]{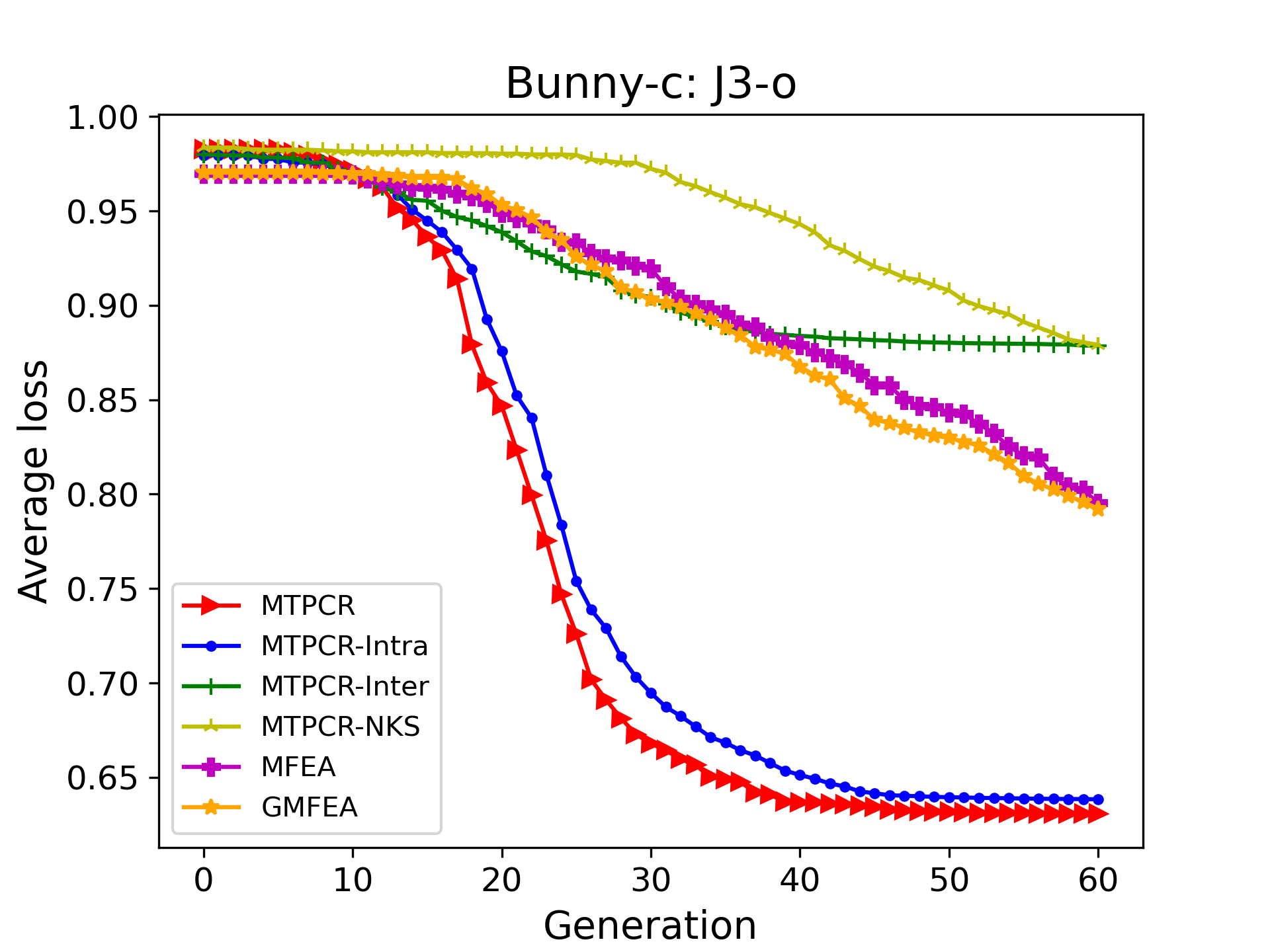}
		\end{minipage}
	}
	\subfloat[]{\label{figConvergenceC2}
		\begin{minipage}{0.46\columnwidth}
			\includegraphics[width=\columnwidth]{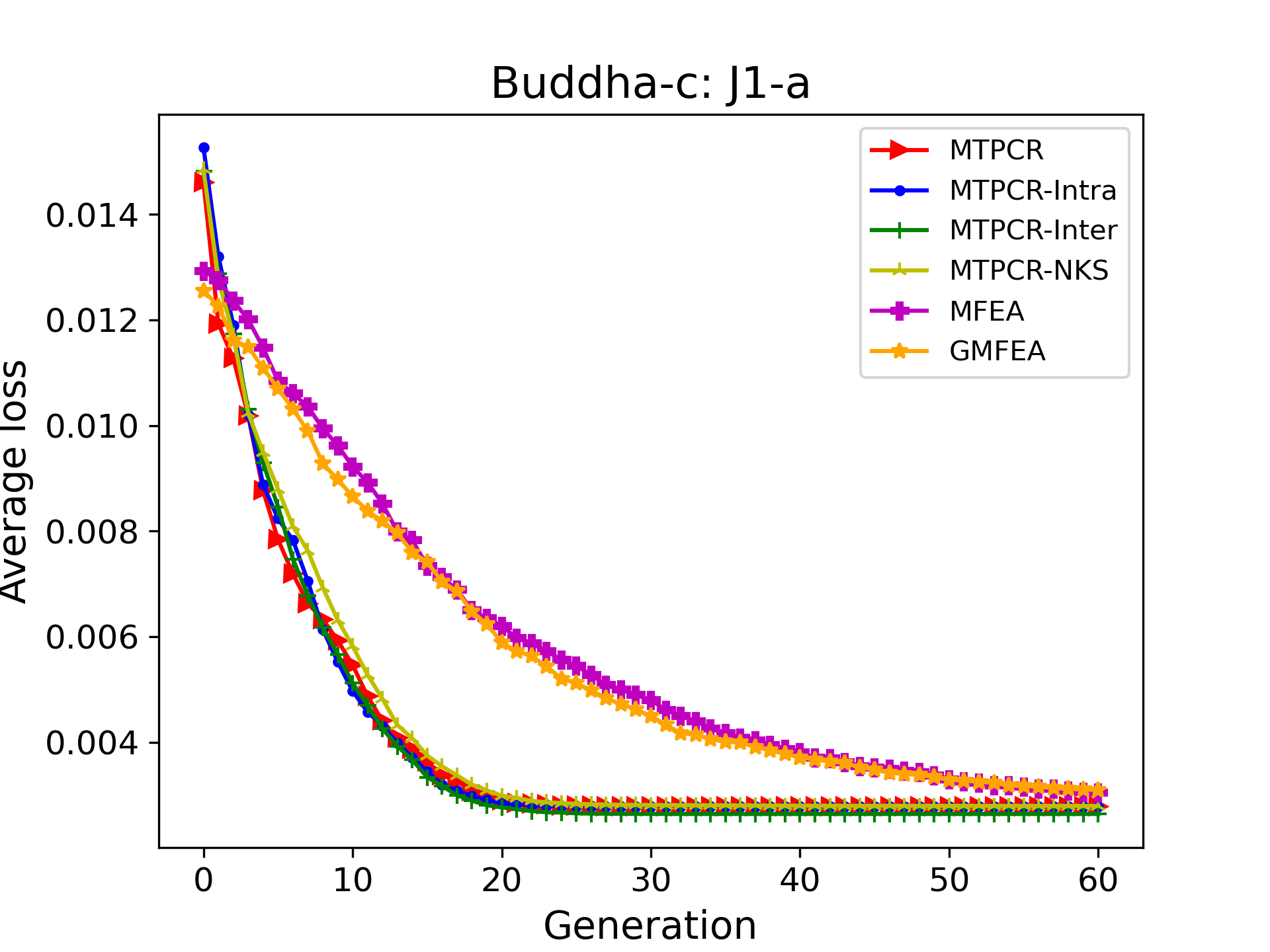}\vspace{1ex}
			\includegraphics[width=\columnwidth]{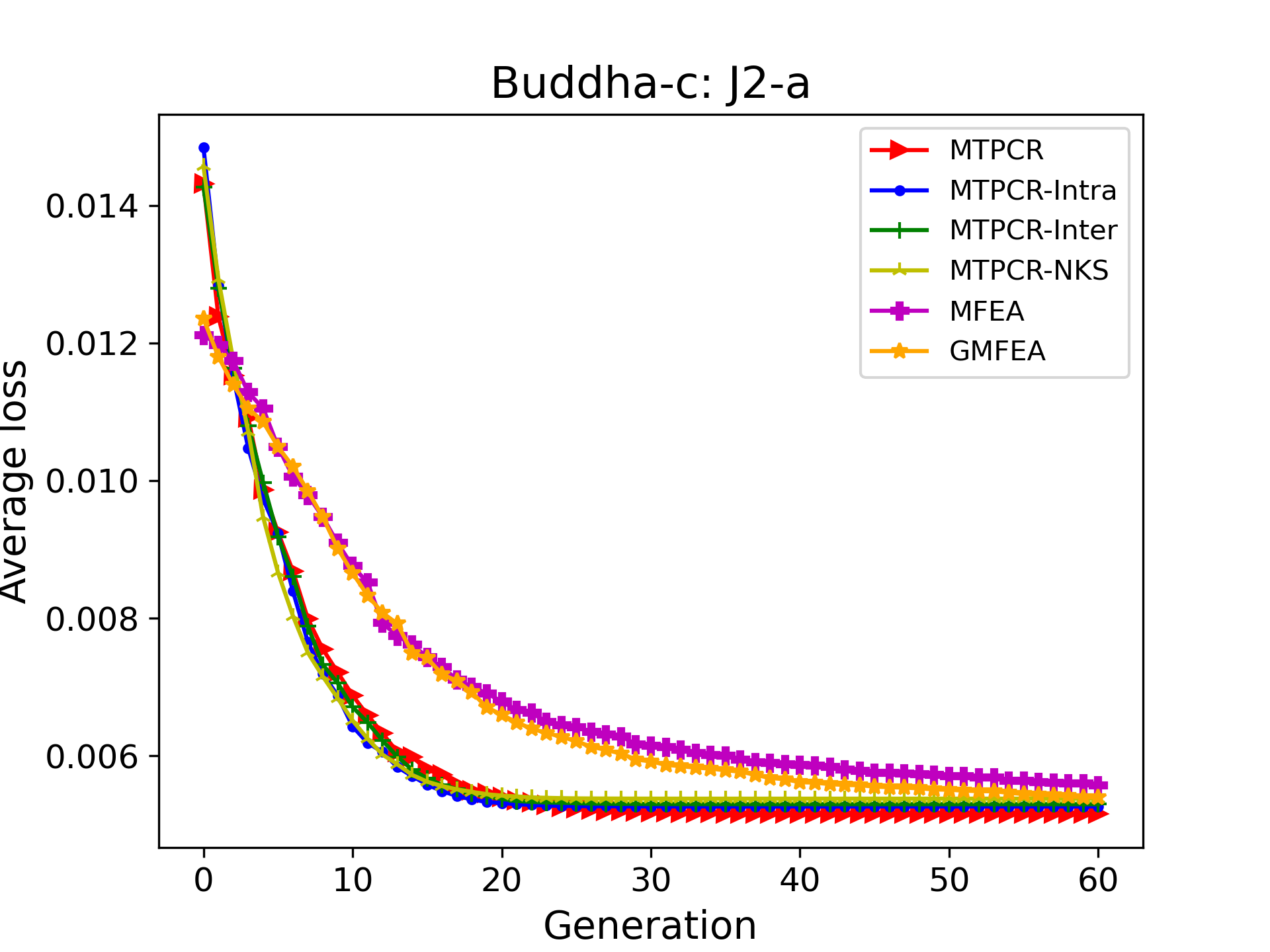}\vspace{1ex}
			\includegraphics[width=\columnwidth]{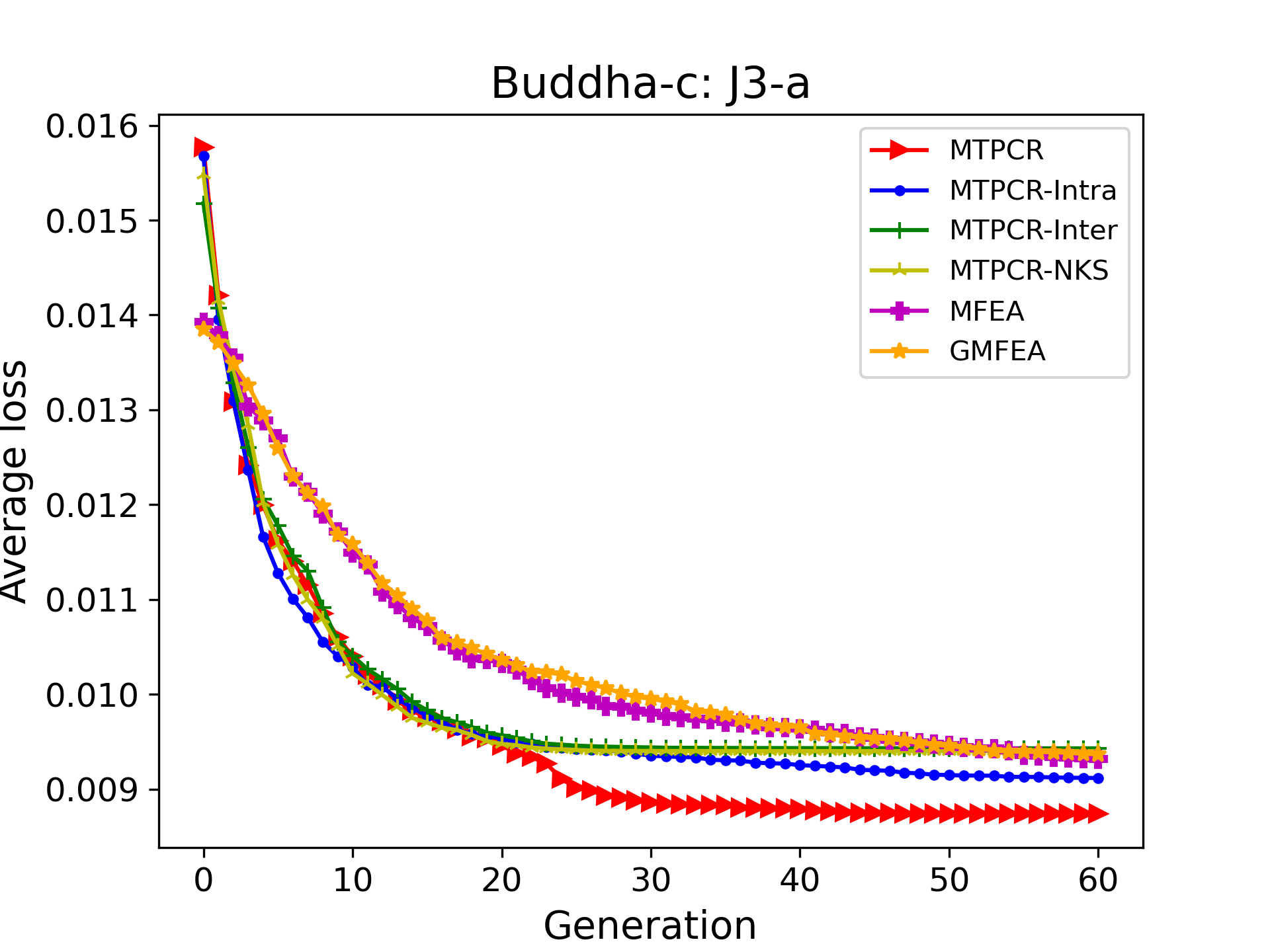}
		\end{minipage}
		\begin{minipage}{0.46\columnwidth}
			\includegraphics[width=\columnwidth]{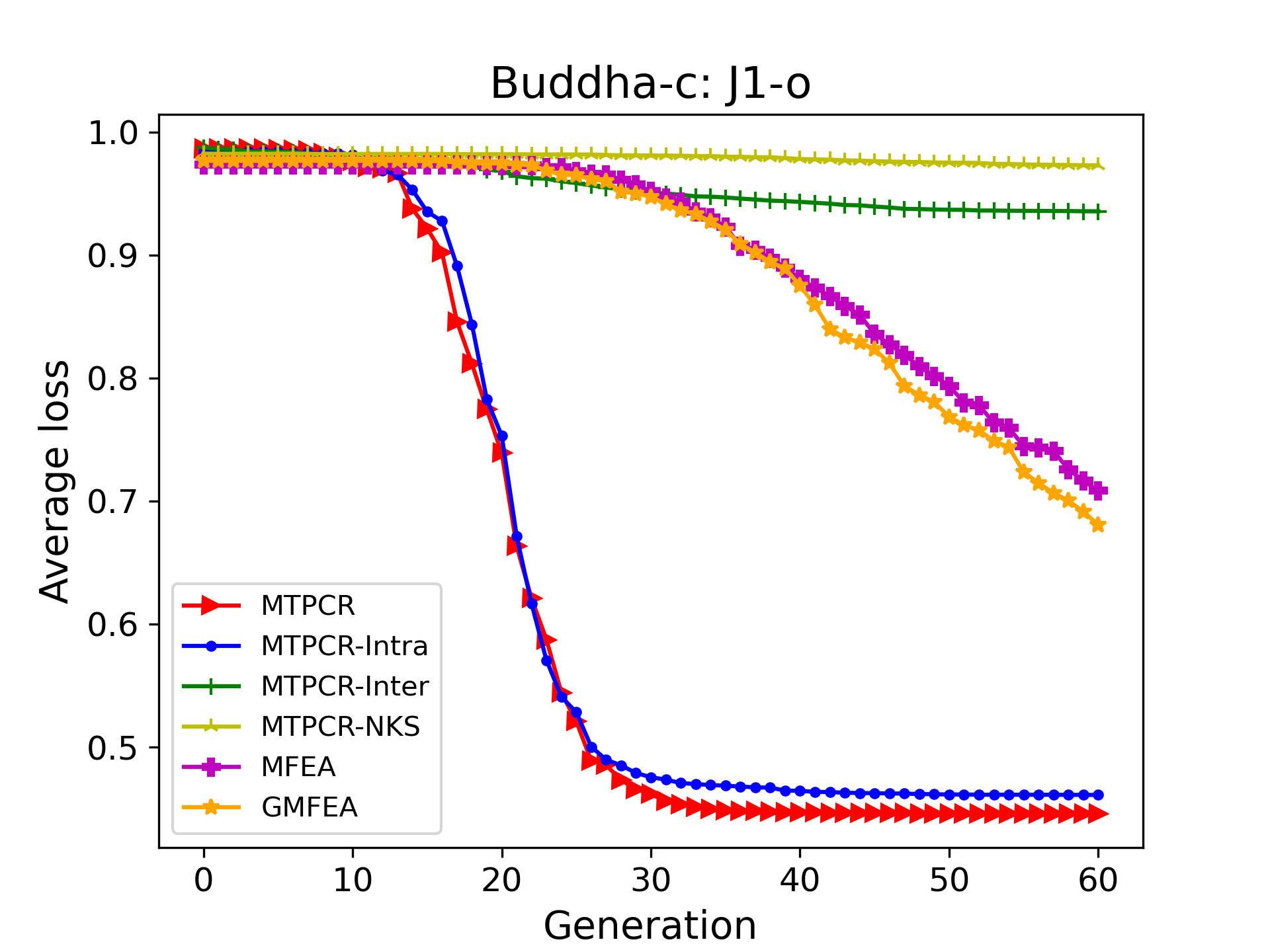}\vspace{1ex}
			\includegraphics[width=\columnwidth]{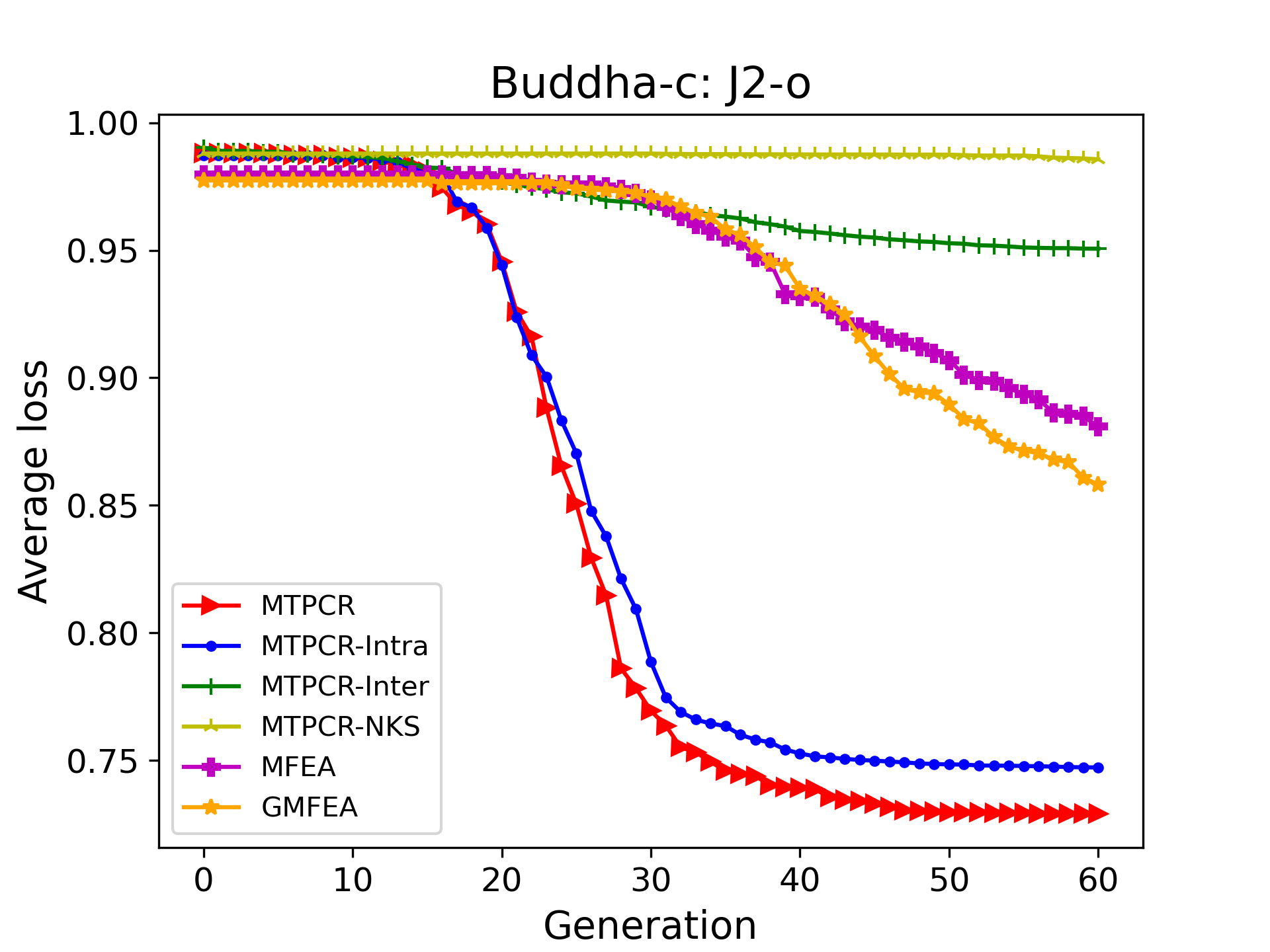}\vspace{1ex}
			\includegraphics[width=\columnwidth]{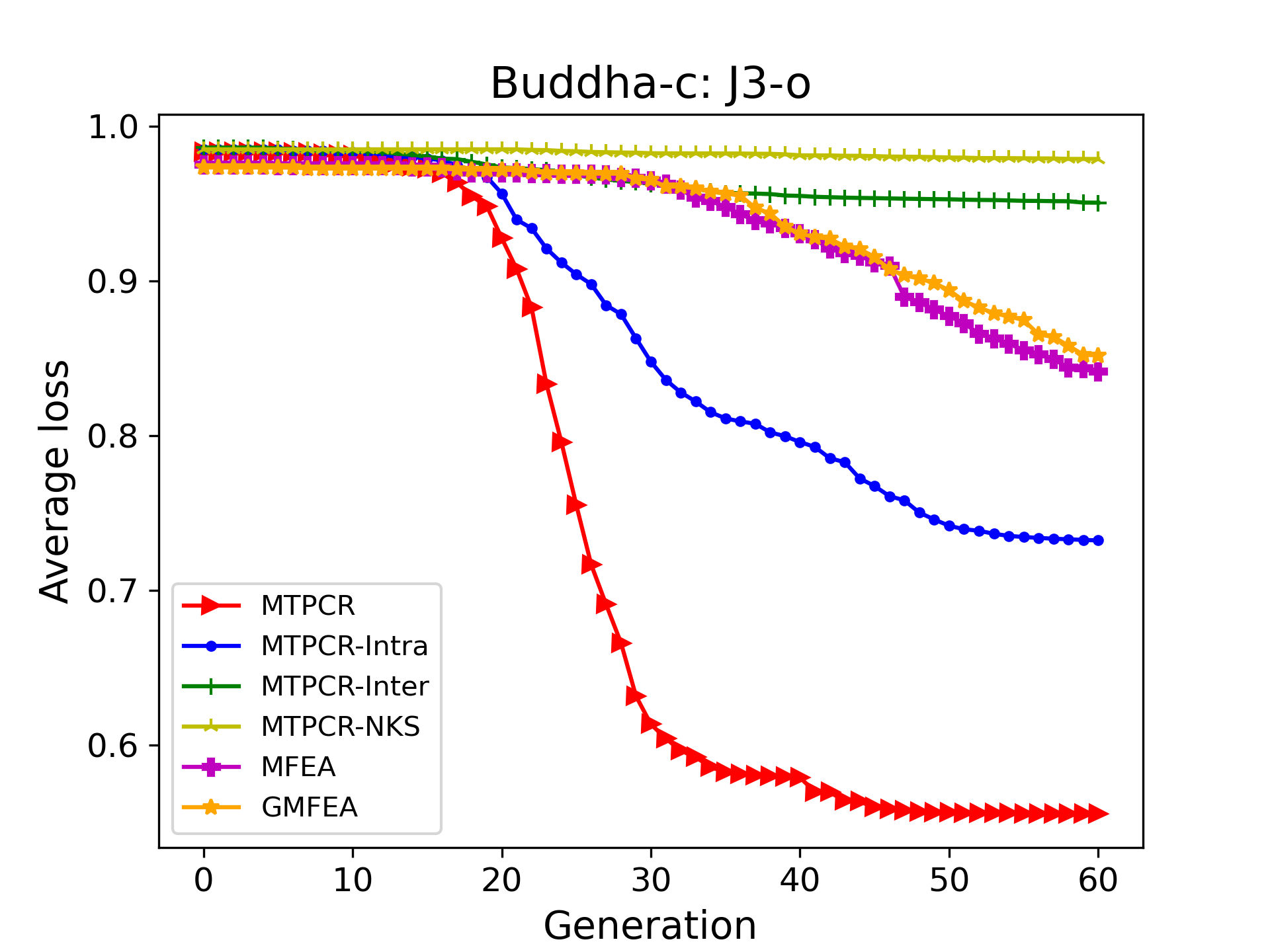}
		\end{minipage}
	}
	\caption{Convergence curves of MTPCR versus MFEA, GMFEA and MTPCR-Intra, MTPCR-Inter, MTPCR-NKS on representative registration tasks. y-axis: loss averaged over 20 independent runs; x-axis: Generation. The left column of each subfigure represents aiding tasks, and the right column of each subfigure represents original tasks. (a) Convergence curves of Bunny-c. (b) Convergence curves of Buddha-c.}
	\label{figConvergenceC}
\end{figure*}

To assess the efficiency of the proposed method with bi-channel knowledge sharing mechanism, Figs.~\ref{figConvergenceA}, \ref{figConvergenceB}, and \ref{figConvergenceC} present the convergence curves of MTPCR, MFEA, GMFEA and MTPCR-Intra, MTPCR-Inter, MTPCR-NKS on representative registration problems averaged over 20 independent runs. In these figures, y-axis shows the actual cost without scaling, and x-axis is the number of generation. Note that there are six subfigures for each registration problem: the three in the left column correspond to the aiding tasks which are much simpler to optimize, and the three in the right column correspond to the original tasks which are hard to solve. Keeping these in mind, the curves can be analyzed as follows.

It can be observed from Fig.~\ref{figConvergenceA} that in terms of convergence curves of the aiding tasks, there is little difference between MTPCR, MTPCR-Intra, MTPCR-Inter and MTPCR-NKS in the first few generations, before they converge to different function values. The similar performances in the beginning can be ascribed to the same parameter settings, while different performances in the later stage are directed by the knowledge sharing mechanism used in the four methods. MFEA and GMFEA, although finds acceptable solutions at the end of their search process, is much slower while evolving. The fast convergence speed of our method may be ascribed to the use of independent population for solving each task, while MFEA and GMFEA both maintain a homogenous population and individuals for each task are distinguished by their skill factors. It is hard to tell whether MFEA or GMFEA performs better than the other from these curves, for they seem to entwine with each other along the evolutionary search. This somewhat indicates the decision variable translation strategy proposed in GMFEA is not that helpful for point cloud registration problems. Different knowledge sharing strategies do play an important role in influencing the evolutionary search process, but it is not obvious in optimizing the aiding tasks. Another point to be noticed in the figures is that MFEA and GMFEA starts from a relatively lower initial loss compared to our proposed method. The reason of this comes from that both MFEA and GMFEA maintain a single population along the evolutionary search, and hence their population size is six times greater than the population of each task in MTPCR to guarantee the same number of fitness evaluations in each generation. Therefore, the initial value of each task in MFEA and GMFEA is picked from 6$n$ individuals, while the initial value of MTPCR is picked from $n$ individuals only.

Performances of these methods diverge greatly when it comes to the optimization of the original tasks. In the right column of Fig.~\ref{figConvergenceA1} and Fig.~\ref{figConvergenceA2}, it can be seen that MTPCR consistently provides the best solutions and exceeds other methods by a large margin. MTPCR-Intra is always at the second place, with MFEA and GMFEA follows. MTPCR-Inter is only a little better than MTPCR-NKS, while both of them provide results far from satisfactory. Unlike smooth convergence curves of the aiding tasks, the cliffhanger results demonstrate the difficulty of optimizing the original tasks. Although MTPCR-NKS performs fairly well in aiding tasks, it shows inability in optimizing the original tasks due to the lack of knowledge sharing. The limited improvement of MTPCR-Inter comes from the increase of population diversity brought by inter-task knowledge sharing, but the designed inter-task knowledge sharing does not work out the way it is supposed to. The main reason is that qualities of solutions from the other two tasks to derive inter-task knowledge are also very poor, leading to some useless or even negative knowledge transfer. Best function values found by MFEA and GMFEA continue to go down after a plateau in the beginning, and it seems that high-quality solutions might be obtained if enough generations are available. However, much efficient methods are required due to limited computational resources. Both the consistent decrease and slow convergence of the two methods result from implicit knowledge transfer they use. On the one hand, implicit knowledge transfer through crossovers among tasks introduces more gene diversity, thus avoiding one task to be trapped in local optima to some extent. On the other hand, there is no guarantee that useful knowledge transfer prevails in the search process. In comparison, the performance of MTPCR-Intra, in which intra-task knowledge sharing is adopted, is much more attracting. Though not the best when optimizing aiding tasks, the results of MTPCR-Intra are satisfying. The best solutions of aiding tasks and their corresponding original tasks are near in the search space, so the direct transfer of solutions help much to solve the original tasks. The potential of MTPCR-Intra can be further explored when combined with MTPCR-Inter, and that is why the MTPCR can always perform the best. Solutions themselves for inter-task knowledge learning are poor when using MTPCR-Inter alone, but things become different when intra-task knowledge sharing involves. The incorporation of intra-task sharing guarantees a fast convergence of the original tasks, but do not ensure the global optimum for two reasons. One is that the optimums of aiding tasks and original tasks are not exactly the same, and the other is that the optimization of aiding tasks itself may also get stuck to local optima. With relatively good solutions in the later stage, positive knowledge sharing prevail in inter-task sharing, which continues to help the search process.

\subsection{Parameter Sensitivity}
\subsubsection{Influence of parameter $\alpha$ used in (\ref{equFinalForm})}
\begin{figure}[!tbh]
	\centering
	\subfloat[]{\label{fig_alpha_a}
		\begin{minipage}{0.32\columnwidth}
			\includegraphics[width=\columnwidth]{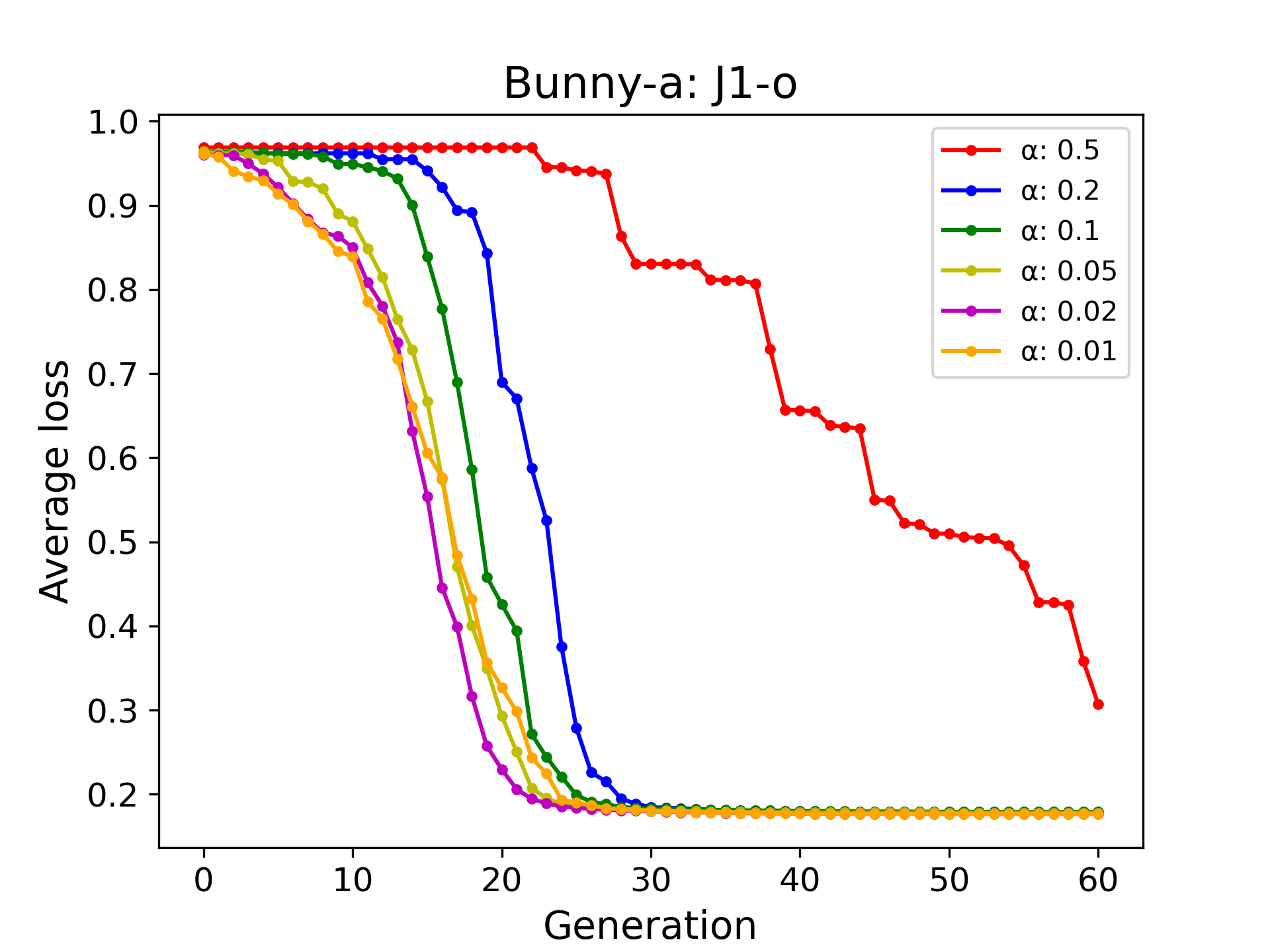}\vspace{1ex}
			\includegraphics[width=\columnwidth]{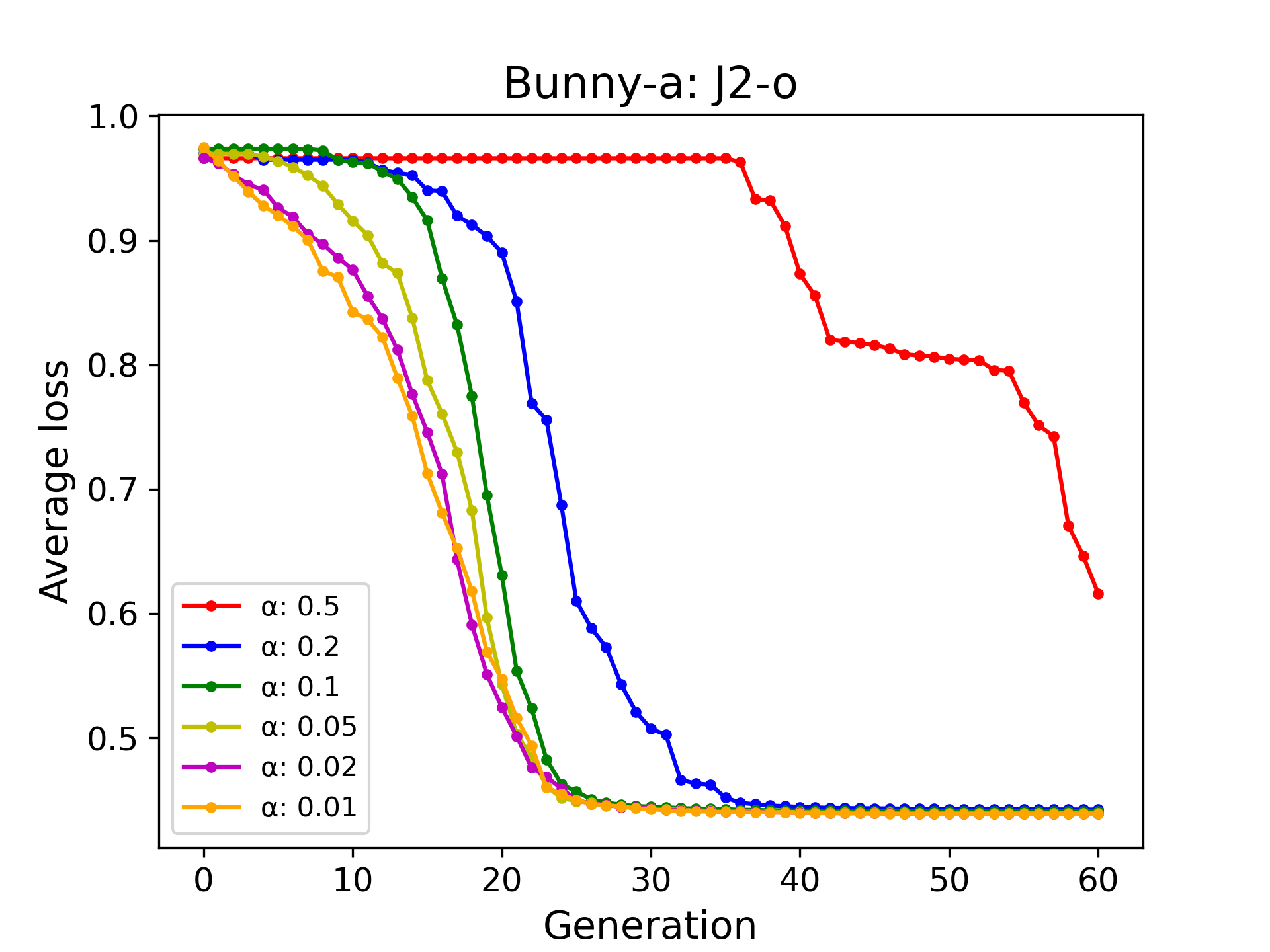}\vspace{1ex}
			\includegraphics[width=\columnwidth]{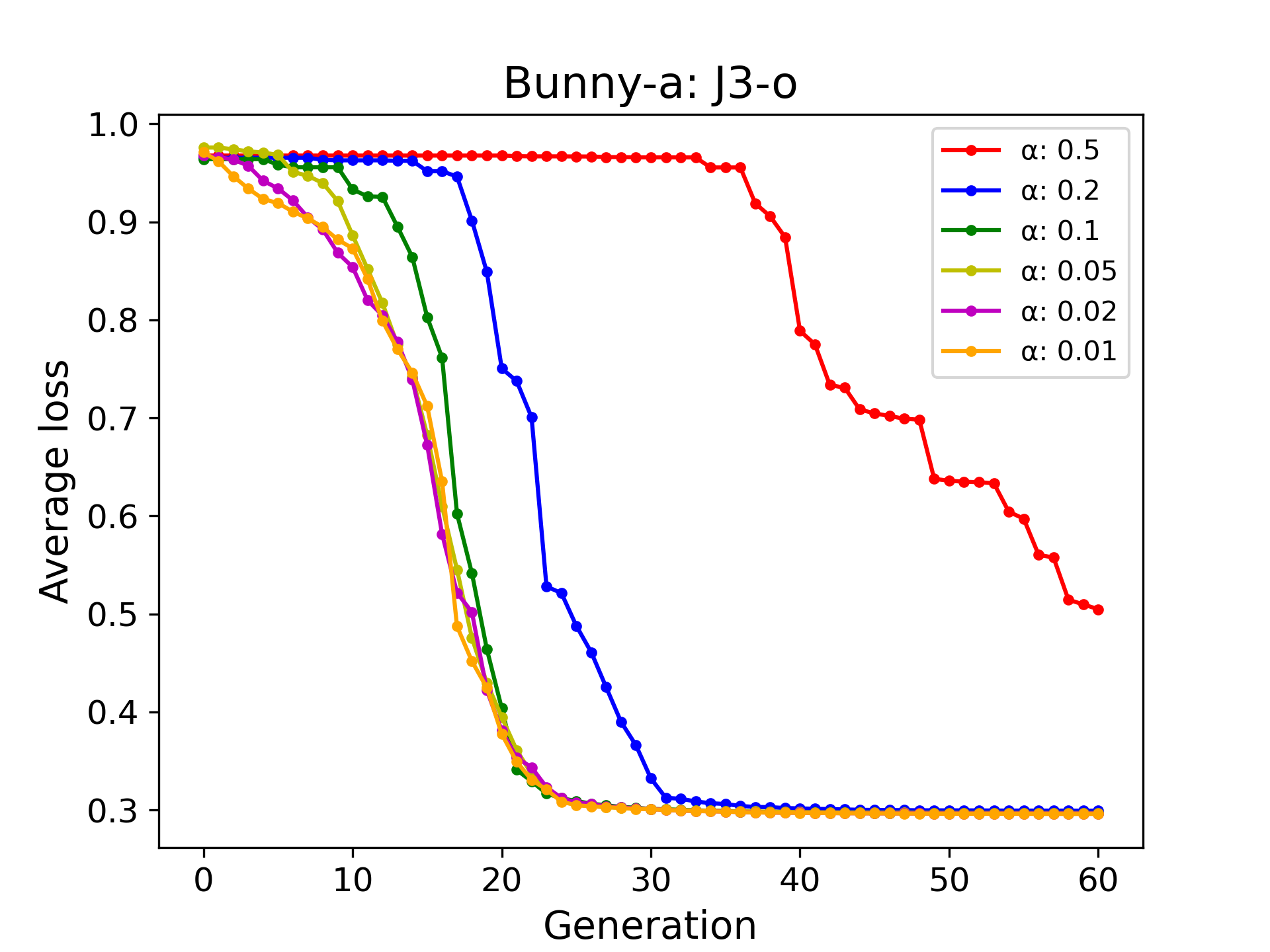}
		\end{minipage}
	}
	\subfloat[]{\label{fig_alpha_b}
		\begin{minipage}{0.32\columnwidth}
			\includegraphics[width=\columnwidth]{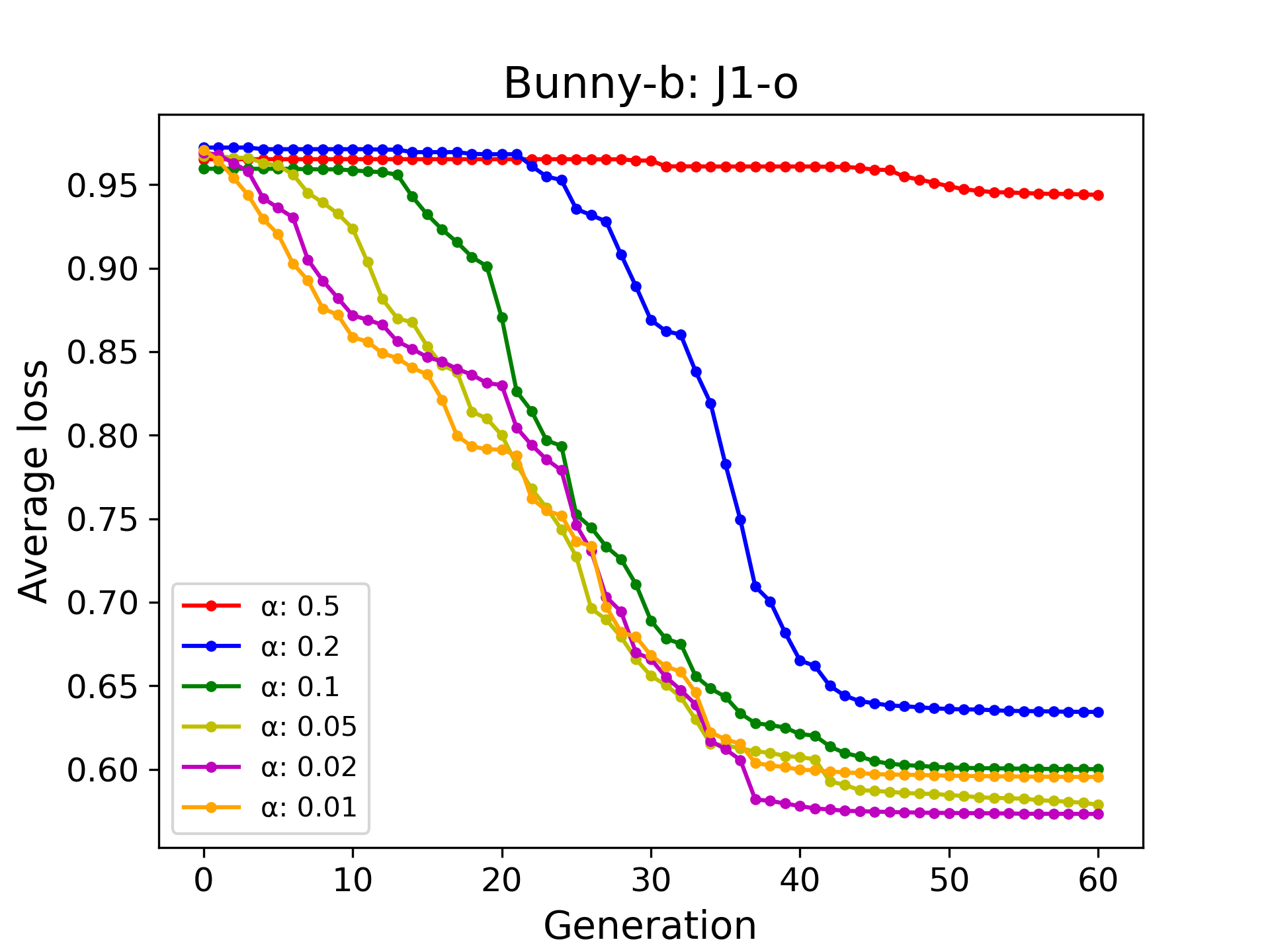}\vspace{1ex}
			\includegraphics[width=\columnwidth]{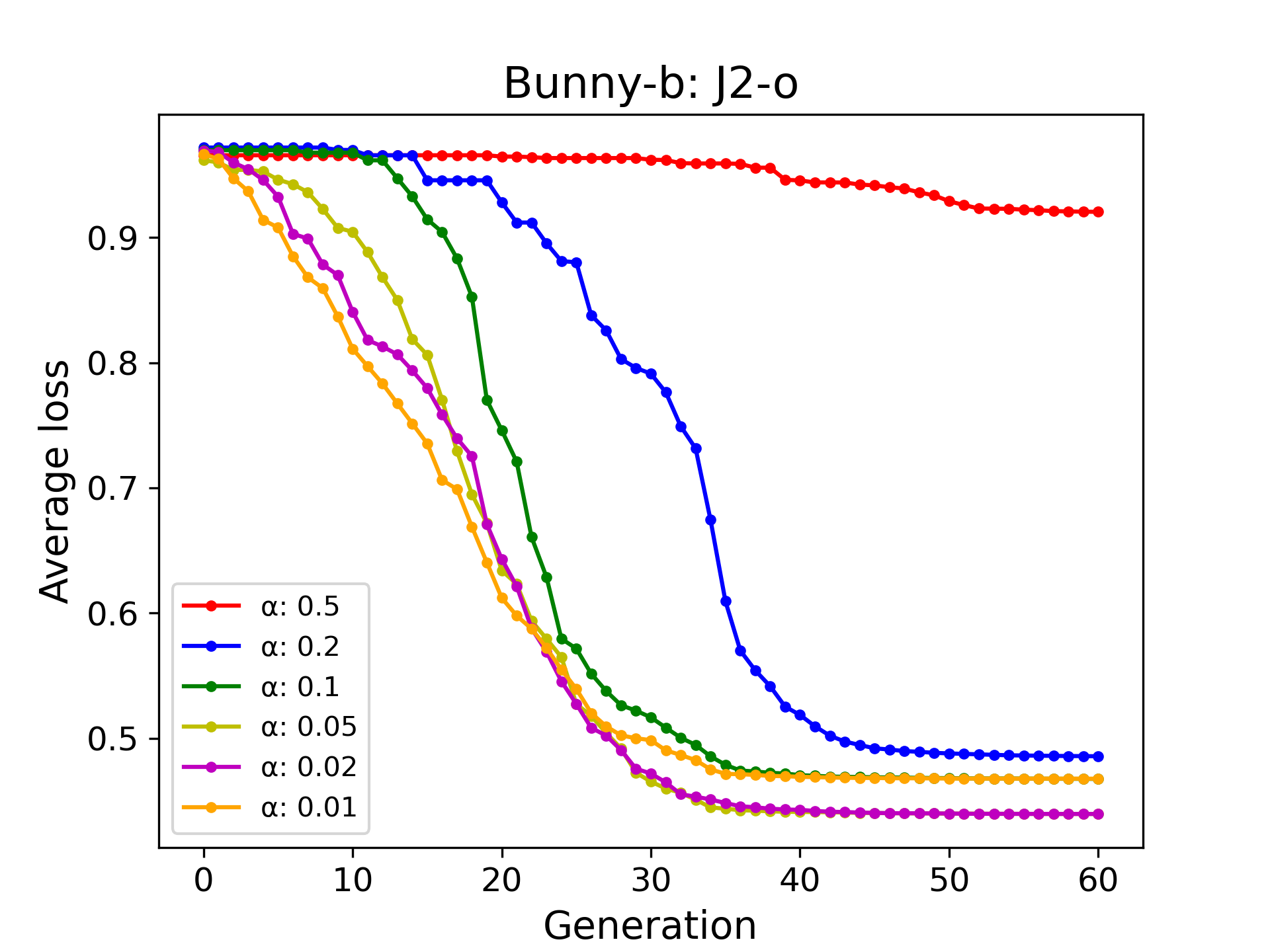}\vspace{1ex}
			\includegraphics[width=\columnwidth]{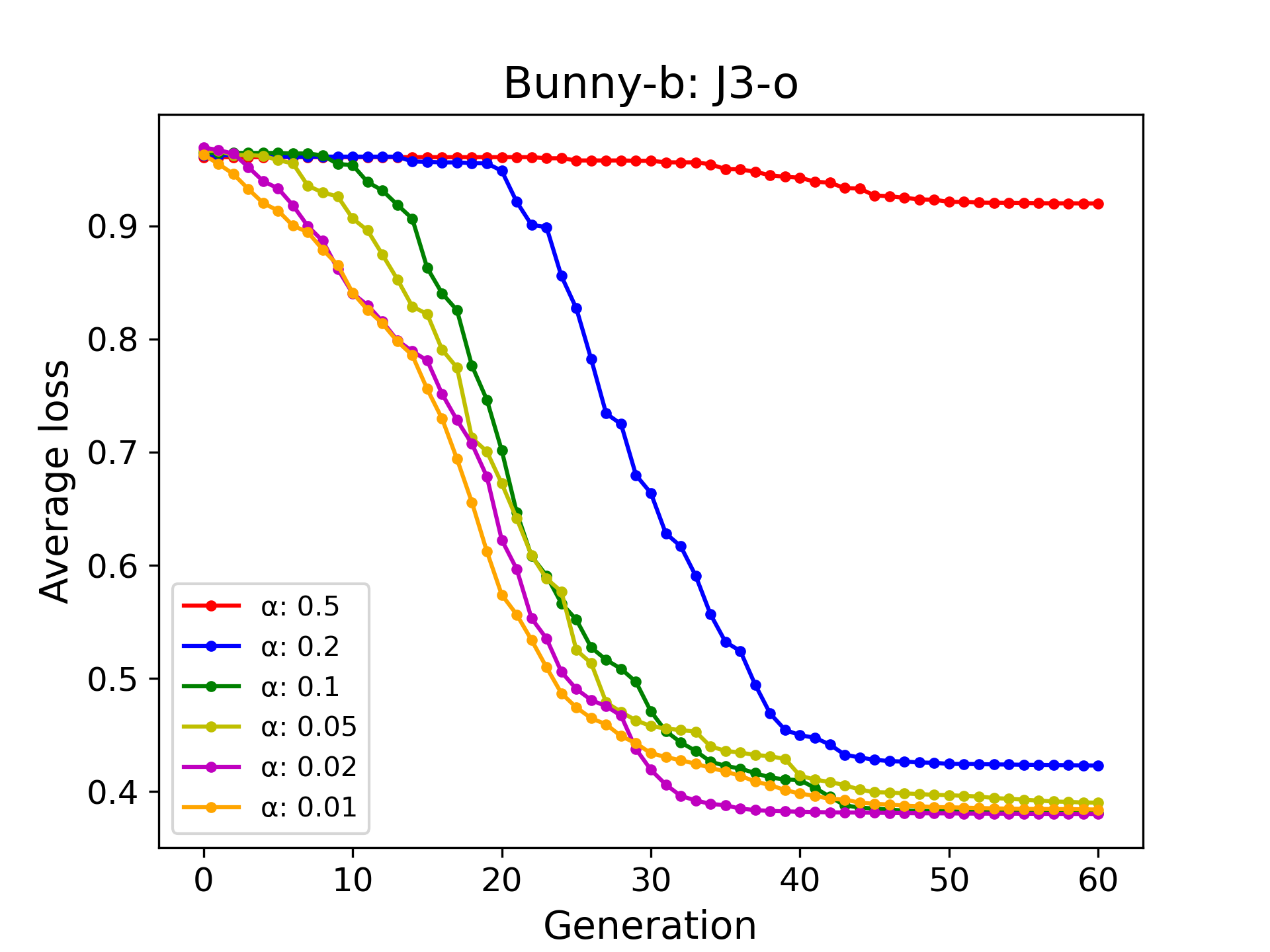}
		\end{minipage}
	}
	\subfloat[]{\label{fig_alpha_c}
		\begin{minipage}{0.32\columnwidth}
			\includegraphics[width=\columnwidth]{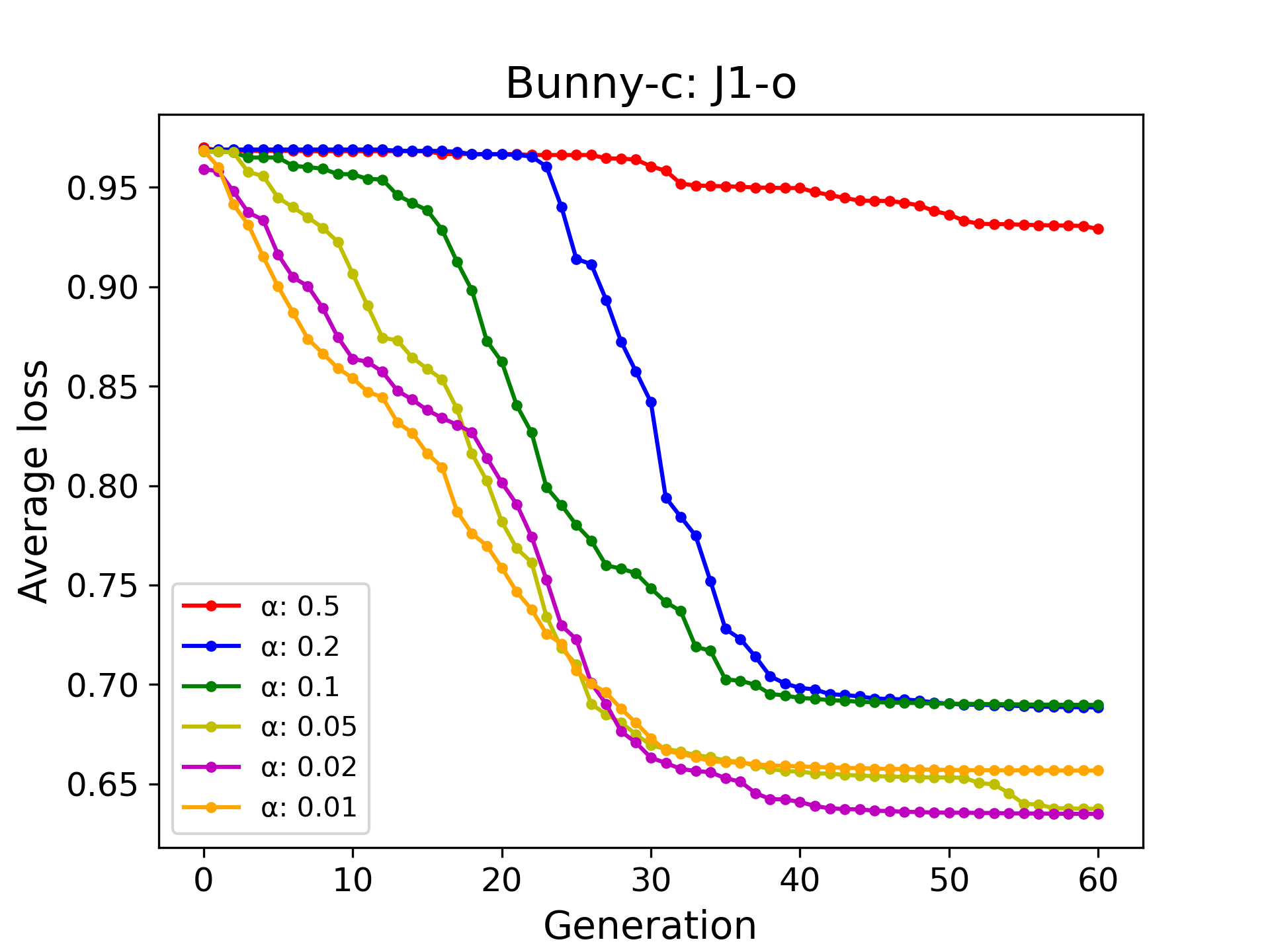}\vspace{1ex}
			\includegraphics[width=\columnwidth]{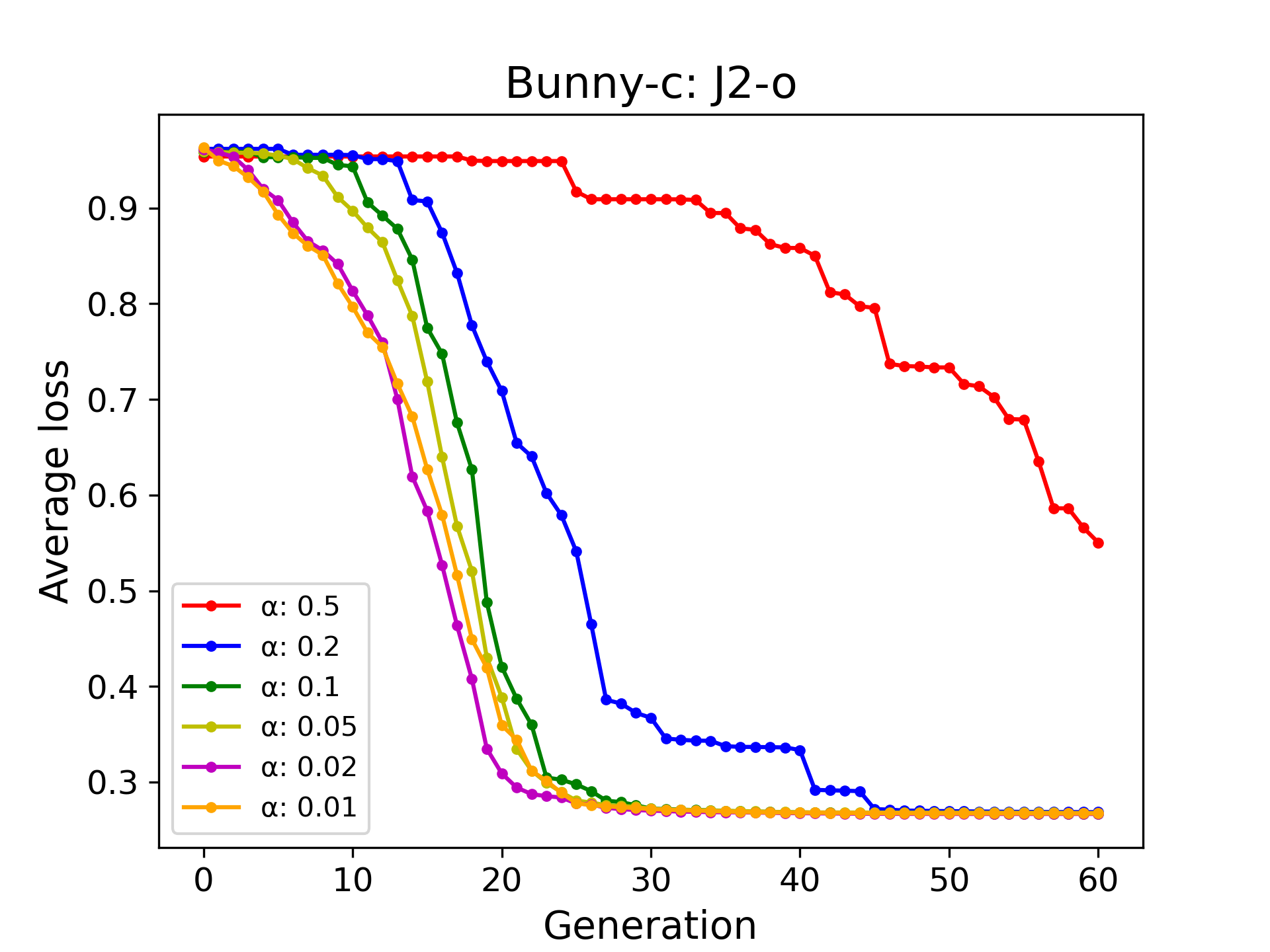}\vspace{1ex}
			\includegraphics[width=\columnwidth]{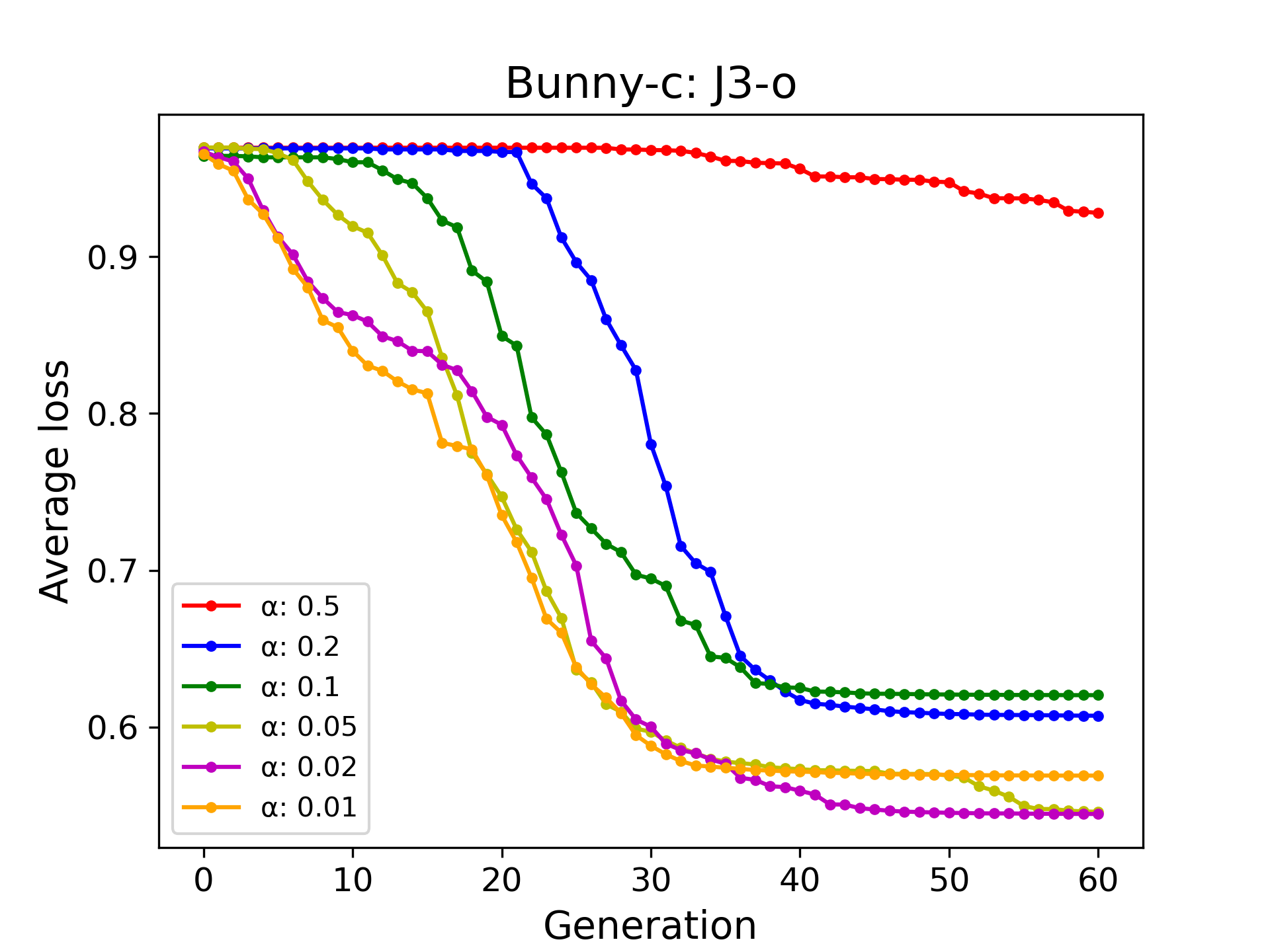}
		\end{minipage}
	}
	\caption{Convergence curves of the original tasks under different values of parameter $\alpha$ used in Equation~(\ref{equFinalForm}). y-axis: loss averaged over 20 independent runs; x-axis: Generation. (a)~Convergence curves of Bunny-a. (b)~Convergence curves of Bunny-b. (c)~Convergence curves of Bunny-c.}
	\label{figParameterAlpha}
\end{figure}

In the final form of fitness function, parameter $\alpha$ is used to tune the proportion of local accuracy and global consistency. The purpose of this is to balance pairwise registration result and the relationship among point clouds in multi-view registration. Therefore, $\alpha$ has a great impact on the performance of multi-view point cloud registration. Fig.~\ref{figParameterAlpha} plots the convergence curves of representative problems under different values of parameter $\alpha$ over 20 independent runs. It can be seen that when $\alpha$ equals 0.5, the convergence characteristics are quite unsatisfying, meaning that the global consistent constraint meddles to much in the evolutionary search. With the decrease of $\alpha$, convergence curves tend to be smoother. However, value 0.2 and 0.1 still seem to overemphasize the importance of global consistency, especially in the early stage. When $\alpha$ comes to 0.05 and below, there is no big difference whatever the value is. Table~\ref{tabParameterAlpha} tabulates the average registration errors under different values of $\alpha$ over 20 indepentend runs, and the results show that the best rotation errors are obtained when $\alpha$ equals 0.5, meaning a better balance compared to other $\alpha$ values. According to the above analysis, a good compromise between local accuracy and global consistency can be achieved when the value of $\alpha$ is around 0.05. Therefore, $\alpha$ equals 0.05 is used in the experiments of this paper.

\begin{table}[ht]
	\caption{Average Registration Errors under Different Values of Parameter $\alpha$ used in Equation~(\ref{equFinalForm}) over 20 Independent Runs. All Values Are Magnified One Thousand Times.}
	\label{tabParameterAlpha}
	\centering
	\renewcommand\arraystretch{1.1}
	\begin{tabular}{|l|l|l|l|l|l|l|}
		\hline
		\multirow{2}{*}{$\alpha$ value} &
		\multicolumn{2}{c|}{Bunny-a} &
		\multicolumn{2}{c|}{Bunny-b} &
		\multicolumn{2}{c|}{Bunny-c} \\
		\cline{2-7}
		& Err\_R & Err\_T & Err\_R & Err\_T & Err\_R & Err\_T \\
		\hline
		0.5	&	141.06 	&	9.78 	&	1657.15 	&	99.23 	&	1622.86 	&	71.65 \\
		0.2	&	6.92 	&	0.27 	&	260.17 	&	15.14 	&	303.77 	&	16.59 \\
		0.1	&	7.22 	&	0.37 	&	105.88 	&	4.09 	&	508.85 	&	21.11 \\
		0.05	&	\textbf{6.71} 	&	0.29 	&	\textbf{10.48} 	&	0.67 	&	\textbf{179.35} 	&	9.39 \\
		0.02	&	8.19 	&	0.32 	&	11.18 	&	0.57 	&	188.29 	&	9.83 \\
		0.01	&	7.72 	&	0.32 	&	105.25 	&	5.63 	&	314.65 	&	7.81 \\
		\hline
	\end{tabular}
\end{table}

\subsubsection{Influence of knowledge sharing probabilities}
\begin{figure}[!tbh]
	\centering
	\subfloat[]{\label{fig_trans_a}
		\begin{minipage}{0.48\columnwidth}
			\includegraphics[width=\columnwidth]{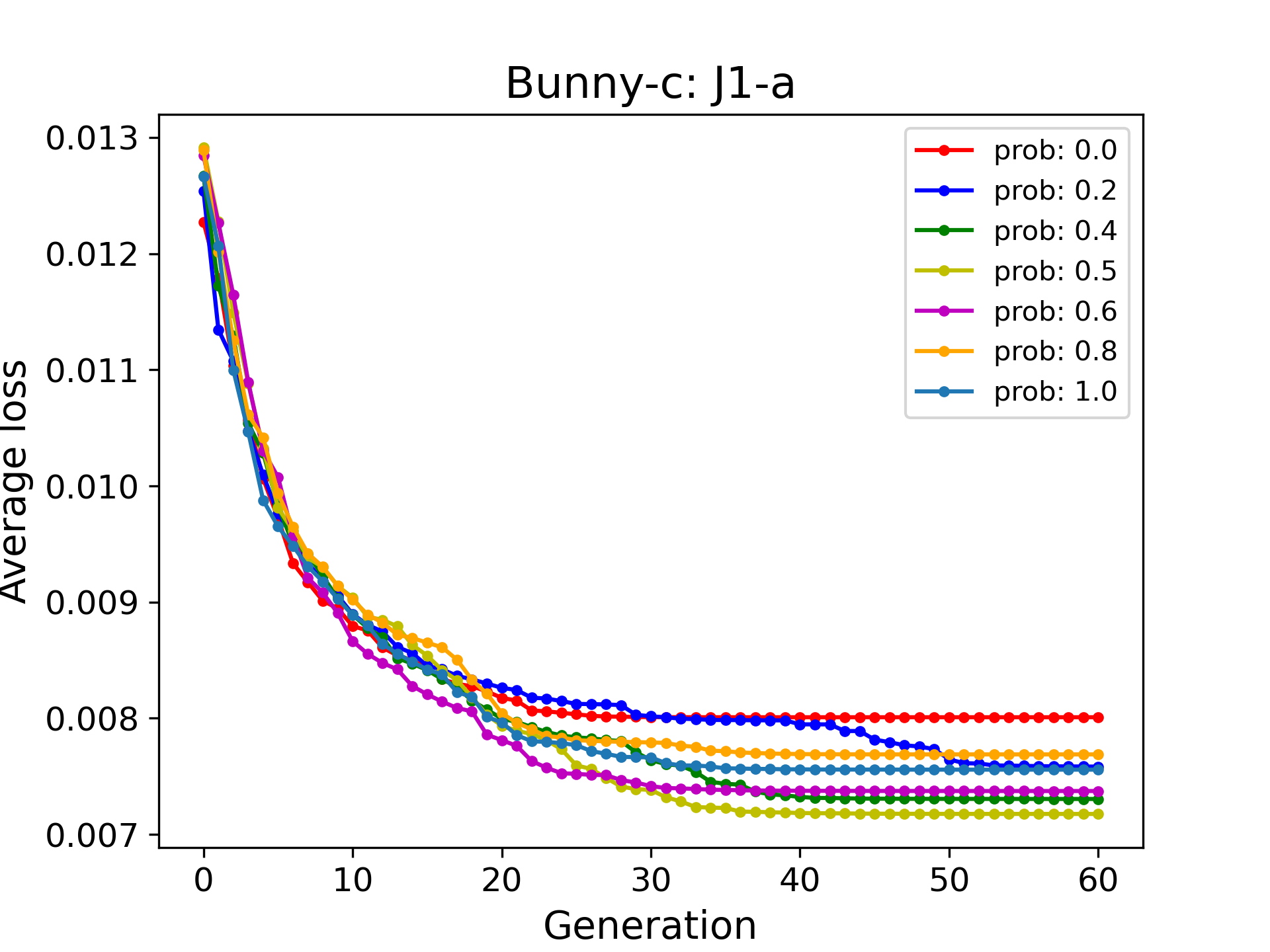}\vspace{1ex}
			\includegraphics[width=\columnwidth]{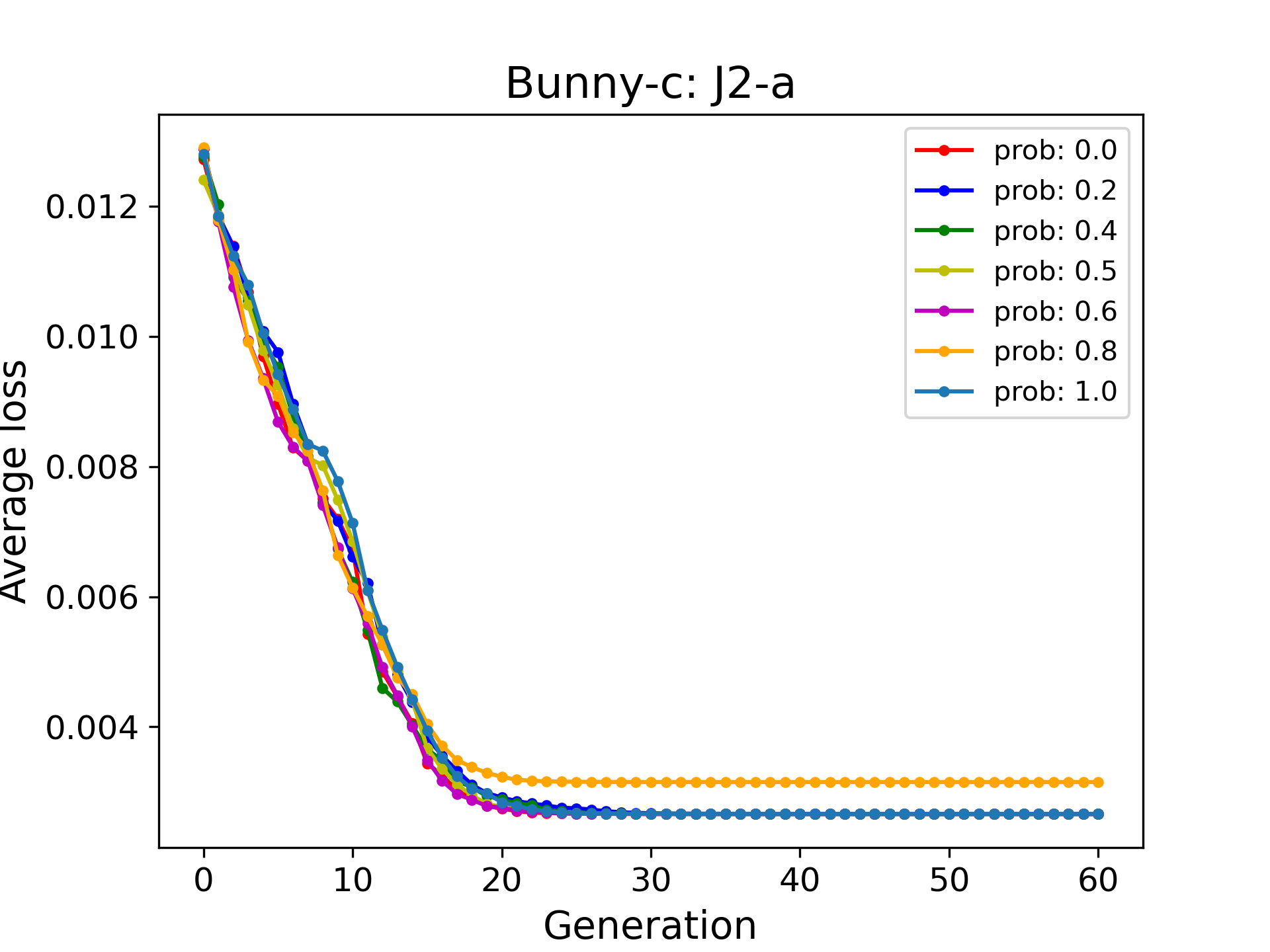}\vspace{1ex}
			\includegraphics[width=\columnwidth]{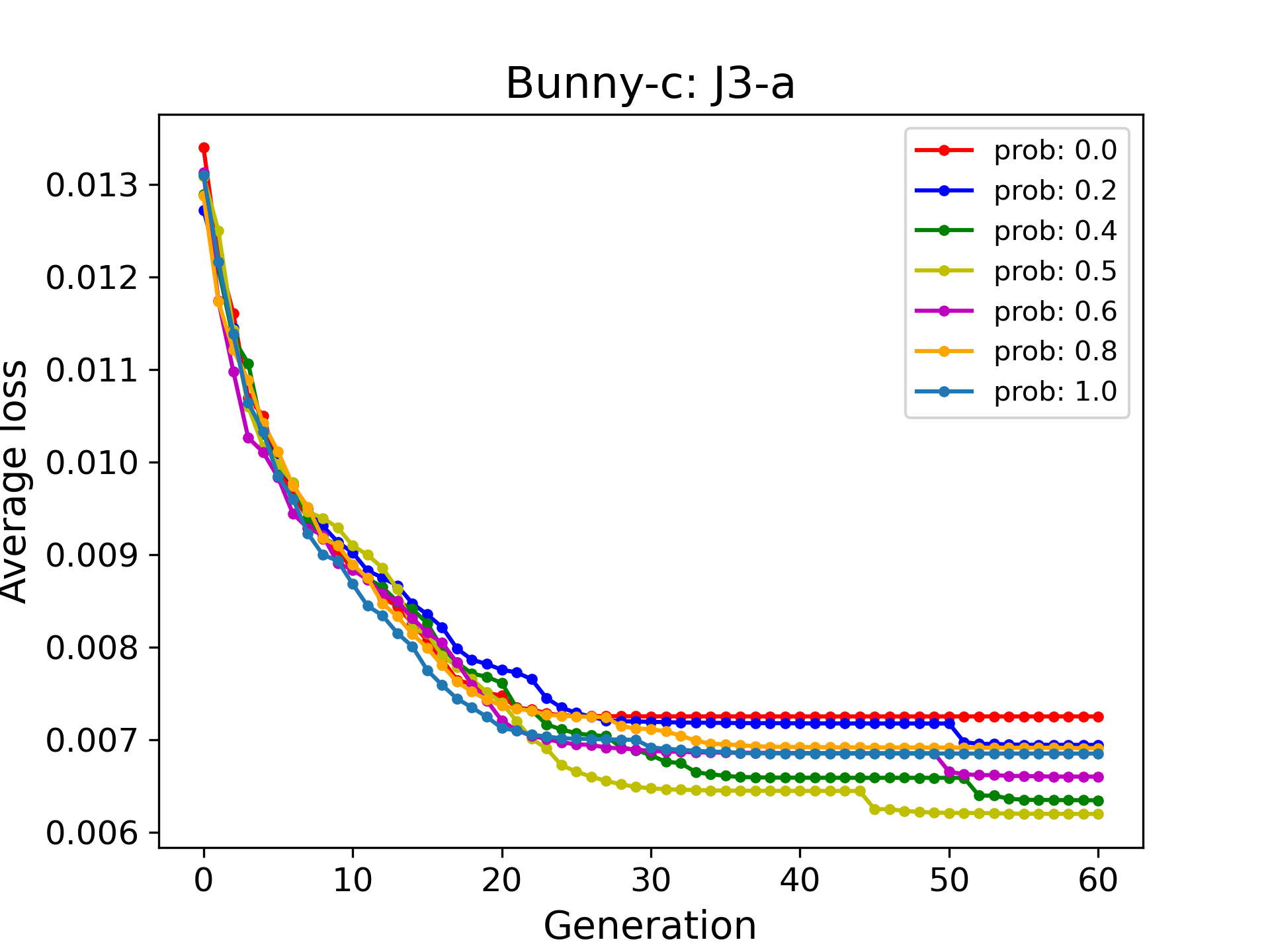}
		\end{minipage}
	}
	\subfloat[]{\label{fig_trans_o}
		\begin{minipage}{0.48\columnwidth}
			\includegraphics[width=\columnwidth]{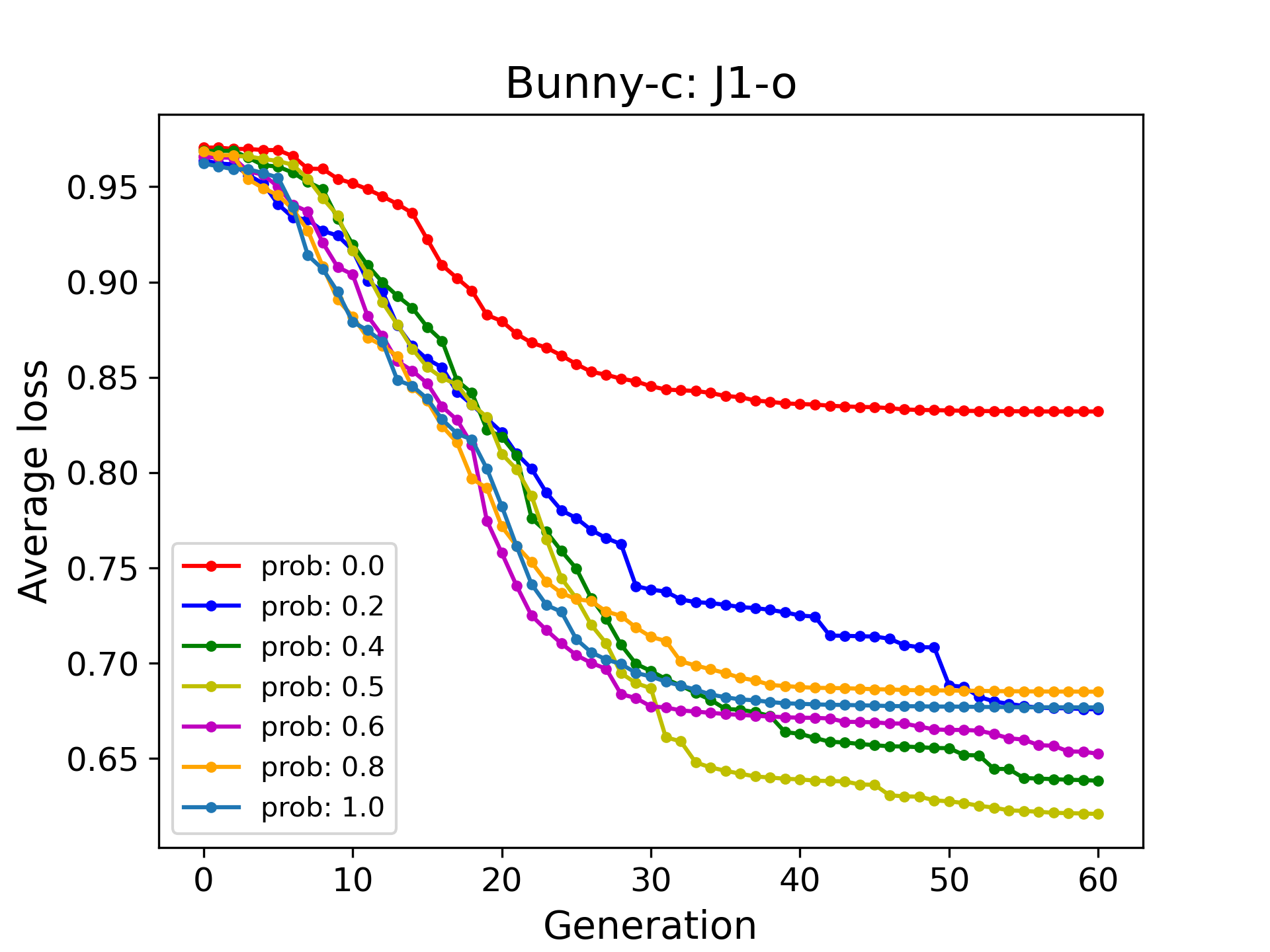}\vspace{1ex}
			\includegraphics[width=\columnwidth]{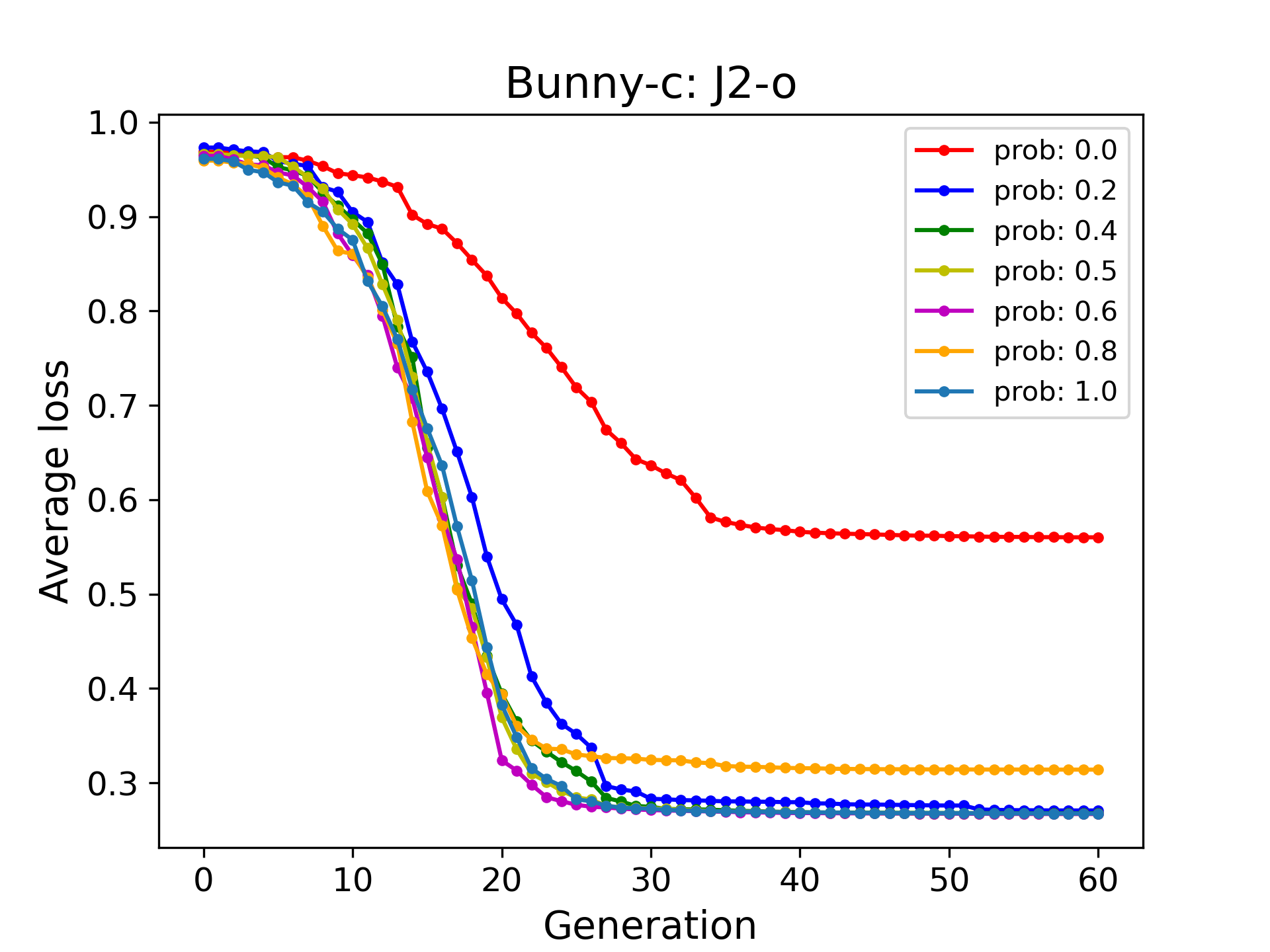}\vspace{1ex}
			\includegraphics[width=\columnwidth]{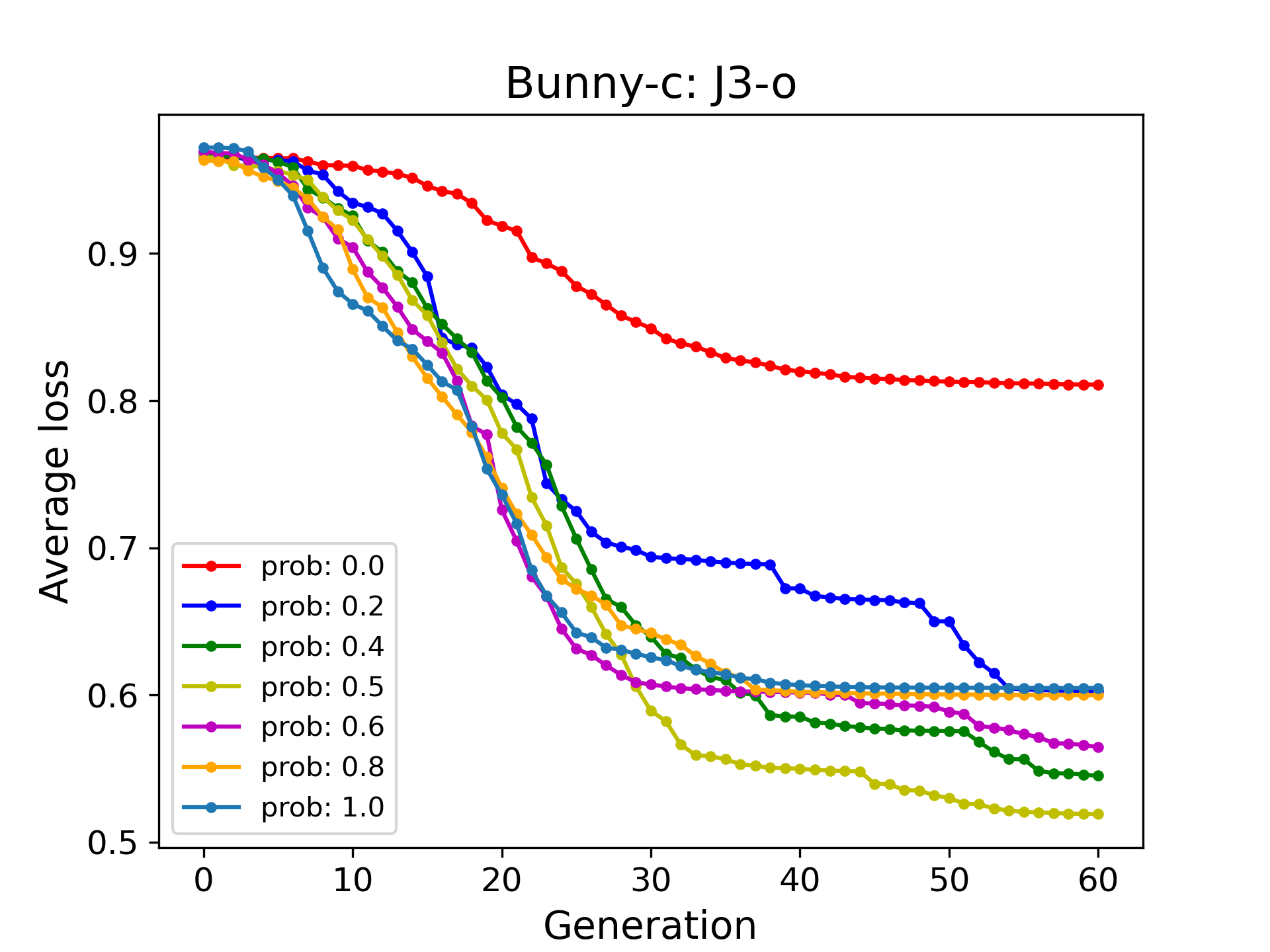}
		\end{minipage}
	}
	\caption{Convergence curves of Bunny-c under different values of intra-task sharing probability $p$-$intra$ and inter-task sharing probability $p$-$inter$. y-axis: loss averaged over 20 independent runs; x-axis: Generation. (a)~Convergence curves of aiding tasks. (b)~Convergence curves of original tasks.}
	\label{figParameterTransfer}
\end{figure}

The intra-task and inter-task knowledge sharing probabilities $p$-$intra$ and $p$-$inter$ are two key parameters of the proposed MTPCR. Therefore, it is reasonable to analyze how they affect the process of evolutionary search. For convenience, the two parameters are set identically in each run. Fig.~\ref{figParameterTransfer} shows the convergence curves of Bunny-c under different knowledge sharing probabilities. The value 0.0 means no knowledge sharing at all, value 1.0 means knowledge sharing happens in every generation, and other values in between. It can be seen that in terms of aiding tasks, there is no big difference whatever the sharing probability is in the first 20 generations, while convergence curves diverge in the later stage, meaning that different sharing probabilities do affect the evolutionary search. When it comes to the original tasks, the influence of varied probabilities tends to be much more apparent. When $p$-$intra$ and $p$-$inter$ are both 0.0, the search process ends up stucking to local optima due to the lack of information exchange. When they are both 0.2, situation turns better, but not good enough due to the insufficient information exchange. However, the search performance does not monotonously increase with the increase of sharing probability. When the probabilities are 1.0 and 0.8, convergence curves end up with similar loss values as that of 0.2, although they are much smoother. This can be ascribed to the excessive knowledge sharing which may meddle too much in the search process, disturbing the local search of one population. By contrast, values in the middle can make a better balance to make use of transferred knowledge and local search of one population. Therefore, sharing probability around 0.5 is preferred, and this paper uses 0.5 in the experiments.

\subsubsection{Influence of sample ratios}
\begin{figure}[!tbh]
	\centering
	\includegraphics[width=0.48\columnwidth]{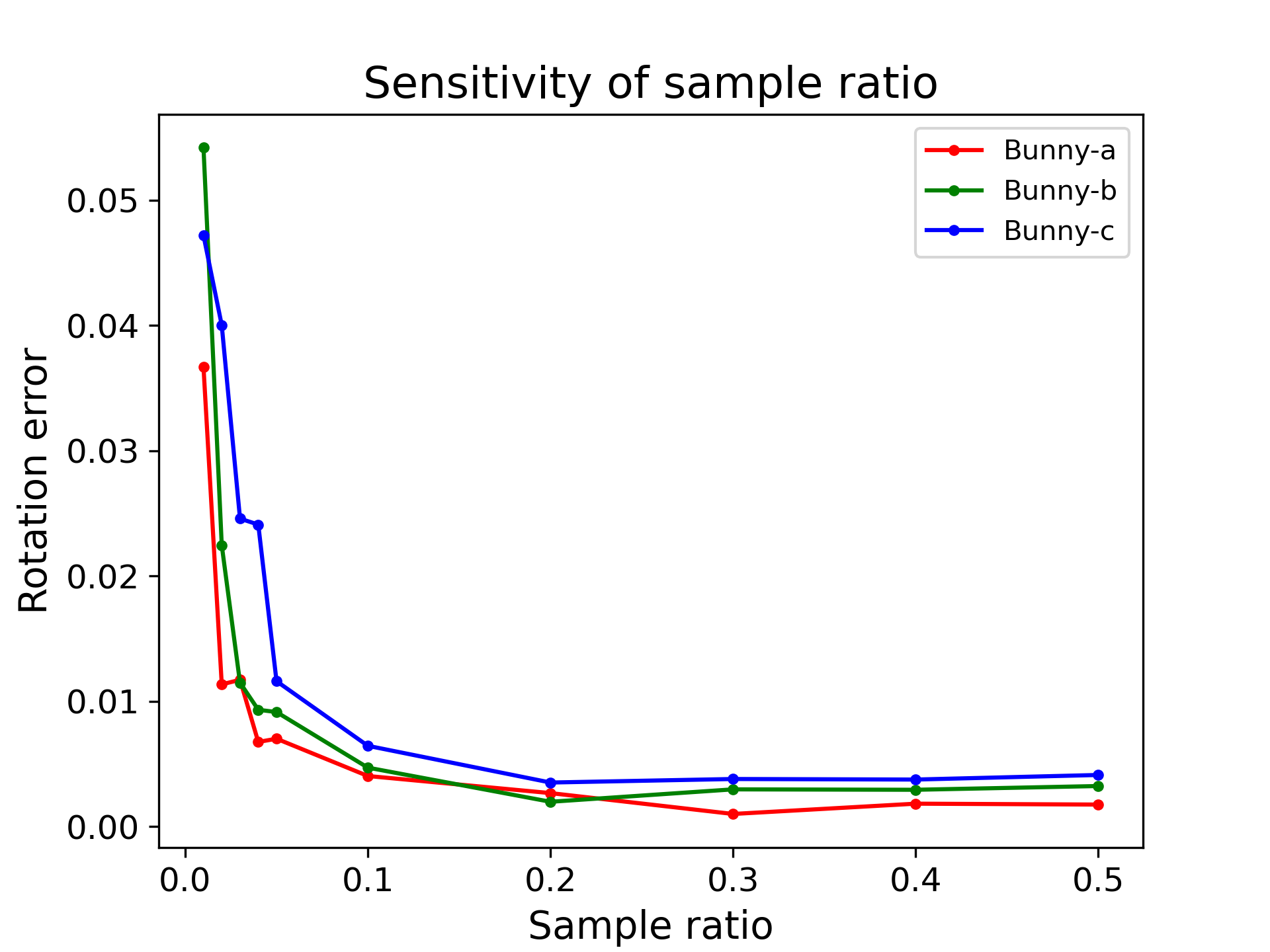}
	\includegraphics[width=0.48\columnwidth]{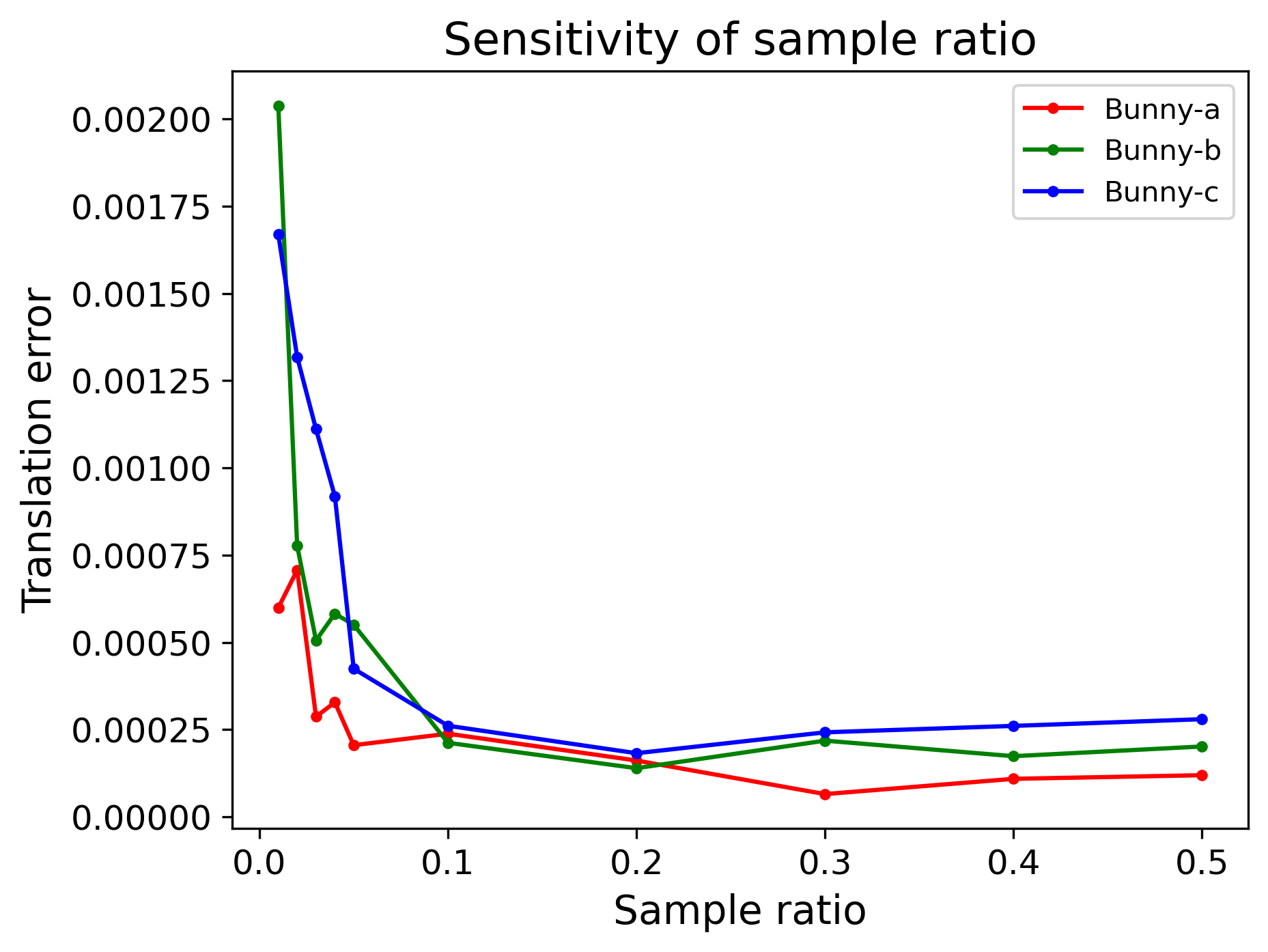}
	\caption{Rotation errors and translation errors under different sample ratios.}
	\label{figSamp}
\end{figure}

Equation~(\ref{equThreshold}) provides a way to calculate the threshold $\varepsilon$ automatically according to different point clouds to be registered, instead of manually set with the change of point clouds. Therefore, it is reasonable to investigate its practicality to different sample ratios. Fig.~\ref{figSamp} depicts the variation of rotation error and translation error with different point cloud sample ratios. The sample ratios from left to right are 0.01, 0.02, 0.03, 0.04, 0.05, 0.1, 0.2, 0.3, 0.4, 0.5 in the figure. It can be seen that when the sample ratio is greater than or equal to 0.1, there is little difference in both rotation error and translation error concerning different sample ratios. Rotation error and translation error grow a little when the sample ratio goes down to 0.05, before they grow rapidly with the decrease of sample ratios below 0.05. The results show that the way of calculating threshold $\varepsilon$ is not sensitive to sample ratios in a wide range, validating its practicality.

\section{Conclusion}
\label{secConclusion}
In this paper, an explicit evolutionary multitasking point cloud registration framework, MTPCR, is proposed. In particular, by simultaneously considering local accuracy and global consistency, a fitness function is derived, and the problem of multi-view point cloud registration is modeled as multi-task optimization. To concurrently solve the optimization tasks effectively and efficiently, a bi-channel knowledge sharing mechanism is developed. The proposed intra-task knowledge sharing guarantees a fast convergence speed by knowledge transfer between aiding tasks and original tasks, and the inter-task knowledge sharing helps to avoid local optima by exploring commonalities among tasks. The two knowledge sharing mechanisms combined together are shown to be of great importance for problem solving. Comprehensive experiments are conducted on model point clouds as well as scene point clouds with varied overlap ratios and different noise levels, confirming the utility of both the fitness function and the proposed bi-channel knowledge sharing mechanism.

\nocite{*}
\bibliography{ref}

\begin{thebibliography}{10}
\providecommand{\url}[1]{#1}
\csname url@samestyle\endcsname
\providecommand{\newblock}{\relax}
\providecommand{\bibinfo}[2]{#2}
\providecommand{\BIBentrySTDinterwordspacing}{\spaceskip=0pt\relax}
\providecommand{\BIBentryALTinterwordstretchfactor}{4}
\providecommand{\BIBentryALTinterwordspacing}{\spaceskip=\fontdimen2\font plus
\BIBentryALTinterwordstretchfactor\fontdimen3\font minus
  \fontdimen4\font\relax}
\providecommand{\BIBforeignlanguage}[2]{{%
\expandafter\ifx\csname l@#1\endcsname\relax
\typeout{** WARNING: IEEEtran.bst: No hyphenation pattern has been}%
\typeout{** loaded for the language `#1'. Using the pattern for}%
\typeout{** the default language instead.}%
\else
\language=\csname l@#1\endcsname
\fi
#2}}
\providecommand{\BIBdecl}{\relax}
\BIBdecl

\bibitem{dai2017scannet}
A.~Dai, A.~X. Chang, M.~Savva, M.~Halber, T.~Funkhouser, and M.~Nie{\ss}ner,
  ``Scannet: Richly-annotated 3d reconstructions of indoor scenes,'' in
  \emph{IEEE conference on computer vision and pattern recognition}, 2017, pp.
  5828--5839.

\bibitem{8099512}
A.~Zeng, S.~Song, M.~Nießner, M.~Fisher, J.~Xiao, and T.~Funkhouser,
  ``3dmatch: Learning local geometric descriptors from rgb-d reconstructions,''
  in \emph{IEEE Conference on Computer Vision and Pattern Recognition}, 2017,
  pp. 199--208.

\bibitem{7368945}
J.~Yang, H.~Li, D.~Campbell, and Y.~Jia, ``Go-icp: A globally optimal solution
  to 3d icp point-set registration,'' \emph{IEEE Transactions on Pattern
  Analysis and Machine Intelligence}, vol.~38, no.~11, pp. 2241--2254, 2016.

\bibitem{7918612}
H.~Lei, G.~Jiang, and L.~Quan, ``Fast descriptors and correspondence
  propagation for robust global point cloud registration,'' \emph{IEEE
  Transactions on Image Processing}, vol.~26, no.~8, pp. 3614--3623, 2017.

\bibitem{7298781}
F.~Yu, J.~Xiao, and T.~Funkhouser, ``Semantic alignment of lidar data at city
  scale,'' in \emph{IEEE Conference on Computer Vision and Pattern
  Recognition}, 2015, pp. 1722--1731.

\bibitem{8461063}
A.~L. Pavlov, G.~W. Ovchinnikov, D.~Y. Derbyshev, D.~Tsetserukou, and I.~V.
  Oseledets, ``Aa-icp: Iterative closest point with anderson acceleration,'' in
  \emph{IEEE International Conference on Robotics and Automation}, 2018, pp.
  3407--3412.

\bibitem{7161358}
A.~Gupta, Y.-S. Ong, and L.~Feng, ``Multifactorial evolution: Toward
  evolutionary multitasking,'' \emph{IEEE Transactions on Evolutionary
  Computation}, vol.~20, no.~3, pp. 343--357, 2016.

\bibitem{Ong2016EvolutionaryMA}
Y.~Ong and A.~Gupta, ``Evolutionary multitasking: A computer science view of
  cognitive multitasking,'' \emph{Cognitive Computation}, vol.~8, pp. 125--142,
  2016.

\bibitem{7850039}
L.~Zhou, L.~Feng, J.~Zhong, Y.-S. Ong, Z.~Zhu, and E.~Sha, ``Evolutionary
  multitasking in combinatorial search spaces: A case study in capacitated
  vehicle routing problem,'' in \emph{IEEE Symposium Series on Computational
  Intelligence}, 2016, pp. 1--8.

\bibitem{8401802}
L.~Feng, L.~Zhou, J.~Zhong, A.~Gupta, Y.-S. Ong, K.-C. Tan, and A.~K. Qin,
  ``Evolutionary multitasking via explicit autoencoding,'' \emph{IEEE
  Transactions on Cybernetics}, vol.~49, no.~9, pp. 3457--3470, 2019.

\bibitem{7464295}
A.~Gupta, Y.-S. Ong, L.~Feng, and K.~C. Tan, ``Multiobjective multifactorial
  optimization in evolutionary multitasking,'' \emph{IEEE Transactions on
  Cybernetics}, vol.~47, no.~7, pp. 1652--1665, 2017.

\bibitem{121791}
P.~Besl and N.~D. McKay, ``A method for registration of 3-d shapes,''
  \emph{IEEE Transactions on Pattern Analysis and Machine Intelligence},
  vol.~14, no.~2, pp. 239--256, 1992.

\bibitem{924423}
S.~Rusinkiewicz and M.~Levoy, ``Efficient variants of the icp algorithm,'' in
  \emph{International Conference on 3-D Digital Imaging and Modeling}, 2001,
  pp. 145--152.

\bibitem{Pomerleau2015ARO}
F.~Pomerleau, F.~Colas, and R.~Y. Siegwart, ``A review of point cloud
  registration algorithms for mobile robotics,'' \emph{Foundations and Trends®
  in Robotics}, vol.~4, pp. 1--104, 2015.

\bibitem{CHETVERIKOV2005299}
D.~Chetverikov, D.~Stepanov, and P.~Krsek, ``Robust euclidean alignment of 3d
  point sets: the trimmed iterative closest point algorithm,'' \emph{Image and
  Vision Computing}, vol.~23, no.~3, pp. 299--309, 2005.

\bibitem{Rusu2008PersistentPF}
R.~B. Rusu, Z.~C. Marton, N.~Blodow, and M.~Beetz, ``Persistent point feature
  histograms for 3d point clouds,'' in \emph{International Conference on
  Intelligent Autonomous Systems}, 2008, pp. 119--128.

\bibitem{5152473}
R.~B. Rusu, N.~Blodow, and M.~Beetz, ``Fast point feature histograms (fpfh) for
  3d registration,'' in \emph{IEEE International Conference on Robotics and
  Automation}, 2009, pp. 3212--3217.

\bibitem{5457637}
Y.~Zhong, ``Intrinsic shape signatures: A shape descriptor for 3d object
  recognition,'' in \emph{IEEE International Conference on Computer Vision
  Workshops}, 2009, pp. 689--696.

\bibitem{Fischler1981RandomSC}
M.~A. Fischler and R.~C. Bolles, ``Random sample consensus: a paradigm for
  model fitting with applications to image analysis and automated
  cartography,'' \emph{Communications of the ACM}, vol.~24, pp. 381--395, 1981.

\bibitem{agenerativemodel}
G.~D. Evangelidis, D.~Kounades-Bastian, R.~Horaud, and E.~Z. Psarakis, ``A
  generative model for the joint registration of multiple point sets,'' in
  \emph{European Conference on Computer Vision}, 2014, pp. 109--122.

\bibitem{7954698}
G.~D. Evangelidis and R.~Horaud, ``Joint alignment of multiple point sets with
  batch and incremental expectation-maximization,'' \emph{IEEE Transactions on
  Pattern Analysis and Machine Intelligence}, vol.~40, no.~6, pp. 1397--1410,
  2018.

\bibitem{9201412}
J.~Zhu, R.~Guo, Z.~Li, J.~Zhang, and S.~Pang, ``Registration of multi-view
  point sets under the perspective of expectation-maximization,'' \emph{IEEE
  Transactions on Image Processing}, vol.~29, pp. 9176--9189, 2020.

\bibitem{9521221}
Y.~Zhang, P.~Tiňo, A.~Leonardis, and K.~Tang, ``A survey on neural network
  interpretability,'' \emph{IEEE Transactions on Emerging Topics in
  Computational Intelligence}, vol.~5, no.~5, pp. 726--742, 2021.

\bibitem{aoki2019pointnetlk}
Y.~Aoki, H.~Goforth, R.~A. Srivatsan, and S.~Lucey, ``Pointnetlk: Robust \&
  efficient point cloud registration using pointnet,'' in \emph{IEEE/CVF
  Conference on Computer Vision and Pattern Recognition}, 2019, pp. 7163--7172.

\bibitem{9009450}
W.~Lu, G.~Wan, Y.~Zhou, X.~Fu, P.~Yuan, and S.~Song, ``Deepvcp: An end-to-end
  deep neural network for point cloud registration,'' in \emph{IEEE/CVF
  International Conference on Computer Vision}, 2019, pp. 12--21.

\bibitem{9157132}
Z.~J. Yew and G.~H. Lee, ``Rpm-net: Robust point matching using learned
  features,'' in \emph{IEEE/CVF Conference on Computer Vision and Pattern
  Recognition}, 2020, pp. 11\,821--11\,830.

\bibitem{9577334}
S.~Huang, Z.~Gojcic, M.~Usvyatsov, A.~Wieser, and K.~Schindler, ``Predator:
  Registration of 3d point clouds with low overlap,'' in \emph{IEEE/CVF
  Conference on Computer Vision and Pattern Recognition}, 2021, pp. 4265--4274.

\bibitem{LOMONOSOV20061201}
E.~Lomonosov, D.~Chetverikov, and A.~Ek{\'a}rt, ``Pre-registration of
  arbitrarily oriented 3d surfaces using a genetic algorithm,'' \emph{Pattern
  Recognition Letters}, vol.~27, no.~11, pp. 1201--1208, 2006.

\bibitem{SILVA2007114}
L.~Silva, O.~R.~P. Bellon, and K.~L. Boyer, ``Multiview range image
  registration using the surface interpenetration measure,'' \emph{Image and
  Vision Computing}, vol.~25, no.~1, pp. 114--125, 2007.

\bibitem{Zhu2014RobustRO}
J.~Zhu, D.~Meng, Z.~Li, S.~Du, and Z.~Yuan, ``Robust registration of partially
  overlapping point sets via genetic algorithm with growth operator,''
  \emph{IET Image Processing}, vol.~8, no.~10, pp. 582--590, 2014.

\bibitem{s17091979}
L.~Yan, J.~Tan, H.~Liu, H.~Xie, and C.~Chen, ``Automatic registration of
  tls-tls and tls-mls point clouds using a genetic algorithm,'' \emph{Sensors},
  vol.~17, no.~9, p. 1979, 2017.

\bibitem{8114198}
A.~Gupta, Y.-S. Ong, and L.~Feng, ``Insights on transfer optimization: Because
  experience is the best teacher,'' \emph{IEEE Transactions on Emerging Topics
  in Computational Intelligence}, vol.~2, no.~1, pp. 51--64, 2018.

\bibitem{8616832}
M.~Gong, Z.~Tang, H.~Li, and J.~Zhang, ``Evolutionary multitasking with dynamic
  resource allocating strategy,'' \emph{IEEE Transactions on Evolutionary
  Computation}, vol.~23, no.~5, pp. 858--869, 2019.

\bibitem{8405560}
J.~Wang, B.~Cen, S.~Gao, Z.~Zhang, and Y.~Zhou, ``Cooperative evolutionary
  framework with focused search for many-objective optimization,'' \emph{IEEE
  Transactions on Emerging Topics in Computational Intelligence}, vol.~4,
  no.~3, pp. 398--412, 2020.

\bibitem{8666053}
X.~Zheng, A.~K. Qin, M.~Gong, and D.~Zhou, ``Self-regulated evolutionary
  multitask optimization,'' \emph{IEEE Transactions on Evolutionary
  Computation}, vol.~24, no.~1, pp. 16--28, 2020.

\bibitem{7969454}
K.~K. Bali, A.~Gupta, L.~Feng, Y.~S. Ong, and T.~P. Siew, ``Linearized domain
  adaptation in evolutionary multitasking,'' in \emph{IEEE Congress on
  Evolutionary Computation}, 2017, pp. 1295--1302.

\bibitem{7969596}
Y.-W. Wen and C.-K. Ting, ``Parting ways and reallocating resources in
  evolutionary multitasking,'' in \emph{IEEE Congress on Evolutionary
  Computation}, 2017, pp. 2404--2411.

\bibitem{astudyofsim}
L.~Zhou, L.~Feng, J.~Zhong, Z.~Zhu, B.~Da, and Z.~Wu, ``A study of similarity
  measure between tasks for multifactorial evolutionary algorithm,'' in
  \emph{Proceedings of the Genetic and Evolutionary Computation Conference
  Companion}, 2018, pp. 229--230.

\bibitem{8231172}
J.~Ding, C.~Yang, Y.~Jin, and T.~Chai, ``Generalized multitasking for
  evolutionary optimization of expensive problems,'' \emph{IEEE Transactions on
  Evolutionary Computation}, vol.~23, no.~1, pp. 44--58, 2019.

\bibitem{8727933}
Y.~Chen, J.~Zhong, L.~Feng, and J.~Zhang, ``An adaptive archive-based
  evolutionary framework for many-task optimization,'' \emph{IEEE Transactions
  on Emerging Topics in Computational Intelligence}, vol.~4, no.~3, pp.
  369--384, 2020.

\bibitem{9027113}
L.~Zhou, L.~Feng, K.~C. Tan, J.~Zhong, Z.~Zhu, K.~Liu, and C.~Chen, ``Toward
  adaptive knowledge transfer in multifactorial evolutionary computation,''
  \emph{IEEE Transactions on Cybernetics}, vol.~51, no.~5, pp. 2563--2576,
  2021.

\bibitem{Li2020MultifactorialOV}
G.~Li, Q.~Lin, and W.~Gao, ``Multifactorial optimization via explicit
  multipopulation evolutionary framework,'' \emph{Information Sciences}, vol.
  512, pp. 1555--1570, 2020.

\bibitem{8967000}
D.~Wu and X.~Tan, ``Multitasking genetic algorithm (mtga) for fuzzy system
  optimization,'' \emph{IEEE Transactions on Fuzzy Systems}, vol.~28, no.~6,
  pp. 1050--1061, 2020.

\bibitem{9195010}
Z.~Tang, M.~Gong, Y.~Wu, W.~Liu, and Y.~Xie, ``Regularized evolutionary
  multitask optimization: Learning to intertask transfer in aligned subspace,''
  \emph{IEEE Transactions on Evolutionary Computation}, vol.~25, no.~2, pp.
  262--276, 2021.

\bibitem{9385398}
C.~Wang, J.~Liu, K.~Wu, and Z.~Wu, ``Solving multi-task optimization problems
  with adaptive knowledge transfer via anomaly detection,'' \emph{IEEE
  Transactions on Evolutionary Computation}, pp. 1--1, 2021.

\bibitem{9023952}
L.~Feng, Y.~Huang, L.~Zhou, J.~Zhong, A.~Gupta, K.~Tang, and K.~C. Tan,
  ``Explicit evolutionary multitasking for combinatorial optimization: A case
  study on capacitated vehicle routing problem,'' \emph{IEEE Transactions on
  Cybernetics}, vol.~51, no.~6, pp. 3143--3156, 2021.

\bibitem{7947122}
L.~Zhang, H.~Pan, Y.~Su, X.~Zhang, and Y.~Niu, ``A mixed representation-based
  multiobjective evolutionary algorithm for overlapping community detection,''
  \emph{IEEE Transactions on Cybernetics}, vol.~47, no.~9, pp. 2703--2716,
  2017.

\bibitem{8338097}
J.~Wang, T.~Weng, and Q.~Zhang, ``A two-stage multiobjective evolutionary
  algorithm for multiobjective multidepot vehicle routing problem with time
  windows,'' \emph{IEEE Transactions on Cybernetics}, vol.~49, no.~7, pp.
  2467--2478, 2019.

\bibitem{Herrera2003ATF}
F.~Herrera, M.~Lozano, and A.~M. S{\'a}nchez, ``A taxonomy for the crossover
  operator for real-coded genetic algorithms: An experimental study,''
  \emph{International Journal of Intelligent Systems}, vol.~18, no.~3, pp.
  309--338, 2003.

\bibitem{4632146}
A.~K. Qin, V.~L. Huang, and P.~N. Suganthan, ``Differential evolution algorithm
  with strategy adaptation for global numerical optimization,'' \emph{IEEE
  Transactions on Evolutionary Computation}, vol.~13, no.~2, pp. 398--417,
  2009.

\bibitem{Deb1995SimulatedBC}
K.~Deb and R.~B. Agrawal, ``Simulated binary crossover for continuous search
  space,'' \emph{Complex systems}, vol.~9, no.~2, pp. 115--148, 1995.

\bibitem{9528069}
A.~Fan, J.~Ma, X.~Jiang, and H.~Ling, ``Efficient deterministic search with
  robust loss functions for geometric model fitting,'' \emph{IEEE Transactions
  on Pattern Analysis and Machine Intelligence}, pp. 1--1, 2021.

\bibitem{ZHU2019205}
J.~Zhu, Z.~Jiang, G.~D. Evangelidis, C.~Zhang, S.~Pang, and Z.~Li, ``Efficient
  registration of multi-view point sets by k-means clustering,''
  \emph{Information Sciences}, vol. 488, pp. 205--218, 2019.

\bibitem{lsgcpd}
W.~Liu, H.~Wu, and G.~S. Chirikjian, ``Lsg-cpd: Coherent point drift with local
  surface geometry for point cloud registration,'' in \emph{IEEE/CVF
  International Conference on Computer Vision}, 2021, pp. 15\,293--15\,302.

\end{thebibliography}

\begin{IEEEbiography}[{\includegraphics[width=1in,height=1.25in,clip,keepaspectratio]{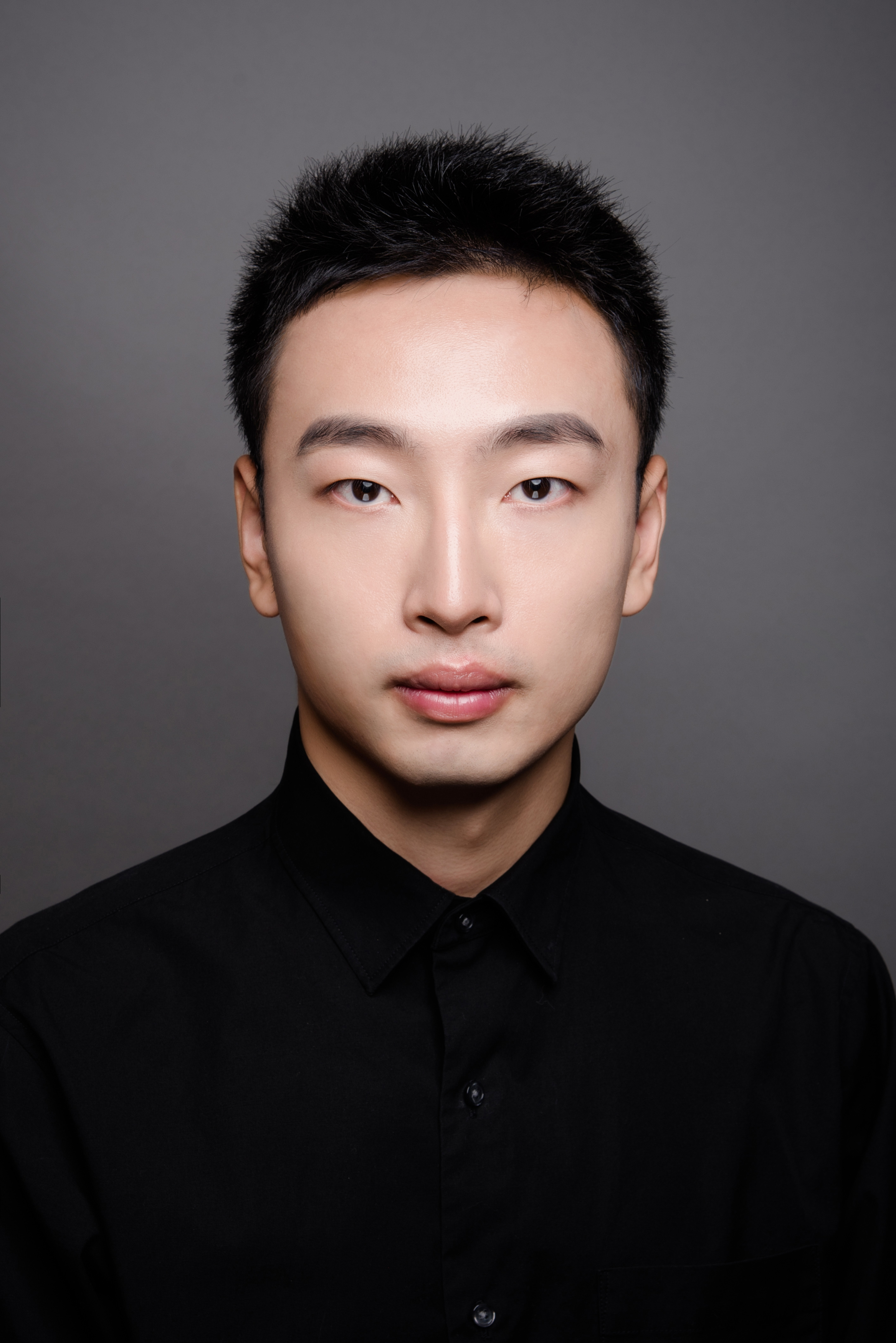}}]{Yue Wu} (Member, IEEE) received the B.Eng. and Ph.D. degrees from Xidian University, Xi’an, China, in 2011 and 2016, respectively. 
	
Since 2016, he has been a Teacher with Xidian University. He is currently an Associate Professor with Xidian University. He has authored or coauthored more than 70 papers in refereed journals and proceedings.

His research interests include computational intelligence and its Applications. He is the Secretary General of Chinese Association for Artificial Intelligence-Youth Branch, Chair of CCF YOCSEF Xi’an, Senior Member of Chinese Computer Federation. He is Editorial Board Member for over five journals, including Remote Sensing, Applied Sciences, Electronics, Mathematics.
\end{IEEEbiography}

\begin{IEEEbiography}[{\includegraphics[width=1in,height=1.25in,clip,keepaspectratio]{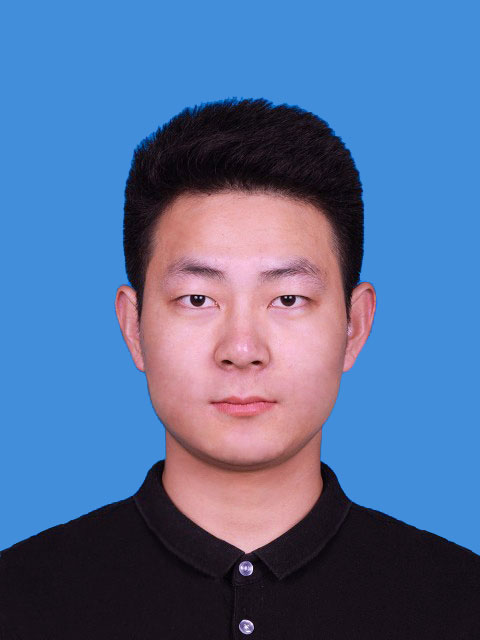}}]{Yibo Liu} received the B.Eng. degree in computer science and technology from North China Electric Power University, Beijing, China, in 2020. He is currently pursuing the master's degree with the Key Laboratory of Big Data and Intelligent Vision, School of Computer Science and Technology, Xidian University, Xi'an, China.
	
His research interests include point cloud registration and computational intelligence.
\end{IEEEbiography}

\begin{IEEEbiography}[{\includegraphics[width=1in,height=1.25in,clip,keepaspectratio]{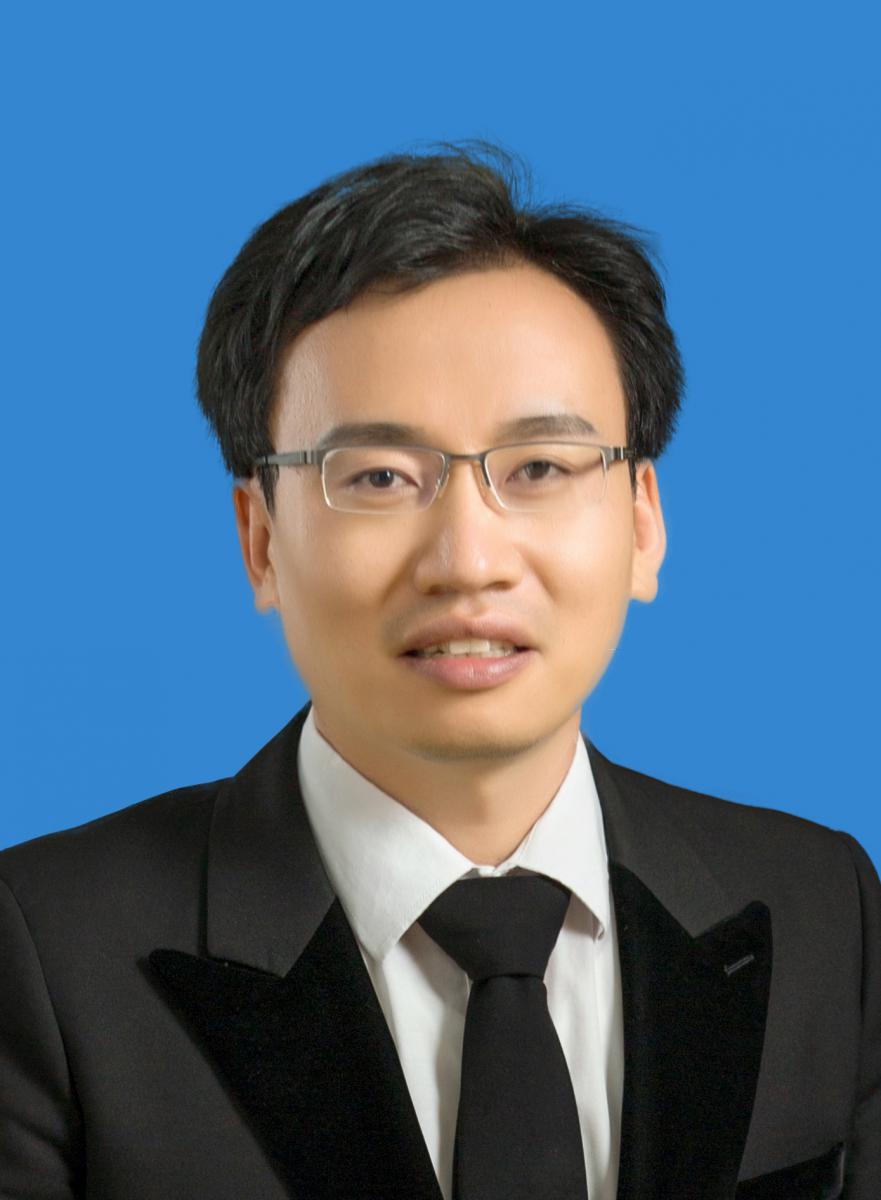}}]{Maoguo Gong} (Senior Member, IEEE) received the B.Eng. and Ph.D. degrees from Xidian University, Xi’an, China, in 2003 and 2009, respectively.

Since 2006, he has been a Teacher with Xidian University. He was promoted to an Associate Professor and a Full Professor, in 2008 and 2010, respectively, with exceptive admission. He has authored or coauthored over 100 articles in journals and conferences. He holds over 20 granted patents as the first inventor. He is leading or has completed over twenty projects as the Principle Investigator, funded by the National Natural Science Foundation of China, the National Key Research and De-velopment Program of China, and others.

His research interests are broadly in the area of computational intelligence, with applications to optimization, learning, data mining, and image understanding. Prof. Gong is the Executive Committee Member of Chinese Association for Artificial Intelligence and a Senior Member of Chinese Computer Federation. He was the recipient of the prestigious National Program for Support of the Leading Innovative Talents from the Central Organization Department of China, the Leading Innovative Talent in the Science and Technology from the Ministry of Science and Technology of China, the Excellent Young Scientist Foundation from the National Natural Science Foundation of China, the New Century Excellent Talent from the Ministry of Education of China, and the National Natural Science Award of China. He is an Associate Editor or an Editorial Board Member for over five journals including the IEEE TRANSACTIONS ON EVOLUTIONARY COMPUTATION and the IEEE TRANSACTIONS ON NEURAL NETWORKS AND LEARNING SYSTEMS.
\end{IEEEbiography}

\begin{IEEEbiography}[{\includegraphics[width=1in,height=1.25in,clip,keepaspectratio]{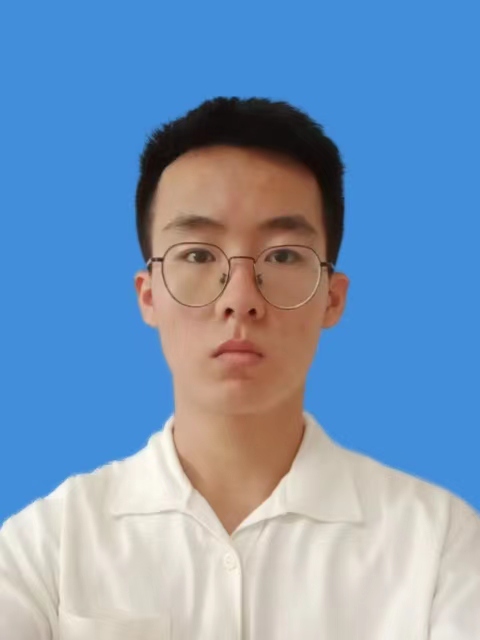}}]{Peiran Gong} received the B.Eng. degree from Nanjing Tech University, Nanjing, China, in 2021. He is currently pursuing the master's degree from the School of Computer Science and Technology, Xidian University, Xi'an, China.

His research interests include evolutionary computation and 3-D registration.
\end{IEEEbiography}

\begin{IEEEbiography}[{\includegraphics[width=1in,height=1.25in,clip,keepaspectratio]{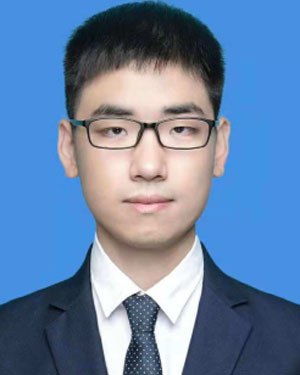}}]{Hao Li} received the B.S. degree in electronic engineering and the Ph.D. degree in pattern recognition and intelligent systems from Xidian University, Xi’an, China, in 2013 and 2018, respectively. He is currently an Associate Professor with the School of Electronic Engineering, Xidian University. His research interests include computational intelligence and machine learning.
\end{IEEEbiography}

\begin{IEEEbiography}[{\includegraphics[width=1in,height=1.25in,clip,keepaspectratio]{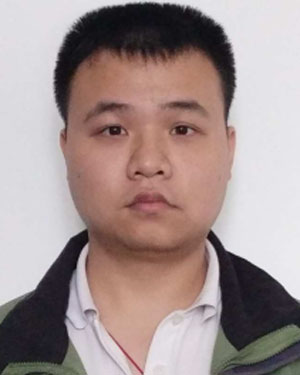}}]{Zedong Tang} received the B.Eng. and Ph.D. degrees from Xidian University, Xi’an, China, in 2014 and 2020, respectively. Since 2020, he has been a Lecturer with the Academy of Advanced Interdisciplinary Research, Xidian University. His current research interests include computational intelligence and machine learning. 
\end{IEEEbiography}

\begin{IEEEbiography}[{\includegraphics[width=1in,height=1.25in,clip,keepaspectratio]{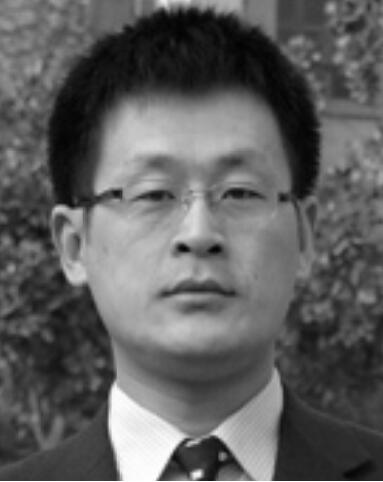}}]{Qiguang Miao} (Senior Member, IEEE) received the M.Eng. and Doctor degrees in computer science from Xidian University, Shaanxi, China.
	
He is currently a Professor with the School of Computer Science and Technology, Xidian University. His research interests include intelligent image processing and multiscale geometric representations for images. 
\end{IEEEbiography}

\begin{IEEEbiography}[{\includegraphics[width=1in,height=1.25in,clip,keepaspectratio]{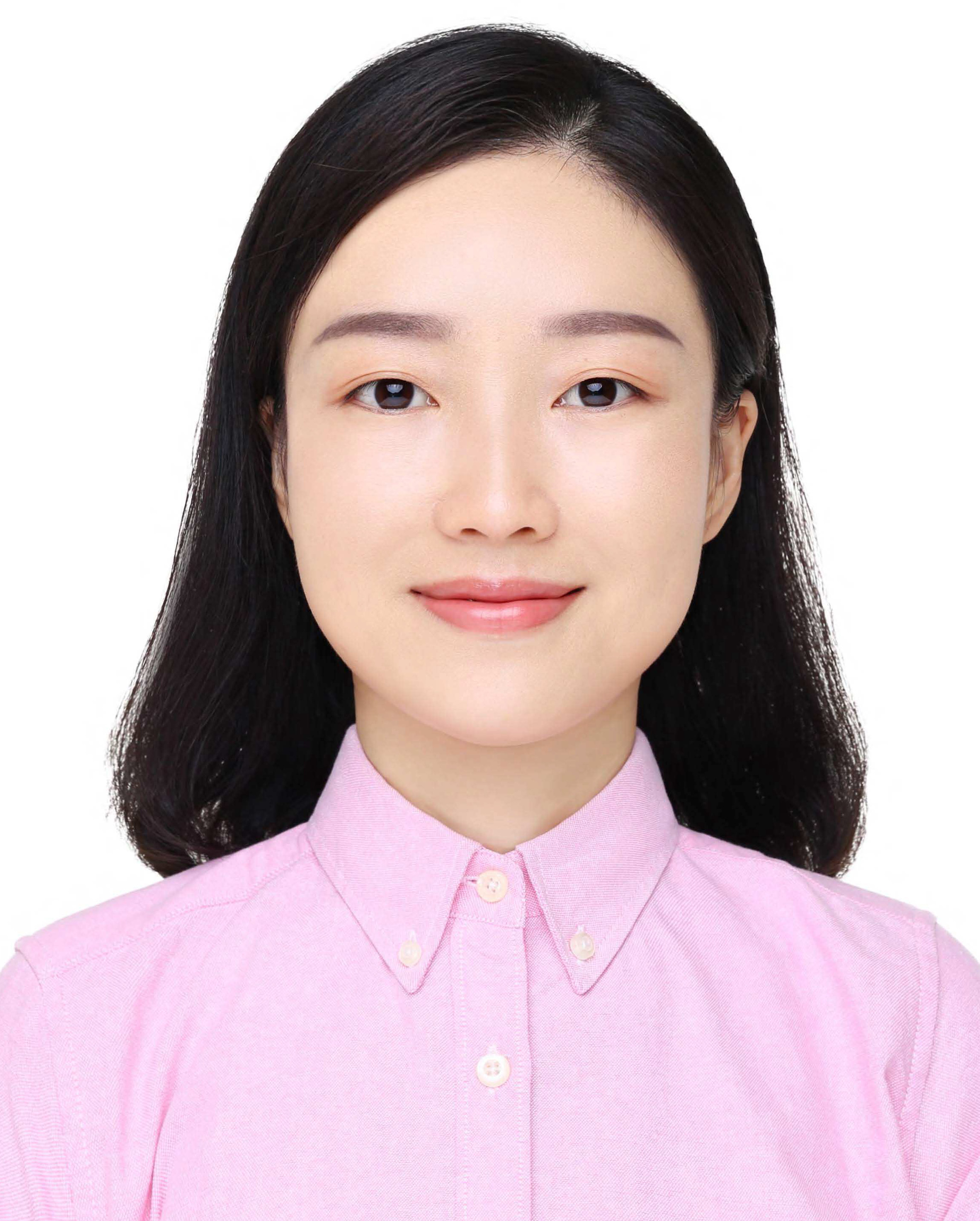}}]{Wenping Ma} (Senior Member, IEEE) received the B.S. degree in computer science and technology and the Ph.D. degree in pattern recognition and intelligent systems from Xidian University, Xi’an, China, in 2003 and 2008, respectively.

She is currently a Professor with the School of Artificial Intelligence, Xidian University. Her research interests include natural computing and intelligent image processing.

Dr. Ma is a member of the Chinese Institute of Electronics.
\end{IEEEbiography}

\end{document}